\documentclass[11pt]{book}
\usepackage{fullname,times,epic,eepic,epsfig,pifont,fancyheadings,doublespace,examples,newalg}

\renewcommand{\Pr}{{\rm P}}
\newcommand{\Prhat}{{\widehat{\Pr}}}
\newcommand{\PrG}{\Pr_{\rm G}}
\newcommand{\scS}{{\mathcal{S}}}
\newcommand{\scSx}{{\cal{S}}}
\newcommand{\scH}{{\cal{H}}}
\newcommand{\lc}[2]{#1\raise 0.1 em \hbox{\small --}#2}
\newcommand{\pt}[2]{#1\raise 0.17em \hbox{\tiny$\backslash$}#2}
\newcommand{\LC}{{\cal LC}}
\newcommand{\FLC}{{\cal FLC}}
\newcommand{\LF}{{\cal LF}}
\newcommand{\CON}{{\cal CN}}

\newcommand{\rewrites}{\;\rightarrow\;}

\newcounter{subequation}[equation]
\renewcommand{\thesubequation}{\alph{subequation}}
\newcommand{\twoparteqno}{\mbox{\theequation\thesubequation}}

\newcommand{\LCX}[1]{\large$\displaystyle\mathop{\mbox{\huge$\Rightarrow$}}_{\LC}$}

\newcommand{\neqno}[1]{ %
  \stepcounter{equation}\stepcounter{subequation}(\twoparteqno) %
  \let\@currentlabel\twoparteqno \label{#1}}

\makeatletter
\newcommand{\nneqno}[1]{ %
  \stepcounter{subequation}(\twoparteqno) %
  \let\@currentlabel\twoparteqno \label{#1}}

\newcommand{\nnxform}[3]{#1 & #2 & \nneqno{#3}}

\newcommand{\textttt}[1]{\textsf{\textsl{\small #1}}}
\newcommand{\textttf}[1]{\textsf{\textsl{#1}}}

\begin{document}
\raggedbottom\
\setlength\topmargin{0in}
\setlength{\headsep}{0.1in}
\setlength{\oddsidemargin}{0.55in}
% if double sided:
\setlength{\evensidemargin}{0in}
% if single sided:
%\setlength{\evensidemargin}{0.55in}

\pagenumbering{roman}

{\pagestyle{empty}

\begin{center}

\vspace*{1.0in}

% 
% TITLE
%
{\LARGE Robust Probabilistic Predictive Syntactic Processing:}\\
{\Large Motivations, Models, and Applications}

\vspace*{1.0in}

%
% NAME AND PREVIOUS DEGREES
%
by\\{\Large Brian Edward Roark}\\
{\large B.A., University of California, Berkeley, 1989}\\
{\large M.S., Claremont Graduate School, 1997}\\

\vspace*{1.3in}

%
% DEGREE AND DEPARTMENT
%
A dissertation submitted in partial fulfillment of the \\
requirements for the degree of Doctor of Philosophy\\
in the Department of Cognitive and Linguistic Sciences\\
at Brown University
\vspace*{0.5in}

%
% DATE
%
May 2001\\
\end{center}
\large
\normalsize
\clearpage}

%
% COPYRIGHT PAGE
%
{\pagestyle{empty}
\vspace*{3in}

%
% NAME AND DATE
%
\begin{center}
\Pisymbol{psy}{211} Copyright\\
by\\
Brian Edward Roark\\
2001
\end{center}
\clearpage}

\pagestyle{plain}
%\setcounter{page}{2}

%
% SIGNATURE PAGE
%
\begin{spacing}{1.1}
\begin{center}
\vspace*{0.05in}

%
% NAME, DEPARTMENT DEGREE
%
This thesis by Brian E. Roark is accepted in its present form by \\
the Department of Cognitive and Linguistic Sciences \\
as satisfying the dissertation requirement for \\
the degree of Doctor of Philosophy\vspace*{0.5in}

\end{center}

{\hspace*{0.5in}Date\makebox[1.0in]{\dotfill}\hspace{0.5in}\makebox[3.0in]{\dotfill}\\}
% NAME OF ADVISOR
\hspace*{2.75in}Mark Johnson, Director\vspace*{0.35in}

\begin{center}
{Recommended to the Graduate Council}\vspace*{0.4in}
\end{center}

{\hspace*{0.5in}Date\makebox[1.0in]{\dotfill}\hspace{0.5in}\makebox[3.0in]{\dotfill}\\}
% COMMITTEE MEMBER #1
\hspace*{2.75in}Eugene Charniak, Reader\vspace*{0.35in}

{\hspace*{0.5in}Date\makebox[1.0in]{\dotfill}\hspace{0.5in}\makebox[3.0in]{\dotfill}\\}
% COMMITTEE MEMBER #2
\hspace*{2.75in}Julie Sedivy, Reader\vspace*{0.35in}

{\hspace*{0.5in}Date\makebox[1.0in]{\dotfill}\hspace{0.5in}\makebox[3.0in]{\dotfill}\\}
% COMMITTEE MEMBER #3
\hspace*{2.75in}Frederick Jelinek, Reader\vspace*{0.35in}

%\hspace{0.1in}{Date\makebox[1.0in]{\dotfill}\hspace{0.5in}\makebox[2.5in]{\dotfill}\\}
\begin{center}
{Approved by the Graduate Council}\vspace*{0.35in}
\end{center}

{\hspace*{0.5in}Date\makebox[1.0in]{\dotfill}\hspace{0.5in}\makebox[3.0in]{\dotfill}\\}
\hspace*{2.75in}Peder J. Estrup\\
\hspace*{2.75in}Dean of the Graduate School and Research\\

\end{spacing}
%\end{center}

\clearpage

{\pagestyle{empty}
\addcontentsline{toc}{chapter}{Abstract}
\vspace*{1in}

Abstract of ``Robust Probabilistic Predictive Syntactic Processing'' by
Brian Edward Roark, Ph.D., Brown University, May, 2001.

\vspace*{.5in}

This thesis presents a broad-coverage probabilistic top-down parser,
and its application to the problem of language modeling for speech
recognition.  The parser builds fully connected derivations
incrementally, in a single pass from left-to-right across the string. 
We argue that the parsing approach that we have adopted is
well-motivated from a psycholinguistic perspective, as a model that
captures probabilistic dependencies between lexical items, as part of
the process of building connected syntactic structures.
The basic parser and conditional probability models are presented, and
empirical results are provided for its parsing accuracy on both
newspaper text and spontaneous telephone conversations.  Modifications
to the probability model are presented that lead to improved
performance.  A new language model which uses the output of the parser
is then defined.  Perplexity and word error rate reduction
are demonstrated over trigram models, even when the trigram is trained
on significantly more data.  Interpolation on a word-by-word basis
with a trigram model yields additional improvements.   
}
\clearpage

\chapter*{Acknowledgments}
\addcontentsline{toc}{chapter}{Acknowledgments}
I would like to thank Mark Johnson for invaluable discussion,
guidance and moral support over the course of my time here at Brown.
It has been a fertile, collaborative atmosphere under his direction,
and the research perspective which he cultivates is one that I will
attempt to carry with me.  Eugene Charniak also contributed greatly to
my work here at Brown, and I'd like to thank him, and Mark, for
maintaining a truly interdisciplinary research group here. 

I have benefited immensely from being in this small, diverse department,
which presented me with so many perspectives on language, computation,
linguistics, and cognitive science in general, many of which I have
internalized.  Thanks so much to Polly Jacobson for being such a
wonderful teacher; she's influenced me more than she thinks.  Also to
Julie Sedivy for kindling an interest in many of the issues addressed
in this thesis; and to Katherine Demuth for encouraging and
collaborating on projects that took me away from my narrow specialty.
Thanks also to Philip Lieberman and Molly Homer.  

The collegiality among graduate students here, and their diversity of
interests, has made for a rewarding work environment.  The
computational students in the department -- including myself,
Massimiliano Ciaramita, Rodrigo Braz, and Yasemin Altun -- have been 
mutually supportive in the extreme, and I appreciate being part of
that.  Thanks also to the graduate and undergraduate students who have
been part of the Brown Laboratory for Linguistic Information Processing
(BLLIP) over the years; in addition to Massi and Yasemin: Don Blaheta,
Sharon Caraballo, Heidi Fox, Niyu Ge, Keith Hall, John Hale, Sharon
Goldwater, Gideon Mann, and Matt Berland.  The weekly group meetings
were among my best learning experiences here at Brown.  Last, but not
least, thanks for that odd intangible that I have received by virtue
of sharing an office with Jesse Hochstadt.

Thanks to our friends from the Center for Language and Speech Processing
at Johns Hopkins -- Frederick Jelinek, Ciprian Chelba, Sanjeev
Khudanpur, and Peng Xu -- for comments and support in a truly
collaborative spirit.

A special additional thanks to both Eugene Charniak and Ciprian Chelba
for providing some of the code used at certain points in the work
reported in this thesis.  They both saved me much time and effort that
would have been spent writing that code.

Some of the results presented in this thesis have appeared, in a
slightly different form, in \namecite{Roark99b} and
\namecite{Roark01}.

Finally, this thesis would not have been written without the tireless
and loving support of my wife Luciana.  Having a family at home has
made all the difference for me, and I recognize that the time that I
have devoted to the research in this thesis has been devoted by her as
well.  She has my love and thanks.  This thesis is dedicated to my
mother, Barbara Jean Harman (b. Jan. 8, 1940; d. Jan. 10, 2000); and
to my son, Max Emilio Roark (b. Jan. 16, 1998).

\tableofcontents

\clearpage

\addcontentsline{toc}{chapter}{List of Figures}
\listoffigures

\clearpage
\listoftables
\addcontentsline{toc}{chapter}{List of Tables}

\clearpage

\pagenumbering{arabic}

\pagestyle{fancy}

% [even {plain}{normal}]{ odd {plain}{normal}}
% plain pages are those that start a chapter

\chead{}

% if double-sided
%\lhead[\fancyplain{}{\rm\thepage}]{\fancyplain{}{}}
%\rhead[\fancyplain{}{}]{\fancyplain{}{\rm\thepage}}

%elseif single-sided
\lhead{}
\rhead[\fancyplain{}{\rm\thepage}]{\fancyplain{}{\rm\thepage}}

\lfoot{}
\cfoot[\fancyplain{\rm\thepage}{}]{\fancyplain{\rm\thepage}{}}
\rfoot{}
\setlength{\headrulewidth}{0pt}

\chapter{Introduction}
Statistical methods have become central within computational
linguistics in the past decade, providing the means by which many
technologies can be applied to freely occurring language.  For
example, broad-coverage parsing of English has become feasible through
the use of statistical techniques and highly ambiguous grammars.
Grammars of natural language syntax (at this stage of their
development) fall into one (or more likely both) of two classes:
those that undergenerate the language they are intended to model --
i.e. those that fail to cover grammatical sentences in the language --
and those that overgenerate the language -- i.e. cover grammatical
sentences and many ungrammatical ones besides.  Grammars that attempt
to cover freely occurring language, i.e. to provide syntactic analyses
for arbitrary strings of the language, must include enough rules to
handle a large variety of potential syntactic structures, and this
leads to both overgeneration and ambiguity -- they provide a great
many possible parses for each grammatical string.  If one parses
freely occurring language with a non-stochastic grammar, one either
fails to parse a large proportion of the sentences (if the grammar
undergenerates) or finds a very large number of parses for each
sentence (if the grammar is ambiguous), and most likely both.
Statistical methods do not change this situation, but 
they do ameliorate the problems associated with ambiguous
grammars, insofar as they can be used to effectively select from among
or prune the set of parses for a particular string.  Broad-coverage
statistical parsers now efficiently achieve 100 percent coverage of
freely occurring language with a very high accuracy.  Similar
advantages have been found in many applications, such as large
vocabulary speech recognition and machine translation.

Statistical methods and increased computing power have combined to
make techniques possible which were unrealistic even a few
years ago.  Some of these techniques rely exclusively on statistics;
for example, word clustering based on co-occurrence, irrespective of
any structural relationship between the words -- so-called bag-of-words
approaches.  Others look to combine structured and stochastic methods,
e.g. the statistical parsing example above.  This combination can and
does make the difference between success and failure in many areas of
computational linguistics, which opens up the possibility that
non-statistical approaches that were considered and discarded as
impractical in the 
past may have the potential to be successfully applied, when
appropriately combined with statistical methods.  This thesis
investigates one such resurrection: incremental top-down parsing.

The principal contribution of this thesis lies in the exploitation of
probabilistic lexico-syntactic information provided by a robust
incremental parser for language modeling.  We present a high accuracy,
efficient incremental parser, and the parsing model will be profitably
applied unmodified as a language model for speech recognition in
multiple test domains.

While the empirical work presented in this thesis will be strictly 
computational, and of primary interest to computational linguists,
it aspires to an interdisciplinary appeal.  In particular, there are
issues in the study of human sentence processing that have been
largely ignored or glossed over within certain currently influential
approaches, and we will spend a chapter discussing these issues.  In
short, people have been repeatedly shown to interpret sentences
incrementally, 
and to use interpretations on-line to influence future interpretation.
While many studies show the contextual sensitivity of on-line
interpretation, and provide models to account for this sensitivity, 
the syntactic implications of these models receive much less
attention.  We will show that this can be more than simply a problem
of leaving certain details of the model less than fully specified, but
one which has the potential to change some models' empirical predictions.

An additional question that might be asked about incremental
interpretation in human sentence processing is:  why?  It is certainly
imaginable that the sentence processing mechanism could have evolved
to interpret sentences only once they have been seen in entirety.  The
computational benefits of delaying the composition of constituents
until those constituents are complete can be very large - see the
discussion of dynamic programming 
below.  Yet this is not the case.  The results in this thesis point
the way to a reason for this question, one which has only just begun
to be investigated in the psycholinguistic literature \cite{Borsky98}.
The ultimate computational payoff to resurrecting top-down parsing
will be in language 
modeling for speech recognition.  The ease with which words can be
incorporated into parses (or interpretations) can be used as a measure
for resolving acoustic ambiguities.  The additional ambiguities
introduced by continuous speech, and the reliance upon many levels of
context (e.g. phonetic, phonological, syntactic) to 
resolve these ambiguities, is one potential reason for incremental
interpretation in human sentence processing.

The thesis will be organized as follows:  this chapter will introduce
the central problems, as well as some
basic terminology; the next chapter will discuss psycholinguistic
models of sentence processing, and their implicit demands upon
syntactic processing; chapter 3 will outline probabilistic top-down
parsing, and report on extensive trials with the basic models; chapter
4 will present refinements to these models, aimed at improving
accuracy, efficiency, and coverage; finally, chapter 5 will present
the application of this parsing model to language modeling for speech
recognition.  

This thesis is intended to be of interest to more than one research
community.  However, not everyone will be uniformly interested in what
is being presented.  For this reason the next sub-section will provide a
roadmap to the contents, complete with advice on what readers
with particular interests can skip.  

\section{Roadmap}
Depending on the reader's primary area of interest, the relevant
chapters in this thesis may differ.  Everyone is invited to read the
thesis in its entirety, but the following three paths through the
thesis are probably sufficient for those with focused interests.

\subsection{Computational linguistics}
For the reader who is primarily interested in computational
linguistics, chapters 3-5 will be of the greatest interest.  For an
informal discussion of parsing, see section \ref{sec:i} through
sub-section \ref{subsec:spar}.  Subsequent sections of chapter one and
all of chapter two deal with models of human sentence processing, and
are not required to begin reading from chapter three.

\subsection{Psycholinguistics}
Chapters one and two present the psycholinguistic claims in this
thesis.  Chapters 3-5 are strictly computational, and
serve the earlier psycholinguistic arguments as something of an
existence proof -- that the kind of parser that we advocate can
perform very well, even with a very ambiguous grammar.  Thus, the
reader who is primarily interested in psycholinguistics should read
chapters one and two in entirety, including the informal introduction
to parsing in the next 
section, since it will introduce a perspective on syntactic parsing
and much vocabulary that will be important to understanding chapter
two. 

\subsection{Speech recognition}
The empirical results in speech recognition are presented in chapter
five.  However, to understand how the parser functions (and hence how
the language model works), it is a good idea for people whose primary
interest is in speech recognition to follow the recommendations for
those interested in computational linguistics above.

\section{Problem statement and background}\label{sec:i}
This thesis deals with the following problem:  incremental
interpretation requires hypothesizing semantic relationships between
constituents before they are necessarily completed; yet,
because of the large number of local ambiguities, the most efficient
way to parse a string into constituent structure is through dynamic
programming.  Jurafsky saw this problem, and made the following
distinction, which we will see again (in a slightly different form) in
the asynchronous model in \namecite{Shieber93}:  
\begin{quote}
\setlength{\baselineskip}{.7\baselineskip}
We assume that the human parser will need to use some such dynamic
programming algorithm for parsing control and for rule-integration.\
\ldots \ However, the efficiency gained by dynamic programming
algorithms for pure syntax may not generalize to the problem of
interpretation. \cite[p. 142]{Jurafsky96}
\end{quote}
From our perspective, this distinction between ``pure syntax'' and
interpretation is unnecessarily complicated.  A simpler model is one
in which syntactic and semantic processing are very tightly coupled,
with fully connected syntactic structures.  Such a model, however,
has been considered problematic for a variety of reasons.  Let us
spend some time making clear why we take the perspective we do, and
the underlying assumptions of both positions. 

The nature of interpretation is not the subject of this thesis, and we
will not provide a model of interpretation.  Any model of
interpretation, however, will take as its starting point some kind of
compositional semantic processing, which productively combines the
meanings of smaller constituents into meanings of larger
constituents.  For example, the meaning of the nominal constituent
\textttt{`green sky'} is derived from the meanings of
\textttt{`green'} and \textttt{`sky'} via some productive rule that
also allows one to derive the meaning of \textttt{`purple sky'}.  This
sort of compositional semantic processing is at least in part
syntactic, to the extent that decisions must be made about which
constituents compose.  Consider, for example, \textttt{`drunken
airplane pilots'}, which is ambiguous between ``drunken pilots of
airplanes'' and ``pilots of drunken airplanes''.  We know that living
beings get drunk, and that pilots, not airplanes, are alive, which
provides a basis for disambiguation.  The semantic composition that
allows us to interpret \textttt{`drunken airplane pilots'} must
decide, or make a guess, at the appropriate compositional structure.

Incremental interpretation will also involve some kind of semantic
composition, resulting in a partial interpretation which may or may not
be correct.  For example, \textttt{`every author read \ldots'} is
ambiguous between what is called the main verb reading and what is
called the reduced relative reading.  The main verb reading would
continue: \textttt{`every author read in the garden.'}  The reduced
relative reading would continue: \textttt{`every author read by the
children voted for Roosevelt.'}  Incremental interpretation involves 
distinguishing the two interpretations, and deciding (perhaps) that
\textttt{`every author'} is a good agent for the verb, thus
immediately favoring a main verb reading of the sentence, in advance
of any truly disambiguating material.  The semantic ambiguity between
the two readings could be represented as follows, using standard
formal semantic notation
\begin{examples}
\item $\lambda$y$\forall$x[author(x) $\rightarrow$ read(x,y)]
\item $\lambda$P$\forall$x[author(x) $\wedge$ $\exists$yread(y,x)
$\rightarrow$P(x)] 
\end{examples}
In both cases, function composition can allow for meaning composition
before the constituents are complete.  Semantically, the two
constituents, the subject NP and the verb, compose to give very
different meanings, which correspond to different interpretations.  As
stated earlier, this incremental semantic composition is a
prerequisite for incremental interpretation.  It involves
hypothesizing compositional structure(s), and perhaps disambiguating
based on considerations such as thematic fit.

Evidence for rapid, incremental interpretation is provided by
garden-path effects.  People have difficulty, for example, with
reduced relative 
clauses, generally preferring the main verb reading, such as in the
above example.  Yet these preferences can vary, depending on many
factors, such as thematic fit.  In contrast to the above example, a
string like 
\textttt{`the newspaper read \ldots'} may not disprefer a reduced
relative reading so dramatically, by virtue of \textttt{`the
newspaper'} being a poor agent for the main verb.  In order to
evaluate these alternatives, they must be hypothesized, i.e. the
potential relationships (compositional structure) between constituents
must be made explicit immediately.

To get back to the distinction made in the Jurafsky quote above, both
parsing and interpretation involve 
hypothesizing relationships between constituents -- composing smaller
constituents to form larger constituents -- and hence it is not
clear how an efficiency gain in ``parsing'' but not in
``interpretation'' would impact the language comprehension process as
a whole.  To the extent that two constituents must be hypothesized
to compose semantically for interpretation, what is the gain in not
hypothesizing this composition syntactically?

Let us be clear from the outset what we mean by ``incremental
interpretation''.  An incremental {\it parser\/} builds syntactic
structure from left-to-right over the string.   There are methods that
couple syntactic and semantic processing in such a way that, to the
extent that the parser is incremental, so is the semantic processing.
Such a processing model can and has been used for improving search, by
pruning certain semantically non-viable analyses --
e.g. \namecite{Dowding94}.  This, however, is not the sense of
incremental interpretation that we are using.  This kind of
incremental parsing produces disconnected fragments of
interpretations, and waits until constituents are complete before
composing them to form larger constituents.  Incremental
interpretation composes meanings before constituents are complete, to
provide a single, connected analyses for the string to that point.
Consider again the example \textttt{`the authors read \ldots'}.  An
incremental bottom-up parser will wait to compose the subject NP and
main VP until they have both been completed, and hence will not
provide a single, connected analysis until the end of the string.

Incremental interpretation requires enough connected constituent
structure to distinguish between competing analyses.  This might be
done via a sentence processing mechanism that performs bottom-up
syntactic parsing, coupled with some additional semantic processing,
such as the mechanism proposed in \namecite{Shieber93}, which will be
discussed in detail in Chapter 2.  In such an architecture, the
syntactic constituent structure is left underspecified, while semantic
connections between constituents and words are hypothesized by a
semantic module.  A simpler architecture than this is one in which
there is a one-to-one correspondence between the syntactic and semantic
compositional structure, and parsing and interpretation are part of
the same process. This is known as the rule-to-rule hypothesis, and it
is standardly assumed in theories of compositional semantics.  Under
such a hypothesis, a top-down parser builds enough structure for
incremental interpretation, since the hypothesized syntactic structure
will directly correspond to a hypothesized semantic structure.  It
should be pointed out that a top-down parser of this sort may build more
structure than is necessary for immediate incremental interpretation.
For example, consider the partial string \textttt{`I saw Jim's
\ldots'}.  Whether or not a link between the verb and 
the constituent begun by the possessive must be hypothesized
immediately to allow for incremental interpretation of the sort that
people appear to perform is unclear.  Certainly by the time the head
of the noun phrase is encountered, such a link must be hypothesized. 
Another predictive parsing approach, left-corner parsing, would also
build enough structure in this case to allow for the kind of
discrimination that is observed in people.  A parsing approach that
delays hypothesizing structure until precisely when it is needed would
meet our criterion for providing enough structure for incremental
interpretation.  In the absence of the knowledge of where that point
is for every construction, we can investigate a top-down parser that
will, in every case, provide enough.

The approach that we will advocate is incremental predictive parsing,
without dynamic programming.  If it can be shown that an incremental
parser that builds enough constituent structure for incremental
interpretation is viable in a highly ambiguous, broad-coverage domain,
then there is no reason to expect that the human sentence processing
mechanism needs to resort to dynamic programming.  Once such a parser
has been constructed, we can ask whether the approach buys us
something computationally in return for not using dynamic
programming.  Hence there is both a psycholinguistic and computational
motivation for the empirical work in this thesis.

The danger in trying to produce something of interest to two research
communities\footnote{In fact, since computational linguists and speech
recognition researchers form largely separate communities, once
recognition results are introduced, the number of distinct audiences
grows to three.} is that one might actually produce something of interest
to nobody at all.  To avoid this fate, we will spend some time
at the outset establishing a common vocabulary, and introducing some
of the fundamental issues at stake for each of the research areas.
This is also intended to delimit the topic a bit, so that we can later 
focus upon just those issues about which our empirical work has
something to say.  First we will discuss the computational background
for parsing.  We will remain informal at this stage, to facilitate
interdisciplinary understanding;  formal computational details will be
provided as the approach is developed in later chapters.  This
computational introduction will be followed by a discussion of certain
central ideas in modeling the human sentence processing mechanism.

\subsection{Parsing}\label{subsec:ipar}
\begin{figure}[t]
\begin{small}
\begin{picture}(396,188)(0,-188)
\put(50,-30){(a)}
\put(53,-52){S}
\drawline(56,-56)(21,-66)
\put(14,-74){NP}
\drawline(21,-78)(7,-88)
\put(-0,-96){DT}
\drawline(7,-100)(7,-110)
\put(1,-118){the}
\drawline(21,-78)(34,-88)
\put(26,-96){NN}
\drawline(34,-100)(34,-110)
\put(24,-118){thief}
\drawline(56,-56)(92,-66)
\put(85,-74){VP}
\drawline(92,-78)(64,-88)
\put(52,-96){VBD}
\drawline(64,-100)(64,-110)
\put(55,-118){saw}
\drawline(92,-78)(120,-88)
\put(114,-96){NP}
\drawline(120,-100)(89,-110)
\put(82,-118){NP}
\drawline(89,-122)(76,-132)
\put(69,-140){DT}
\drawline(76,-144)(76,-154)
\put(70,-162){the}
\drawline(89,-122)(101,-132)
\put(93,-140){NN}
\drawline(101,-144)(101,-154)
\put(94,-162){cop}
\drawline(120,-100)(152,-110)
\put(146,-118){PP}
\drawline(152,-122)(129,-132)
\put(123,-140){IN}
\drawline(129,-144)(129,-154)
\put(119,-162){with}
\drawline(152,-122)(175,-132)
\put(169,-140){NP}
\drawline(175,-144)(156,-154)
\put(149,-162){DT}
\drawline(156,-166)(156,-176)
\put(149,-184){the}
\drawline(175,-144)(195,-154)
\put(184,-162){NNS}
\drawline(195,-166)(195,-176)
\put(173,-184){binoculars}
\put(227,-8){(b)}
\put(230,-30){S}
\drawline(233,-34)(186,-44)
\put(179,-52){NP}
\drawline(186,-56)(172,-66)
\put(165,-74){DT}
\drawline(172,-78)(172,-88)
\put(166,-96){the}
\drawline(186,-56)(199,-66)
\put(191,-74){NN}
\drawline(199,-78)(199,-88)
\put(189,-96){thief}
\drawline(233,-34)(280,-44)
\put(273,-52){VP}
\drawline(280,-56)(229,-66)
\put(217,-74){VBD}
\drawline(229,-78)(229,-88)
\put(221,-96){saw}
\drawline(280,-56)(267,-66)
\put(260,-74){NP}
\drawline(267,-78)(254,-88)
\put(247,-96){DT}
\drawline(254,-100)(254,-110)
\put(248,-118){the}
\drawline(267,-78)(279,-88)
\put(272,-96){NN}
\drawline(279,-100)(279,-110)
\put(272,-118){cop}
\drawline(280,-56)(330,-66)
\put(324,-74){PP}
\drawline(330,-78)(307,-88)
\put(301,-96){IN}
\drawline(307,-100)(307,-110)
\put(297,-118){with}
\drawline(330,-78)(354,-88)
\put(347,-96){NP}
\drawline(354,-100)(334,-110)
\put(327,-118){DT}
\drawline(334,-122)(334,-132)
\put(328,-140){the}
\drawline(354,-100)(373,-110)
\put(362,-118){NNS}
\drawline(373,-122)(373,-132)
\put(351,-140){binoculars}
\end{picture}
\end{small}
\caption{Two parse structures for an ambiguous string: (a) NP-attached
structure; and (b) VP-attached structure}\label{fig:parse1}
\end{figure}
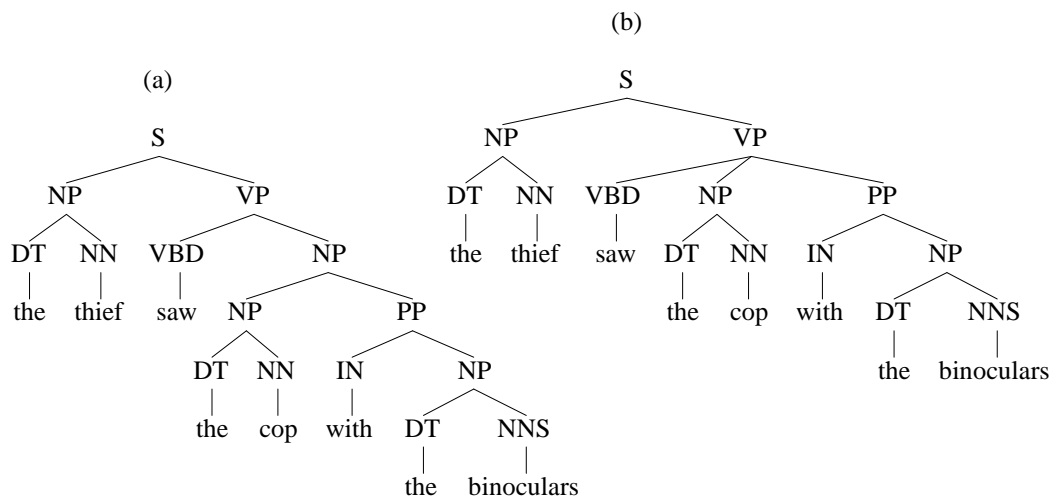

Parsing can be decomposed into several sub-tasks.  The first is
identification of constituents.  For example, in figure
\ref{fig:parse1}, the substring \textttt{`the thief'} in the string
\textttt{`the thief saw the cop with the binoculars'} is quite
uncontroversially a constituent, whatever its internal structure may
be.  We can identify the constituent by a beginning word
(\textttt{`the'}), an ending word (\textttt{`thief'}), and 
some kind of label.  Less clear, and hence more controversial, are
the appropriate labels for constituents, and their internal
structure.  The labels and structures shown in figure
\ref{fig:parse1} come from the Penn Treebank \cite{Marcus93}, whose
conventions we have adopted for no theoretical reason, but simply
because our approach requires a treebank, and this is the only one of
an adequate size available.

Note that the internal structure of the NP constituents in the Penn
Treebank is flat, with no $\bar{\mathrm{N}}$
constituents or compounding structure.  While this decision (and
others) about the constituent structure are perhaps theoretically
unsatisfying, they do not seriously impact the results in this thesis,
insofar as we are investigating whether a parser of a certain type can
find good parses in a very large search space.  If we collect all of
the context-free grammar rules from the trees of the Penn Treebank,
the resulting grammar is very ambiguous.  In addition, it is a
heavily left-recursive grammar, which is a well-known problem for
top-down parsers.  Hence this particular parsing domain provides
a more-than-suitable test bed for an incremental predictive parser.
By using this grammar, we are not making any claims 
about the actual constituent structure of English, although we take
the existence of such basic constituents as noun phrases (NP), verb
phrases (VP), prepositional phrases (PP), and clauses (S) as
uncontroversial.

At this point it is convenient to begin discussing constituents as
nodes in a tree representation, such as that presented in figure
\ref{fig:parse1}.  A tree is a labelled, ordered, directed graph consisting of
nodes and arcs between nodes.  The arc represents the parent/child
relationship between nodes or constituents: the arc goes from the
parent to the child.  By definition every node in a tree, excepting
the root, has exactly one parent node (see e.g. \egcite{Aho86}.
Nodes in the tree with no outgoing arcs -- i.e. no children -- are
called {\it leaves\/}.  There are two kinds of node labels: (i) those
that are words in the language, which are sometimes called {\it
terminal items\/} or {\it terminals\/}, by virtue of the fact that 
they terminate that branch of the tree (they cannot have children);
and (ii) {\it non-terminals\/}, which are not part of the string.  If
a leaf node of the tree has a non-terminal label, it is called an {\it
empty node\/}, since it does not produce any terminals in the string.
By convention, 
in the Penn Treebank there are two disjoint sets of non-terminals,
those which can and those which cannot be the parent of a terminal
item.  Those which can be the parent of a terminal item are called
{\it pre-terminals\/}, or {\it parts-of-speech\/} (POS), and they can
only be the 
parent of one terminal item at a time.  These include things like
determiner (DT), singular noun (NN), and preposition (IN).  Again,
we have no theoretical reason for adopting the convention that POS
constituents are disjoint from other non-terminal constituents, but we
follow the Penn Treebank for purely pragmatic reasons.  

Given the pervasive use of trees as objects of interest in their own
right in linguistic and psycho-linguistic theory, let us make clear
that we consider trees a notational convenience and not an end-product
of the parsing and interpretation process.  Trees represent the
compositional structure of a string of words, i.e. how smaller
constituents compose to form larger constituents.  Ideally,
composition of constituents is both syntactic and semantic; both
parsing and interpretation involve hypothesizing that particular constituents 
compose to form larger constituents.  Trees are a convenient graphical 
representation for this compositional, hierarchical constituent
structure, and (from our perspective) nothing else\footnote{We must
offer a caveat to these remarks, that the trees in the Penn treebank
are not suited to the kind of rule-to-rule syntactic/semantic
processing that we would favor.  For example, the flat NP constituents
underspecify the rich compositional structure of noun phrases in
English.  The parsers, however, that we will be investigating do not
depend on a particular kind of constituent structure, but can handle any
constituent structure encoded in a treebank, including grammars more
appropriate for a rule-to-rule correspondence.  Once again, however, our
hands are tied by the lack of large treebanks apart from the ones that
we are using.}.

The two parse structures in figure \ref{fig:parse1} represent two
hypotheses of the constituent structure of the sentence.  These
correspond to two interpretations: (a) the cop had the binoculars; or
(b) the thief had the binoculars, and used them to see the cop.  In
terms of beginning and ending locations of constituents, and their
labels, the two hypotheses are identical, except for an extra NP
constituent in structure (a), spanning the substring \textttt{`the cop
with the binoculars'}.  These two structures do more, however, than
simply hypothesize constituent boundaries -- they also hypothesize
dominance relationships between constituents.  It is possible to have
two distinct structural hypotheses with exactly the same hypothesized
constituents, if two of the constituents span exactly the same
substring.  This will happen with unary productions, or with empty
nodes.  Something more must be said, then, about these dominance
relationships.

A constituent $X$ {\it dominates\/} a constituent $Y$ if and only if
either (i) $X$ is the 
parent of $Y$; or (ii) $X$ is the parent of a constituent $Z$ that
dominates $Y$.  The parent relationship is also called {\it immediate
dominance\/}, since there is no intermediate node that is dominated by
the parent and dominates the child.  Each non-terminal node in the
tree dominates a string of zero or more terminal items, which is
called its {\it 
terminal yield\/}.  Each constituent can be characterized by its label
and its terminal yield, as was done when we introduced the notion of
constituency.  Each parse structure (or constituent structure) can be
characterized by its constituents and the dominance relationships between
constituents.  This characterization allows us to state the difference
between the two hypothesized constituent structures in figure
\ref{fig:parse1} in a way that is closer to our understanding of the
differences in interpretation: the parent of the PP is either (a) the
object NP or (b) the VP.

We have two hypothesized constituent structures in figure
\ref{fig:parse1} for the given string:  how were they found?  Well, in
this case, we relied on the author's knowledge of
lowest-common-denominator constituency and intuitions about English.  
In the absence of this homunculus, some search strategy must be
adopted to find the constituent structures from among the space of
possible structures, and for selecting them as ``good'' once they are
found.  First, some sort of grammar must be provided,
which specifies the sets of terminals and non-terminals (including the
root non-terminal), along with rules specifying how larger
constituents can be composed of smaller constituents. For example, an
NP constituent in English can consist of a DT followed by an NN, but
not by an NN  
followed by a DT.  By far the most common representation of such a
grammar is through context-free rules\footnote{It is worth pointing
out that there are grammars of constituent structure that do not use
phrase structure rules of this kind, but rather embed larger fragments
of syntactic structure in the lexicon.  For example, Tree Adjoining
Grammars \cite{Joshi75} and Categorial Grammars
\cite{Barhillel53,Lambek58,Steedman87} make use of lexical categories
that specify precisely what kinds of categories the word can compose
with.  Note,  however, that these approaches also require, in addition
to access of these syntactic ``part-of-speech'' tags, syntactic
operations to combine them.  Hence, although the ``division of labor''
if you will has been shifted so that lexical access provides more of
the syntactic structure, most of the points
that will be made about syntactic processing also hold for these
lexicalized formalisms.\label{foot:lex}} of the form
NP~$\rightarrow$~DT~NN, which can be interpreted as, ``an NP
constituent can immediately 
dominate a DT constituent, followed by an NN constituent and nothing
else.''

In addition to the grammar, we must also consider the direction of
hypothesis search.  Two extremes in the continuum of parsing
strategies are pure top-down and pure bottom-up. A pure top-down
parsing strategy starts with the root of the tree and builds the
structure down towards the terminals.  In this strategy, parent nodes
are hypothesized before any of their children. A pure bottom-up
parsing strategy starts with the terminals and builds structure up
towards the root.  In this strategy, parent nodes are hypothesized
after all of their children.  There are many potential hybrid
strategies, in which some of the structure is recognized top-down and
some bottom-up, or some compromise between the two.  One well-known
hybrid strategy that will come up several times in the course of this
thesis is called {\it left-corner\/} parsing \cite{Rosenkrantz70}, in
which (to over-simplify) parent nodes are hypothesized after their
leftmost child has been fully built, but before the rest of their
children are built.

\begin{figure}[t]
\begin{small}
\begin{picture}(382,144)(0,-144)
\put(42,-8){(a)}
\put(45,-30){S}
\drawline(48,-34)(20,-44)
\put(13,-52){NP}
\drawline(20,-56)(7,-66)
\put(0,-74){DT}
\drawline(7,-78)(7,-88)
\put(1,-96){the}
\drawline(20,-56)(32,-66)
\put(25,-74){NN}
\drawline(32,-78)(32,-88)
\put(25,-96){cop}
\drawline(48,-34)(76,-44)
\put(69,-52){VP}
\drawline(76,-56)(62,-66)
\put(50,-74){VBD}
\drawline(62,-78)(62,-88)
\put(54,-96){saw}
\drawline(76,-56)(90,-66)
\put(83,-74){NP}

\put(180,-8){(b)}
\put(156,-52){NP}
\drawline(163,-56)(150,-66)
\put(143,-74){DT}
\drawline(150,-78)(150,-88)
\put(144,-96){the}
\drawline(163,-56)(176,-66)
\put(168,-74){NN}
\drawline(176,-78)(176,-88)
\put(168,-96){cop}
\put(194,-74){VBD}
\drawline(205,-78)(205,-88)
\put(197,-96){saw}

\put(327,-8){(c)}
\put(330,-30){S}
\put(332,-42){\vdots}
\put(330,-52){S}
\drawline(333,-56)(305,-66)
\put(298,-74){NP}
\drawline(305,-78)(292,-88)
\put(285,-96){DT}
\drawline(292,-100)(292,-110)
\put(286,-118){the}
\drawline(305,-78)(317,-88)
\put(310,-96){NN}
\drawline(317,-100)(317,-110)
\put(310,-118){cop}
\drawline(333,-56)(361,-66)
\put(354,-74){VP}
\put(359,-86){\vdots}
\put(354,-96){VP}
\drawline(361,-100)(347,-110)
\put(335,-118){VBD}
\drawline(347,-122)(347,-132)
\put(338,-140){saw}
\drawline(361,-100)(375,-110)
\put(368,-118){NP}
\end{picture}
\end{small}
\caption{Partial structures built when the word \textttt{`saw'} is
incorporated into the analysis: (a) top-down; (b) bottom-up; (c)
left-corner (dotted lines indicate underspecified links)}\label{fig:parse2}
\end{figure}
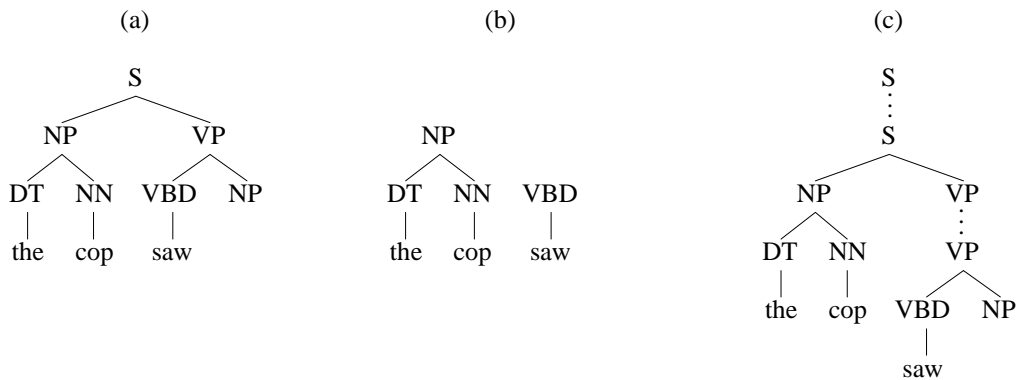

Let us illustrate how the hypothesis search would work with each of
these three strategies, by stepping through a parse of the first few
words of the sentence \textttt{`the cop saw the thief with the
binoculars'} incrementally, left-to-right.  For simplicity, suppose
that the root of the tree is always S, and that the only rules in the
grammar are those in the derivations in figure \ref{fig:parse2}.
These approaches are being presented informally, to facilitate
discussions of their differences.

The top-down parser begins with S, and expands it to NP followed by
VP.  This is a left-to-right parser, so it always works on the
leftmost unexpanded non-terminal, in this case NP.  The parser expands
the NP into DT followed by NN.  The DT expands to the terminal item
\textttt{`the'}, and the NN expands to the terminal item
\textttt{`cop'}.  At this point, the leftmost unexpanded non-terminal
is VP, which is expanded to VBD followed by NP.  The VBD expands to
the terminal item \textttt{`saw'}, at which point the leftmost
unexpanded non-terminal is NP.  And so on.

The bottom-up parser begins with the leftmost terminal item
\textttt{`the'}.  This can have the parent DT.  The parser then moves
to the next terminal item \textttt{`cop'}.  This can have the parent
NN.  A DT followed by an NN can have the parent NP.  The parser then moves
to the next terminal item \textttt{`saw'}.  This can have the parent
VBD.  And so on.  Eventually a sequence of non-terminals can combine
to form an S.

The left-corner parser begins with S and the leftmost terminal item
\textttt{`the'}.  The terminal item can have the parent DT, which is a
valid left-corner category for S.  After having found a DT, the parser
predicts the parent NP and the remaining child of the NP, NN.  The NN
expands to the next terminal item \textttt{`cop'}, completing the NP.
Having found an NP, the parser predicts the parent S and the remaining
child of the S, VP.  What is left underspecified is whether the
originally predicted S and the newly predicted S are the same
constituent.  The next terminal item is \textttt{`saw'}, which can
have the POS label VBD.  Having found a VBD, the parser predicts its
parent VP and the remaining child of the VP, NP.  Again, whether the
newly predicted VP and the one predicted by virtue of expanding the S
are the same is left underspecified.  These decisions are made later
in the string.

To understand the relevance of the parsing strategies to
incremental interpretation, let us examine the structure built by
different hypothesis search strategies when applied incrementally to a
string.  Figure \ref{fig:parse2} shows the amount of structure that
will be built at the point when the word \textttt{`saw'} is integrated
into the analysis, for a pure top-down, pure bottom-up, and standard
left-corner parser, each moving incrementally from left-to-right, as
in the informal examples given.  The
top-down parser builds a fully connected tree to the left of the word;
the bottom-up parser, because it does not build a non-terminal until
all of its children have been built, has no structure built above the
subject NP and the main verb.  The left corner parser will have
predicted an S constituent, after having built its left-corner
constituent (the subject NP) and a VP constituent, after having
built its left-corner constituent (the main verb).  A standard
left-corner algorithm will not, however, have attached the top-down
and bottom-up predictions at this point in the string; an ``eager
attachment'' version of the left-corner algorithm, in which this
attachment is made earlier, will be discussed
later in the thesis.  The basic point to make here is that, if
incremental interpretation involves hypothesizing that \textttt{`the
thief'} is an NP constituent (as opposed to \textttt{`the thief saw'},
which could be a nominal compound -- some kind of tool), and that it
is the subject of the main verb \textttt{`saw'}, then additional
links will need to be hypothesized beyond those of the bottom-up
parser.  In other words, some kind of link between the constituents --
which the top-down partial parse already provides -- is needed for
incremental interpretation.

Keep in mind that the partial parses in figure \ref{fig:parse2}
represent one potential constituent structure, yet there might
(and most often will) be many such hypotheses for a given string at
any given point.  One can either maintain all such alternatives, or have
some metric for deciding that certain alternatives are better than
others, and so discard the bad ones.  Such metrics can come in many
forms, from structural biases to memory constraints to probabilities.
Of course, there is a danger to throwing away hypotheses in an
incremental parser: garden pathing.  It may very well be the case that
a dispreferred analysis at an intermediate point turns out to be the
correct one.  If all viable analyses have been discarded, then either
the parser fails or some sort of repair strategy must be pursued.  

\begin{figure}[t]
\begin{picture}(375,144)(0,-144)
\put(21,-8){(a)}
\put(24,-30){\small VP}
\drawline(31,-34)(12,-44)
\put(-0,-52){\small VBD}
\drawline(12,-56)(12,-66)
\put(3,-74){saw}
\drawline(31,-34)(50,-44)
\put(44,-52){\small NP}
\drawline(50,-56)(37,-66)
\put(30,-74){\small DT}
\drawline(37,-78)(37,-88)
\put(30,-96){the}
\drawline(50,-56)(64,-66)
\put(56,-74){\small NN}
\drawline(64,-78)(64,-88)
\put(54,-96){thief}
\put(123,-8){(b)}
\put(126,-30){\small VP}
\drawline(133,-34)(112,-44)
\put(101,-52){\small VBD}
\drawline(112,-56)(112,-66)
\put(104,-74){saw}
\drawline(133,-34)(154,-44)
\put(148,-52){\small NP}
\drawline(154,-56)(137,-66)
\put(131,-74){\small NP}
\drawline(137,-78)(124,-88)
\put(117,-96){\small DT}
\drawline(124,-100)(124,-110)
\put(117,-118){the}
\drawline(137,-78)(151,-88)
\put(143,-96){\small NN}
\drawline(151,-100)(151,-110)
\put(141,-118){thief with $\ldots$}
\drawline(154,-56)(171,-66)
\put(165,-74){\small PP}
\put(243,-8){(c)}
\put(246,-30){\small VP}
\drawline(253,-34)(231,-44)
\put(220,-52){\small VBD}
\drawline(231,-56)(231,-66)
\put(223,-74){saw}
\drawline(253,-34)(274,-44)
\put(267,-52){\small NP}
\drawline(274,-56)(256,-66)
\put(249,-74){\small NP}
\drawline(256,-78)(239,-88)
\put(232,-96){\small NP}
\drawline(239,-100)(226,-110)
\put(219,-118){\small DT}
\drawline(226,-122)(226,-132)
\put(219,-140){the}
\drawline(239,-100)(253,-110)
\put(245,-118){\small NN}
\drawline(253,-122)(253,-132)
\put(242,-140){thief of Baghdad with \ldots}
\drawline(256,-78)(273,-88)
\put(267,-96){\small PP}
\drawline(274,-56)(292,-66)
\put(286,-74){\small PP}
\put(347,-8){(d)}
\put(350,-30){\small VP}
\drawline(357,-34)(343,-44)
\put(331,-52){\small VBD}
\drawline(343,-56)(343,-66)
\put(335,-74){saw}
\drawline(357,-34)(371,-44)
\put(364,-52){\small NP}
\put(369,-64){\vdots}
\put(364,-74){\small NP}
\drawline(371,-78)(358,-88)
\put(351,-96){\small DT}
\drawline(358,-100)(358,-110)
\put(351,-118){the}
\drawline(371,-78)(385,-88)
\put(377,-96){\small NN}
\drawline(385,-100)(385,-110)
\put(375,-118){thief \ldots}
\end{picture}
\caption{Incrementally building left-recursive structures: (a)-(c)
distinct top-down parses; (d) underspecified left-corner parse (dotted
lines indicate underspecified links)}\label{fig:tdvslc} 
\end{figure}
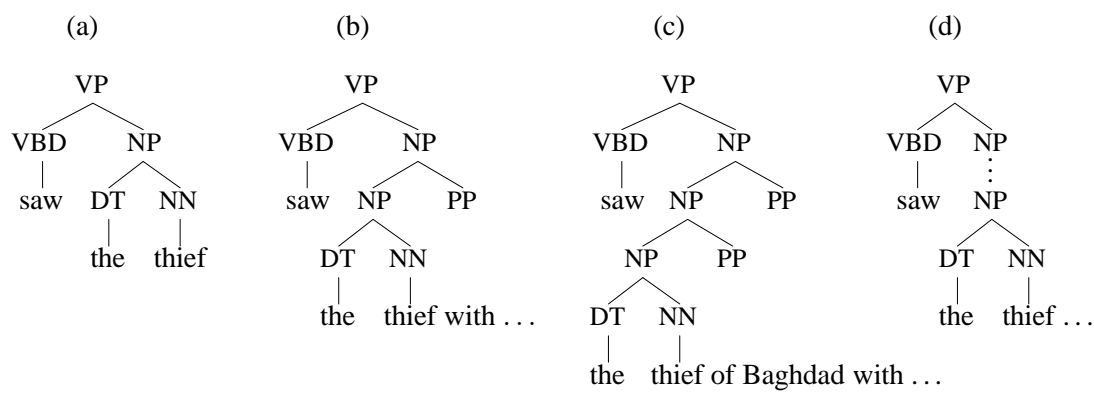

In some cases, it is impossible to enumerate all of the possible
partial parses, in particular if they are infinite.  If we follow a
top-down parsing strategy with a grammar that contains left-recursion,
i.e. where there are rules in the grammar of the form
A~$\rightarrow$~A~$\alpha$, in which the first child on the right-hand
side of the 
rule is of the same category of the parent, there are an infinite
number of possible partial parses.  Consider figure \ref{fig:tdvslc}a-c,
where we have shown three possible top-down partial parses of the verb
phrase \textttt{`saw the thief'}, each providing for a different number
of PP attachments.  In order to allow for an arbitrary number of PP
attachments, we would have to maintain an arbitrary number of distinct
parses.  A left-corner derivation, in contrast, by underspecifying the
immediate dominance links, can handle this by delaying the point at
which a decision must be made about the relationship between the
predicted NP and the found NP.  This difficulty with left-recursion is
one of the major criticisms of top-down parsing, and one of the large
benefits of left-corner and bottom-up parsing over top-down.  The
issue of left recursion and top-down parsing will be examined in
detail in Chapter 4.

Even if some analyses are effectively pruned from the search space or
choice points delayed, the number of
distinct partial parses that must be retained may be large enough to
benefit from dynamic programming.  There is some flexibility with
respect to the specific parsing strategy: standard chart parsing uses
dynamic programming in a strictly bottom-up fashion; an Earley parser
\cite{Earley70} incrementally uses dynamic programming with top-down
filtering.  A top-down parser cannot make use of
dynamic programming techniques because of the exponential complexity
that results from moving through the chart in that particular order.
The next topic that we will cover in this introductory section on
parsing is dynamic programming.

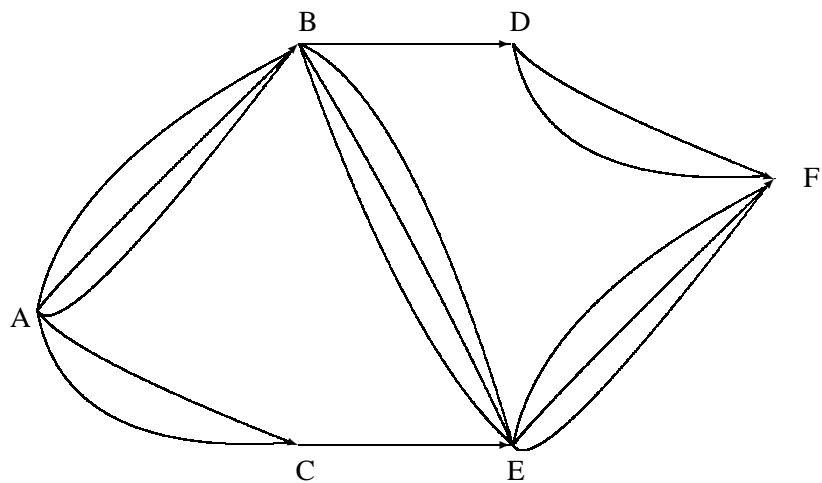
\begin{figure}[t]
\begin{center}
\begin{picture}(382,174)(0,-104)
\put(10,-50){A}
\put(119,62){B}
\put(119,57){\vector(1,1){0}}
\qbezier(20,-44)(30,10)(116,54)
\qbezier(20,-44)(30,-30)(116,54)
\qbezier(20,-44)(30,-60)(116,54)

\put(118,-108){C}
\put(119,-95){\vector(3,-1){0}}
\qbezier(20,-44)(30,-100)(116,-94)
\qbezier(20,-44)(30,-60)(116,-94)

\put(198,-108){E}
\put(119,-95){\vector(1,0){80}}
\put(119,57){\vector(1,0){80}}
\put(199,62){D}

\put(200,-95){\vector(1,-2){0}}
\qbezier(119,57)(160,-10)(200,-95)
\qbezier(119,57)(160,-60)(200,-95)
\qbezier(119,57)(160,40)(200,-95)

\put(299,6){\vector(1,1){0}}
\qbezier(200,-95)(210,-41)(296,3)
\qbezier(200,-95)(210,-81)(296,3)
\qbezier(200,-95)(210,-111)(296,3)

\put(310,3){F}
\put(299,6){\vector(3,-1){0}}
\qbezier(200,57)(210,1)(296,7)
\qbezier(200,57)(210,41)(296,7)

\end{picture}
\end{center}
\caption{Dynamic programming example: finding the shortest path from A
to F}\label{fig:dp}
\end{figure}

The basic idea behind dynamic programming is to use the structure of a
problem to collapse partial solutions to the problem.  For example,
consider the graph in figure \ref{fig:dp}.  Each arc is intended to
represent a directional path from one point to another.  Suppose that
we want to evaluate all paths from point A to point F, to find the
shortest.  This could be done by explicitly evaluating all of the
distinct paths, which in this case number 39.  Or this could be done
by simply 
evaluating each of the smaller intermediate links en-route from A to
F.  For example, to get from A to F through B, we must first go from A
to B, then from B to F.  The path choice from B to F is independent of
which path was taken from A to B.  The shortest path from A to F
through B is the combination of the shortest path from A to B and the
shortest path from B to F.   Hence, this problem can be represented and
processed efficiently by considering the 15 sub-paths, rather than all
39 complete paths.  Thus there is a benefit even in this very sparsely
connected case.

The same idea holds in parsing, insofar as every possible tree traces
a path between the root node and the leaves.  A chart can be
constructed, every cell of which represents a possible constituent
location.  A chart parser will populate cells in the chart with {\it
edges\/}, which are entries that indicate a constituent label and
location.  Under the context-free assumption, the path from the leaves
to that constituent is independent 
of the path from that constituent to the root, in precisely the same
way as in the example problem above.  Hence, through dynamic
programming over a chart, we can efficiently enumerate and evaluate
all possible parses for a string given a context-free grammar.

\begin{figure}[t]
\begin{center}
\begin{picture}(321,122)(0,-122)
\put(15,-8){(a)}
\put(17,-30){O}
\drawline(21,-34)(4,-44)
\put(-0,-52){O}
\drawline(21,-34)(38,-44)
\put(35,-52){O}
\drawline(38,-56)(21,-66)
\put(17,-74){O}
\drawline(38,-56)(56,-66)
\put(52,-74){O}
\put(88,-52){$\Rightarrow$}
\put(145,-8){(b)}
\put(147,-30){O$^1$}
\drawline(151,-34)(134,-44)
\put(127,-52){$^2$O}
\drawline(140,-48)(163,-48)
\drawline(151,-34)(169,-44)
\put(165,-52){O$^3$}
\drawline(169,-56)(151,-66)
\put(144,-74){$^4$O}
\drawline(157,-70)(181,-70)
\drawline(169,-56)(187,-66)
\put(183,-74){O$^5$}
\put(218,-52){$\Rightarrow$}
\put(276,-8){(c)}
\put(278,-30){O}
\drawline(282,-31)(264,-44)
\put(260,-52){O}
\drawline(282,-31)(282,-66)
\drawline(262,-53)(281,-66)
\drawline(282,-31)(317,-66)
%\put(295,-52){O}
%\drawline(299,-56)(281,-66)
\put(277,-74){O}
\drawline(270,-49)(313,-67)
\drawline(287,-70)(311,-70)
%\drawline(299,-56)(317,-66)
\put(313,-74){O}
\end{picture}
\end{center}
\caption{Transforming the tree structure in (a) to the undirected,
triangulated graph in (b), and the structure of the dependencies (c) after
performing the dynamic programming calculation on node 3}\label{fig:dp2}
\end{figure}
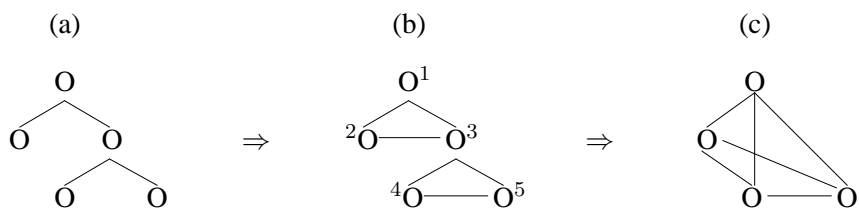

To see why dynamic programming does not work for top-down parsing, one
must understand what determines the complexity of the computation.
Very briefly, to perform dynamic programming over a directed acyclic
graph (DAG), one first removes the direction on the arcs, then creates
new arcs between nodes with the same parent (triangularization).  This
is represented in figure \ref{fig:dp2}, in the transition from the
tree structure in (a) to the undirected triangulated graph in (b).
The graph in 
(b) consists of two cliques of size three.  A clique is a subset of
the nodes in the graph that are fully connected: nodes 1, 2, and 3
form a clique, as do 3, 4, and 5.  If the maximum clique size in a
dynamic programming problem is $n$, the complexity of the computation
is on the order of $|v|^{n-1}$, where $|v|$ is the space of possible
values at each node (for a full presentation of dynamic programming
see, e.g., \egcite{Corman90}.  After performing the dynamic programming
calculation on any node, all remaining nodes that have an arc to that
node are now linked.  Hence, if we choose node 3 in the graph in
(b), the resulting graph structure after performing dynamic
programming on that node is shown in (c).  This is now a fully
connected graph, i.e. there is now a clique of size 4, so the
complexity of computation is $|v|^3$.  If we had chosen to visit nodes
in the graph in the order 2, 4, 5, 3, 1, for example, then the maximum
clique size at any stage would be 3, and hence the complexity
$|v|^2$.  One can see from this that it is a good idea to begin with
the leaves of the tree and move up, to minimize the maximum clique
size.  Dynamic programming with a top-down node visitation schedule
results in a very large maximum clique size.  It is worthwhile noting
that there are dynamic programming versions of other predictive
parsing strategies, such as left-corner \cite{Sikkel97}, which do not
suffer this problem so dramatically.

As Jurafsky points out, the internal structure of a constituent can
make a difference to interpretation.  In other words, as has been
pointed out repeatedly, the context-free assumption is quite false for
natural language.  Consider, for example, a noun phrase such as
\textttt{`the House Ways and Means Committee'} as opposed to
\textttt{`the Oakland A's and Detroit Tigers'}.  There are several
ways to carve up each of these NP constituents, which change the way
they would be interpreted.  By representing such a constituent as an
NP, with an unspecified internal structure, these distinctions in
interpretation would be lost.  As pointed out in \namecite{Martin81},
the number of possible analyses for constructions such as PP
attachment inside of noun phrases (e.g. \textttt{`the boy with the dog
with the leash'}) grows exponentially according to the catalan numbers.
The chart, however, with its context-free independence assumptions,
contains on the order of $n^2$ cells, where $n$ is the length of the
string.  Hence, it packs this exponential number of analyses into a
polynomial space.  Differences in interpretation can hinge upon
distinctions lost in the chart.

In sum, the critical requirement for incremental interpretation is
that enough of the structure must be specified to allow for alternate
interpretations to be 
distinguished. Insofar as one interpretation is ``preferred'' over
another to the extent that the expectation for constituents that are
consistent with that interpretation is increased at the expense of
other constituents, such a processing mechanism can be called
``predictive''.  We will argue that top-down and left-corner parsing,
both of which can be implemented in a top-down architecture, fit the
requirements for such a processing mechanism.

\subsection{Statistical parsing}\label{subsec:spar}
In this thesis, we will present a novel statistical parser that
performs comparably with state-of-the-art statistical parsers, while
following an incremental, top-down parsing strategy.  It is thus
important, in this introductory section, to explain what is
state-of-the-art in statistical parsing, and how our parser differs
from this.  

The past five years have seen enormous improvements in
broad-coverage parsing accuracy, through the use of statistical
techniques.  The parsers that perform at the highest level of
accuracy (Charniak \shortcite{Charniak97,Charniak00}; Collins
\shortcite{Collins97,Collins00}; \egcite{Ratna97} use
probabilistic models with a very large number of parameters,
including, critically, lexical head-to-head dependencies.  Each of
these parsers proceed in multiple stages:  the Charniak and Collins
parsers both prune the chart of edges that fall below some threshold
score, before using their full models on the trees that remain packed
in the chart;  the Ratnaparkhi parser first runs a part-of-speech
tagger, followed by a shallow ``chunker'', which finds flat
constituents given the input and part-of-speech tags, and finally
a procedure which builds and evaluates fully connected structures.

The specific probabilistic models differ from approach to approach,
with different parameters and different ways of mixing and smoothing
the probabilistic evidence.  At a very general level, however, these
approaches share some key characteristics.  In each of these
approaches, scores or weights are calculated for events, e.g. edges or
other structures, or perhaps constituent/head or even head/head
relations. The scores for these events are compared and 
``bad'' events, i.e. events with relatively low scores, are either 
discarded (as in beam search) or sink to the bottom of a heap
(as in best-first).  In fact, this general characterization is
basically what goes on at each of the stages in multi-stage parsers,
although the events that are being weighted, and the models by which
they are scored, may change.  In each parser's final stage, the parse
which emerges with the best score is returned for evaluation. 
These parsers get between 86 and 90 percent labeled precision and
recall on standard test sets.

This is a general characterization of the best statistical parsers in
the literature.  Let us focus upon one of these three approaches, and
give more details about how it works.  The other approaches differ in
details, but all of them involve pruning the search space and scoring
alternatives in some way.  We will give a brief outline of
the Charniak parser \cite{Charniak97,Charniak00}, which we will
refer to as the EC parser.  The EC parser first prunes the search
space by building a chart containing only the most likely edges.  Each
new edge is assigned a score, which is called its figure-of-merit
({\small FOM}), and pushed onto a priority queue\footnote{See,
e.g., \namecite{Corman90} for an introduction to data structures such
as priority queues and stacks}.  The {\small 
FOM} is the product of the probability of the constituent given the
simple probabilistic context-free grammar ({\small PCFG}) and certain 
boundary statistics, which are scores measuring the likelihood of
the constituent integrating with its surrounding context.  Edges that
are taken from the priority queue (highest score first) are put into
the chart, and standard chart building occurs, with new edges being
pushed onto the heap.  This 
process continues until a complete parse is found; hence this is a 
best-first approach.  Of course, the chart building does not
necessarily need to stop when the first parse is found;  it can
continue until some stopping criterion is met.  The
criterion that was used by Charniak is
a multiple of the number of edges that were present in the chart when
the first parse was found.  Thus, if the parameter is 1, the parser
stops when the first parse is found; if the parameter is 10, the
parser stops when the number of edges in the chart is ten times the
number that were in the chart when the first parse was found.  

This is the first stage of the parser.  The second stage takes all of
the parses packed in the chart that are above a certain probability
threshold given the {\small PCFG}, and assigns a score using the full
probability model.  To evaluate the probability of each parse, the
evaluation proceeds from the top down.  Given a particular constituent, it
first evaluates the probability of the part-of-speech of the head of
that constituent, conditioned on a variety of information
from the context.  Next, it evaluates the probability of the head
itself, given the part-of-speech that was just predicted (plus other
information).  Finally, it evaluates the probability of the rule
expansion, conditioned on, among other things, the {\small POS} of the 
head and the head.  It then moves down the tree to evaluate the newly
predicted constituents.  See \namecite{Charniak00} for more details on
the specifics of the parser.

The parser that we will present in later chapters shares some
characteristics with these parsers, but differs in certain fundamental
ways. Our parser will also condition the probabilities of events on a
large number of contextual parameters in more-or-less the way Charniak
does.  It also will use boundary statistics to assign partial
structures a figure-of-merit, which is the product of the probability
of the structure in its own right and a score for its likelihood of
integrating with its surrounding context.  The parser will differ,
however, in that it will parse incrementally in a single pass, from
left to right.  This will mean, among other things, that the lexical
head of a constituent may not be available at a particular point in
the analysis, to condition the probabilities of other subordinate
heads.  Hence, while our conditional probability model will share many
parameters with, say, the Charniak model, a good number of important
features will not be available, given our incremental parsing
strategy.

Another statistical parser that should be mentioned is the
probabilistic shift-reduce parser of \namecite{Chelba98a}.  This
parser differs from the parsers mentioned above as ours does: by
following an incremental, single pass parsing strategy.  It differs
from our model by following a bottom-up parsing strategy.  Accuracy
results have not been made available for this parser; it has been
evaluated as a 
language model for speech recognition.  Our top-down parsing strategy
will make it easy for us to capture certain top-down conditioning
parameters that are used in the parsers mentioned earlier, but which
are harder for the bottom-up Chelba and Jelinek parser to make use of.
We will go into more detail about the specifics of their parser in
chapter five, and compare our parser with theirs as a language model
for statistical speech recognition.

Now we will discuss certain central ideas in the human sentence
processing literature, in an attempt to (i) establish a common
vocabulary, and (ii) focus the issues upon those which this thesis can
potentially address. 

\subsection{Human sentence processing}\label{subsec:isen}
Ever since the publication of \namecite{Bever70}, the study of human
sentence processing has largely focused on situations where the
process is burdened or fails altogether.  His famous example 
\begin{examples}
\item The horse raced past the barn fell\label{ex:0}
\end{examples}
illustrates the difficulty that people can have interpreting
grammatical sentences.  Sentences such as \ref{ex:0} have been called
{\it garden path\/} sentences, because people appear to commit to the
wrong interpretation at early points of local ambiguity, and cannot
recover.  People have trouble with many kinds of grammatical
constructions, although this is usually exhibited by an increase in
processing time, rather than by a complete failure to interpret.  A
common experimental paradigm is to present subjects with sentences
that contain local ambiguities, and measure their performance at or
near the point of disambiguation.  If a minimal change in the stimulus
items can produce a measurable, significant difference in the
comprehension process, then this can be taken as evidence of the
nature of the comprehension mechanism.

There have been a number of centrally important issues in the sentence
processing literature over the past three decades, and we will
identify a number of them in this section, in an attempt to narrow the
discussion to those issues upon which we feel the present thesis
bears.  

First on this list is {\it modularity\/}, or the degree to which
different ``levels'' of linguistic processing and general cognition
are mutually inaccessible.  Early models of human sentence processing
\cite{Frazier78,Frazier79,Forster79} made very explicit predictions based on
highly modular architectures, in which, for example, the only
information that could pass between the syntactic processor and
the semantic processor were syntactic hypotheses in one direction and 
accept/reject in the other direction.  This is a finer-grained
modularity than that called for in \namecite{Fodor83}.  His
distinction was between automatic processes, such as language
processing, and general cognition.  Insofar as a ``level'' of
linguistic processing is automatic, requiring no general inference,
then there is nothing in Fodor's conception of modularity that would
preclude that level interacting with other levels of automatic
language processing.  It may be convenient because of traditional
divisions in linguistic theory to conceive of language
processing as involving, sequentially: phonetic, phonological,
morphological, syntactic, and semantic processing.  This convenience,
however, is a perhaps less-than-compelling reason for hypothesizing a
modular division of labor.  One might very well hold a modular view of
the language faculty without further decomposing this module into the
above-mentioned sub-modules.

Recent evidence of the rapid use of the discourse and visual context
in disambiguation -- e.g. \namecite{Tanen95} -- may lead some to
question modularity in language processing, even in the very general
sense of Fodor.  Whether or not inferential processes are involved in
the immediate disambiguation processes, or some kind of dumb proxy (as
Fodor himself has suggested), such as contextually sensitive
probabilities, might be an empirical question, although it remains to
be seen if they really are empirically differentiable. 
In the models considered in this thesis, the kind of information that is
guiding syntactic processing in our model is of the dumb sort, yet
it is quite consistent with a richer inferential model.  Whether 
the preferences are probabilities estimated from surface co-occurrence
of lexical items and constituents, or some kind of
inferentially-driven weighted biases, a processing mechanism of the
sort we advocate would apply.  In other words, we will not be making
any claims regarding the modularity of processing.

A second issue that has claimed many pages in the literature is serial
versus parallel processing.  In the sentence processing domain, the
question is: are multiple hypotheses about the syntactic structure
pursued simultaneously (in parallel), or is one ``preferred''
hypothesis pursued until it fails, prompting a reanalysis of the
sentence?  In the former position, longer reading times in the face of 
ambiguity are typically said to arise as the result of some kind of
weighting process -- often involving competition among analyses.  In
such a model, some fixed amount of resources (e.g. activation in a
neural network) must be spread among analyses, which compete via
mutual inhibition for a share of the resources; this competition is
what accounts for the longer reading times.  In the serial position,
these longer reading times reflect a reanalysis 
of the structure of the sentence.  In fact, there is nothing
inconsistent with reanalysis in a parallel model: one can imagine a
sentence processing mechanism that keeps some number of analyses in
parallel, but which follows some kind of reanalysis procedure under
certain circumstances.  For example, suppose that there is a fixed
limit on the number of analyses that can be kept active, say two.
This would involve pursuing two ``preferred'' hypotheses instead of
just one.  If both of them end up failing, a reanalysis procedure
could be followed, in just the same way that it would be followed in
the face of the failure of a single preferred analysis.  Hence the
debate between serial and parallel is not a debate between competition
and reanalysis.  Nevertheless, the debate is most often framed
in terms of the presence or absence of reanalysis.

These serial and parallel positions are potentially empirically
differentiable, as was shown in \namecite{Mendelsohn99}, which
reported a reading time increase when a dis-preferred analysis became
implausible.  In a serial model, this secondary analysis would never
have been constructed, so it cannot account for such an effect.  Given 
that the serial/parallel debate most often takes the form of
reanalysis versus re-ranking, however, even if such a result is
replicated and verified, it is unlikely to resolve lingering questions
about how to account for increases in reading times.  As is pointed
out in a discussion of these issues in \namecite{Gibson00}, a limited
parallel parser could involve reanalysis as well.  A limited
parallel parser that can consider, say, up to two analyses in parallel
(but no more) with reanalysis is consistent with these results, and might
account for many sentence processing effects in just the way a serial
parser would.  Furthermore, reanalysis can mean different things to
different people, and until there is a general theory of syntactic
reanalysis, it will be difficult to generate truly falsifiable
hypotheses.

While the parser that we will implement is, in these terms, a
parallel, non-backtracking model, it could be extended to include some 
sort of reanalysis.  One possible way that this could be
straightforwardly done is to narrow the number of analyses that can
be simultaneously considered, and when some triggering event occurs --
either a complete failure to extend active analyses, or the active
analyses become very unlikely given the probabilistic model --
reanalysis and repair 
strategies can be pursued.  Hence, we will have little to say about this
ongoing controversy.  Difficult questions that would face such an
extension -- such as when to reanalyze and how to go about it -- are
the same as would be faced by any large scale reanalysis
implementation.  As far as we know, there are no broad-coverage
parsers that follow such a strategy.

Two other central issues at play in the sentence processing
literature are frequently confounded: lexically-driven models with
interacting constraints, and connectionist models.  The reason that
they are confounded is because many of the lexically-driven models
were simulated with artificial neural networks; because
lexically-driven models shift the division of labor in parsing 
to include a larger role for the lexicon (see footnote \ref{foot:lex}
above), local non-hierarchical processing mechanisms, such as neural
nets, are able to simulate certain syntactic disambiguation
processes.  Yet as \namecite{Steedman99} points out, the syntactic
disambiguation carried out in neural networks is akin to
part-of-speech tagging, ignoring as it does the hierarchical structure 
required for the kind of compositionality that goes on in syntactic
and semantic processing.  While some of the work on enabling neural
networks to handle hierarchical structure is quite interesting
for the novel perspectives on the relationship between syntactic
structure and structures that a neural network can handle, such as
fractals \cite{Tabor01}, structured stochastic models of the sort
investigated in this thesis offer many of the benefits of neural networks
(e.g. graceful degradation and probabilistic weights) without limits
on the kinds of structures that can be processed.  Hence we will
discuss lexically-driven models with interacting constraints
independently from a connectionist framework. 

Perhaps the best way to see how syntactic ambiguity resolution can be
recast as lexical ambiguity is through the main verb / reduced
relative ambiguity, illustrated in example \ref{ex:0} above.  This can 
be thought of as an ambiguity between two possible syntactic
structures -- the verb \textttt{`raced'} is either analyzed as the main
verb of the clause or as a past participle (as in \textttt{`was
raced'}), attaching via a relative clause to the subject NP.
Alternatively, this can be thought of as a lexical ambiguity, between
a main verb and a past participle.  Each of these lexical items is
consistent with only one of the two syntactic structures, hence this
strictly lexical ambiguity can account for the syntactic ambiguity.
Such a model stands in contrast to one that stipulates a syntactic
module that operates on strictly structural principles.  As stated in
\namecite{MacDonald94} 
\begin{quote}
\setlength{\baselineskip}{.7\baselineskip}
Reinterpreting syntactic ambiguity resolution as a form of lexical
ambiguity resolution obviates the need for special parsing principles
to account for syntactic interpretation preferences \ldots
\cite[p. 676]{MacDonald94} 
\end{quote}

The ``special parsing principles'' referenced in this quote are those
used in so-called garden pathing models, such as {\it minimal
attachment\/} and {\it late closure\/} \cite{Frazier79}, whereby 
structural principles guide the syntactic processor's initial choice
of analysis.  For example, the principle of minimal attachment
dictates that attachments are preferred that introduce the fewest
nodes into the tree.  This leads to, among other things, a preference
for VP attachment of prepositional phrases in standard PP attachment
ambiguities.  A lexical model of the sort advocated by
\namecite{MacDonald94} would have different preferences depending on
specific lexical items in the string, as well as other things.

Our perspective on the shift from the predominant garden pathing
models of the late seventies and eighties to the probabilistic
constraint models of the last decade will be outlined in the next
chapter.  At this point we will just say that this is one central
issue in the human sentence processing literature where our work does
have something to say.  While the garden pathing models may be
unmotivated, this does not mean that syntactic processing, distinct
from lexical processing, is unmotivated.  On the contrary,
hypothesizing links between constituents is a prerequisite
to incremental sentence processing models, and lexical disambiguation
cannot do all of the work.  One simple example of this is multiple PP
attachments, e.g.
\begin{examples}
\item the thief from the city with the narrow streets
\item the thief from the city with the narrow waist
\end{examples}
The ambiguity when the second preposition is encountered is not a
lexical ambiguity, but rather is a pure attachment ambiguity -- which
NP is being modified?  Disambiguating material may be encountered
downstream, but if incremental interpretation is taking place, 
certain syntactic, non-lexical decisions must be made.  The point is
simply that there is a role for syntactic processing, even in models
that rely heavily on the lexicon.

\subsection{Broad-coverage parsing and human sentence processing}
What can research about broad-coverage parsing possibly say about
human sentence processing?  Potentially many things, including making
predictions about what kinds of constructions are likely to be
difficult and what kind of information is likely to be useful to
people.  Also, as was mentioned above and will be discussed further
in the next chapter, explicit computational modeling can unearth
interesting confounds in experimental data, and suggest ways of
removing them.  One additional way that parsing research of the sort
that will be pursued in this thesis can be psycholinguistically
relevant is by underlining an aspect of language comprehension that is
simply ignored in most current computational models of human
sentence processing:  that it is robust and productive in the
extreme.  Computational psycholinguistics in the sentence processing
domain these days is typically focused on models that can be shown to
experience difficulty on examples that people have difficulty with,
taken from a small, hand-built set of sentences.  The fact that these
models fail to extend much beyond this small set, including to
sentences that are extremely easy for people to process, is largely
unremarked upon\footnote{An encouraging exception is in
\namecite{Brants00}, which demonstrates the ability of an incremental
probabilistic chart parser to find ``good'' parses, even with severe
memory limitations.  The fact that a chart parser is not sufficient
for incremental interpretation should be noted, but their comments
about computational models being focused on behavior when confronted
with toy input is in
accordance with ours.}.  This has resulted in computational
psycholinguistics being of interest in recent years almost exclusively
to psycholinguists, far less to computational linguists\footnote{An
exception to this is Supertagging -- see \namecite{Srini99} and
\namecite{Kim01} -- which has some currency in both communities,
although this is not, strictly speaking, a parsing approach.}.

This is in large part an issue of evaluation: what is the standard 
by which we decide whether the model has explanatory power?  Complete
explanatory adequacy is met when the model is restrictive enough to
explain why people fail or struggle to parse certain strings, yet
permissive enough to explain why they do not have difficulty with
other strings.  Given that no model even begins to approach complete
explanatory adequacy at this point, psycholinguists have largely
chosen to evaluate models on their restrictiveness; while
computational linguists are largely interested in adequate coverage,
i.e. they evaluate on a model's ability to analyze arbitrary,
naturally occurring strings in the language.  Hence, models such as
those advocated by \namecite{Stevenson93}, \namecite{Sturt96}, 
\namecite{Lewis98}, \namecite{Gibson98}, and \namecite{Vosse00} all
present evaluations of their models in terms of a small number of
simulations (either manual or computed) on exemplars of construction
types common in the experimental literature, which exhibit attested
patterns of performance.  In contrast, parsing models in the
computational literature -- too many to list here, but including all
of the parsers that will be mentioned in later chapters -- are
evaluated on what percentage of sentences in some test set are parsed,
and with what accuracy.

The problem is scalability.  As mentioned above, the experimental
paradigm in human sentence processing typically involves pathological
linguistic examples constructed expressly for the particular study by
the researchers.  If we think of a probability distribution over
strings in a language, these stimuli very often inhabit the tail of the
distribution, i.e. they are quite often rare constructions, such as
the reduced relative or center-embedding.  If the explanation of
parser behavior in these pathological circumstances relies in some
critical way on a specific parser architecture -- e.g. reanalysis
operations as a part of the parser as in \namecite{Sturt96} and
\namecite{Lewis98}; or competition processes as in
\namecite{Stevenson93} and \namecite{Vosse00} -- then the onus is upon
the researchers to show that these architectural specifics that
restrict processing appropriately in the handful of examples given do
not restrict processing inappropriately in naturally occurring
language.  Whether or not these models can scale up to handle freely
occurring language is an open question, and the burden of proof is on
their proponents.

While the \namecite{Gibson98} paper has a similar focus on
pathological constructions, its predictions are based on
well-formedness metrics that are to a certain extent independent of
the processing mechanism, insofar as a wide variety of mechanisms
could use the metric.  This is akin to probabilistic constraints,
which can be brought to bear on processing in any number of ways.  By
the same argument, the kinds of constraints on the model in
\namecite{Jurafsky96} are quite general, although he makes a
stronger assumption about the nature of the processing mechanism,
namely a parallel beam search, to account for garden path phenomenon.
In a sense, this thesis is a demonstration that a processing mechanism
of the sort proposed in \namecite{Jurafsky96} is permissive enough to
cover freely occurring natural language, while potentially being
restrictive enough to account for certain empirical facts.

We will hence do three things in this thesis that are relevant to
psycholinguistic models of human sentence processing.  In the next
chapter, we will examine some of the computational modeling and
experimental literature in a further attempt to justify our approach,
as well as expose 
possible confounds in some of the empirical results.  Our empirical
computational work will establish the applicability of predictive
models to the general problem of language comprehension.  Also, our
investigation into conditional probability models for parsing will
point out certain distributional regularities that might be exploited
by people for disambiguation.

\section{Chapter summary}
In this chapter, we have introduced the central problems to be dealt
with by this thesis, as well as fundamental notions in both parsing
and human sentence processing.  The main points made in the chapter
were: (i) links between constituents -- of the sort provided by top-down
parsers -- are required for incremental interpretation; (ii) lexical
approaches grammars, while accounting for some syntactic
disambiguation via local lexical disambiguation, do not obviate the
need for syntactic processing; and (iii) processing 
models that have been shown to be restrictive in ways that humans are,
must also be capable of robust, productive processing in ways that
humans are.  

The next chapter will explore in detail some of the changes in models
of human sentence processing over the past decade and a half, and the
role of parsing within these models.  We will show that some of the
modular architectures that were proposed complicated their models to
avoid committing to top-down parsing. 
We will also demonstrate that more than one model is consistent
with many recent empirical results -- one in which constituent
structure plays a role and one in which it does not.  A failure to be
explicit about the parsing mechanism underlying models of
disambiguation is to blame for this, and we will provide some
potential stimuli to resolve the empirical questions that arise from
this underspecification.  Finally, we will discuss specific ways in which
robust parsing of the sort we are investigating in this thesis can be
relevant to the study of human language processing.

\chapter{Psycholinguistic models of sentence processing}
A major turning point in the study of how humans process sentences
came with the study by \namecite{Crain85} and its follow up
\namecite{Altmann88}.  In these studies, it was demonstrated that the
referential context could immediately influence the preferred
interpretation of certain ambiguities that involved restrictive noun
phrase modification, evidenced by reading times in the region of
interest.  In the case that referential ambiguity exists
(the definite noun phrase describes more than one discourse entity)
then definite noun phrase modification is facilitated; otherwise VP
modification is preferred.  For example, consider the following
four sentences, taken from the two cited papers above.

\begin{examples}
\item The psychologist told the woman that he was having trouble with
{\it to visit him again}\label{ex:2}
\item The psychologist told the woman that he was having trouble with
{\it her husband}\label{ex:1}
\item The burglar blew open the safe {\it with the diamonds}\label{ex:3}
\item The burglar blew open the safe {\it with the dynamite}\label{ex:4}
\end{examples}

In sentences \ref{ex:2} and \ref{ex:1}, there is a local ambiguity at
the word {\it that\/} between NP modification (as in sentence
\ref{ex:2}), and a sentential argument to the verb (as in sentence
\ref{ex:1}).  The definite article of the object in each of the
sentences carries with it a presupposition of uniqueness.  In cases
where the context within which the sentence is presented contains more
than one entity consistent with the base noun phrase (e.g. two women),
then this presupposition is violated, unless some restriction resolves
the ambiguity.  \namecite{Crain85} showed a preference for the NP
modification reading in such a case.  Similarly, sentences \ref{ex:3}
and \ref{ex:4} have a local 
prepositional phrase attachment ambiguity between NP attachment (as in
sentence \ref{ex:3}) and VP attachment (as in sentence \ref{ex:4}).
\namecite{Altmann88} showed a facilitation of the NP attachment
reading relative to the VP attachment reading in cases where
referential ambiguity existed.   This effect has been replicated
repeatedly with a variety of constructions -- e.g. \namecite{Britt94}
-- and the preference shown for NP attachment in the case of
referential ambiguity is immediate \cite{Tanen95}.

This is strong evidence for incremental interpretation of sentences,
insofar as a bias such as those reported requires some kind of
mechanism by which hypothesized NP constituents are immediately
matched with existing discourse entities.  In the case that the
hypothesized NP is definite, yet could potentially match more than one
entity (e.g. \textttt{`the cop'} in a context within which more than
one cop has already been introduced), an expectation for NP
modification can be formed.  The facilitation reported in Tanenhaus et
al. above indicates that the attachment is immediate, i.e. the
constituent structure that is being hypothesized is a connected
structure; if this were not the case, then one would not see the
difference between conditions until later in the sentence.

Up to the point of these studies, the dominant perspective in the
sentence processing literature involved the so-called ``garden
pathing'' models mentioned in the previous chapter, in which a single
initial syntactic hypothesis is generated via purely structural
principles, such as minimal attachment, and then reanalyzed in the
case of a mis-parse.  In the face of the immediate influence of
discourse context on reading times, such a model must involve a rapid
(and very smart) reanalysis procedure, which takes essentially no
time (at least not measurable experimentally) in the case that there
is referential support for the as-yet-unbuilt dispreferred NP
attachment, but which garden paths in the absence of referential
support.  This would require the interpretation module to know the
presence of a syntactic alternative to the one that it has been given
by the syntactic module, without having seen it. The module would then
have to reject the VP attachment, despite that analysis being perfectly
acceptable to that point.  If the NP attachment that presumably
results from the 
reanalysis fails, a new reanalysis would be triggered -- would this
second reanalysis be able to find the previously discarded VP
analysis?  The complications that would be required to maintain the
garden pathing models led some
researchers to adopt alternative architectures: a parallel, weakly
interactive modular architecture \cite{Altmann88,Steedman89}, or 
non-modular architectures -- e.g. \namecite{MacDonald94}.  The rest of
this chapter will focus upon these two ways of dealing with immediate
contextual influence on parsing.

We will make two main arguments in the course of this chapter.  First,
that the brief literature on modular approaches to incremental
interpretation at the beginning of the 1990s tried to address the
parsing issue, but in the end bypassed the ``connected structure''
problem by assuming that a separate ``interpretation'' module would do
that work.  Our parser obviates the need for such complication, by
allowing connected syntactic structures to be built.
Second, that the literature on non-modular approaches to
incremental interpretation, by which we mean models making use of
interacting probabilistic constraints, have just plain ignored parsing
as a legitimate concern, choosing to remain ``agnostic'' about the extent
of structure building going on, focusing instead exclusively on
disambiguation, and hence failing to make relatively
important distinctions in their models.  In the end, regardless of the
processing model one adopts -- parallel, serial, with or without
reanalysis -- the basic requirement of incrementally built connected
structure remains.  That will be the topic of the subsequent
chapters of the thesis.

\section{Parsing for incremental interpretation}
One result of the evidence for incremental interpretation coming out
of the psycholinguistic literature was a small set of papers from
interested computational linguists, dealing explicitly with the
ramifications for parsing and grammars.  Two papers
\cite{Abney89,Steedman89} made different claims about the
ramifications of incremental interpretation; then \namecite{Stabler91}
attempted to show that neither of these claims was in fact true; and
finally \namecite{Shieber93} showed that the criticism of
\namecite{Abney89} in \namecite{Stabler91} was unmotivated, but that
\namecite{Abney89} was wrong anyhow.  Our position will be that, in
fact, \namecite{Abney89} was right, at least insofar as his claim was
that connected representations provide the sort of information
required for interpretation.  The specific arguments will be discussed
next.

Steedman's argument was an attempt to show that Combinatory Categorial
Grammar (CCG) is the appropriate form of the competence grammar, by
virtue of the 
fact that it allows for three desirable properties of sentence
processing to co-exist: (i) ``strong competence'', i.e. that the
competence grammar is what people use for syntactic processing, as
opposed to some approximation or simplification of the competence
grammar; (ii) incremental interpretation; and (iii) right-branching
constituent structures in languages like English.  His argument is
that, in a right-branching constituent structure, many constituents
will not be complete before the end of the string, hence either the
constituent structure is wrong, or the grammar is not used directly
for sentence processing, or the string is not incrementally
interpreted.  In CCG, however, a mechanism exists (function
composition) for composing what in standard grammars are incomplete
constituents into interpretable units, hence CCG can avoid the
``paradox'' above.

Abney's argument was that the parser must be top-down, because a
bottom-up parser does not have enough structure built to allow for
incremental interpretation.  This is basically the argument that we
have been advocating here, under the assumption that syntactic and
semantic processing are part of the same derivational process.
Although Abney makes
the point about top-down parsing in his paper, he does not take any
position about the feasibility of using such a parser in practice.

One other point about \namecite{Steedman89} before moving on to
discuss the other papers.  By using function composition to change
right-branching constituent structures into left-branching structures,
a CCG derivation of the sort Steedman is arguing for is a fully
connected structure of the sort that we advocate.  A left-branching
structure with complex CCG slash categories corresponds to a
fully-connected tree using standard categories.  He recognizes that
this brings along with it the processing difficulties of any such
parser.
\begin{quote}
\setlength{\baselineskip}{.7\baselineskip}
The proliferation of possible analyses that is induced by the
inclusion of function composition seems at first glance to have
disastrous implications for processing efficiency \ldots
\cite[p. 481]{Steedman89}
\end{quote}
This underlines a point that we have made and will make later in the
chapter, that lexicalized grammars can embed much syntactic
information in the lexicon, but the syntactic processing difficulties
involved in hypothesizing connected trees are just as profound as with
standard phrase-structure grammars.  This is a point that is well
understood by syntacticians and computational linguists, but
apparently not by some psycholinguists.

Stabler dismisses Steedman's argument for CCG by showing that one can
easily interpret partial syntactic structures.  In fact, one can see
this by simply noting that function composition works for CCG, so it
can certainly work for some interpretive mechanism, based on
incomplete constituents.  Function composition allows for a partial
evaluation, which is precisely what interpretation over incomplete
constituents would involve.
What Steedman proposed was not that, for example, the
right-branching constituent structure of English favored by linguists
is in fact left-branching, but simply that this constituent structure
could be derived in a left-branching way, through function
composition.  He was working in a framework that tightly couples
syntactic and semantic processing, but one can imagine a scenario in
which they are decoupled, and his point would still hold for
interpretation, even if the syntax were something other than CCG.

Stabler attempts to dismiss Abney's argument as well, but as
\namecite{Shieber93} subsequently pointed out, he does not succeed.
His point was that, with the appropriate formal apparatus, each syntactic
operation can correspond to a semantic operation, and to the extent
that we can uniquely identify a syntactic analysis by listing the
constituents in the order that they are identified (presuming we know
the search strategy), then we can also provide a semantic analysis by
listing the semantic operations in the appropriate order.  This notion
of incremental interpretation, however, is not the one that we are
working with, as mentioned in the introductory chapter, so let us be
very explicit about the differences.  Our next points are related to
those made in \namecite{Shieber93}. 

We will define an incremental parser as a parser that moves from left
to right across the string $w_{0}w_{1}\ldots w_{n}$, and which
completes all nodes that end at $w_k$ before $w_{k+1}$ is incorporated
into the derivation.  Under this definition, a
top-down, left-to-right parser is incremental, and so is a bottom-up,
left-to-right parser, such as a shift-reduce parser.  An incremental
semantic processor, insofar as semantic processing is a way of
combining constituents in some way, can also be incremental by
following either a bottom-up or top-down search path.  Incremental
interpretation, however, in the way that we are using it, is more than
simply moving from left-to-right over the string; it involves
identifying key dependencies between words and constituents in the
string immediately, often before the last word in the constituent is
reached.  For example, there are (at least) 
two potential interpretations of a string that begins \textttt{`the
horse raced'} -- the main verb interpretation and the reduced
relative interpretation.  Incremental interpretation in the way that
we are speaking of it here involves discriminating these two potential
interpretations, and perhaps even discarding one in favor of the
other.  In any case, it involves {\it explicitly\/} identifying the
relevant distinctions between these two interpretations, which has to
do with the syntactic and semantic dependency between the verb form
and the subject noun phrase.  This is what psycholinguists mean when
they say that people incrementally interpret sentences -- they
identify and then prefer or disprefer alternative interpretations.  

The point that Stabler made is simply that interpretation can be seen
as being fundamentally related to parsing; hence, insofar as a
bottom-up parser can be incremental, so can incremental
interpretation.  While we agree that there is a fundamental
similarity between the kinds of hypotheses that need to be made
syntactically and those that need to be made for interpretation, we
will argue in the opposite direction:  since incremental
interpretation involves hypothesizing relationships between
constituents, so should incremental parsing, in this other sense of
incremental.

\namecite{Shieber93} criticize Stabler's argument against
\namecite{Abney89} on more-or-less the grounds that we have put
forward here, but they provide another criticism against Abney's view.
In a nutshell, their argument is as follows:  to the 
extent that one stipulates a modular architecture for the human
sentence processing mechanism (an asynchronous processor, in their
parlance), the output of a bottom-up parser is an acceptable input to
an interpretation mechanism, even one that proceeds incrementally in
the way that we have defined it.  In other words, even though the
constituents constructed on the {\it stack\/} of a bottom-up parser
does not have sufficient information (as Stabler claimed) for
incremental interpretation: 
\begin{quote}
\setlength{\baselineskip}{.7\baselineskip}
There is also a finite amount of state information.  As it turns out,
the state of an LR parser finitely encodes a set of possible
left-contexts for the stack items.  This set of contexts has a regular
structure, and corresponding to that regular structure of syntactic
left contexts, there is a regular structure of functors over the
completed constituent meanings (from the stack).  Since these functors
are incrementally computable from the LR state, they are accessible by
the interpreter, and hence available for incremental
interpretation. \cite[p. 305]{Shieber93} 
\end{quote}
The parser, hence, underspecifies the left context, i.e. the stack
represents the set of all left contexts that are consistent with it.
The interpreter can then take the stack and derive the potential
semantic relationships from this.  They give an example, using
synchronous Tree-Adjoining Grammar, that demonstrates how, in certain
circumstances, connected syntactic structure is not required for
incremental interpretation.  

Note that the semantic derivation which feeds interpretation requires
a derivation strategy just as the parser does for the syntactic
derivation.  The semantic derivation, however, cannot be strictly
bottom up, since it requires the composition of incomplete
constituents.  This asynchronous model is similar to Jurafsky's in
this respect, in that the parser gets the efficiency gain of a
bottom-up parser, which allows for dynamic programming, while
interpretation, which operates simultaneously with parsing, must
follow a semantic derivation strategy that is not strictly bottom-up,
but which composes incomplete constituents.  Under a rule-to-rule
assumption, in which syntactic and semantic derivations make use of
corresponding rules, it is not clear what advantage such a model would
have for the interpretive process as a whole in following a bottom-up
search path in parsing; however, it would force a certain amount of
duplicate effort, to the extent that two separate mechanisms do what
amounts to the same structure building.

In the absence of this rule-to-rule assumption, one could have a big
efficiency win with an asynchronous model of this sort, if a large
number of syntactic alternatives correspond to a single semantic
alternative.  Their example is VP adjunction: suppose that VP 
modification occurred with left recursive VP rules of the form 
VP~$\rightarrow$~VP~AP.  When the main verb is encountered, a top-down
parser is faced with the problem mentioned in the previous chapter,
namely that there are an infinite number of parses corresponding to
the infinite number of possible adjunctions.  Regardless of the
number of these adjunctions, however, the semantic relationship between
the subject NP and the main verb will be the same, i.e. the number of
semantic analyses corresponding to this infinite number of syntactic
analyses is one.  Of course, the opposite is likely to be true as well
-- that single syntactic analyses correspond to a large number of
semantic analyses.  Consider noun compounds: the compounds
\textttt{`ostrich burger'} and \textttt{`kid burger'} appear 
syntactically identical, but (hopefully) semantically quite
different.  Here a single syntactic derivation would correspond to
multiple possible semantic derivations.

We will follow Steedman and Stabler (and many others), and make the
simplifying assumption of a rule-to-rule correspondence between
syntactic and semantic rules.  Under this assumption, if the semantic
derivation is required, for incremental interpretation, to compose
constituents, the syntactic derivation might as well compose the
constituents as well.  If we assume such a correspondence, there is no
overall gain in delaying the choice points in one derivation and not
the other, so the two derivations may as well move along in 
lockstep.  In other words, under such an assumption, the syntactic
derivation needs to specify enough of the structure for adequate
semantic processing to occur to allow incremental interpretation.

There are a number of factors that caused Shieber and Johnson to
go to the lengths they did in allowing for bottom-up parsing.  First,
and most obvious, is left-recursion, which is widely known
to be a major problem with the top-down approach.  We will discuss
this problem at length in the next chapter, in the context of our top-down
parser, and will further examine the issue in chapter 4.  We advocate
a probabilistic beam-search, in which low 
probability partial parses are discarded.  This sort of mechanism was
also advocated by \namecite{Jurafsky96} as a way to account for garden
pathing phenomenon in a parallel parser.  In such an approach, each
left-recursive rule has a probability associated with it, so that
every time it is used, the probability of the analysis drops.
Eventually, the probability will fall outside of the beam, and the
parser will stop building these categories.  The capacity of people to
interpret left-recursive structures is presumably not unbounded, so
this may or may not be an appropriate way to handle these
constructions.  One alternative to this that will be discussed in the next
chapters will be selective left-corner parsing, which can be
simulated in our parser, delaying choice points until later in
the string, and thus avoiding problems with left-recursion.  In
practice, we achieve good results with pure top-down parsing, in spite
of frequent left-recursion.

Shieber and Johnson also assume parsing to be a deterministic serial
process, so that all parsing choices 
involve a commitment to a particular structure.  Thus, to the extent
that a top-down parser builds structure, it rules out everything
inconsistent with that structure.  
\begin{quote}
\setlength{\baselineskip}{.7\baselineskip}
The choice to expand the NP top-down as a determiner (Det) followed by
a noun (N) -- as opposed to, say, an NP followed by a reduced relative
clause -- essentially commits the top-down parser to this reading, and
that choice occurs at the beginning of the constituent, i.e. the
beginning of the sentence.  In general, a top-down parser makes such
commitments too early. \cite[p. 301]{Shieber93} 
\end{quote}
Steedman, in contrast, posits a
parallel parser, which builds all of the alternatives, and sends them
off to be interpreted.  As discussed above, these two alternatives are
extremely difficult to distinguish empirically; basing an argument for
one model over another based on an assumption of serialism is
questionable.  

Another assumption is about the choice point associated with
top-down parsing: when a commitment to constituents is made within one
particular analysis.  We can first note that their particular example
in the quote above is not necessarily a good example of early
commitment.   A reduced relative clause, as a restrictive NP modifier,
is likely to fall under an $\bar{\mathrm N}$ constituent, so their
point for this particular example is based on an NP structure that
many linguists would argue against.  If the modification falls within
the $\bar{\mathrm N}$ constituent, even given their pure top-down
parsing assumptions, the choice point would not be at the beginning of
the sentence, but later, once the head noun is in the look-ahead.  Of
course, this is still very early in the sentence. 

\begin{figure}
\begin{center}
\begin{picture}(275,100)(0,-100)
\put(35,-8){(a)}
\put(37,-30){S}
\drawline(40,-34)(21,-44)
\put(14,-52){NP}
\drawline(21,-56)(7,-66)
\put(0,-74){DT}
\drawline(7,-78)(7,-88)
\put(1,-96){the}
\drawline(21,-56)(35,-66)
\put(25,-74){NNS}
\drawline(40,-34)(59,-44)
\put(52,-52){VP}

\put(120,-8){(b)}
\put(122,-30){S}
\drawline(125,-34)(112,-44)
\put(105,-52){NP}
\drawline(112,-56)(101,-66)
\put(94,-74){DT}
\drawline(101,-78)(101,-88)
\put(94,-96){the}
\drawline(112,-56)(122,-66)
\drawline(125,-34)(139,-44)

\put(236,-8){(c)}
\put(238,-30){S}
\drawline(241,-34)(212,-44)
\put(205,-52){NP}
\drawline(212,-56)(178,-66)
\put(171,-74){DT}
\drawline(178,-78)(178,-88)
\put(171,-96){the}
\drawline(212,-56)(211,-66)
\put(200,-74){NNS}
\drawline(211,-78)(211,-88)
\put(195,-96){flowers \hspace*{.6in}sent}
\drawline(212,-56)(245,-66)
\put(232,-74){STOP}
\drawline(241,-34)(271,-44)
\end{picture}
\end{center}
\caption{(a) Fully specified top-down prediction; (b) Underspecified
top-down prediction; (c) Delayed prediction of constituent boundary}\label{fig:tdpred}
\end{figure}
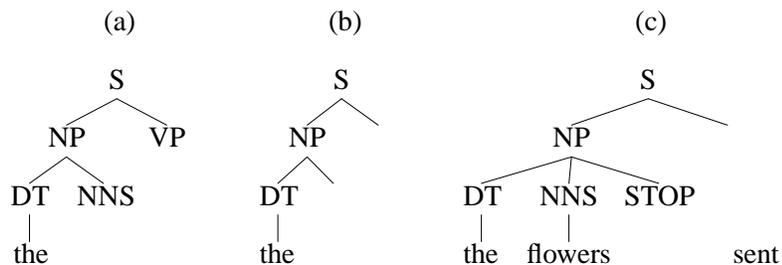

That aside, their point about when the commitment must take place in a
top-down parser is based on one potential top-down algorithm, which
has full specification and prediction to the extreme.  There has been
much work on variation of choice points in parsing strategies
\cite{Demers77,Nijholt80,Abney91}, with pure top-down and pure
bottom-up occupying the extremes of a continuum of possible
strategies. Choices about node labels and rules can be delayed to 
points where they can more reasonably be made, without going to the
extreme of a pure bottom-up strategy.   Figure
\ref{fig:tdpred}a shows the structure that a fully specified top-down
derivation will build to incorporate the first word of the string.
Figure \ref{fig:tdpred}b shows the same structure, but with the
predictions about categories that do not dominate the first word left
underspecified.  In fact, both the predicted categories and the number
of subsequent children of each predicted constituent can be left
underspecified.  Further, figure \ref{fig:tdpred}c shows the structure
that will be built when constituents are explicitly closed with an
additional empty final child.  A simple modification to top-down
parsing, which preserves constituent structures and maintains a
connected left-context, is to predict one child at a time in a
leftmost manner, and to explicitly predict the end of a constituent at
the point when it ends.  In such a way, at the beginning of the sentence
(\textttt{`the'}), an NP is predicted, consisting of a determiner (and
maybe more).  When the next word is encountered (\textttt{`flowers'}),
a noun is added to the NP constituent, but it is not necessarily
closed (there may be modification, or noun compounding).  When the
third word (\textttt{`sent'}) is seen, the NP can be explicitly closed,
and the word attached through another constituent.  This is a top-down
parsing algorithm, just one that uses a certain amount of
underspecification of that which is yet to be encountered to delay
disambiguation to the appropriate point.  We will achieve this effect
through grammar factorization.  Shieber and Johnson go much farther in
trying to delay the disambiguation points:
\begin{quote}
\setlength{\baselineskip}{.7\baselineskip}
To get incremental interpretation to work with disambiguation requires
postponement of choice points; a bottom-up parser is a natural way to
achieve this postponement. \cite[p. 301]{Shieber93} 
\end{quote}

While an extreme top-down algorithm may be prey to their criticisms,
very simple and natural modifications make a top-down parser perfectly
consistent with their architecture, and one that would provide connected
structures for interpretation.  A problem which remains is
left-recursion, but we shall see that, in a probabilistic setting,
this is not such a problem in practice.

While their point is valid, that connected syntactic structures are
not necessarily required for incremental interpretation, the question
remains as to whether the complications that must be introduced into
the model -- a modular architecture and additional semantic computation
over and above syntactic parsing -- is worth it.  It may be true that
connected structures can be bypassed, but such structures do make incremental
interpretation far simpler, and they allow for a cleaner link between
syntactic and semantic processing, without strong modular assumptions.

\section{Lexical and syntactic disambiguation}
A sentence processing modeling perspective that has become popular in
the wake of empirical evidence for rapid on-line integration and 
interpretation is that of interacting probabilistic constraints.  As
was mentioned in the last chapter, these models are very often
associated with neural network models, and tend to focus on local
lexical constraints as opposed to longer distance or non-lexical
dependencies.  In this section, we will take the position that the
extreme lexicalist position is untenable, and that experimental stimuli
that have been developed to be modeled in this way ignore crucial
distinctions that may change the kinds of predictions they 
make.  Further, there seems to be some confusion of terminology, so
that lexical influences on interpretation are taken to be `bottom-up',
when in fact the probabilistic effect can be quite well modeled in a
top-down fashion.  Let us first clear up this potential terminological
confusion, before turning to the issue of local vs. non-local
modeling.

The top-down parsing strategy that we will advocate involves a
look-ahead word (the next word in the string), and the parser is
guided from predicted categories down to the new lexical item.  This
sort of look-ahead is consistent with garden pathing phenomena, since
it is used only to help guide its own integration into the constituent
structure, not the integration of previous
words\footnote{\namecite{Weinberg87} criticized parsing models with
look-ahead for not being able to account for certain kinds of garden
pathing phenomena.  For example, if the parser can look to the word
{\it after\/} the verb in a reduced relative construction, it can in
certain cases disambiguate (e.g. \textttf{`the horse raced fell'}) yet
people still seem to garden path.  Our model is not prey to this
criticism, since we only use a look-ahead word to guide its own
integration, after the previous words have already been
incorporated.}. The goal is to generate a set of likely partial
constituent structures that incorporate the word.  One can think of it
as growing the structure down towards the word:  at each word, some
new extended piece of structure is added to existing structure,
incorporating the new word, and making new predictions about what is
likely to come. Given an existing fully connected partial analysis and
a new word, the search space for the ways of incorporating the new
word into the connected structure is the same whether one grows the
new structure down from the existing structure to the new word, or up
from the word to the existing structure.  Hence, to the extent that
psycholinguists speak of integrating words into the syntactic
structure, they are speaking (computationally) of a top-down process.

An incremental parser that is guided by the input is not
necessarily bottom-up.  To the extent that connected structures are
being incrementally built, in the manner we are advocating, then
the parser is not strictly bottom-up.  In the following quote,
however, the bottom-up/top-down distinction seems to be denoting more
of a purely-lexical/purely-syntactic distinction, rather than the
derivation strategies that we have been discussing:
\begin{quote}
\setlength{\baselineskip}{.7\baselineskip}
The results favor a constraint-based model of sentence processing in
which the bottom-up input computes, in parallel, the possible
syntactic alternatives at the point of ambiguity, and contextual
constraints provide immediate support for one or another of those
alternatives. \cite[p. 228-9]{Spivey95} 
\end{quote}
The phenomenon at issue in this paper is PP attachment ambiguities,
where the attachment preferences are observed at the preposition.
According to the model that they are advocating, the various potential
attachments are hypothesized and then evaluated with respect to a
certain number of contextual constraints, including things like
referential ambiguity and lexical bias.  As we can see, the terms
top-down and bottom-up mean something quite different for
psycholinguists than for computational linguists. In the terminology
that we have been using, standard in the computational literature,
what they describe is not necessarily 
a bottom-up process, but is rather consistent with predictive parsing,
in which incomplete constituents, such as the new prepositional
phrase, is composed with earlier constituents, such as the NP or VP.
The crucial question is: what is the nature of the lexical preferences,
and are these preferences mediated by the compositional structure of 
the string. 

Before moving on to discuss specific studies, let us be clear what is
meant in this domain by ``lexical'' models.  Lexicalized grammar
formalisms, such as Categorial Grammar, Tree-Adjoining Grammars (TAG),
and Lexical-Functional Grammar (LFG), have been around for decades.
They embed syntactic information into rich lexical entries, and
account for such lexically dependent phenomena as subcategorization by
including the relevant information in each lexical entry.  Other
formalisms, such as phrase-structure based theories like GPSG and HPSG,
also include a role for lexical influence on the syntax.  All of these
theories, however, deal with categorical syntactic information, not
with lexical preferences or biases, which is more the province of
psycholinguistic and computational models.  It is this latter form of
lexical model that we are discussing here.

To be sure, lexical items can carry with them strong structural
preferences, e.g. verb subcategorization preferences (which may
involve a probabilistic distribution over possible frames) or preposition
attachment preferences.  Sometimes these preferences can immediately 
affect the likelihood of new links -- such as selectional preferences
exerted by a main verb on a hypothesized subject.  Other times, these
preferences drive processing downstream -- such as subcategorization,
which in English can be modeled (as we will show later in the thesis)
as a top-down predictive bias towards particular argument structures
downstream.  These are the sorts of preferences 
that are being investigated in these sentence processing models -- as
well as by statistically-oriented computational linguists such as
Charniak, Collins, and many others.

While the garden path models advocated by Frazier and colleagues
occupy one extreme on the continuum of syntactic and lexical models of
processing, by virtue of giving no role to lexical preferences in
guiding the parsing process, other more recent approaches
\cite{MacDonald94,Trueswell96} have taken the other extreme,
advocating models that can be interpreted as giving no role to
syntactic processing outside of the lexicon.  In these models, all
relevant information and structure is projected out of the lexicon,
and processing is driven by purely local constraints.  
\begin{quote}
\setlength{\baselineskip}{.7\baselineskip}
It is further proposed that most, if not all, syntactic ambiguities
hinge upon one or more lexical ambiguities present in a phrase or
sentence. \cite[p. 566]{Trueswell96}
\end{quote}
Incidentally, many of these models are implemented as neural networks,
to simulate the time-course of disambiguation.  These local models are
straightforward to simulate in a neural network.  Unfortunately,
models which make use of hierarchical structures, such as those
typically used to model syntactic and semantic composition, are quite
difficult for neural networks to handle.  Hence, to the degree that
the ability to model phenomena with neural networks is desirable to a
particular researcher, this is an additional incentive for modeling
disambiguation as a local process.

The point that we want to make in this section is that this sort of model
goes too far; one can pack as much of the structure as one likes into
the lexicon, and there will still be a significant amount of syntactic
processing required.  This is a point that is recognized by linguists
(see, e.g., the Steedman quote in the previous section), but not
apparently by all psycholinguists.  Furthermore, not all constraints
on disambiguation are strictly lexical.  For example, the
presuppositional constraints that have been shown to lead to an NP
modification preference in \namecite{Altmann88}.  Or consider the
following ambiguous sentence:
\begin{examples}
\item Mary told the man John saw\label{ex:nonlex}
\end{examples}
One reading of this is as a response to the question: what did Mary
tell the man?  Another reading is as a response to the question: Who
did Mary tell about it?  It is not clear how one might cast this
ambiguity in terms of any particular lexical ambiguity; rather it is
an ambiguity about how different constituents compose to form an
interpretable sentence.

In fact, some of the constraints that are used to simulate results in
local, lexical approaches -- so-called {\it configurational\/}
biases -- are quite simply non-lexical syntactic constraints.  To take
two recent examples of studies in this paradigm:
\begin{quote}
\setlength{\baselineskip}{.7\baselineskip}
The fourth constraint was a configurational bias favoring the main
clause over the relative clause.  A sentence-initial sequence of
``noun phrase verbed'' is typically the beginning of a main clause.
For present purposes, we remain agnostic about whether the main clause
bias is best characterized at the structural level or whether it
emerges from other more local constraints, including argument
structure preferences or other lexically-triggered constraints.
\cite[p. 288]{Mcrae98}

For present purposes, we remain agnostic about whether this
configurational bias is best characterized at a structural level or
whether it emerges from other more local
constraints. \cite[p. 1521]{Spivey98} 
\end{quote}

Both of these studies were looking at the reduced relative / main verb
ambiguity, attempting to understand the pattern of preferences and how
they can be accounted for in a model of interacting constraints.
Another constraint in the McRae et al. study quoted above is the
thematic fit of the subject NP and either the agent role of the main
verb or the patient role of the reduced relative.  Hence, in this study
at least, there is a clear role of constituency in the
model.  Indeed, the agnosticism expressed in the above papers
distinguishes them from the more extreme position expressed in, for
example, \namecite{MacDonald94}.  These approaches seem instead to
presuppose some syntactic processing mechanism.

An interesting feature of simple English noun phrases without phrasal
modifiers is
that the head is almost always the rightmost noun.  For this reason,
to test the influence of thematic fit, a stimulus is typically
constructed that has the head noun and the ambiguous verb adjacent to
one another.  For example, the following were taken from the items of
the McRae et al. study:
\begin{examples}
\item The snakes devoured by the tribesmen had been roasting over the
coals all afternoon. \label{ex:sn}
\item The rabbits devoured by the tribesmen had been roasting over the
coals all afternoon. \label{ex:ra}
\end{examples}
To the extent that a garden path (evidenced by increased reading time)
is avoided when the subject is a good patient of the verb, this is
taken as evidence for the immediate influence of thematic fit.  There
are two models, however, by which this influence can be exerted, given
the previous items: (i) the probability of the reduced relative
reading given that the previous word is rabbit is greater than the
probability of the reduced relative reading given that the previous
word is snake; or (ii) the probability of the reduced relative
reading given that the subject NP is a rabbit is greater than the
probability of the reduced relative reading given that the the subject
NP is a snake.  The first model is akin to an n-gram model, of the
sort that have been shown to be quite powerful for part-of-speech
tagging and speech recognition.  If the hypothesis is that all of
syntactic disambiguation is locally, lexically driven, then this is
the appropriate kind of model.  Such a model, however, will make very
different predictions from the second.  Consider, for example:
\begin{examples}
\item The snakes, brown and unusually plump, devoured by the tribesmen \ldots \label{ex:sn2}
\item The rabbits, brown and unusually plump, devoured by the tribesmen \ldots \label{ex:ra2}
\end{examples}
The above appositives (with commas included to aid disambiguation)
remove the head of the 
noun phrase from its position adjacent to the verb, without
introducing another attachment ambiguity.  Now the two models would make
different predictions.  The locally driven (n-gram) model would
predict that the thematic role constraint would not contribute to
disambiguation in this case; a model which instead made hypotheses
about the subject NP of the utterance could make use of the thematic
role information.  In addition, it is not clear how the {\it
configurational\/} constraint of string initial NP V$^\prime$ed --
which we will write for convenience S/[NP,V] -- holds up under more
complex NP structures.

A similar point can be made about items from another study within this
 lexicalist paradigm, \namecite{Tabor97}.  Here the ambiguity
is between the complementizer and determiner reading of
\textttt{`that'} when it appears post-verbally and sentence initially.
For example, here are sample items from that study:
\begin{examples}
\item The lawyer insisted that cheap hotel was clean and comfortable.
\item The lawyer insisted that cheap hotels were clean and comfortable.
\item The lawyer visited that cheap hotel to stay for the night.
\item That cheap hotel was clean and comfortable to our surprise.
\item That cheap hotels were clean and comfortable surprised us.
\end{examples}
Here again we have configurational constraints (post-verbal
vs. sentence initial), and the verb, which can select for the
syntactic category of its complement, is adjacent to the ambiguous 
item.  The claim is that interacting constraints (the syntactic
environment and the lexical bias) will determine preferences of
interpretation.  The most natural model, given just the above items,
is again a strictly local Markov model that conditions probabilities
based on adjacent words (where the beginning of
the sentence is treated as a `word').  This, of course, also makes it
straightforward to model in a neural network\footnote{As was
previously mentioned, \namecite{Steedman99} points out that the
syntactic disambiguation carried out in neural networks is akin to
part-of-speech tagging.}.  The predictions of this locally driven
model would be, for example, that the probability of \textttt{`that'} as
a determiner following \textttt{`visited'} is much higher than the
probability of \textttt{`that'} as a complementizer following
\textttt{`visited'}.  

An alternative model might make the following
claim: that the probability of an NP complement of \textttt{`visited'}
is much higher than the probability of a sentential complement, hence
the probability of a complementizer is very low, since complementizers
begin sentential constituents, not NP constituents.  This model
would by necessity involve some syntactic processing, since the
prediction being made is not for the specific lexical item, but for a
constituent, which in turn makes predictions of its own.  Of course,
both kinds of predictions (local and syntactic) could play a role.

In order to differentiate between these two models, the critical
modification that must be made to the stimuli is to move the
ambiguities out of the local influence of the specific lexical items. 
If the lexical influence is active despite the intervention of
arbitrary constituents, then very simple local models will 
prove inadequate to model such a process.  Interestingly, these kinds
of stimuli have been around from the beginning of this line of
psycholinguistic research.  The stimuli from the \namecite{Crain85}
study that were mentioned above, and which we reprint here for ease of
reference, do exactly this, in separating the ambiguous
\textttt{`that'} from the verb via an earlier NP argument:
\begin{examples}
\item The psychologist told the woman that he was having trouble with
{\it to visit him again}
\item The psychologist told the woman that he was having trouble with
{\it her husband}
\end{examples}
One interesting line of empirical research would be to enrich the
intervening NP with an arbitrary amount of lexical material -- through
noun compounding and other forms of modification -- to examine whether
the verb can sustain its influence.

Another modification that can be made to try to tease apart whether or
not there is a role for processing over and above lexical processing
is to remove the lexicon from the process -- either by using
non-lexical ambiguities of the sort illustrated above in example
\ref{ex:nonlex}, or by the use of non-words, which do not have a
lexical entry.  For example
\begin{examples}
\item The man snarmed the larimonious klarm of the vite.
\end{examples}
To the extent that there are attachment preferences independent of the
heads of the phrases to which the PP is attaching, this is evidence
for the kinds of statistical generalizations that a structural,
constituent-based model can capture.

\section{Robust sentence processing models}
The parsing model that we will present in the following chapters is
one that conditions the probabilities (weights) of structures on many
features in the syntactic and lexical context.  It shares with the
models outlined in the previous section the ability to model certain
lexical preferences such as verb subcategorization and PP attachment
biases. Like these models, our parsing model is parallel; further, it
builds the connected structures which are assumed as an input into
many of
these models.  It allows, however, for syntactic generalizations that
are not available in the strictly lexical models -- such as a general
NP attachment bias for specific prepositions, e.g. \textttt{`of'}.
Most importantly, the model scales up to process freely occurring
sentences; i.e. the processing model is robust enough to accurately
process arbitrary strings of the language.

Why should such a project be of interest to psycholinguists?  Most
psycholinguists will be interested only to the extent that the parser
can, or potentially can, help to understand how people process
language.  Apart from arguments about the relative simplicity of
extending this model to include incremental interpretation, and its
ability to capture abstract, syntactic dependencies that would
elude extreme local, lexicalist models, there are a couple of ways in
which this project could potentially shed light on human sentence
processing.  The first is through falsifiable hypotheses that
are explicit in the model as it is implemented.  The second is as a
testbed for investigating models of interacting probabilistic
constraints beyond a set of toy examples.  We will discuss both of
these in this section.

Before we do, however, let us first reiterate our focus on the robust
nature of human language processing, and the importance of having
models of sentence processing that can scale up to handle language
robustly.  For example, a serial sentence processing model with
reanalysis can make very explicit predictions based on the failure of
a preferred interpretation at certain points in the string, and the
subsequent processing required to generate a second analysis.  It
would be of great interest to see an effective robust parsing
implementation of such a processing mechanism, because it would of
necessity require an explicit definition of failure (what triggers
reanalysis) that is sufficiently sensitive to cause reanalysis when
needed, but sufficiently conservative not to cause reanalysis when it
is not needed.  It would also require a reanalysis procedure that
would be general enough to handle the very large number of
constructions that occur in natural language.  Our broad coverage
parser may be considered as an existence proof of a robust limited
parallel model.  Such a proof may be of limited interest to 
psycholinguists, but the level of explicitness that is required for
the implementation goes beyond what has been provided for other
models: to become robust, other models would certainly have to extend the
coverage of their grammars, and thus increase the ambiguity of the
space within which their mechanism searches.  This would almost
certainly force changes to the model, and make explicit the
predictions of their mechanism.

For our parsing model, we have made a number of explicit definitions
that could result in testable hypotheses about how people process
sentences.  In particular, the parser follows a top-down search path,
which carries with it certain advantages and disadvantages relative to
other parsing strategies,  which will be discussed at length in the
chapters to come.  In order to make this parsing strategy work, we
followed a probabilistic beam search, in which parses falling outside
of the beam were discarded.  This combination of top-down and
probabilistic beam-search makes some strong predictions.
We will mention a couple here, with the proviso that this is not the
central work of this thesis, and that more work would be necessary to
fully flesh out these predictions into testable hypotheses.  We
will give a sense of these predictions, but do not claim to be in a
position to evaluate them at this time.

One prediction our robust parser makes is with regards to the depth of
embedding.  It is generally believed that center embedding is costlier
than purely left- or right-branching structures, and this has been
used to argue for a left-corner parsing mechanism
\cite{Abney91,Resnik92}, which we will discuss in detail in the next
chapter.  A left-corner parser, unlike our top-down parser, would
predict a potentially unlimited depth of embedding along the left edge
of the tree, i.e. at the first word.  In a nutshell, the left-corner parser
delays prediction of the parent node until the leftmost child has been
constructed, and hence will only build embedded structures at decision
points later in the string.  Our parsing model, in contrast, would
predict that 
embedded structure beyond a certain depth, even at the left edge of
the tree, would be pruned by the beam search.  To the extent that
unlimited embedding is not acceptable to people sentence initially,
this could be explained by our mechanism.

Interestingly, the specific definition of the beam threshold that we
use in the parser, which was chosen for both efficiency and accuracy,
does provide an explanation for increased difficulty for
deep center embedding versus embedded structure along the left or
right periphery of the tree.  To understand the specifics of the
threshold definition, the reader is referred to the next chapter,
where it is explained in depth.  The basic idea is that all parses
within some probability range of the best parse are kept; all those
with a probability that falls too far below the best probability are
discarded.  The breadth of the probability range, however, is variable,
depending on the number of alternatives that fall within the range: if
many parses fall within the original range, it is narrowed, so
that more of them fall outside of the threshold.  Of course, as new
ambiguities are introduced in the string, more relatively probable
parses are introduced into the beam.  This increases the density of
alternatives within the original probability range, causing the beam
to shrink.  In other words, further along the string, predicted
embedded structures fall outside of the beam much faster than at the
left edge of the string.  This would predict more difficulty for center
embedding than for sentence-initial embedding.

While the above predictions are for the specific version of beam
search that we implemented, any incremental beam search approach
will result in a garden pathing model, as is noted in
\namecite{Jurafsky96}.  The parser that we outline in the next chapter
does garden path on some percentage of the sentences, although this
seems to be more the result of poor probabilistic models due to sparse
data -- the specific examples 
of garden path sentences are not particularly instructive.  Our
approach to parameter estimation -- relative frequency estimation from
a treebank corpus -- yields an effective yet quite stupid probability
model, by which we mean that it encodes little in the way of explicit
knowledge.  All that the model encodes is observed co-occurrence,
which is a far cry from the sophisticated linguistic and real-world
dependencies that humans presumably encode and exploit in language
processing.  As a result, it is difficult to make general predictions
based upon the particular probability model that our parser uses.
Hence, if the parser garden paths on a particular sentence, it seems
to be more the result of a problem with the probabilistic model.

That said, one potential research program would be to engineer a
robust probabilistic grammar that encodes probabilistic dependencies
that are psycholinguistically well-motivated or attested.  That is not
to say that the dependencies that we encode do not exist -- on the
contrary, the certainly seem to capture important regularities, ones
which may have a psycholinguistic correlate.
However, given that people have a much richer model of these
dependencies, perhaps certain engineered probabilistic models would
be better as psycholinguistic models.  Since the
parsing approach that we advocate is independent of the grammar and
probability model, such engineered grammars could be tested to see how
they perform with standard garden path sentences, as well as how they
scale up to deal with what should be unproblematic strings.  One
cannot underestimate the importance of this last test.  Finding
generalizations that buy a syntactic processing mechanism the right
bias in a handful of cases is easy; those same generalizations can
often cause the parser to go very wrong in other unexpected
circumstances. 

Another potentially interesting direction would be to look at adding
inhibitory competition between alternatives.  To the extent that
competition for limited resources is seen to drive the timecourse of
sentence processing (e.g. \egcite{Tabor97}, our parser, with additional
competitive processing, could make predictions about the timecourse of
human sentence processing.  As with the previous suggestion of
engineering a well-motivated grammar, the results of such experiments
would be very dependent on the parameters that are adopted, both for
the probabilistic model and the parsing model.  

If one were confident in the psychological reality of the parameters
of the probabilistic model, another interesting potential model for
the timecourse of sentence processing would be in terms of the speed
of lexical access.  We will demonstrate in this thesis the
applicability of a probabilistic parser of this sort to language
modeling, i.e. predicting each subsequent word from context.  With the
appropriate head-to-head dependencies, such a language model can be
quite peaked; in other words, it makes very strong predictions about
what is likely to follow.  If the following word has a high
probability (a high activation), it should be processed faster than an
unexpected (low probability or activation) word.  To the extent that a
parsing model such as the one that we are using can capture the kinds
of lexico-syntactic dependencies that are argued for in the
psycholinguistic literature -- such as verb subcategorization and
selectional preferences, as well as other head-to-head dependencies --
it could provide detailed timecourse predictions based on probabilistic
expectations.  A similar approach was taken in \namecite{Hale01}, with
a probabilistic Earley parser.

All of this is quite speculative.  The research program that will be
presented in the remainder of the thesis is focused upon building a
robust parser, not on building an explicit model of human sentence
processing. The key contributions of this thesis are computational,
not psycholinguistic.  Nevertheless, the model is consistent with several
critical aspects of human sentence processing, and such a parser could
serve as a fruitful testbed for richer psychological models.

\section{Chapter summary}
The basic claim that is being made here is that connected syntactic
structures, of the sort provided by a top-down parser, facilitate
incremental interpretation.
Shieber and Johnson showed that a bottom-up parser could be sufficient
for incremental interpretation, provided that it is coupled
asynchronously with a semantic processor which can derive, from these
bottom-up syntactic derivations, the semantic relationships.  This
does result in sufficient information to allow for incremental
interpretation, yet such an approach requires an additional 
derivational mechanism at some point in the process that is capable of
producing this information.  A simpler model is one that tightly
couples the syntactic and semantic processing, hence requiring
sufficient syntactic structure to be built to allow for distinctions
in interpretation to be made. 
The basic problem that will be addressed in subsequent chapters will
be: how can this be done?  The rest of this thesis will show that
a probabilistic approach not only makes top-down parsing possible in a
very ambiguous search space, but also an efficient alternative to
other parsing strategies, with computational benefits in its own right.

In presenting the parser that follows, we are not making claims about
the syntactic representations that people maintain in processing
sentences.  Rather, we are making the claim that the kind of
parsing strategy that we are investigating is
consistent with models of human sentence processing, and results in a
simpler model than would result from another parsing strategy.
Further, many of the phenomena modeled by
local, lexically driven models, such as PP attachment preferences or
verb subcategorization biases, are captured by this model as well.
Yet those models tend to emphasize disambiguation and ignore real
issues of hypothesis search that our approach handles.

\chapter{Probabilistic top-down and left-corner parsing}
This chapter will introduce, outline, and evaluate a broad-coverage
probabilistic parsing strategy that maintains fully connected trees in
the left context.  In such a way, 
this parser is consistent with many diverse models of human sentence
processing in a way that no other broad-coverage parser is.  Certainly 
there have been parsing models proposed that fit the demands of
on-line interpretation -- e.g. \namecite{Abney91},
\namecite{Stabler91}, \namecite{Jurafsky96},
\namecite{Stevenson93}, and \namecite{Vosse00}.  Yet these models at 
best are implemented with toy grammars and toy examples, and there has
been no attempt to investigate how they scale up to deal with freely
occurring language.  Our model will be shown to scale up very well,
and to provide results that are comparable with the best
broad-coverage parsers in the literature.

The chapter will be structured as follows.  We will first provide the 
requisite computational background for the algorithm that we will be
discussing, and formally present the algorithm.  We will then give
some empirical results 
for parsers that take POS labels as input, rather than words.  The
model will then be extended to take words as input, and to use more
robust conditional probability models.  Empirical results will then be 
presented for this extended parser, and it will be compared with other 
parsing results from the literature.

\section{Background}
This section will introduce probabilistic (or stochastic) context-free
grammars (PCFGs), as well as
such notions as derivations, trees, and c-command, which
will be important in defining our language model later in the thesis.
In addition, we will explain several grammar transformations 
that will be used.  

\subsection{Grammars, derivations, and trees}\label{sec:gramm}
Here we will formally introduce context-free grammars, which form the
basis for the PCFGs that we will use in parsing.  The presentation here
follows closely that in \namecite{Aho86}.  A CFG $G =
(V,T,P,S^\dag)$, consists of a set of non-terminal symbols $V$, a set
of terminal symbols $T$, a start symbol $S^\dag \in V$, and a set of
rule productions $P$ of the form: $A~\rightarrow~\alpha$, where $A \in
V$ and $\alpha \in (V \cup T)^{\ast}$. These context-free 
rules can be interpreted as rewrite rules, whereby the non-terminal
$A$ on the left-hand side of the rule is rewritten as (or replaced by)
the $\alpha$ on the right-hand side of the rule.  Note that $\alpha$
is a member of $(V \cup T)^{\ast}$, so that it may be the empty string
$\epsilon$.  Productions with an empty
right-hand side are called epsilon productions, and they are usually
written $A~\rightarrow~\epsilon$.  Such a rule says that $A$ can be
re-written as nothing at all.

A sequence of replacements is called a derivation, and we 
represent one step of a derivation -- i.e. one replacement via a
context-free rule -- with the symbol $\Rightarrow$.  Thus, if we have
a rule $A~\rightarrow~\alpha$, then
$\beta~A\gamma~\Rightarrow~\beta\alpha\gamma$.  To denote a derivation
in zero or more steps, we 
use $\stackrel{\ast}{\Rightarrow}$.  To denote a derivation in one or
more steps, we use $\stackrel{+}{\Rightarrow}$.

A CFG $G$ defines a language $L_G$, which is a subset of $T^{\ast}$,
consisting of only those strings that can be derived in one or
more steps from the 
start symbol, i.e. $\alpha$ such that $\alpha \in T^{\ast}$ and
$S^\dag~\stackrel{+}{\Rightarrow}~\alpha$.
We will denote strings either as $w$ or as $w_{0}w_{1}\ldots w_{n}$,
where $w_n$ is understood to be the last terminal symbol in the string.
For simplicity in displaying equations, from this point forward let
$w_{i}^{j}$ be the substring $w_{i}\ldots w_{j}$. 

Consider the following simple example of noun compounding.  Suppose
that the symbol N is our non-terminal start symbol, and our grammar
consists of the following rules: 
\begin{examples}
\item N $\rightarrow$ N N
\item N $\rightarrow$ \textttt{dog}
\item N $\rightarrow$ \textttt{food}
\item N $\rightarrow$ \textttt{can}
\end{examples}
If we want to show that \textttt{`dog food can'} is in the language
generated by this grammar, we can derive it starting from N and
replacing non-terminals via these rewrite rules.  Here are three
possible derivations of the string.
\begin{examples}
\item N $\Rightarrow$ N N $\Rightarrow$ N \textttt{can} $\Rightarrow$
N N \textttt{can} $\Rightarrow$ N \textttt{food can} $\Rightarrow$
\textttt{dog food can}  \label{dev1}
\item N $\Rightarrow$ N N $\Rightarrow$ N N N $\Rightarrow$
N N \textttt{can} $\Rightarrow$ N \textttt{food can} $\Rightarrow$
\textttt{dog food can}  \label{dev2}
\item N $\Rightarrow$ N N $\Rightarrow$ \textttt{dog} N $\Rightarrow$
\textttt{dog} N N $\Rightarrow$ \textttt{dog food} N $\Rightarrow$
\textttt{dog food can}  \label{dev3}
\end{examples}
Notice that in the second step of each derivation, we followed three
different derivation paths.  Derivation \ref{dev1} replaced the
rightmost non-terminal with a terminal item, whereas derivation
\ref{dev3} replaced the leftmost non-terminal with a
terminal item.  A derivation that always replaces the
rightmost non-terminal is called {\it rightmost\/}, and a derivation
that always replaces the leftmost non-terminal is called {\it
leftmost\/}.  It is ambiguous in derivation \ref{dev2} whether the
non-terminal replaced in the second step of the derivation was the
left or rightmost.  To remove this ambiguity, we can introduce
brackets, which delimit the beginning and ending symbols of the
replaced non-terminals.  We will omit the outermost brackets and
brackets around terminal items for convenience.  The three derivations
with such bracketing are:
\begin{examples}
\item N $\Rightarrow$ N N $\Rightarrow$ N \textttt{can} $\Rightarrow$
(N N) \textttt{can} $\Rightarrow$ (N \textttt{food}) \textttt{can}
$\Rightarrow$ (\textttt{dog food}) \textttt{can}  \label{bdev1}
\item N $\Rightarrow$ N N $\Rightarrow$ N (N N) $\Rightarrow$
N (N \textttt{can}) $\Rightarrow$ N (\textttt{food can}) $\Rightarrow$
\textttt{dog} (\textttt{food can})  \label{bdev2}
\item N $\Rightarrow$ N N $\Rightarrow$ \textttt{dog} N $\Rightarrow$
\textttt{dog} (N N) $\Rightarrow$ \textttt{dog} (\textttt{food} N)
$\Rightarrow$ \textttt{dog} (\textttt{food can})  \label{bdev3}
\end{examples}
Derivations \ref{bdev1} and \ref{bdev2} are rightmost derivations,
while derivation \ref{bdev3} is leftmost.  Note, however, that the two
rightmost
derivations do not end up with the same bracketing.  This corresponds
to an ambiguity in the grammar.  

We will sometimes speak of top-down leftmost derivations in terms
of a {\it pushdown automaton\/} (PDA).  A PDA is a finite-state
automaton with a stack.  A stack is a data structure within which
entries can be inserted ({\it pushed\/}) and removed ({\it popped\/}) in
a last-in-first-out manner.  A derivation can be thought of as
follows: first, we place the start symbol of the grammar onto the
stack; then, until the stack is empty, pop the last non-terminal from
the stack, and if it is a non-terminal, push the symbols on the
right-hand side of the rule onto the stack, from right-to-left.  This
order of pushing symbols onto the stack ensures that the next symbol
popped from the stack will be the leftmost child.

The bracketing that we have provided can be enhanced to provide more
information about the derivation, including the label of the
non-terminal that has been 
expanded.  Let us take derivation \ref{bdev3} as an example.  To make
this clear, the outermost brackets will now be included, and all
non-terminals will be bracketed, even before they are expanded.  
\begin{examples}
\item (N) $\Rightarrow$ (N (N) (N)) $\Rightarrow$ 
(N (N \textttt{dog}) (N)) $\Rightarrow$ (N (N \textttt{dog}) (N (N)
(N))) $\Rightarrow$ \\(N (N \textttt{dog}) (N (N \textttt{food}) (N))) 
$\Rightarrow$ (N (N \textttt{dog}) (N (N  \textttt{food}) (N
\textttt{can})))  \label{bdev4}
\end{examples}
With the exception of the new bracketing and labeling, derivation
\ref{bdev4} is identical to derivation \ref{bdev3}.  

The labeled bracketing at the end of derivation \ref{bdev4} is
equivalent to a tree.  A tree represents a derivation, without any
information about the order of replacements.  Thus derivations
\ref{bdev2} and \ref{bdev3} would be represented by the same tree,
despite following different derivation strategies.
Our top-down parser will be following a leftmost derivation strategy,
so the ordering of the rewriting will be fixed, which implies one derivation
per tree.  Hence we will sometimes speak of trees and sometimes of
derivations.  

A complete derivation is one in which there are no
more non-terminals left to be replaced.  A complete tree is the tree
representation of a complete derivation.  The terminal yield of any
derivation is the sequence of terminal items in the output of the
derivation. 

A PCFG is a CFG with a probability assigned to each
rule;  specifically, each right-hand side has a probability given the
left-hand side of the rule.  The probability of a parse tree is
the product of the probabilities of each rule in the tree.  Provided a
PCFG is consistent (or tight), which it always will be in the
approach we will be advocating\footnote{A PCFG is consistent or tight
if there is no probability mass reserved for infinite trees.
\namecite{Chi98} proved that any PCFG estimated from a treebank with
the maximum likelihood relative frequency estimator is tight.  All of
the PCFGs that are 
used in this thesis are estimated using the relative frequency
estimator.}, this defines a proper probability distribution over completed
trees.  

A PCFG also defines a probability distribution over strings of
words (terminals) in the following way.  Let $T_w$ be the set of all
complete trees rooted at the start symbol, with the string of
terminals $w_{0}^{n}$ as the terminal yield.   Then
\begin{eqnarray}
\Pr(w_{0}^{n}) &=& \sum_{t \in T_w}\Pr(t) \label{eq:prst}
\end{eqnarray}
The intuition behind equation \ref{eq:prst} is that, if a string
is generated by the PCFG, then it will be produced
if and only if it is the terminal yield of one of the trees in the set
$T_w$.  Hence, the probability of the string occurring in the space of
all possible strings of the language is the probability of the set $T_w$,
i.e. the sum of its members' probabilities.

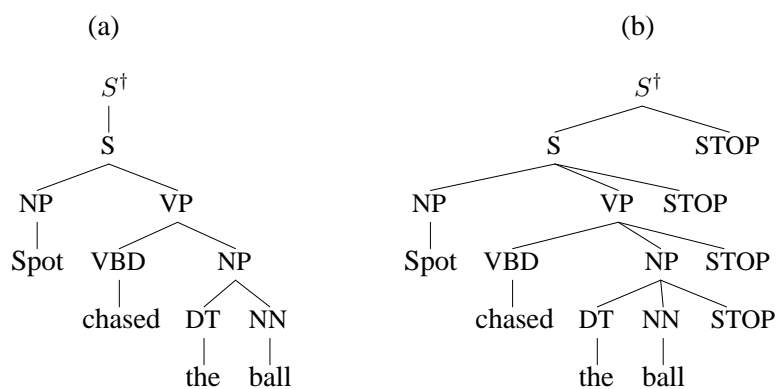
\begin{figure}[t]
\begin{picture}(106,145)(0,-125)
\put(29,15){(a)}
\put(34,-8){\small $S^\dag$}
\drawline(37,-12)(37,-22)
\put(34,-30){\small S}
\drawline(37,-34)(10,-44)
\put(3,-52){\small NP}
\drawline(10,-56)(10,-66)
\put(-0,-74){Spot}
\drawline(37,-34)(63,-44)
\put(56,-52){\small VP}
\drawline(63,-56)(41,-66)
\put(30,-74){\small VBD}
\drawline(41,-78)(41,-88)
\put(27,-96){chased}
\drawline(63,-56)(85,-66)
\put(78,-74){\small NP}
\drawline(85,-78)(73,-88)
\put(66,-96){\small DT}
\drawline(73,-100)(73,-110)
\put(66,-118){the}
\drawline(85,-78)(98,-88)
\put(90,-96){\small NN}
\drawline(98,-100)(98,-110)
\put(90,-118){ball}
\end{picture}
\begin{picture}(119,145)(-40,-125)
\put(82,15){(b)}
\put(87,-8){\small $S^\dag$}
\drawline(90,-12)(57,-22)
\put(54,-30){\small S}
\drawline(57,-34)(10,-44)
\put(3,-52){\small NP}
\drawline(10,-56)(10,-66)
\put(-0,-74){Spot}
\drawline(57,-34)(81,-44)
\put(74,-52){\small VP}
\drawline(81,-56)(41,-66)
\put(30,-74){\small VBD}
\drawline(41,-78)(41,-88)
\put(27,-96){chased}
\drawline(81,-56)(97,-66)
\put(91,-74){\small NP}
\drawline(97,-78)(73,-88)
\put(66,-96){\small DT}
\drawline(73,-100)(73,-110)
\put(66,-118){the}
\drawline(97,-78)(98,-88)
\put(90,-96){\small NN}
\drawline(98,-100)(98,-110)
\put(90,-118){ball}
\drawline(97,-78)(122,-88)
\put(116,-96){\small STOP}
\drawline(81,-56)(121,-66)
\put(114,-74){\small STOP}
\drawline(57,-34)(104,-44)
\put(98,-52){\small STOP}
\drawline(90,-12)(123,-22)
\put(110,-30){\small STOP}
\end{picture}
\caption{Two parse trees: (a) a complete parse tree; (b) a complete
parse tree with an explicit stop symbol for rules}\label{fig:itree}
\end{figure}

Finally, let us introduce the term \textttt{c-command}.  We will use
this notion in our conditional probability model, and it is also
useful for understanding some of the previous work in this area.  
Recall that a node $A$ dominates a node $B$ in a tree if  and only if
either (i) $A$ is the parent of $B$; or (ii) $A$ is the parent of a
node $C$ that dominates $B$.  A node $B$ is {\it lower\/} in the tree
than $A$ if $A$ dominates $B$.  The simple definition of c-command
that we will be using in this thesis is the following:  a node $A$
c-commands a node $B$ if and  only if (i) $A$ does not dominate $B$;
and (ii) the lowest branching node  (i.e. non-unary node) that
dominates $A$ also dominates $B$.  Thus in
figure \ref{fig:itree}(a), the subject NP and the VP
each c-command the other, because neither dominates the other and the
lowest branching node above both (the S) dominates the
other.  Notice that the subject NP c-commands the object
NP, but not vice versa, since the lowest branching node that
dominates the object NP is the VP, which does not
dominate the subject NP.

In certain circumstances it will be useful for us to think of each
rule expansion in the tree as having an explicit stop symbol.  In the
incremental algorithms that we will be presenting, rules will
typically be predicted one child at a time.  We can leave the
possibility open for subsequent children by not predicting that the
rule stops.  One can do this by including an explicit empty STOP
category to every production, as in figure \ref{fig:itree}(b).  The
syntactic structures that are represented in the two trees in figure
\ref{fig:itree} are the same.

\subsection{Grammar transforms}\label{sec:fact}
\namecite{Nijholt80} characterized parsing strategies in terms of two 
{\it announce points\/}: the point at which a parent category is
announced (identified) relative to its children, and the point at
which the rule expanding the parent is identified.  \namecite{Abney91}
expanded on this, adding a third announce point, namely when the arc
is announced between a parent and a particular child.  In 
pure top-down parsing, a parent category and the rule expanding it are
announced {\it before\/} any of its children. In pure bottom-up parsing, they
are identified {\it after\/} all of the children. Standard left-corner
parsers announce a parent category and its expanding rule {\it after\/} its
leftmost child has been completed, but {\it before\/} any of the other
children.

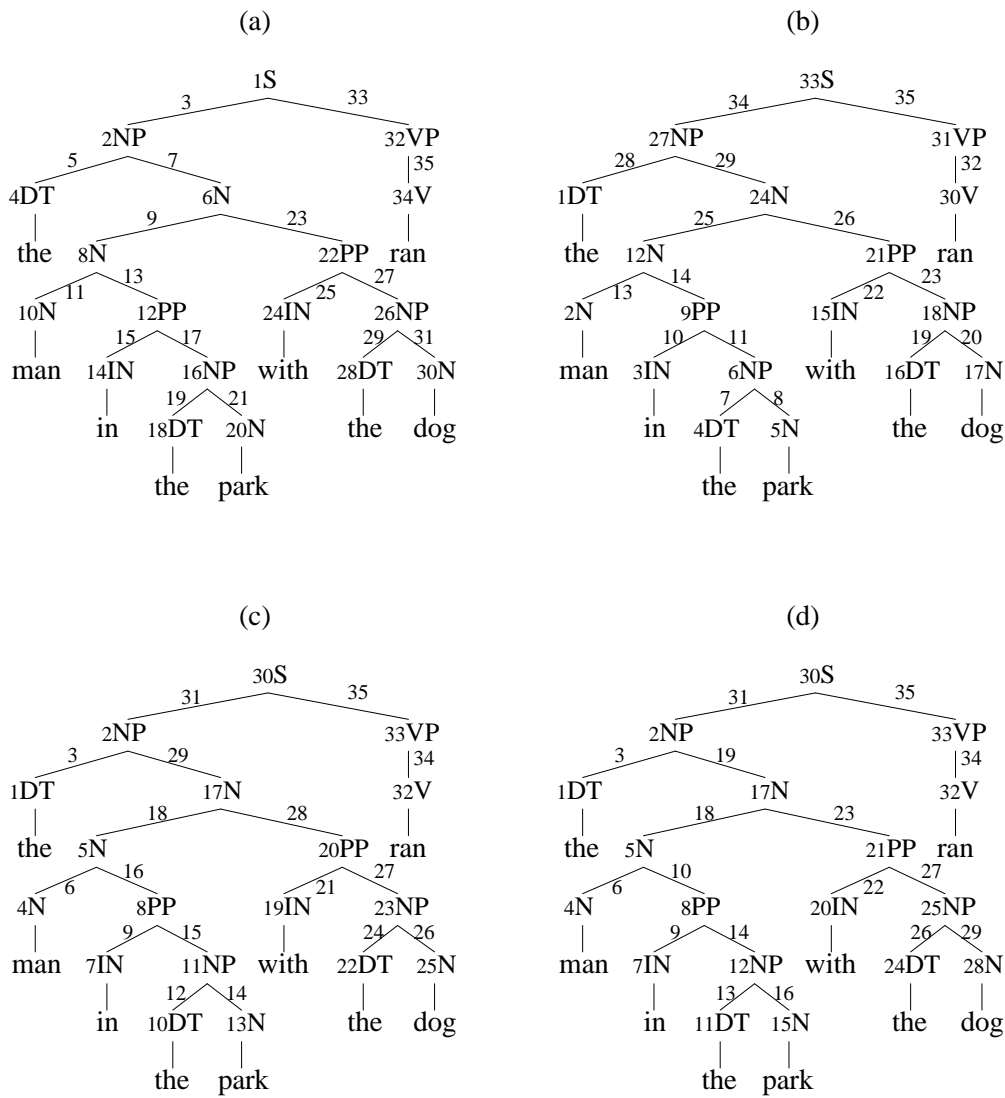
\begin{figure}[t]
\begin{picture}(344,188)(0,-188)
\put(87,-8){(a)}
\put(92,-30){\small {\scriptsize 1}S}
\drawline(98,-34)(45,-44)
\put(65,-38){\scriptsize 3}
\put(35,-52){\small {\scriptsize 2}NP}
\drawline(45,-56)(10,-66)
\put(22,-60){\scriptsize 5}
\put(-0,-74){\small {\scriptsize 4}DT}
\drawline(10,-78)(10,-88)
\put(3,-96){the}
\drawline(45,-56)(80,-66)
\put(60,-60){\scriptsize 7}
\put(73,-74){\small {\scriptsize 6}N}
\drawline(80,-78)(33,-88)
\put(52,-82){\scriptsize 9}
\put(26,-96){\small {\scriptsize 8}N}
\drawline(33,-100)(10,-110)
\put(21,-110){\scriptsize 11}
\put(3,-118){\small {\scriptsize 10}N}
\drawline(10,-122)(10,-132)
\put(1,-140){man}
\drawline(33,-100)(56,-110)
\put(43,-104){\scriptsize 13}
\put(48,-118){\small {\scriptsize 12}PP}
\drawline(56,-122)(37,-132)
\put(40,-127){\scriptsize 15}
\put(29,-140){\small {\scriptsize 14}IN}
\drawline(37,-144)(37,-154)
\put(33,-162){in}
\drawline(56,-122)(75,-132)
\put(65,-127){\scriptsize 17}
\put(65,-140){\small {\scriptsize 16}NP}
\drawline(75,-144)(62,-154)
\put(59,-150){\scriptsize 19}
\put(52,-162){\small {\scriptsize 18}DT}
\drawline(62,-166)(62,-176)
\put(55,-184){the}
\drawline(75,-144)(88,-154)
\put(83,-150){\scriptsize 21}
\put(82,-162){\small {\scriptsize 20}N}
\drawline(88,-166)(88,-176)
\put(79,-184){park}
\drawline(80,-78)(126,-88)
\put(105,-82){\scriptsize 23}
\put(117,-96){\small {\scriptsize 22}PP}
\drawline(126,-100)(104,-110)
\put(116,-110){\scriptsize 25}
\put(96,-118){\small {\scriptsize 24}IN}
\drawline(104,-122)(104,-132)
\put(94,-140){with}
\drawline(126,-100)(147,-110)
\put(138,-104){\scriptsize 27}
\put(138,-118){\small {\scriptsize 26}NP}
\drawline(147,-122)(134,-132)
\put(134,-127){\scriptsize 29}
\put(124,-140){\small {\scriptsize 28}DT}
\drawline(134,-144)(134,-154)
\put(128,-162){the}
\drawline(147,-122)(161,-132)
\put(153,-127){\scriptsize 31}
\put(154,-140){\small {\scriptsize 30}N}
\drawline(161,-144)(161,-154)
\put(153,-162){dog}
\drawline(98,-34)(151,-44)
\put(128,-36){\scriptsize 33}
\put(142,-52){\small {\scriptsize 32}VP}
\drawline(151,-56)(151,-66)
\put(153,-61){\scriptsize 35}
\put(145,-74){\small {\scriptsize 34}V}
\drawline(151,-78)(151,-88)
\put(144,-96){ran}
\put(294,-8){(b)}
\put(299,-30){\small {\scriptsize 33}S}
\drawline(305,-34)(252,-44)
\put(272,-38){\scriptsize 34}
\put(242,-52){\small {\scriptsize 27}NP}
\drawline(252,-56)(217,-66)
\put(229,-60){\scriptsize 28}
\put(207,-74){\small {\scriptsize 1}DT}
\drawline(217,-78)(217,-88)
\put(210,-96){the}
\drawline(252,-56)(286,-66)
\put(267,-60){\scriptsize 29}
\put(280,-74){\small {\scriptsize 24}N}
\drawline(286,-78)(240,-88)
\put(259,-82){\scriptsize 25}
\put(233,-96){\small {\scriptsize 12}N}
\drawline(240,-100)(217,-110)
\put(228,-110){\scriptsize 13}
\put(210,-118){\small {\scriptsize 2}N}
\drawline(217,-122)(217,-132)
\put(208,-140){man}
\drawline(240,-100)(263,-110)
\put(250,-104){\scriptsize 14}
\put(254,-118){\small {\scriptsize 9}PP}
\drawline(263,-122)(244,-132)
\put(247,-127){\scriptsize 10}
\put(236,-140){\small {\scriptsize 3}IN}
\drawline(244,-144)(244,-154)
\put(240,-162){in}
\drawline(263,-122)(282,-132)
\put(272,-127){\scriptsize 11}
\put(272,-140){\small {\scriptsize 6}NP}
\drawline(282,-144)(269,-154)
\put(269,-150){\scriptsize 7}
\put(259,-162){\small {\scriptsize 4}DT}
\drawline(269,-166)(269,-176)
\put(262,-184){the}
\drawline(282,-144)(295,-154)
\put(289,-150){\scriptsize 8}
\put(288,-162){\small {\scriptsize 5}N}
\drawline(295,-166)(295,-176)
\put(285,-184){park}
\drawline(286,-78)(333,-88)
\put(312,-82){\scriptsize 26}
\put(324,-96){\small {\scriptsize 21}PP}
\drawline(333,-100)(311,-110)
\put(323,-110){\scriptsize 22}
\put(303,-118){\small {\scriptsize 15}IN}
\drawline(311,-122)(311,-132)
\put(301,-140){with}
\drawline(333,-100)(354,-110)
\put(345,-104){\scriptsize 23}
\put(345,-118){\small {\scriptsize 18}NP}
\drawline(354,-122)(341,-132)
\put(341,-127){\scriptsize 19}
\put(331,-140){\small {\scriptsize 16}DT}
\drawline(341,-144)(341,-154)
\put(334,-162){the}
\drawline(354,-122)(368,-132)
\put(360,-127){\scriptsize 20}
\put(361,-140){\small {\scriptsize 17}N}
\drawline(368,-144)(368,-154)
\put(360,-162){dog}
\drawline(305,-34)(358,-44)
\put(335,-36){\scriptsize 35}
\put(349,-52){\small {\scriptsize 31}VP}
\drawline(358,-56)(358,-66)
\put(360,-61){\scriptsize 32}
\put(352,-74){\small {\scriptsize 30}V}
\drawline(358,-78)(358,-88)
\put(351,-96){ran}
\end{picture}

\vspace*{.5in}

\begin{picture}(344,188)(0,-188)
\put(87,-8){(c)}
\put(92,-30){\small {\scriptsize 30}S}
\drawline(98,-34)(45,-44)
\put(65,-38){\scriptsize 31}
\put(35,-52){\small {\scriptsize 2}NP}
\drawline(45,-56)(10,-66)
\put(22,-60){\scriptsize 3}
\put(-0,-74){\small {\scriptsize 1}DT}
\drawline(10,-78)(10,-88)
\put(3,-96){the}
\drawline(45,-56)(80,-66)
\put(60,-60){\scriptsize 29}
\put(73,-74){\small {\scriptsize 17}N}
\drawline(80,-78)(33,-88)
\put(52,-82){\scriptsize 18}
\put(26,-96){\small {\scriptsize 5}N}
\drawline(33,-100)(10,-110)
\put(21,-110){\scriptsize 6}
\put(3,-118){\small {\scriptsize 4}N}
\drawline(10,-122)(10,-132)
\put(1,-140){man}
\drawline(33,-100)(56,-110)
\put(43,-104){\scriptsize 16}
\put(48,-118){\small {\scriptsize 8}PP}
\drawline(56,-122)(37,-132)
\put(43,-127){\scriptsize 9}
\put(29,-140){\small {\scriptsize 7}IN}
\drawline(37,-144)(37,-154)
\put(33,-162){in}
\drawline(56,-122)(75,-132)
\put(65,-127){\scriptsize 15}
\put(65,-140){\small {\scriptsize 11}NP}
\drawline(75,-144)(62,-154)
\put(59,-150){\scriptsize 12}
\put(52,-162){\small {\scriptsize 10}DT}
\drawline(62,-166)(62,-176)
\put(55,-184){the}
\drawline(75,-144)(88,-154)
\put(82,-150){\scriptsize 14}
\put(82,-162){\small {\scriptsize 13}N}
\drawline(88,-166)(88,-176)
\put(79,-184){park}
\drawline(80,-78)(126,-88)
\put(105,-82){\scriptsize 28}
\put(117,-96){\small {\scriptsize 20}PP}
\drawline(126,-100)(104,-110)
\put(116,-110){\scriptsize 21}
\put(96,-118){\small {\scriptsize 19}IN}
\drawline(104,-122)(104,-132)
\put(94,-140){with}
\drawline(126,-100)(147,-110)
\put(138,-104){\scriptsize 27}
\put(138,-118){\small {\scriptsize 23}NP}
\drawline(147,-122)(134,-132)
\put(134,-127){\scriptsize 24}
\put(124,-140){\small {\scriptsize 22}DT}
\drawline(134,-144)(134,-154)
\put(128,-162){the}
\drawline(147,-122)(161,-132)
\put(153,-127){\scriptsize 26}
\put(154,-140){\small {\scriptsize 25}N}
\drawline(161,-144)(161,-154)
\put(153,-162){dog}
\drawline(98,-34)(151,-44)
\put(128,-36){\scriptsize 35}
\put(142,-52){\small {\scriptsize 33}VP}
\drawline(151,-56)(151,-66)
\put(153,-61){\scriptsize 34}
\put(145,-74){\small {\scriptsize 32}V}
\drawline(151,-78)(151,-88)
\put(144,-96){ran}
\put(294,-8){(d)}
\put(299,-30){\small {\scriptsize 30}S}
\drawline(305,-34)(252,-44)
\put(272,-38){\scriptsize 31}
\put(242,-52){\small {\scriptsize 2}NP}
\drawline(252,-56)(217,-66)
\put(229,-60){\scriptsize 3}
\put(207,-74){\small {\scriptsize 1}DT}
\drawline(217,-78)(217,-88)
\put(210,-96){the}
\drawline(252,-56)(286,-66)
\put(267,-60){\scriptsize 19}
\put(280,-74){\small {\scriptsize 17}N}
\drawline(286,-78)(240,-88)
\put(259,-82){\scriptsize 18}
\put(233,-96){\small {\scriptsize 5}N}
\drawline(240,-100)(217,-110)
\put(228,-110){\scriptsize 6}
\put(210,-118){\small {\scriptsize 4}N}
\drawline(217,-122)(217,-132)
\put(208,-140){man}
\drawline(240,-100)(263,-110)
\put(250,-104){\scriptsize 10}
\put(254,-118){\small {\scriptsize 8}PP}
\drawline(263,-122)(244,-132)
\put(250,-127){\scriptsize 9}
\put(236,-140){\small {\scriptsize 7}IN}
\drawline(244,-144)(244,-154)
\put(240,-162){in}
\drawline(263,-122)(282,-132)
\put(272,-127){\scriptsize 14}
\put(272,-140){\small {\scriptsize 12}NP}
\drawline(282,-144)(269,-154)
\put(267,-150){\scriptsize 13}
\put(259,-162){\small {\scriptsize 11}DT}
\drawline(269,-166)(269,-176)
\put(262,-184){the}
\drawline(282,-144)(295,-154)
\put(289,-150){\scriptsize 16}
\put(288,-162){\small {\scriptsize 15}N}
\drawline(295,-166)(295,-176)
\put(285,-184){park}
\drawline(286,-78)(333,-88)
\put(312,-82){\scriptsize 23}
\put(324,-96){\small {\scriptsize 21}PP}
\drawline(333,-100)(311,-110)
\put(323,-110){\scriptsize 22}
\put(303,-118){\small {\scriptsize 20}IN}
\drawline(311,-122)(311,-132)
\put(301,-140){with}
\drawline(333,-100)(354,-110)
\put(345,-104){\scriptsize 27}
\put(345,-118){\small {\scriptsize 25}NP}
\drawline(354,-122)(341,-132)
\put(341,-127){\scriptsize 26}
\put(331,-140){\small {\scriptsize 24}DT}
\drawline(341,-144)(341,-154)
\put(334,-162){the}
\drawline(354,-122)(368,-132)
\put(360,-127){\scriptsize 29}
\put(361,-140){\small {\scriptsize 28}N}
\drawline(368,-144)(368,-154)
\put(360,-162){dog}
\drawline(305,-34)(358,-44)
\put(335,-36){\scriptsize 35}
\put(349,-52){\small {\scriptsize 33}VP}
\drawline(358,-56)(358,-66)
\put(360,-61){\scriptsize 34}
\put(352,-74){\small {\scriptsize 32}V}
\drawline(358,-78)(358,-88)
\put(351,-96){ran}
\end{picture}
\caption{Order of announce points for (a) top-down; (b) bottom-up; (c)
arc-standard left-corner; and (d) arc-eager left-corner parsing strategies}\label{fig:announce}
\end{figure}

Let us illustrate this graphically, by indicating the point at which
nodes and arcs in a particular tree would be identified under
different strategies.  Figure \ref{fig:announce} shows the sequence of
node and arc announcements for the same tree in four different parsing
strategies. 
There are a couple of key points that can be made from these detailed
diagrams.  First, consider the second word of the string,
\textttt{`man'}, which is the head noun of the noun phrase, with two
PP adjuncts.  To allow for the adjunction, three N constituents are
built in a left-recursive chain.  The top-down parsing strategy
(figure \ref{fig:announce}a) first announces the topmost N node, then
announces the nodes down this left-recursive chain until it reaches
the terminal item.  This search strategy is widely known to encounter
problems in dealing with left-recursive structures.  If the potential 
depth of this chain is unconstrained, the top-down parser can continue
to announce new N nodes as the left-child of the previously announced N
node, ad infinitum.  Enumeration of all of these alternative analyses
leads to non-termination\footnote{\namecite{Johnson95} points out
that, with appropriate memoization of what structures have been built,
a top-down parser can be built that will terminate in the face of
left-recursion.}.

The bottom-up strategy (figure \ref{fig:announce}b) waits until all of
the children constituents have been announced before building parent
constituents, so it does not have the problem with left-recursion
that a top-down parser does.  Neither do the two variants of
left-corner parsing, what \namecite{Abney91} term {\it arc-standard\/}
(figure \ref{fig:announce}c) and {\it arc-eager\/} (figure
\ref{fig:announce}d), although these avoid non-termination by first
recognizing just the left-child of each node before the node itself,
rather than all of the children, as in bottom-up parsing.  These two
left-corner strategies differ not in the order of node identification,
but in the order of arc identification.  The arc-standard algorithm
waits for non-leftmost children to be fully built before announcing
the arc between the parent and these children, whereas the arc-eager
announces these arcs as soon as the child node is announced.  One can
see this difference in the first PP adjunction.  In arc-standard
left-corner, the arc between the N and the PP is the 16th
announcement, after the entire PP has been built.  In contrast, the
arc-eager strategy announces the arc between the N and the PP
immediately after the PP is announced.

In terms of tree traversal, top-down is a {\it pre-order\/} traversal,
i.e. it visits the parent first, before the children.  Bottom-up is a
{\it post order\/} traversal, visiting the children before the
parent.  Left-corner parsing is an example of {\it in-order\/}
traversal, which goes from the leftmost child to the parent to the
remaining children.

Recall from the previous chapter the argument made by
\namecite{Shieber93} about the problems in top-down parsing with
respect to early commitment to certain structures.  An incremental
pure top-down parser has the same announce point for
both parent and rule (future children of the parent) -- in particular,
before any children have been built.  They used this criticism to
advocate a bottom-up parser, which announces both parent and rule
after all of the children are built.  We argued for a variant, which
handles their objection about early commitment, that has separate
announce points for the parent and for the arcs that collectively
constitute the rule.  This is the top-down parsing strategy that is
outlined in figure \ref{fig:announce}a, where the arc between the root
S node and the VP is only announced after the entire NP has been
built. As before, the
parent is announced before the children; but the specific rule is not
announced until after all of the children have been built.  For an
incremental parser, it is of critical importance to delay announce
points of rule expansions, to enable as much potentially
disambiguating information to enter either the left-context or the
look-ahead.  

To see this point, suppose that the category on the top of the stack
is an NP and there is a determiner (DT) in the look-ahead. In such a
situation, there is no information to distinguish between the rules 
NP~$\rightarrow$~DT~JJ~NN and NP~$\rightarrow$~DT~JJ~NNS. 
If the decision  can be delayed, however, until such a time as the 
relevant pre-terminal is in the look-ahead, the parser can make a more
informed decision. 

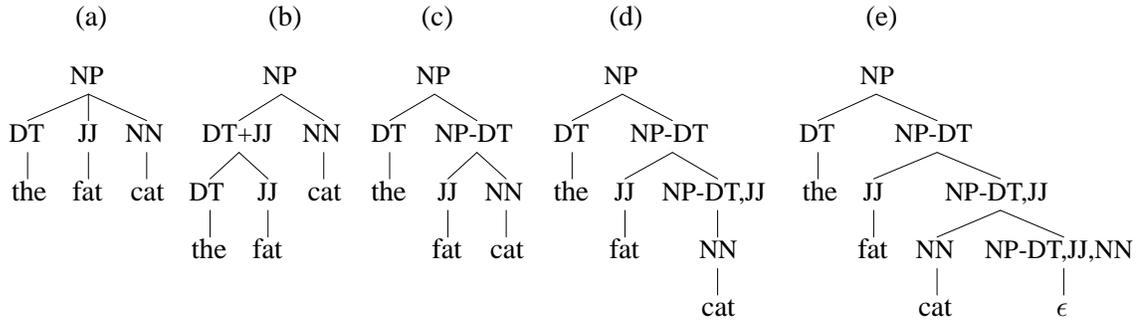
\begin{figure}
\begin{picture}(429,144)(0,-144)
\put(25,-8){(a)}
\put(23,-30){\small NP}
\drawline(30,-34)(7,-44)
\put(-0,-52){\small DT}
\drawline(7,-56)(7,-66)
\put(1,-74){the}
\drawline(30,-34)(30,-44)
\put(26,-52){\small JJ}
\drawline(30,-56)(30,-66)
\put(24,-74){fat}
\drawline(30,-34)(52,-44)
\put(44,-52){\small NN}
\drawline(52,-56)(52,-66)
\put(46,-74){cat}
\put(98,-8){(b)}
\put(96,-30){\small NP}
\drawline(103,-34)(87,-44)
\put(73,-52){\small DT+JJ}
\drawline(87,-56)(76,-66)
\put(68,-74){\small DT}
\drawline(76,-78)(76,-88)
\put(69,-96){the}
\drawline(87,-56)(98,-66)
\put(94,-74){\small JJ}
\drawline(98,-78)(98,-88)
\put(92,-96){fat}
\drawline(103,-34)(119,-44)
\put(111,-52){\small NN}
\drawline(119,-56)(119,-66)
\put(113,-74){cat}
\put(156,-8){(c)}
\put(154,-30){\small NP}
\drawline(161,-34)(144,-44)
\put(137,-52){\small DT}
\drawline(144,-56)(144,-66)
\put(137,-74){the}
\drawline(161,-34)(177,-44)
\put(161,-52){\small NP-DT}
\drawline(177,-56)(166,-66)
\put(162,-74){\small JJ}
\drawline(166,-78)(166,-88)
\put(160,-96){fat}
\drawline(177,-56)(188,-66)
\put(180,-74){\small NN}
\drawline(188,-78)(188,-88)
\put(182,-96){cat}
\put(227,-8){(d)}
\put(225,-30){\small NP}
\drawline(232,-34)(213,-44)
\put(206,-52){\small DT}
\drawline(213,-56)(213,-66)
\put(206,-74){the}
\drawline(232,-34)(251,-44)
\put(235,-52){\small NP-DT}
\drawline(251,-56)(233,-66)
\put(229,-74){\small JJ}
\drawline(233,-78)(233,-88)
\put(227,-96){fat}
\drawline(251,-56)(268,-66)
\put(247,-74){\small NP-DT,JJ}
\drawline(268,-78)(268,-88)
\put(261,-96){\small NN}
\drawline(268,-100)(268,-110)
\put(262,-118){cat}
\put(324,-8){(e)}
\put(322,-30){\small NP}
\drawline(328,-34)(306,-44)
\put(299,-52){\small DT}
\drawline(306,-56)(306,-66)
\put(300,-74){the}
\drawline(328,-34)(351,-44)
\put(335,-52){\small NP-DT}
\drawline(351,-56)(327,-66)
\put(323,-74){\small JJ}
\drawline(327,-78)(327,-88)
\put(321,-96){fat}
\drawline(351,-56)(375,-66)
\put(354,-74){\small NP-DT,JJ}
\drawline(375,-78)(351,-88)
\put(343,-96){\small NN}
\drawline(351,-100)(351,-110)
\put(344,-118){cat}
\drawline(375,-78)(399,-88)
\put(369,-96){\small NP-DT,JJ,NN}
\drawline(399,-100)(399,-110)
\put(396,-118){$\epsilon$}
\end{picture}
\caption{Factored trees:  (a) unfactored original; (b) right factored
({\small RF}); (c) left factored to binary ({\small LF2}); (d) left
factored to unary ({\small LF1}); (e) left factored to nullary
({\small LF0})}\label{fig:bin} 
\end{figure}

Grammar factorization is one way to do this, by allowing the parser to
use a rule like NP~$\rightarrow$~DT~NP-DT, where the new
non-terminal NP-DT can expand into anything that follows a DT in an
NP.  With a top-down parser, the expansion of NP-DT occurs only 
after the next pre-terminal is in the look-ahead.  We will first give
an informal intuition for the factorization via some examples, then
explicitly define it.  There are actually several ways to factor a
grammar, some of which are better than others for a top-down
search. The first distinction that can be 
drawn is between what we will call {\it left\/} factorization ({\small
LF}) versus {\it right\/} factorization ({\small RF}, see figure
\ref{fig:bin}). In the former, the rightmost items 
on the right-hand side of each rule are grouped together; in the
latter, the leftmost items on the right-hand side of the rule are
grouped together.  Within {\small LF} transforms, however, there is
some variation, with respect to how long rule underspecification is
maintained. One method 
is to have the final underspecified category rewrite as a binary rule
(hereafter {\small LF2}, see figure \ref{fig:bin}c). Another is to
have the final underspecified category rewrite as a unary rule
({\small LF1}, figure \ref{fig:bin}d). The last is to have the final
underspecified category rewrite as a nullary rule ({\small LF0},
figure \ref{fig:bin}e). 

We will show some trials demonstrating the effect of these different
factorizations on our parser, but we will ultimately settle on LF0,
which we formalize here -- the other LF factorizations are easy
modifications of this.  The {\em left-factorization transform} of a
CFG $G=(V,T,P,S)$ is the CFG $\LF(G) = (V_1, T, P_1, S)$, where:
\[
 V_1 \; = \; V \cup \{ \lc{D}{\beta} : D \in V, \beta \in (V \cup T)^+ \}
\]
and $P_1$ contains all instances of the schemata \ref{lf1}.  The
$\lc{D}{\beta}$ are new 
nonterminals; informally, they encode the left-hand side of a rule (D),
and the sequence of children categories ($\beta$) to the left in the
rule, so
   $\lc{D}{\beta} \mathop{\Rightarrow}_{\LF(G)}^* \gamma$
only if $D \mathop{\Rightarrow}_{G}^* \beta \gamma$.
\[ \refstepcounter{equation}\label{lf1}
\begin{array}{lll}
\nnxform{D \rewrites B\; \lc{D}{B}}{\mbox{where } D \rightarrow
B\,\gamma \in P}{lf1a} \\
\nnxform{\lc{D}{\beta} \rewrites B\; \lc{D}{\beta B}}{\mbox{where } D
\rightarrow \beta B\gamma \in P}{lf1b} \\
\nnxform{\lc{D}{\beta} \rewrites \epsilon}{\mbox{where } D
\rightarrow \beta \in P}{lf1c} \\
\nnxform{D \rewrites \epsilon}{\mbox{where } D
\rightarrow \epsilon \in P}{lf1d} \\

\end{array}
\]

This factorization results in all productions being binary, except
epsilon productions, even originally unary productions\footnote{In practice,
if the non-terminal set is split into disjoint sets of pre-terminals
and other non-terminals, such as in the Penn Treebank, the factored
pre-terminal productions are uniformly unary, so no factorization need
take place.}.  For a left-to-right, top-down parser, this
delays predictions about what non-terminals we expect 
later in the string until we have seen more of the string.  In effect,
this is an underspecification of some of the predictions that our
top-down parser is making about the rest of the string.  

This underspecification of the non-terminal predictions (e.g. NP-DT in
the example in figure \ref{fig:bin}, as opposed to JJ) allows
constituents to become part of the left-context before their siblings
are announced.  By virtue of being in the left-context of a specific
top-down derivation, conditioning information, such as lexical heads,
can be extracted for use in the conditional probability distribution.
For example, inside of a VP constituent, once the head verb has been
found, the probability of other children within the VP constituent can
be conditioned on that verb.  This would provide a specific verb's
subcategorization preferences.  In addition, this underspecification
means that words further downstream will be in look-ahead at the
announce point of later children.  For example, suppose that the
grammar allows
NP modification with either relative clauses or prepositional
phrases.  By underspecifying further children of the NP, the decision
about what category is modifying the NP is delayed until the look-ahead word
is the word after the head noun.  If the next word is a
preposition, then the modification is likely to be a prepositional
phrase, hence we obtain additional guidance by delaying this decision.

These transforms have a couple of very nice properties.  First, they are
easily reversible, i.e. every parse tree built with $\LF(G)$ corresponds
to a unique parse tree built with $G$.  Second, if we use the relative
frequency estimator for our production probabilities, the probability
of a tree built with $\LF(G)$ is identical to the probability of the
corresponding tree built with $G$.  

Left-corner ({\small LC}) parsing \cite{Rosenkrantz70} is a
well-known strategy that uses both bottom-up evidence (from the left
corner of a rule) and top-down prediction (of the rest of the
rule).  To do this, one makes a distinction between (top-down) predicted
categories and (bottom-up) found categories.  Three actions can be
followed by the 
parser: (i) shift: put the next word of the string onto the top of the
stack; (ii) predict: if the topmost item on the stack is the first
category on the right-hand side of a rule in the grammar, pop it, and
push the remaining right-hand side categories of the rule onto the
stack, marking them as {\it predicted\/}, followed by the parent
category of the rule; and (iii) attach: if a category is found on the
top of the stack followed by an identical category that is marked {\it
predicted\/}, then both can be popped from the stack.  
Useless non-terminals, which can never match a predicted category, can
be filtered out by building a left-corner table, and
checking to make sure that prediction only builds categories that can
occur at the left-corner of a predicted category or attach to it.

\namecite{Demers77} defined {\it generalized\/} left-corner parsing
(GLC), in which prediction occurs, not necessarily after the first 
category is found on the right-hand side, but after some pre-specified
number of categories.  Thus standard left-corner parsing is an
instance of GLC, where all prediction takes place after the first
child (after 1 category); and top-down parsing is also an instance of
GLC, where prediction takes place before the first child (after 0
categories).  Note that one can also vary the announce points for arcs
as well as nodes, and that the announce points could vary from rule to
rule.

\namecite{Rosenkrantz70} showed how to transform a context-free
grammar into a grammar that, when used by a top-down parser, announces
nodes in the same order as an {\small LC} parser would with the
original grammar.  As mentioned earlier, left-corner parsing
has been advocated by virtue of the fact that it does not face the
same non-termination problem with left-recursive grammars that
top-down parsing does.  It has also been advocated because eager
attachment left-corner parsing places psychologically plausible
demands on memory \cite{Abney91,Resnik92}, since the stack only grows
when there are center-embedded structures.  We will investigate
left-corner parsing by performing the $\LC$ grammar transform, then
using the transformed grammar with 
the same top-down parser as for other trials.  This provides a way of
comparing the two approaches, without worrying about the impact of
differences in implementation.

The {\em left-corner transform} of a CFG $G=(V,T,P,S)$ 
is the CFG $\LC(G) = (V_1, T, P_1, S)$, where:
\[
 V_1 \; = \; V \cup \{ \lc{D}{X} : D \in V, X \in V \cup T \}
\]
and $P_1$ contains all instances of the schemata \ref{lc1}.  In these
schemata, $D \in V$, $w \in T$, and lower case
Greek letters range over $(V \cup T)^*$.  The $\lc{D}{X}$ are new
nonterminals; informally they encode a parse state in which a $D$ is
predicted top-down and an $X$ has been found left-corner, so
   $\lc{D}{X} \mathop{\Rightarrow}_{\LC(G)}^* \gamma$
only if $D \mathop{\Rightarrow}_{G}^* X \gamma$.
\[ \refstepcounter{equation}\label{lc1}
\begin{array}{lll}
\nnxform{D \rewrites w \; \lc{D}{w}}{}{lc1a} \\
\nnxform{\lc{D}{B} \rewrites \beta \; \lc{D}{C}}{\mbox{where } C \rightarrow B\,\beta \in L}{lc1b} \\
\nnxform{\lc{D}{D} \rewrites \epsilon}{}{lc1c}
\end{array}
\]

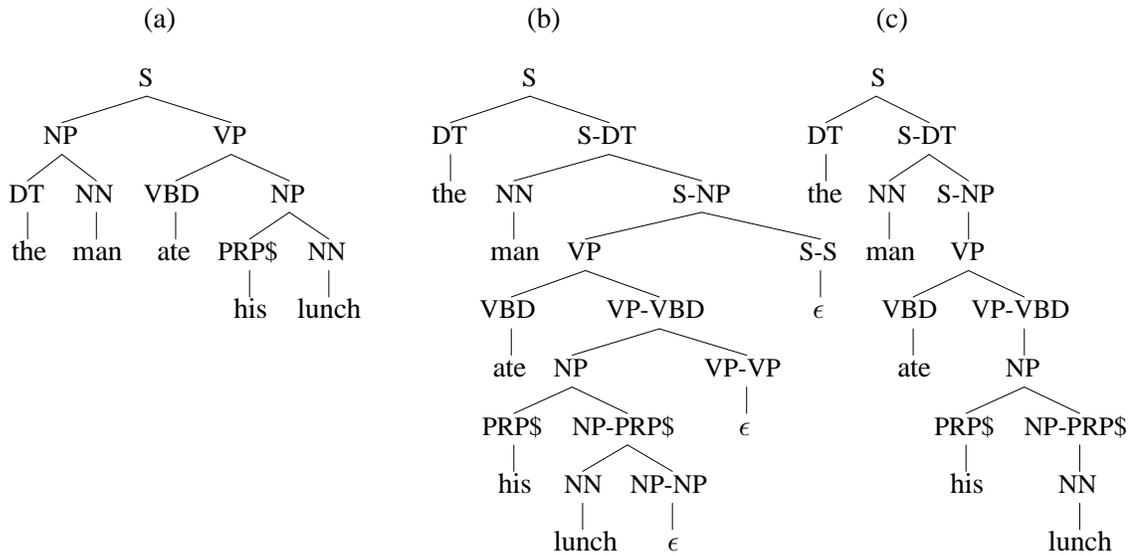
\begin{figure}
\begin{picture}(314,232)(0,-232)
\put(51,-8){(a)}
\put(49,-30){\small S}
\drawline(52,-34)(20,-44)
\put(13,-52){\small NP}
\drawline(20,-56)(7,-66)
\put(-0,-74){\small DT}
\drawline(7,-78)(7,-88)
\put(1,-96){the}
\drawline(20,-56)(33,-66)
\put(25,-74){\small NN}
\drawline(33,-78)(33,-88)
\put(24,-96){man}
\drawline(52,-34)(84,-44)
\put(77,-52){\small VP}
\drawline(84,-56)(63,-66)
\put(51,-74){\small VBD}
\drawline(63,-78)(63,-88)
\put(56,-96){ate}
\drawline(84,-56)(106,-66)
\put(99,-74){\small NP}
\drawline(106,-78)(91,-88)
\put(79,-96){\small PRP\$}
\drawline(91,-100)(91,-110)
\put(85,-118){his}
\drawline(106,-78)(121,-88)
\put(113,-96){\small NN}
\drawline(121,-100)(121,-110)
\put(109,-118){lunch}
\put(196,-8){(b)}
\put(194,-30){\small S}
\drawline(197,-34)(167,-44)
\put(160,-52){\small DT}
\drawline(167,-56)(167,-66)
\put(160,-74){the}
\drawline(197,-34)(227,-44)
\put(215,-52){\small S-DT}
\drawline(227,-56)(191,-66)
\put(184,-74){\small NN}
\drawline(191,-78)(191,-88)
\put(182,-96){man}
\drawline(227,-56)(262,-66)
\put(251,-74){\small S-NP}
\drawline(262,-78)(218,-88)
\put(211,-96){\small VP}
\drawline(218,-100)(190,-110)
\put(178,-118){\small VBD}
\drawline(190,-122)(190,-132)
\put(183,-140){ate}
\drawline(218,-100)(246,-110)
\put(226,-118){\small VP-VBD}
\drawline(246,-122)(213,-132)
\put(206,-140){\small NP}
\drawline(213,-144)(191,-154)
\put(179,-162){\small PRP\$}
\drawline(191,-166)(191,-176)
\put(185,-184){his}
\drawline(213,-144)(234,-154)
\put(213,-162){\small NP-PRP\$}
\drawline(234,-166)(217,-176)
\put(210,-184){\small NN}
\drawline(217,-188)(217,-198)
\put(206,-206){lunch}
\drawline(234,-166)(251,-176)
\put(235,-184){\small NP-NP}
\drawline(251,-188)(251,-198)
\put(249,-206){$\epsilon$}
\drawline(246,-122)(279,-132)
\put(263,-140){\small VP-VP}
\drawline(279,-144)(279,-154)
\put(276,-162){$\epsilon$}
\drawline(262,-78)(307,-88)
\put(299,-96){\small S-S}
\drawline(307,-100)(307,-110)
\put(304,-118){$\epsilon$}
\put(328,-8){(c)}
\put(326,-30){\small S}
\drawline(328,-34)(309,-44)
\put(302,-52){\small DT}
\drawline(309,-56)(309,-66)
\put(302,-74){the}
\drawline(328,-34)(348,-44)
\put(336,-52){\small S-DT}
\drawline(348,-56)(333,-66)
\put(325,-74){\small NN}
\drawline(333,-78)(333,-88)
\put(324,-96){man}
\drawline(348,-56)(363,-66)
\put(351,-74){\small S-NP}
\drawline(363,-78)(363,-88)
\put(356,-96){\small VP}
\drawline(363,-100)(342,-110)
\put(330,-118){\small VBD}
\drawline(342,-122)(342,-132)
\put(336,-140){ate}
\drawline(363,-100)(384,-110)
\put(364,-118){\small VP-VBD}
\drawline(384,-122)(384,-132)
\put(377,-140){\small NP}
\drawline(384,-144)(362,-154)
\put(350,-162){\small PRP\$}
\drawline(362,-166)(362,-176)
\put(356,-184){his}
\drawline(384,-144)(405,-154)
\put(384,-162){\small NP-PRP\$}
\drawline(405,-166)(405,-176)
\put(397,-184){\small NN}
\drawline(405,-188)(405,-198)
\put(393,-206){lunch}
\end{picture}
\caption{The same structure with (a) the original grammar; (b) the
left-corner transformed grammar; and (c) the left-corner transformed
grammar with $\epsilon$-removal} \label{fig:lctree}
\end{figure}

This transform converts left-recursion to right-recursion, which is
not a problem for top-down parsers \cite{Johnson98a}.
The effect of the transform can be seen in figure \ref{fig:lctree}.
The transformed trees have the same root, but the topmost production
jumps immediately to the left-corner terminal item of the original
tree.  Once a left-corner is recognized, the next production predicts
the rest of the children of the original production and recognizes the
parent.  The epsilon productions in \ref{fig:lctree}(b) represent an
attachment of the predicted category and the recognized category.
Eager attachment can be effected by removing these
$\epsilon$-productions, i.e. composing \ref{lc1b} and \ref{lc1c},
which would leave \lc{D}{D} categories only in the 
case of a $D$ constituent being at the left-corner of another $D$
constituent.  This is eager attachment because these epsilon
productions correspond to 
the {\it attach\/} move of the left-corner production; by composing
this with the rule in which the \lc{D}{D} occurs on the right-hand
side, the parser must make the decision about attachment earlier.
Full $\epsilon$-removal yields the grammar given by the schemata below. 
\[ \refstepcounter{equation}\label{slc1eps}
\begin{array}{lll}
% \nnxform{A \rewrites \alpha}{\mbox{where } A \rightarrow \alpha \in P - L}{slc1a} \\
\nnxform{D \rewrites w \; \lc{D}{w}}{}{slc1epsa1} \\
\nnxform{D \rewrites w}{\mbox{where } D \Rightarrow_L^{+} w}{slc1epsa2} \\
\nnxform{\lc{D}{B} \rewrites \beta \; \lc{D}{C}}{\mbox{where } C \rightarrow B\,\beta \in L}{slc1c1} \\
\nnxform{\lc{D}{B} \rewrites \beta}{\mbox{where } D \Rightarrow_L^{\star} C, C \rightarrow B\,\beta \in L}{slc1c2}
\end{array}
\]
This transform results in trees like that in figure
\ref{fig:lctree}(c).  

A variant of the standard left-corner transform that will be explored
in later chapters is the selective left-corner transform
\cite{Johnson00}.  In such a transform, some productions, but not
necessarily all, are recognized left-corner, while the rest are
recognized top-down.  Such a transform could be used to eliminate all,
or the most likely, left-recursive structures from the grammar.
The selective left-corner transform takes as input a CFG $G=
(V,T,P,S)$ and a set of {\em left-corner productions} $L \subseteq P$,
which contains no epsilon productions; the non-left-corner productions
$P-L$ are called {\em top-down productions}.  The {\em standard
left-corner transform} is obtained by setting $L$ to the set of all
non-epsilon productions in $P$.  
The {\em selective left-corner transform} of $G$ with respect
to $L$ is the CFG $\LC_L(G) = (V_1, T, P_1, S)$, where, again:
\[
 V_1 \; = \; V \cup \{ \lc{D}{X} : D \in V, X \in V \cup T \}
\]
and $P_1$ contains all instances of the schemata \ref{slc1}.
\[ \refstepcounter{equation}\label{slc1}
\begin{array}{lll}
\nnxform{D \rewrites w \; \lc{D}{w}}{}{slc1a} \\
\nnxform{D \rewrites \alpha \; \lc{D}{A}}{\mbox{where } A \rightarrow \alpha \in P - L}{slc1b} \\
\nnxform{\lc{D}{B} \rewrites \beta \; \lc{D}{C}}{\mbox{where } C \rightarrow B\,\beta \in L}{slc1c} \\
\nnxform{\lc{D}{D} \rewrites \epsilon}{}{slc1d}
\end{array}
\]
The schemata function as follows.  The productions introduced by
schema \ref{slc1a} start a left-corner parse of a predicted
nonterminal $D$ with its leftmost terminal $w$, while those introduced
by schema \ref{slc1b} start a left-corner parse of $D$ with a
left-corner $A$, which is itself found by the top-down recognition of
production $A~\rightarrow~\alpha~\in~P-L$.  Schema \ref{slc1c} extends
the current left-corner $B$ up to a $C$ with the left-corner
recognition of production $C~\rightarrow~B\,\beta$.  Finally,
schema \ref{slc1d} matches the top-down prediction with the recognized
left-corner category.

An LC grammar can also benefit from factorization.  We use transform
composition to apply first one transform, then another to the output
of the first. We denote this {\small A} $\circ$ {\small B} where
({\small A} $\circ$ {\small B})(t) = {\small B} ({\small A}
(t)). After applying the left-corner transform, we then factor the
resulting grammar, i.e. {\small LC} $\circ$ {\small LF}.  If we have
$\epsilon$-productions in the 
left-corner grammar, the use of LF0 is not needed, i.e. nullary
productions need only be introduced from one source.  Thus
with standard left-corner, left factorization is always to unary (LF1),
while $\epsilon$ removed left-corner grammars are typically factored
to nullary (LF0).

\begin{figure}
\begin{center}
\begin{picture}(328,78)(0,-78)
\put(160,-8){A}
\drawline(164,-12)(29,-22)
\put(25,-30){B$_0$}
\drawline(29,-34)(12,-44)
\put(0,-52){w$_0$}
\drawline(29,-34)(46,-44)
\put(34,-52){\ldots w$_j$}
\drawline(164,-12)(96,-22)
\put(93,-30){B$_1$}
\drawline(96,-34)(79,-44)
\put(68,-52){w$_{j+1}$}
\drawline(96,-34)(113,-44)
\put(101,-52){\ldots w$_k$}
\put(160,-30){\ldots}
\put(160,-52){\ldots}
\drawline(164,-12)(231,-22)
\put(228,-30){B$_{i-1}$}
\drawline(231,-34)(214,-44)
\put(203,-52){w$_l$}
\drawline(231,-34)(248,-44)
\put(236,-52){\ldots w$_m$}
\drawline(164,-12)(299,-22)
\put(295,-30){B$_i$}
\drawline(299,-34)(282,-44)
\put(270,-52){w$_{m+1}$}
\drawline(299,-34)(316,-44)
\put(304,-52){\ldots w$_n$}
\end{picture}
\begin{tabular}{|l|l|l||l|l|}
\hline
Transform & \multicolumn{2}{|c||}{Announce A} & \multicolumn{2}{|c|}{Announce A
$\rightarrow$ B$_0$ \ldots B$_i$}\\\hline
None & before B$_0$ & w$_0$ in look-ahead & before B$_0$ & w$_0$ in look-ahead\\\hline
LF1 & before B$_0$ & w$_0$ in look-ahead & before B$_i$ & w$_{m+1}$ in look-ahead\\\hline
LF0 & before B$_0$ & w$_0$ in look-ahead & after B$_i$ & w$_{n+1}$ in look-ahead\\\hline
LC & after B$_0$ & w$_{j+1}$ in look-ahead & after B$_0$ & w$_{j+1}$ in look-ahead\\\hline
LC $\circ$ LF & after B$_0$ & w$_{j+1}$ in look-ahead & after B$_i$ & w$_{n+1}$ in look-ahead\\\hline
\end{tabular}
\caption{Announce points for different transformations of the
context-free grammar, when used with a top-down parser}\label{fig:ann}
\end{center}
\end{figure}
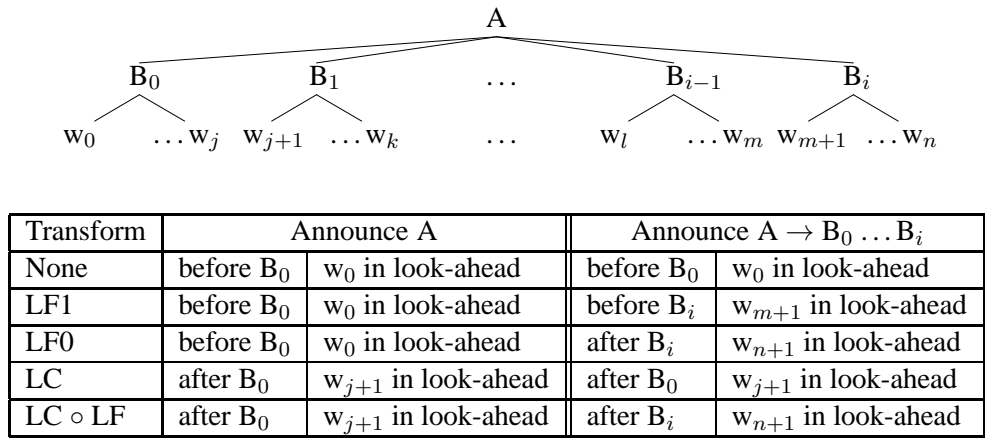

Figure \ref{fig:ann} shows the effect of various grammar transforms on
the announce points when used by an incremental top-down parser.
These announce points are given, not simply in relation to the
immediate children of the constituent, but also in relation to the
word in the look-ahead.  The guidance provided by the look-ahead --
not to mention the lexical items incorporated into the left-context --
can make a large difference in the efficiency with which plausible
alternatives are identified, as was illustrated above with the base NP
expansion via NP~$\rightarrow$~DT~JJ~NN versus NP~$\rightarrow$~DT~JJ~NNS.  

There are a couple of ways that the grammar transforms can be used in
treebank parsing.  Because the grammar is induced from a corpus of
trees, the transform can be performed either before or after the
grammar induction.  In other words, the trees in the corpus can be
transformed using the grammar transform, and this new, transformed
corpus can be used for grammar induction; or the grammar can be
induced, and the transform applied to the induced grammar.  Some of
the transforms -- such as left and right factorization -- can be
applied either way, and the resulting grammar is the same.
Transforming the corpus may be slightly easier in this case, since the
rule probabilities can be estimated by the same relative frequency
estimation technique used with the original grammar, whereas the
grammar transform, performed after rule induction, would have to
explicitly gather probability mass for each new transformed rule.
\namecite{Abney99} describes how to compute the transformed rule
probabilities.  See \namecite{Johnson98b} for details of the
transform/de-transform paradigm.

For the left-corner transform, the grammars produced by the two
methods are, in fact, different.  Performing the grammar transform
before rule induction results in fewer rules than if the
transform is applied to the grammar after induction, sometimes even an
order of magnitude fewer.  This is because the grammar transform
produces all possible left-corner rules, whereas the tree transform
produces only the observed left-corner rule instances.  This has the
benefit of introducing more contextual guidance to the grammar (see
discussion below about the differences in performance), but the
disadvantage of reducing coverage.  See \namecite{Johnson00} for more
discussion of the differences between these two methods of inducing a
left-corner grammar.  Unless otherwise specified, all transformations
are performed on the corpus of trees, prior to grammar induction.

\section{Top-down probabilistic parsing}
This parser is essentially a stochastic version of the top-down parser 
described in \namecite{Aho86}.  To present the parser, we will first
present their deterministic algorithm, then discuss how to handle the
non-determinism.  The parser will be presented as taking strings of
words as input, but it can also be applied to strings of POS tags,
with the obvious changes to look-ahead calculations.

Deterministic top-down parsing (see the algorithm\footnote{Algorithms
in this thesis are formatted according to the style in
\namecite{Corman90}.  Although we do not, in general, refer to the
line numbers, they are provided in accordance with this standard
style.} in figure 
\ref{fig:algasu}) is effected via a parsing table M$_G$,
which takes a non-terminal category $X$ and the look-ahead word (the next
word in the string) $w_i$, and returns either a rule expanding the
non-terminal or a fail symbol.  Consider the set of productions from a
very simple context-free grammar\footnote{We consider rules expanding 
pre-terminals to terminals as rules in the grammar, rather than a
lexicon separate from the productions.}:
\begin{examples}
\item S $\rightarrow$ NP VP\label{rule:S}
\item NP $\rightarrow$ DT NN\label{rule:NP1}
\item VP $\rightarrow$ V NP
\item DT $\rightarrow$ \textttt{the}
\item NN $\rightarrow$ \textttt{moon}
\item NN $\rightarrow$ \textttt{sun}
\item V  $\rightarrow$ \textttt{is}
\end{examples}
This grammar is very limited, but can handle strings like
\textttt{`the moon is the moon'}. The parsing table will have an entry
for the pair M$_G$(S,\textttt{the}) = S~$\rightarrow$~NP~VP, since 
rule \ref{rule:S} is the only possible path, given the above grammar,
from S to \textttt{the}.  Given a 
grammar G, a parsing table M$_G$ can be built, by finding,
for every non-terminal/terminal pair, all rules that can be the first
step in a path
from the non-terminal to the terminal.  If a parsing table can be
built where each entry in the table is unique, i.e. in which there is
no ambiguity about which rule to apply with any pair (such as the
above grammar), then the grammar
is said to be LL(1), where LL stands for left-to-right and leftmost,
and the 1 means that there is one terminal item in look-ahead.  With
such a parsing table, one can deterministically parse the input
top-down, with the algorithm in figure \ref{fig:algasu}.

\begin{figure}
\begin{algorithm}{Top-down-parser}{S^\dag,M_G,w = w_0\ldots w_n\langle /s\rangle}
i \= 0\\
\scSx = S^\dag \$ \hspace*{.25in}\COMMENT{Let $\mathcal{S}$ be the stack, and \$ the end-of-stack marker}\\
\begin{REPEAT}
\COMMENT{let {\it X\/} be the top stack symbol on $\mathcal{S}$}\\
\CALL{pop} \ X \text{from} \scSx\\
\begin{IF}{X \in T}
\begin{IF}{X = w_i}
i \= i+1
\ELSE \CALL{error}
\end{IF}
\ELSE
\begin{IF}{M_G[X,w_i] = X \rightarrow Y_1 \ldots Y_k}
\CALL{push} \ Y_1 \ldots Y_k \text{onto} \scSx\\
\CALL{output}(X \rightarrow Y_1 \ldots Y_k)
\ELSE \CALL{error}
\end{IF}
\end{IF}
\end{REPEAT}X = \$\\
\begin{IF}{w_i \neq \langle /s\rangle\hspace*{.2in}\COMMENT{if
look-ahead is not the end-of-string}} \CALL{error}
\end{IF}
\end{algorithm}
\caption{A deterministic top-down parsing algorithm, modified from
Aho, Sethi, and Ullman (algorithm 4.3), taking a start symbol
$S^\dag$, a parsing table M$_G$, and an input string $w$ as 
arguments.  The symbol $\triangleright$ precedes comments.}\label{fig:algasu} 
\end{figure}

Suppose that we were to enrich our small toy grammar with a couple of
rules to handle NP modification with prepositional phrases, to be able
to handle strings like \textttt{`the moon is the sun of the night'}.
The rules might look something like:
\begin{examples}
\item NP $\rightarrow$ NP PP\label{rule:NP2}
\item PP $\rightarrow$ IN NP
\item IN $\rightarrow$ \textttt{of}
\item NN $\rightarrow$ \textttt{night}
\end{examples}
With the introduction of these rules, the grammar is no longer LL(1),
because for certain non-terminal/terminal pairs, there is more than
one rule in the table.  For example, in this case
M$_G$(NP,\textttt{the}) has two entries: rules \ref{rule:NP1} and
\ref{rule:NP2}.  Grammars sufficient to cover freely occurring
strings of English, as mentioned in earlier chapters, are typically
massively ambiguous, so any top-down approach must be able to handle
such non-determinism.  

\begin{figure}[t]
\begin{algorithm}{above-threshold}{C = (D,\scS,P_D,F,w_i^n\langle /s\rangle),\scH_{i+1},\gamma,f}
\scH_{i+1}[0] =
(D^\prime,\scS^\prime,P_{D^\prime},F^\prime,w_{i+1}^n\langle
/s\rangle)\hspace*{.25in}\COMMENT{Heap provides the best scoring
entry}\\
\begin{IF}{F > P_{D^\prime}*f(\gamma,|\scH_{i+1}|)} \RETURN \CALL{TRUE}
\ELSE \RETURN \CALL{FALSE}
\end{IF}
\end{algorithm}
\begin{algorithm}{ND-Top-down-parser}{G=(V,T,P,S^\dag),\Rightarrow,w = w_0\ldots w_n\langle /s\rangle,\gamma,f}
i \= 0\\
\scH_i[0] \= (\langle\rangle,S^\dag\$,1,1,w_0^n\langle /s\rangle)
\hspace*{.25in}\COMMENT{Let $\mathcal{H}_i$ be the priority queue
for $w_i$}\\
\begin{FOR}{i \= 0 \TO n}
\begin{WHILE}{\CALL{above-threshold}(\scH_i[0],\scH_{i+1},\gamma,f)}
C \= \scH_i[0] = (D,\scS,P_D,F,w_i^n\langle /s\rangle)\\
\CALL{pop} \ C \text{from} \scH_i\\
\COMMENT{let {\it X\/} be the top stack symbol on $\mathcal{S}$}\\
\begin{IF}{X \in T}
\forall C^\prime \ \text{such that} C \Rightarrow C^\prime:\ 
\CALL{push} \ C^\prime \text{onto} \scH_{i+1}
\ELSE
\forall C^\prime \ \text{such that} C \Rightarrow C^\prime:\ 
\CALL{push} \ C^\prime \text{onto} \scH_i\\
\end{IF}
\end{WHILE}
\end{FOR}
\hspace*{-.9in}\COMMENT{At the end of the string, we must empty the stack to complete
the derivation}\\
\begin{WHILE}{\CALL{above-threshold}(\scH_{n+1}[0],\scH_{n+2},\gamma,f)}
C \= \scH_{n+1}[0] = (D,\scS,P_D,F,\langle /s\rangle)\\
\CALL{pop} \ C \text{from} \scH_{n+1}\\
\COMMENT{let {\it X\/} be the top stack symbol on $\mathcal{S}$}\\
\begin{IF}{X = \$}
\CALL{push} \ C \text{onto} \scH_{n+2}
\ELSE
\forall C^\prime \ \text{such that} C \Rightarrow C^\prime:\ 
\CALL{push} \ C^\prime \text{onto} \scH_{n+1}
\end{IF}
\end{WHILE}\\
\begin{IF}{\CALL{empty}(\scH_{n+2}[0])\hspace*{.2in}\COMMENT{if
no analysis made the final heap}} \CALL{error}
\end{IF}
\end{algorithm}
\caption{A non-deterministic top-down parsing algorithm, taking a
context-free grammar G, a derives relation $\Rightarrow$, an input
string $w$, a base beam-factor $\gamma$, and a threshold function $f$ as
arguments. The symbol $\Rightarrow$ denotes our derives relation
defined on page 72. The symbol $\triangleright$ precedes
comments.}\label{fig:algbeam}
\end{figure}

Our basic approach will be to
keep many separate derivations, each of which follows a search path
akin to the deterministic parser just outlined.  This will involve
assigning each partial derivation a {\it figure-of-merit\/}, or a
score of how good the derivation is.  The goal is to work on just the
promising ones, and discard the rest. Finding an appropriate
figure-of-merit can be difficult, because of issues of comparability.
Two 
competing analyses may be at different points in their derivation, so
this figure-of-merit must be able to, in a sense, normalize the scores
with respect to the extent of the derivation.  This can be done by
including in the score the probability of the derivation to that
point, as well as some estimate of how much probability the
analysis is going to spend to extend the derivation.  We do not
compare derivations with different terminal yields, but rather extend
the set of competing derivations to the current word before moving on
to the next word; hence the derivations are more comparable than they
might otherwise be. Each derivation
probability is monotonically decreasing, i.e. every rule added to the
derivation decreases its probability; yet each rule also brings the
existing derivation closer to the look-ahead word, so that the amount
of probability that will have to be spent, for promising
analyses, to reach the look-ahead word will offset the drop in
probability.  Thus attention is appropriately focused on these
promising derivations.

To introduce our parsing algorithm, 
we will first define {\it candidate analysis\/} (i.e. a partial parse),
and then a {\it derives\/} relation between candidate analyses.  We
will then present the algorithm in terms of this relation.

The input to the parser is a string $w_0^n$ and a PCFG $G$.  The
parser's basic data structure is a priority queue of
candidate analyses.  A candidate analysis $C = (D,\scS,P_D,F,w_i^n)$
consists of a partial derivation $D$, a stack $\scS$, a derivation
probability $P_D$, a 
figure-of-merit $F$, and a string $w_i^n$ remaining to be parsed.  The
first word in the string remaining to be parsed, $w_i$, we will call the
{\it look-ahead\/} word. 
The derivation $D$ consists of a sequence of rules used from $G$.
The stack $\scS$ contains a sequence of non-terminal symbols that need
to be accounted for, and an
end-of-stack marker \$ at the bottom.  The probability
$P_D$ is the product of 
the probabilities of all rules in the derivation $D$.
$F$ is the product of $P_D$ and a look-ahead probability,
LAP($\scS$,$w_i$), which is a measure of the likelihood
of the stack $\scS$ rewriting with $w_i$ at its left corner.  Exactly
how the LAP is calculated is described on page 73.

We can define a {\it derives\/} relation, denoted $\Rightarrow$,
between two candidate analyses as follows. $(D,\scS,P_D,F,w_i^n)
\Rightarrow
(D^\prime,\scS^\prime,P_{D^\prime},F^\prime,w_j^n)$ if
and only if\footnote{The + in (i) denotes
concatenation. To avoid confusion between sets and sequences,
$\emptyset$ will not be used for empty strings or sequences, rather
the symbol $\langle\rangle$ will be used. Note that the script $\scS$
is used to denote stacks, while $S^{\dag}$ is the start symbol.} 
\pagebreak
\newcounter{yylist}
\begin{list}{\roman{yylist}.}{\setlength{\itemsep}{0in} \usecounter{yylist}}
\item $D^\prime = D\ +\ A \rightarrow \beta$
\item $\scS$ = $A\alpha$\$;
\item either $\scS^\prime = \beta\alpha$\$ and $j$ = $i$\\
or $\beta = w_i$, $j$ = $i$+1, and $\scS^\prime = \alpha$\$;
\item $P_{D^\prime} = P_D\Pr(A \rightarrow
\beta)$; and
\item $F^\prime = P_{D^\prime}\mathrm{LAP}(\scS^\prime,w_j)$
\end{list}

The parse begins with a single candidate analysis on the priority queue:
($\langle\rangle$,$S^\dag$\$,1,1,$w_0^n$).  It then proceeds as
follows.  The top ranked candidate 
analysis, $C = (D,\scS,P_D,F,w_i^n)$, is popped from the priority queue.
If $\scS$ = \$ and $w_i$ = $\langle$/s$\rangle$, then the analysis is
complete. Otherwise, all $C^\prime$ such that $C \Rightarrow C^\prime$
are pushed onto the priority queue.

We implement this as a beam search.  For each word position $i$, we have 
a separate priority queue, $\scH_i$, of analyses with look-ahead $w_i$. When there are
``enough'' analyses by some criteria (which we will discuss below) on
priority queue $\scH_{i+1}$, all candidate analyses remaining on $\scH_{i}$ are
discarded.  Since $w_n$ = $\langle$/s$\rangle$, all parses that are pushed onto
$\scH_{n+1}$ are complete.  The parse on $\scH_{n+1}$ with the
highest probability is returned for evaluation.  In the case that no
complete parse is found, a partial parse is returned and evaluated.
Figure \ref{fig:algbeam} presents the algorithm formally.

The LAP is the probability of a particular terminal being the 
next left-corner of a particular analysis.  The terminal may be the
left-corner of the top-most non-terminal on the stack of the analysis
or it might be 
the left-corner of the {\it nth} non-terminal, after the top
$n$--1 non-terminals have rewritten to $\epsilon$. Of course, we
cannot expect to have adequate statistics for each non-terminal/word
pair that we encounter, so we smooth to the POS.  Since we do
not know the POS 
for the word, we must sum the LAP for all POS
labels\footnote{Equivalently, we can split the analyses at this point,
so that there is one POS per analysis.  If the POS label is given by
the input string, then, obviously, this does not need to occur.}.  

For a PCFG $G$, a stack $\scS = A_{0} \dots A_{n}$\$ (which we will write
$A_0^n$\$) and a look-ahead terminal item $w_i$, we define the look-ahead
probability as follows: 
\begin{equation}
\mathrm{LAP}(\scS,w_i) = \sum_{\alpha \in (V \cup T)^{*}} \PrG(A_0^n
\stackrel{\ast}{\Rightarrow} w_i\alpha)
\end{equation}  
We recursively estimate this with two empirically observed conditional
probabilities for every non-terminal $A_{i}$:
$\Prhat (A_{i} \stackrel{\ast}{\Rightarrow} w_i\alpha)$
and $\Prhat (A_{i} \stackrel{\ast}{\Rightarrow} \epsilon)$.
The same empirical probability, $\Prhat (A_{i} \stackrel{\ast}{\Rightarrow} X\alpha)$, is collected for every
pre-terminal $X$ as well. The LAP approximation for a given
stack state and look-ahead terminal is:
\begin{eqnarray}
\PrG(A_j^n \stackrel{\ast}{\Rightarrow} w_i\alpha) 
& \approx &  \PrG(A_{j}\stackrel{\ast}{\Rightarrow} w_i\alpha) + \Prhat (A_{j} \stackrel{\ast}{\Rightarrow} \epsilon)
\PrG(A_{j+1}^n \stackrel{\ast}{\Rightarrow}
w_i\alpha)  
\end{eqnarray}
where
\begin{equation}
\PrG(A_{j} \stackrel{\ast}{\Rightarrow} w_i\alpha) 
\approx  \lambda_{A_{j}} \Prhat (A_{j}
\stackrel{\ast}{\Rightarrow} w_i\alpha) + (1-\lambda_{A_{j}}) \sum_{X \in V}  \Prhat (A_{j}
\stackrel{\ast}{\Rightarrow} X\alpha) \Prhat (X
\rightarrow w_i) 
\end{equation}
The $\lambda_{A_{j}}$ mixing coefficients for interpolation are a
function of the frequency of the non-terminal $A_{j}$, and are
estimated in the standard way using held-out training data
\cite{Jelinek80}.

We have identified three pieces of information that are potentially
useful in deciding when ``enough'' parses have been collected for the
substring $w_0^i$ -- in other words, that it is likely that the parse
which will ultimately have the highest score is in the set already
collected:
(i) the number of analyses that have successfully reached
$w_i$; (ii) the number of analyses that have been pushed back on the
heap without having reached $w_i$; and (iii) the highest probability
from among the analyses that have reached $w_i$, which can be used to
define a probability range as a beam threshold.  A couple of
considerations are relevant when considering which of these scores to
use for beam thresholding.  First, if (ii) is ignored, there is no
assurance of termination with a left-recursive grammar, since it is
possible that no analysis ever reaches $w_i$.   Second, the density of
competing analyses within a fixed probability range can vary
dramatically depending on the syntactic context.  If the threshold is
defined by a target number for factor (i) above, and the density is
very low, the parser could spend a lot of time searching for parses
with an extremely low probability in an attempt to find enough of them
to fill the beam.  If the threshold is defined by a target range based
on factor (iii) above, and the density is very high, the parser could
spend a lot of time following all of the paths within that range.
Based on these considerations, all three factors should play a role in
deciding when ``enough'' analyses have been found, i.e. in setting the
threshold below which analyses are discarded.  One way to give each
factor a role is to simply set three thresholds, and stop expanding
whenever one of them is crossed.  We shall instead define two
thresholds, one of which is defined as a function of two of the
factors. 

The beam threshold at word $w_{i}$ is a function of the probability of
the top ranked candidate analysis, $\scH_{i+1}[0]$, on priority queue
$\scH_{i+1}$ and the number, $|\scH_{i+1}|$, of candidate analyses on
$\scH_{i+1}$.  The basic idea is that we want the beam to be very wide
if there are few analyses that have been added to $\scH_{i+1}$, but
relatively narrow if many analyses have been advanced.  If $\tilde{p}$
is the probability of the highest ranked analysis on $\scH_{i+1}$,
then all other analyses are discarded if their probability falls  
below $\tilde{p}f(\gamma,|\scH_{i+1}|)$, where $\gamma$ is an initial
parameter, which we call the {\it base beam factor\/}.  For the
initial study, reported in the next section, which parsed strings of
POS tags, $\gamma$ was $10^{-4}$, and $f(\gamma,|\scH_{i+1}|) =
\gamma|\scH_{i+1}|$.  In this case, if 100 analyses have already been
pushed onto $\scH_{i+1}$, then a candidate analysis must have a
probability above $10^{-2}\tilde{p}$ to avoid being pruned.  When
$|\scH_{i+1}|$ = 1000 candidate analyses, the beam is narrowed
to $10^{-1}\tilde{p}$.  For the later study, reported beginning on
page 93, which parsed strings of words,  $\gamma$ was
$10^{-11}$, unless otherwise noted, and $f(\gamma,|\scH_{i+1}|) =
\gamma|\scH_{i+1}|^{3}$.  This function has the effect of having a very
wide beam early, but closing much faster than in the early study.  Thus, 
if 100 analyses have already been pushed onto $\scH_{i+1}$, then a
candidate analysis must have a probability above $10^{-5}\tilde{p}$ 
to avoid being pruned.  After 1000 candidate analyses, the beam has
narrowed to $10^{-2}\tilde{p}$.  There is also a 
maximum number of allowed analyses on $\scH_{i}$, in case the parse
fails to advance an analysis to $\scH_{i+1}$.  This was typically 10,000
in both studies.

\section{Empirical results I}
\subsection{Evaluation}
Statistical parsers are typically evaluated for accuracy at the
constituent level, rather than simply whether or not the parse that
the parser found is completely correct or not.  A constituent for
evaluation purposes consists of a non-preterminal non-terminal 
label (e.g. NP) and a span (beginning and ending word positions).  For
example, in figure \ref{fig:itree}(a), there is a VP that spans the
words ``\textttt{chased the ball}''.  Evaluation is carried out on a
hand-parsed test corpus \cite{Marcus93}, and the manual parses are
treated as correct.   We will call the manual parse GOLD and the parse
that the parser returns TEST.  Precision is
the number of common constituents in GOLD and TEST
divided by the number of constituents in TEST.  Recall is the
number of common constituents in GOLD and TEST
divided by the number of constituents in GOLD. Following
standard practice, we will be reporting scores only for
non-part-of-speech constituents, which are called labeled recall
(LR) and labeled precision (LP).  Also following standard practice, we
will ignore punctuation altogether, and treat ADVP and PRN as
equivalent.  Sometimes we will present average labeled precision and
recall, and also what can be termed the parse error, which is one
minus their average.

LR and LP are part of the standard set of {\small PARSEVAL} measures
of parser quality \cite{Black91}.  For the preliminary empirical
results we will focus upon LR and LP as measures of accuracy, but when
the full-blown model is investigated, we will also include, from this
set of measures, the crossing bracket scores: average crossing brackets 
(CB), percentage of sentences with no crossing brackets (0 CB), and
the percentage of sentences with two crossing brackets or fewer
($\leq$ 2 CB).  In addition, to measure efficiency, we will show the
average number of rule expansions considered per word, i.e. the number
of rule expansions for which a probability was calculated -- see
\namecite{Roark00b} -- and the average number of analyses advanced to
the next priority queue per word. 

This is an incremental parser with a pruning strategy and no
backtracking.  In such a model, it is possible to commit to a set of
partial analyses at a particular point that cannot be completed given
the rest of the input string (i.e. the parser can {\it garden
path\/}).  In such a case, the parser fails to return a complete
parse.  For the preliminary results, we performed evaluation upon
those sentences for which a parse is found.  For the full-blown model,
in the event that no complete parse is found, the highest
initially ranked parse on the last non-empty priority queue is
returned.  All unattached words are then attached at the highest level
in the tree. 
In such a way we predict no new constituents and all incomplete
constituents are closed.  This structure is evaluated for precision
and recall, which is entirely appropriate for these incomplete as well
as complete parses.  If we fail to identify nodes later in the parse,
the recall will suffer, and if our early predictions were bad, both
precision and recall will suffer.  Of course, the percentage of these
failures are reported as well.

\subsection{Delaying rule identification through factorization}
\begin{table*}
\begin{small}
\begin{tabular}{|l|c|c|c|c|c|c|}
\hline
{Transform} &
{Rules in} &
{Percent of} &
{Avg. Rule} &
{Avg. LP} &
{Avg. MLP} &
{Ratio of Avg.}\\
{} &
{Grammar} &
{Sentences} &
{Expansions} &
{and LR${}^{\dag}$} &
{LP and LR${}^{\dag}$} &
{Prob to Avg.} \\
&&
{Parsed${}^{\ast}$} &
{Considered} &
&&
{MLP Prob${}^{\dag}$} \\\hline
{None} &
{14962} &
{34.16} &
{19270} &
{.65521} &
{.76427} &
{.001721} \\\hline
{RF} &
{37955} &
{33.99} &
{96813} &
{.65539} &
{.76095} &
{.001440} \\\hline
{LF1} &
{29851} &
{91.27} &
{10140} &
{.71616} &
{.72712} &
{.340858} \\\hline
{LF0} &
{41084} &
{97.37} &
{13868} &
{.73207} &
{.72327} &
{.443705} \\\hline
\end{tabular}
\end{small}\\
{\footnotesize Beam Factor = $10^{-4}$ \hspace*{.18in}
${}^{\ast}$Length $\leq$ 40 (2245 sentences
in F23 - Avg. length = 21.68)\\
${}^{\dag}$Of those sentences parsed}
\caption{The effect of different approaches to
factorization}\label{tab:bin}
\end{table*}

Table \ref{tab:bin} summarizes some trials demonstrating the effect of
different factorization approaches on parser performance. The grammars were
induced from sections 2-21 of the Penn Wall St. Journal Treebank
\cite{Marcus93}, and tested on section 23. The parser was applied to
strings of given POS tags, not words.  For each transform
tested, every tree in the training corpus was transformed before
grammar induction, resulting in a transformed {\small PCFG} and
look-ahead probabilities estimated in the standard way. Each parse
returned by the parser was de-transformed for evaluation. The parser
used in each trial was identical, with a base 
beam factor $\gamma = 10^{-4}$, and $f(\gamma,|\scH_{i+1}|) =
\gamma|\scH_{i+1}|$. The performance 
is evaluated using these measures: (i) the percentage of candidate
sentences for which a parse was found (coverage); (ii) the average
number of rule expansions considered per candidate
sentence (efficiency); and 
(iii) the average labeled precision and recall of those sentences for
which a parse was found (accuracy). We also used the same grammars
with an exhaustive, bottom-up {\small CKY} parser, to ascertain both the
accuracy and probability of the maximum likelihood parse ({\small MLP}). We
can then additionally compare the parser's performance to the {\small MLP}'s
on those same sentences. 

As expected, {\it right\/} factorization conferred no benefit to our
parser.  No factorization and right factorization resulted in very low
coverage -- those sentences that they do cover are apparently easier
to parse, given that their maximum likelihood parse precision and
recall scores are higher than for the test set as a whole.  {\it
Left\/} factorization, in contrast, improved performance across the
board. {\small LF0} provided a substantial improvement in coverage and
accuracy over {\small LF1}, with something of a decrease in
efficiency. This efficiency hit is partly attributable to the fact that 
the same tree has more nodes with {\small LF0}. Indeed, the efficiency
improvement with left factorization over the standard grammar is even
more interesting in light of the great increase in the size of the
grammars. 

It is worth noting at this point that, with the {\small LF0} grammar,
this parser is now a viable 
broad-coverage statistical parser, with good coverage, accuracy, and
efficiency\footnote{The very efficient bottom-up statistical parser
detailed in \namecite{Charniak98} measured efficiency in terms of total 
edges {\it popped\/}.  An edge (or, in our case, a parser state) is
{\it considered\/} when a probability is calculated for it, and we
felt that this was a better efficiency measure than simply those
popped.  As a baseline, their parser {\it considered\/} an average of
2216 edges per sentence in section 22 of the WSJ corpus (p.c.).}.
Left-recursion, typically a great problem for top-down parsers, does
not seem to be fatal here, despite being relatively probable.
Next we considered the left-corner parsing strategy.

\subsection{Left-corner parsing}
We will be investigating left-corner parsing as a grammar transform,
as discussed in the previous section.  Multiple transforms can be
applied to the grammar in sequence, through function composition.  For
example, the output of a left-corner transform can be left-factored.
Recall that we will denote this kind of transform composition (A
$\circ$ B)($\tau$) = B(A($\tau$)).  Recall also that, since
$\epsilon$-productions can be introduced from either transform in a
redundant location, when composing them the factorization will be to
unary rather than nullary.

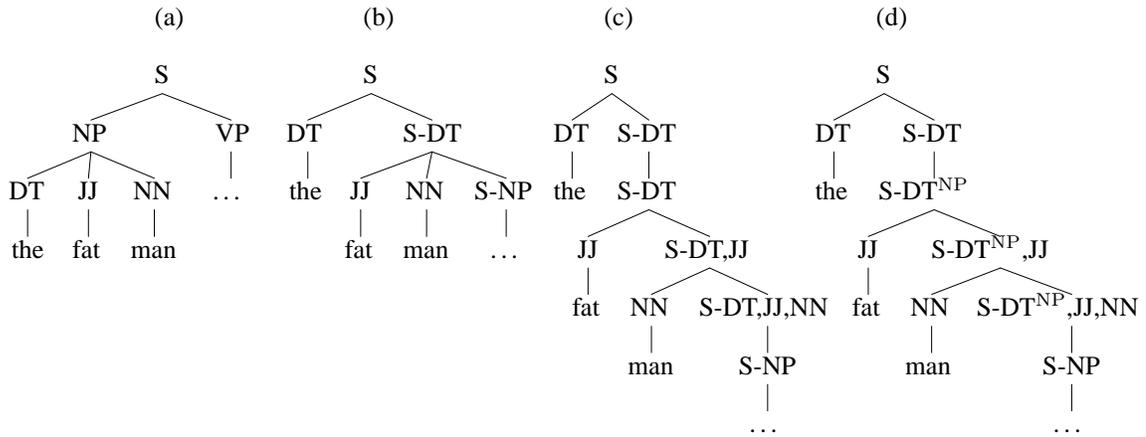
\begin{figure}[t]
\begin{small}
\begin{picture}(453,170)(0,-160)
\put(55,-8){(a)}
\put(55,-30){S}
\drawline(58,-34)(31,-44)
\put(24,-52){NP}
\drawline(31,-56)(7,-66)
\put(-0,-74){DT}
\drawline(7,-78)(7,-88)
\put(1,-96){the}
\drawline(31,-56)(30,-66)
\put(26,-74){JJ}
\drawline(30,-78)(30,-88)
\put(24,-96){fat}
\drawline(31,-56)(55,-66)
\put(47,-74){NN}
\drawline(55,-78)(55,-88)
\put(46,-96){man}
\drawline(58,-34)(84,-44)
\put(78,-52){VP}
\drawline(84,-56)(84,-66)
\put(77,-74){\ldots}

\put(134,-8){(b)}
\put(134,-30){S}
\drawline(137,-34)(113,-44)
\put(105,-52){DT}
\drawline(113,-56)(113,-66)
\put(106,-74){the}
\drawline(137,-34)(160,-44)
\put(149,-52){S-DT}
\drawline(160,-56)(133,-66)
\put(129,-74){JJ}
\drawline(133,-78)(133,-88)
\put(127,-96){fat}
\drawline(160,-56)(158,-66)
\put(150,-74){NN}
\drawline(158,-78)(158,-88)
\put(149,-96){man}
\drawline(160,-56)(188,-66)
\put(176,-74){S-NP}
\drawline(188,-78)(188,-88)
\put(182,-96){\ldots}

\put(225,-8){(c)}
\put(225,-30){S}
\drawline(228,-34)(213,-44)
\put(206,-52){DT}
\drawline(213,-56)(213,-66)
\put(206,-74){the}
\drawline(228,-34)(242,-44)
\put(230,-52){S-DT}
\drawline(242,-56)(242,-66)
\put(230,-74){S-DT}
\drawline(242,-78)(219,-88)
\put(215,-96){JJ}
\drawline(219,-100)(219,-110)
\put(213,-118){fat}
\drawline(242,-78)(265,-88)
\put(248,-96){S-DT,JJ}
\drawline(265,-100)(243,-110)
\put(235,-118){NN}
\drawline(243,-122)(243,-132)
\put(234,-140){man}
\drawline(265,-100)(287,-110)
\put(261,-118){S-DT,JJ,NN}
\drawline(287,-122)(287,-132)
\put(276,-140){S-NP}
\drawline(287,-144)(287,-154)
\put(280,-162){\ldots}

\put(328,-8){(d)}
\put(328,-30){S}
\drawline(331,-34)(312,-44)
\put(305,-52){DT}
\drawline(312,-56)(312,-66)
\put(305,-74){the}
\drawline(331,-34)(350,-44)
\put(338,-52){S-DT}
\drawline(350,-56)(350,-66)
\put(329,-74){S-DT$^\mathrm{NP}$}
\drawline(350,-78)(325,-88)
\put(321,-96){JJ}
\drawline(325,-100)(325,-110)
\put(319,-118){fat}
\drawline(350,-78)(375,-88)
\put(349,-96){S-DT$^\mathrm{NP}$,JJ}
\drawline(375,-100)(349,-110)
\put(341,-118){NN}
\drawline(349,-122)(349,-132)
\put(339,-140){man}
\drawline(375,-100)(402,-110)
\put(366,-118){S-DT$^\mathrm{NP}$,JJ,NN}
\drawline(402,-122)(402,-132)
\put(391,-140){S-NP}
\drawline(402,-144)(402,-154)
\put(395,-162){\ldots}
\end{picture}
\end{small}
\caption{(a) The original structure; (b) after left-corner transform (LC);
(c) after left-corner transform and left-factorization (LC $\circ$
LF); and (d) after left-corner transform, left-factorization, and
parent announce annotation (LC $\circ$ LF $\circ$ ANN)}\label{fig:anntran} 
\end{figure}

Another probabilistic {\small LC} parser investigated \cite{Manning97},
which utilized an {\small LC} parsing architecture (not a transformed
grammar), also got a performance boost through left
factorization. Since that involved feeding an LF grammar to an LC parser,
this is equivalent to {\small LF} $\circ$ 
{\small LC}, which is a very different grammar from {\small LC}
$\circ$ {\small LF}. Given our two factorization orientations ({\small
RF} and {\small LF}), there are four possible compositions of 
factorization and {\small LC} transforms: 
\begin{center}\begin{small}
(a) RF $\circ$ LC (b) LF $\circ$ LC
(c) LC $\circ$ RF  (d) LC $\circ$ LF 
\end{small}\end{center}
Table \ref{tab:left} shows left-corner results over various
conditions\footnote{Option (c) is not the appropriate kind of
factorization for our parser, as argued in the previous section, and so 
is omitted.}. Interestingly, options (a) and (d) encode the same
information, leading to nearly identical performance\footnote{The
difference is due to the introduction of vacuous unary rules with
LF.}. As stated before, left factorization moves the rule announce
point from before to after all of the children. The {\small LC} transform is
such that {\small LC} $\circ$ {\small LF} 
also delays {\it parent\/} identification until after all of the
children. The transform {\small LC} $\circ$ {\small LF} $\circ$
{\small ANN} moves the parent announce
point back to the left corner by introducing unary rules at the left
corner that simply identify the parent of the factored rule. Figure
\ref{fig:anntran} shows a tree, and the sequential effects of the
transforms: tree (b) shows the left-corner transform; tree (c) shows
the structure after tree (b) has been left-factored.  Notice in tree
(c) that the category NP does not appear until after the last word of
the NP.  In tree (d), the NP prediction is annotated onto a category
that is predicted after the first child has been built.  This
allows us to test the effect of the position of the parent announce
point on the performance of the parser. As we can see, however, the
effect is slight, with similar performance on all measures. 

{\small LF} $\circ$ {\small LC} performs with higher accuracy than the others when used with
an exhaustive parser, but seems to require a massive beam in order to
even approach performance at the {\small MLP} level. \namecite{Manning97}
used a beam width of 40,000 parses on the success heap at each input
item, which 
must have resulted in several orders of magnitude more rule expansions
than what we have been considering up to now, and yet their average
labeled precision and recall (.7875) still fell well below what we
found to be the {\small MLP} accuracy (.7987) for the grammar.  This
is most likely due to sparse data, which the relatively narrow search
makes our parser particularly susceptible to, and apparently also
makes the search in Manning and Carpenter fall short as well.  The
chart parser, while also affected by sparse data, is not a garden
pathing model, so a poor local estimate of probabilities may not
derail the parser in the way that ours or Manning and Carpenter's may
be. 

\begin{table*}[t]
\begin{small}
\begin{tabular}{|l|c|c|c|c|c|c|}
\hline
{Transform} &
{Rules in} &
{Percent of} &
{Avg. Rule} &
{Avg. LP} &
{Avg. MLP} &
{Ratio of Avg.}\\
{} &
{Grammar} &
{Sentences} &
{Expansions} &
{and LR${}^{\dag}$} &
{LP and LR${}^{\dag}$} &
{Prob to Avg.} \\
&&
{Parsed${}^{\ast}$} &
{Considered} &
&&
{MLP Prob${}^{\dag}$} \\\hline
{LC} &
{21797} &
{91.75} &
{9000} &
{.76399} &
{.78156} &
{.175928} \\\hline
{RF $\circ$ LC} &
{53026} &
{96.75} &
{7865} &
{.77815} &
{.78056} &
{.359828} \\\hline
{LC $\circ$ LF} &
{53494} &
{96.7} &
{8125} &
{.77830} &
{.78066} &
{.359439} \\\hline
{LC $\circ$ LF $\circ$ ANN} &
{55094} &
{96.21} &
{7945} &
{.77854} &
{.78094} &
{.346778} \\\hline
{LF $\circ$ LC} &
{86007} &
{93.38} &
{4675} &
{.76120} &
{.80529} &
{.267330} \\\hline
\end{tabular}
\end{small}\\
{\footnotesize Beam Factor = $10^{-4}$ \hspace*{.18in}
${}^{\ast}$Length $\leq$ 40 (2245 sentences
in F23 - Avg. length = 21.68)\\
% \hspace*{.14in}
${}^{\dag}$Of those sentences parsed}
\caption{Left Corner Results}\label{tab:left}
\end{table*}

Sparse data occurs when the parameters of the model become too large
to be accurately estimated from the limited amount of training data.
In this case, the parameters are the rules, and the estimation
procedure (relative frequency estimation) used in this preliminary
experiment requires rules to be observed to give them any probability
mass.  Beyond this, in order for the estimate for any particular
parameter to converge to the true probability, many observations are
required.  Thus, for example, the {\small LF} $\circ$ {\small LC}
grammar has over 86,000 rules, as opposed to less than 54,000 for {\small
LC} $\circ$ {\small LF}, i.e. many more parameters and more-or-less
the same number of observations, which leads to fewer observations per
parameter to be estimated.  The reason for the large increase in
grammar size is that in the {\small LF} $\circ$ {\small LC} grammar,
even the composite non-terminals introduced by factorization are
recognized left-corner, leading to ancestor/left-corner pairs that
involve an indefinite number of base categories in combination.  This
same problem of sparse data will come up with non-local annotation in
the next section, and even more severely with the full model later in
this chapter.

\subsection{Non-local annotation}

\namecite{Johnson98b} discusses the improvement of {\small PCFG}
models via the annotation of non-local information onto non-terminal
nodes in the trees of the training corpus. One simple example is to
copy the parent node onto every non-terminal (called {\it parent
annotation\/} below), e.g. the rule S~$\rightarrow$~NP~VP at the root
of the tree becomes
S$^{\uparrow}S^\dag~\rightarrow$~NP$^{\uparrow}$S~VP$^{\uparrow}$S.
The idea here is that  
the distribution of rules of expansion of a particular non-terminal
may differ depending on the non-terminal's parent. Indeed, it was
shown that this additional information improves the {\small MLP}
accuracy dramatically.

\begin{figure}[t]
\begin{small}
\begin{picture}(399,122)(0,-122)
\put(87,-8){(a)}
\put(87,-30){S}
\drawline(90,-34)(61,-44)
\put(54,-52){NP}
\drawline(61,-56)(34,-66)
\put(27,-74){NP}
\drawline(34,-78)(7,-88)
\put(0,-96){DT}
\drawline(7,-100)(7,-110)
\put(1,-118){the}
\drawline(34,-78)(33,-88)
\put(25,-96){NN}
\drawline(33,-100)(33,-110)
\put(24,-118){man}
\drawline(34,-78)(61,-88)
\put(51,-96){POS}
\drawline(61,-100)(61,-110)
\put(57,-118){'s}
\drawline(61,-56)(89,-66)
\put(81,-74){NN}
\drawline(89,-78)(89,-88)
\put(81,-96){dog}
\drawline(90,-34)(118,-44)
\put(111,-52){VP}
\drawline(118,-56)(118,-66)
\put(107,-74){VBD}
\drawline(118,-78)(118,-88)
\put(112,-96){ate}

\put(221,-8){(b)}
\put(221,-30){S}
\drawline(224,-34)(196,-44)
\put(183,-52){NP$\uparrow$S}
\drawline(196,-56)(169,-66)
\put(152,-74){NP$\uparrow$NP}
\drawline(169,-78)(142,-88)
\put(134,-96){DT}
\drawline(142,-100)(142,-110)
\put(135,-118){the}
\drawline(169,-78)(168,-88)
\put(160,-96){NN}
\drawline(168,-100)(168,-110)
\put(158,-118){man}
\drawline(169,-78)(195,-88)
\put(186,-96){POS}
\drawline(195,-100)(195,-110)
\put(192,-118){'s}
\drawline(196,-56)(223,-66)
\put(215,-74){NN}
\drawline(223,-78)(223,-88)
\put(215,-96){dog}
\drawline(224,-34)(253,-44)
\put(240,-52){VP$\uparrow$S}
\drawline(253,-56)(253,-66)
\put(241,-74){VBD}
\drawline(253,-78)(253,-88)
\put(246,-96){ate}

\put(356,-8){(c)}
\put(356,-30){S}
\drawline(359,-34)(330,-44)
\put(318,-52){NP$\uparrow$S}
\drawline(330,-56)(303,-66)
\put(291,-74){NP$\uparrow$S}
\drawline(303,-78)(276,-88)
\put(269,-96){DT}
\drawline(276,-100)(276,-110)
\put(270,-118){the}
\drawline(303,-78)(302,-88)
\put(294,-96){NN}
\drawline(302,-100)(302,-110)
\put(293,-118){man}
\drawline(303,-78)(330,-88)
\put(320,-96){POS}
\drawline(330,-100)(330,-110)
\put(326,-118){'s}
\drawline(330,-56)(358,-66)
\put(350,-74){NN}
\drawline(358,-78)(358,-88)
\put(350,-96){dog}
\drawline(359,-34)(387,-44)
\put(380,-52){VP}
\drawline(387,-56)(387,-66)
\put(376,-74){VBD}
\drawline(387,-78)(387,-88)
\put(381,-96){ate}
\end{picture}
\end{small}
\caption{(a) Original tree; (b) with parent annotation; and (c) with
left-corner annotation}\label{fig:parann}
\end{figure}
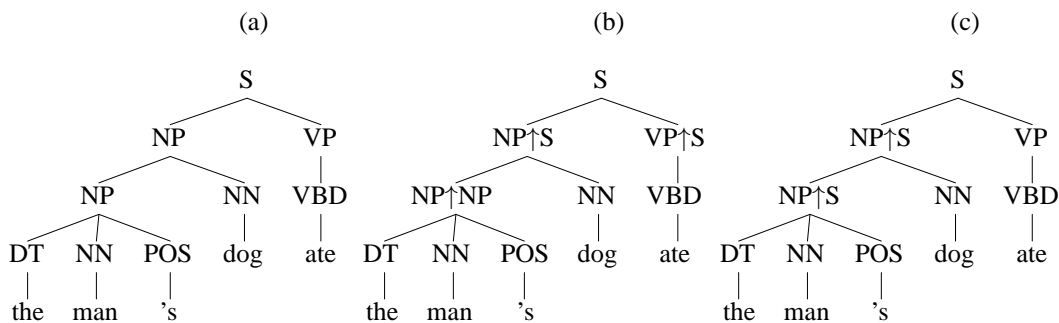

We looked at two kinds of
non-local information annotation: parent ({\small PA}) and left-corner
({\small LCA}). Left-corner parsing gives improved accuracy over top-down or
bottom-up parsing with the same grammar. Why? One reason may be that
the ancestor category is a useful piece of information in estimating a
probability distribution over likely rules, just as the parent
category is. To test this, we annotated the left-corner ancestor
category onto every leftmost non-terminal category. Figure
\ref{fig:parann} shows each of these two annotation transforms.
The results of our annotation trials are shown in table
\ref{tab:ann}. 

\begin{table*}
\begin{small}
\begin{tabular}{|l|c|c|c|c|c|c|}
\hline
{Transform} &
{Rules in} &
{Percent of} &
{Avg. Rule} &
{Avg. LP} &
{Avg. MLP} &
{Ratio of Avg.}\\
{} &
{Grammar} &
{Sentences} &
{Expansions} &
{and LR${}^{\dag}$} &
{LP and LR${}^{\dag}$} &
{Prob to Avg.} \\
&&
{Parsed${}^{\ast}$} &
{Considered} &
&&
{MLP Prob${}^{\dag}$} \\\hline
{LF0} &
{41084} &
{97.37} &
{13868} &
{.73207} &
{.72327} &
{.443705} \\\hline
{PA $\circ$ LF0} &
{63467} &
{95.19} &
{8596} &
{.79188} &
{.79759} &
{.486995} \\\hline
{LC $\circ$ LF} &
{53494} &
{96.7} &
{8125} &
{.77830} &
{.78066} &
{.359439} \\\hline
{LCA $\circ$ LF0} &
{58669} &
{96.48} &
{11158} &
{.77476} &
{.78058} &
{.495912} \\\hline
{PA $\circ$ LC $\circ$ LF} &
{80245} &
{93.52} &
{4455} &
{.81144} &
{.81833} &
{.484428} \\\hline
\end{tabular}
\end{small}\\
{\footnotesize Beam Factor = $10^{-4}$ \hspace*{.18in}
${}^{\ast}$Length $\leq$ 40 (2245 sentences
in F23 - Avg. length = 21.68)\\
${}^{\dag}$Of those sentences parsed}
\caption{Non-local annotation results}\label{tab:ann}
\end{table*}

There are two important points to notice from
these results. First, with {\small PA} we get not only the previously reported
improvement in accuracy, but additionally a fairly dramatic decrease
in the number of rule expansions that must be visited to find a
parse. That is, the non-local information not only improves the final
product of the parse, but it guides the parser more quickly to the
final product. The annotated grammar has 1.5 times as many rules, and
would slow a bottom-up {\small CKY} parser proportionally. Yet our parser
actually considers far fewer rule expansions en route to the more accurate
parse. 

Second, {\small LC}-annotation gives nearly all of the accuracy gain of
left-corner parsing, in support of the hypothesis that the ancestor 
information was responsible for the observed accuracy
improvement. This result suggests that if we can determine the
information that is being annotated by the troublesome (apparently
because of sparse data given the size of the grammar) {\small LF}
$\circ$ {\small LC} transform, we may be able to get the accuracy
improvement with a relatively narrow beam. Parent-annotation before
the {\small LC} transform gave us the best performance of all, with
very few rule expansions considered on average, and excellent accuracy
for a non-lexicalized grammar.  

\subsection{Accuracy/Efficiency tradeoff}
\begin{figure*}
\begin{center}
\epsfig{file=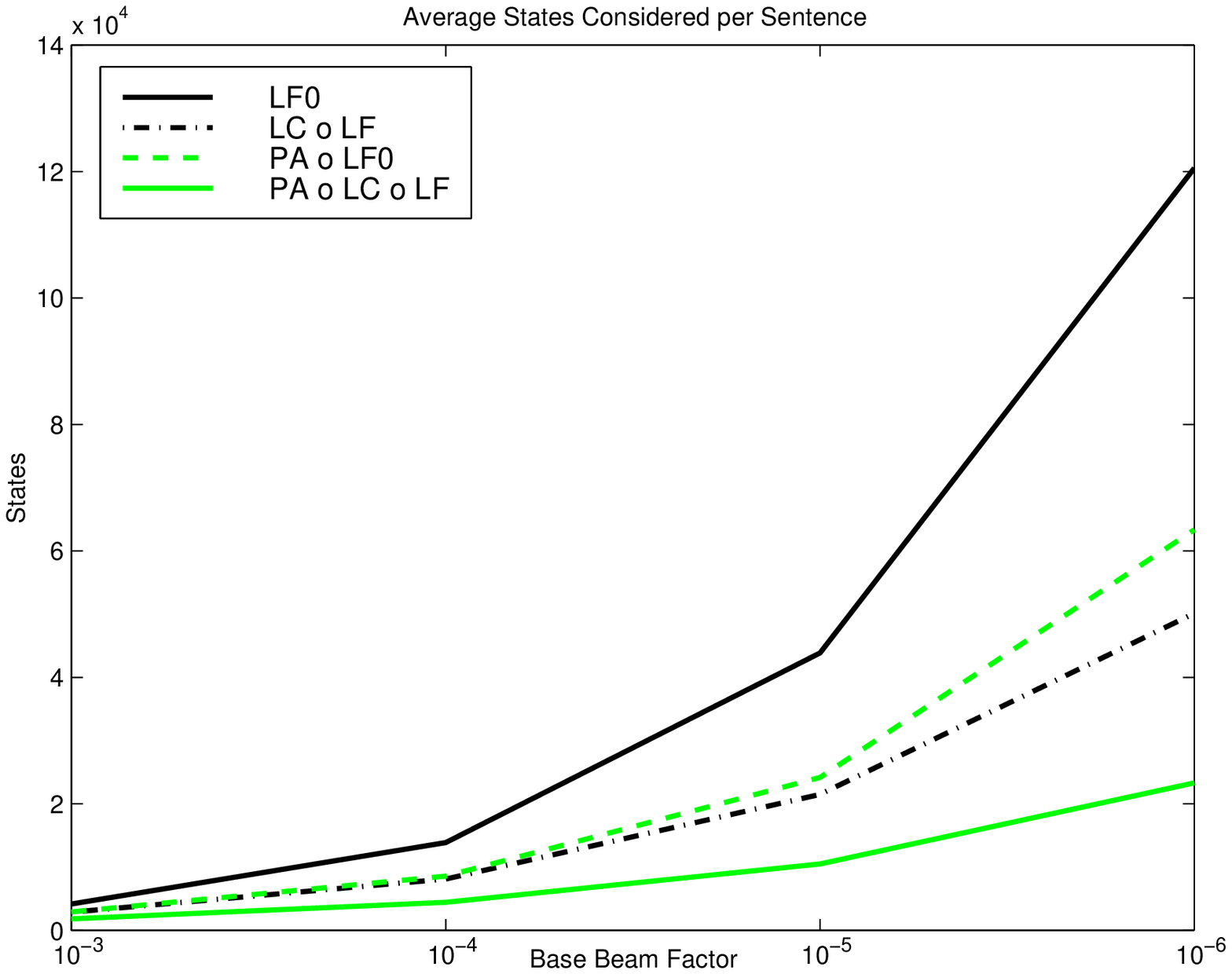, width=5.1in}\vspace*{.05in}\\
\epsfig{file=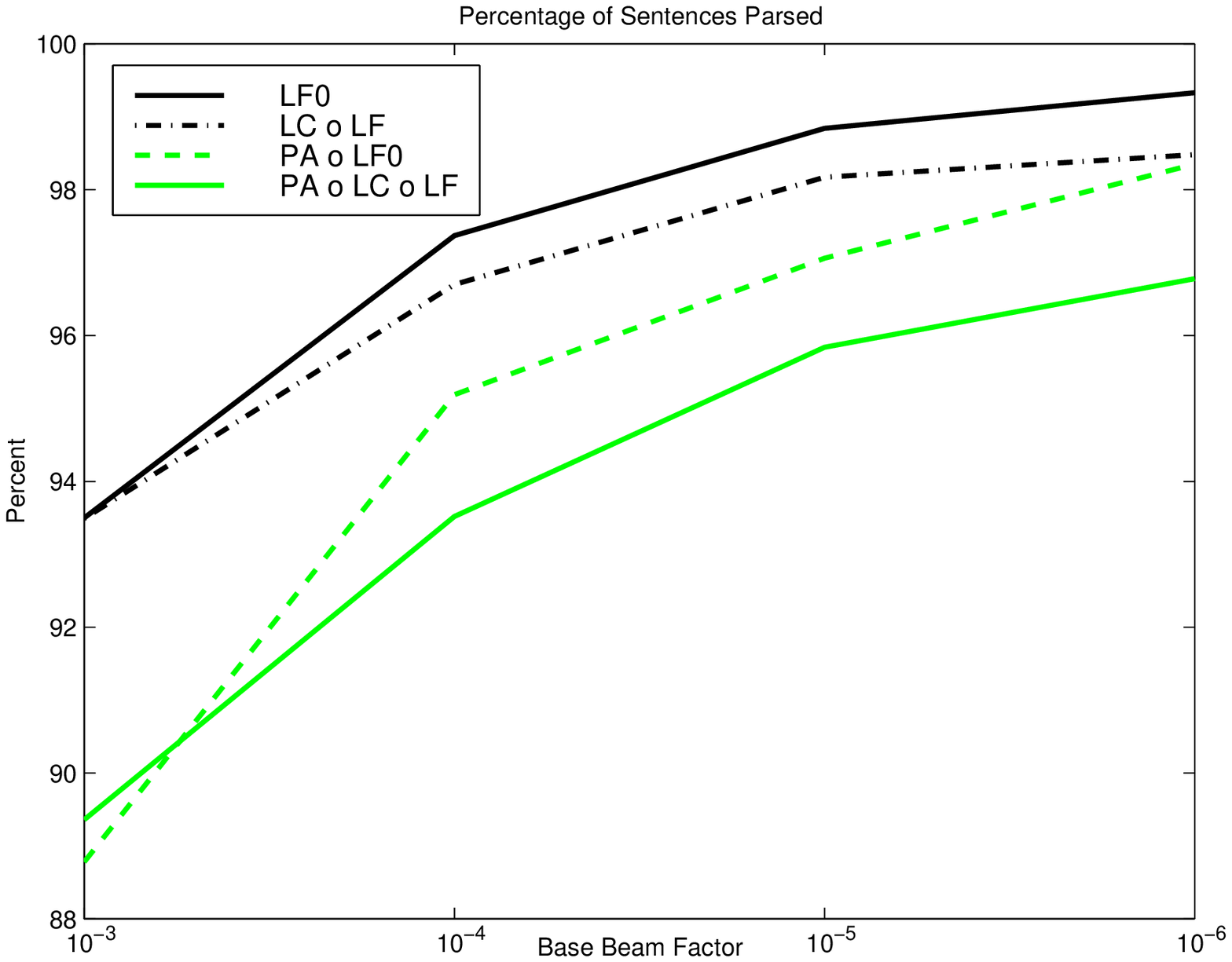, width=5.1in}
\end{center}
\caption{Changes in performance with beam factor variation} \label{fig:ef1}
\end{figure*}

\begin{figure*}
\begin{center}
\epsfig{file=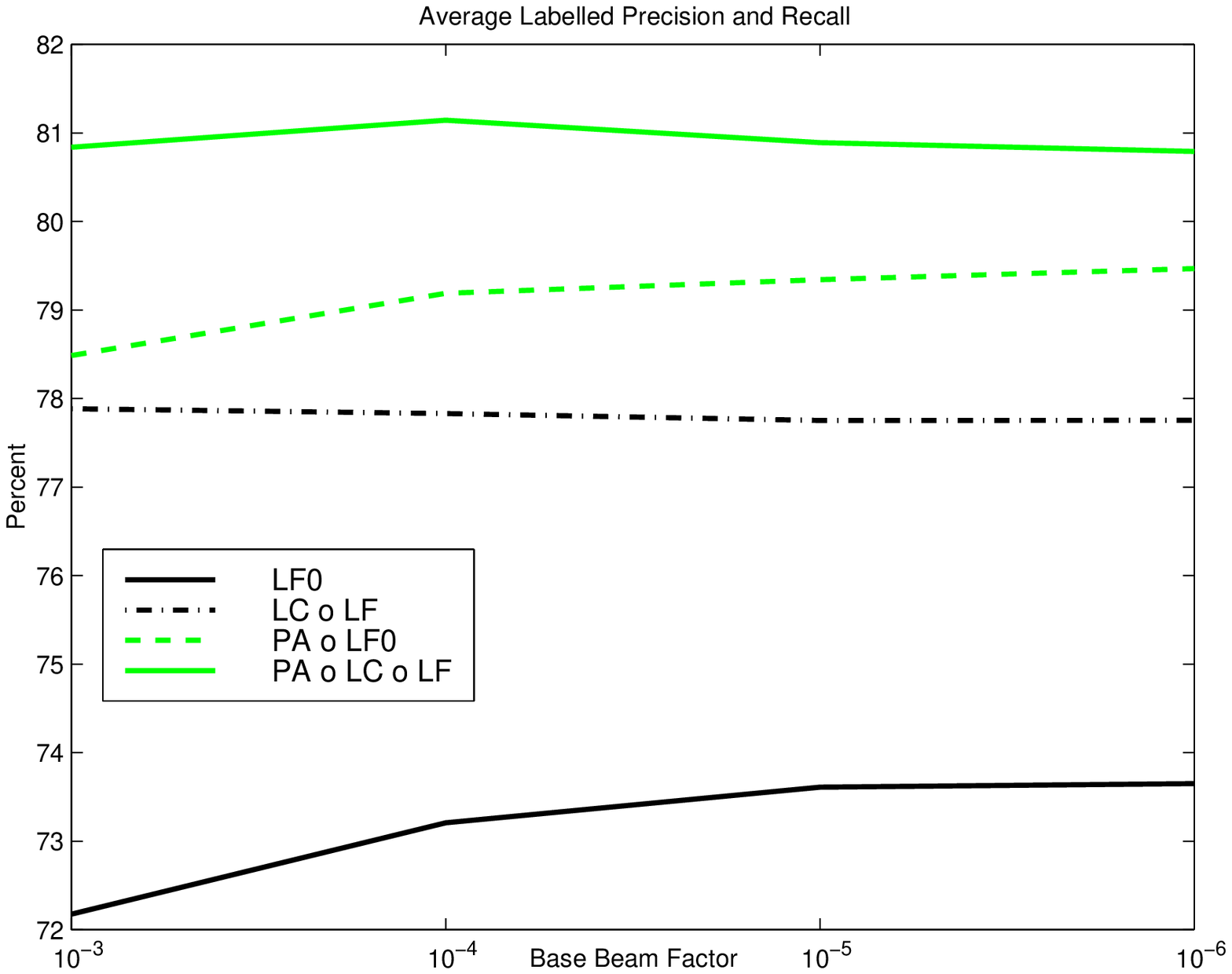, width=5.1in}\vspace*{.05in}\\
\epsfig{file=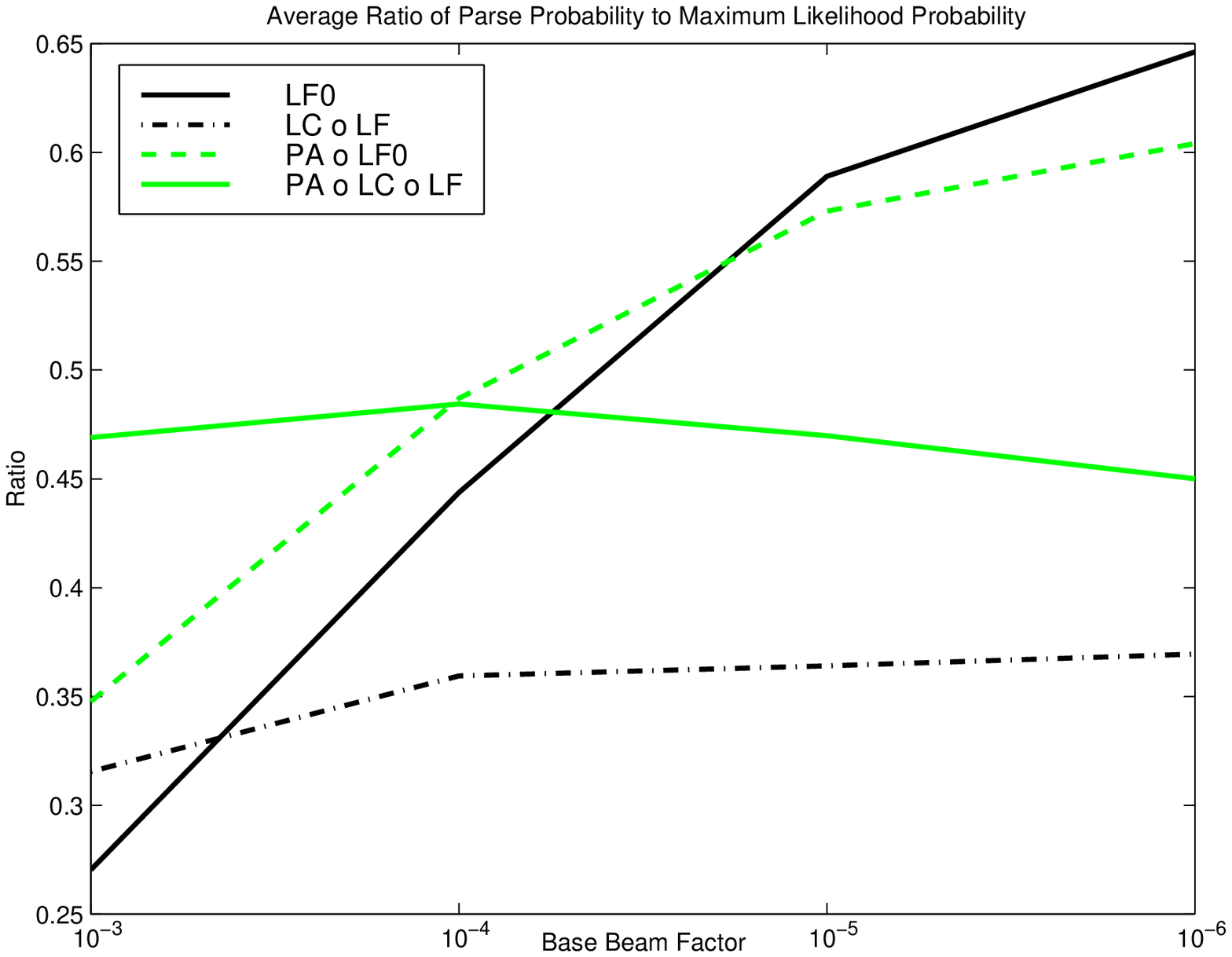, width=5.1in}
\end{center}
\caption{Changes in performance with beam factor variation} \label{fig:ef2}
\end{figure*}

One point that deserves to be made is that there is something of an
accuracy/efficiency tradeoff with regards to the base beam factor. The
results given so far were at $10^{-4}$, which functions pretty well for the
transforms we have investigated. Figures \ref{fig:ef1} and
\ref{fig:ef2} show four performance 
measures for four of our transforms at base beam factors of $10^{-3}$, 
$10^{-4}$, $10^{-5}$, and $10^{-6}$.  There is a dramatically increasing
efficiency burden as 
the beam widens, with varying degrees of payoff. With the top-down
transforms ({\small LF0} and {\small PA} $\circ$ {\small LF0}), the ratio of the average probability
to the {\small MLP} probability does improve substantially as the beam grows,
yet with only marginal improvements in coverage and
accuracy. Increasing the beam seems to do less with the left-corner
transforms. 

\subsection{Summary of results}
There are several key results in this section.  First, when the
grammar is appropriately factored, it is possible to navigate this
very large search space to find parses with nearly the same accuracy
as the maximum likelihood parse, with fairly high coverage.  This is
achieved with our top-down parser despite a heavily left-recursive
grammar.  Second,
non-local annotation -- of parent category or left-corner ancestor --
not only improves the accuracy of the parses found, which is
consistent with previous results, but also improves the efficiency
with which they are found, measured by the number of distinct rule
expansions that need to be considered.  Left-corner parsing provides
an accuracy improvement over simple top-down, but we have shown that
this is by virtue of the annotation of the ancestor category in the
transform.  Nevertheless, there is an efficiency improvement of over
25 percent through the use of a left-corner grammar versus a
left-corner annotated grammar, which is unsurprising, given that the
left-corner grammar jumps over left-recursive cycles.

These are promising results.  Sparse data, however, was experienced
through lower coverage (higher percentage of failure to find a parse)
whenever additional non-local information was 
encoded into the grammar.  This will become even more acute when we
move to parsing words instead of POS tags, and when lexical
information is ``annotated'' onto the grammars.  Smoothing of the
grammars will be employed, so that more reliable estimates can be
obtained of conditional rule probabilities.  Hence, while the parsing
algorithm will remain largely unchanged, the probability model will
change dramatically.

\section{Lexicalized conditional probability model}
A simple PCFG conditions rule probabilities on the
left-hand side of the rule.  It has been shown repeatedly --
e.g. \namecite{Briscoe93}, \namecite{Charniak97},
\namecite{Collins97}, \namecite{Inui97}, \namecite{Johnson98b} --
that conditioning the probabilities of structures on the
context within which they appear, for example on the lexical head of a
constituent \cite{Charniak97,Collins97}, on the label of its parent
non-terminal \cite{Johnson98b}, or, ideally, on both and many other
things besides, leads to a much better parsing model and results in
higher parsing accuracies.

One way of thinking about conditioning the probabilities of
productions on contextual information, e.g. the label of the parent of
a constituent or the lexical heads of constituents, is as annotating
the extra conditioning information onto the labels in 
the context-free rules.  Examples of this are bilexical grammars -- see
e.g. \namecite{Eisner99}, \namecite{Charniak97}, \namecite{Collins97} --
where the lexical heads of each constituent are annotated on both the
right- and left-hand sides of the context free rules, under the
constraint that every constituent inherits the lexical head from
exactly one of its children, and the lexical head of a POS is its
terminal item.  Thus the rule S~$\rightarrow$~NP~VP becomes, for
instance, S[{\it barks}]~$\rightarrow$~NP[{\it dog}]~VP[{\it barks}].
One way to estimate the probabilities of these rules is to annotate
the heads onto the constituent labels
in the training corpus, and simply count the number of times
particular productions occur (relative frequency estimation). This
procedure yields conditional probability distributions of 
constituents on the right-hand side with their lexical heads, given
the left-hand side constituent and 
its lexical head.  The same procedure works if we annotate parent
information onto constituents.  This is how \namecite{Johnson98b}
conditioned the probabilities of productions: the left-hand side is no
longer, for example, S, but rather S$^\uparrow$SBAR,
i.e. an S with SBAR as parent.  This means that such a grammar is
still a PCFG, yet with a much larger non-terminal set, i.e. we are
growing the number of potential states of the pushdown automaton.  

Notice, however, that in the case of parent annotation, the
annotations on the right-hand side are predictable from the label
on the left-hand side (unlike, for example, bilexical grammars), so
that the relative frequency estimator yields conditional probability
distributions of the original rules, given the parent of the left-hand
side.  Let us make this explicit.  Consider the rule S~$\rightarrow$~NP~VP:
\begin{eqnarray}
\Pr(\mathrm{S}\ \rightarrow\ \mathrm{NP\ \ VP}) &=& \Pr(\mathrm{NP,VP} | \mathrm{lhs=S})
\end{eqnarray}
Now consider parent annotation; for example,
S$^\uparrow$SBAR~$\rightarrow$~NP$^\uparrow$S~VP$^\uparrow$S:
\begin{eqnarray}
\Pr(\mathrm{S}^\uparrow\mathrm{SBAR}\ \rightarrow\
\mathrm{NP}^\uparrow\mathrm{S}\ \mathrm{VP}^\uparrow\mathrm{S}) &=&
\Pr(\mathrm{NP}^\uparrow\mathrm{S},\mathrm{VP}^\uparrow\mathrm{S} |
\mathrm{lhs=S}^\uparrow\mathrm{SBAR})\nonumber\\
&=&\Pr(\mathrm{NP,VP,lhs=S} | \mathrm{lhs=S},
\mathrm{par=SBAR})\nonumber\\ 
&=& \Pr(\mathrm{NP,VP} | \mathrm{lhs=S},\mathrm{par=SBAR})
\end{eqnarray}
Hence the probability of the rule is the probability of the children
given the left-hand side of the rule, plus some additional
information.  The probability of a rule in a bilexical grammar does
not simplify in the same way, since the conditioned variables must
also include a novel lexical head.  For example, the probability of
S[{\it barks}]~$\rightarrow$~NP[{\it dog}]~VP[{\it barks}] is
formulated as $\Pr(\mathrm{NP,VP,np({\it dog})} |
\mathrm{lhs=VP,vp({\it barks})})$, which involves the prediction of an
additional event (namely the head of the NP) beyond the categories of
the children.

All of the conditioning information that we will be considering will
be such that the only novel predictions being made by rule
expansions are the 
node-labels of the constituents on the right-hand side.  Everything
else is already specified by the left-context.  We use the relative
frequency estimator, and smooth our production probabilities by
interpolating the relative frequency estimates with those obtained by
``annotating'' less contextual information.  As mentioned in the
previous section, rich unsmoothed models suffer greatly from sparse
training data, and these smoothing methods are a way of giving
probability mass to infrequent events.

This perspective on conditioning production probabilities makes it
easy to see that, in essence, by conditioning these probabilities, we
are expanding the state space.  That is, the number of distinct
non-terminals grows to include the composite labels; so does the
number of distinct productions in the grammar.  In a top-down parser,
each rule expansion 
is made for a particular candidate parse, which carries with it the
entire rooted derivation to that point; in a sense, the
left-hand side of the rule is annotated with the entire left-context,
and the rule probabilities can be conditioned on any aspect of this
derivation.

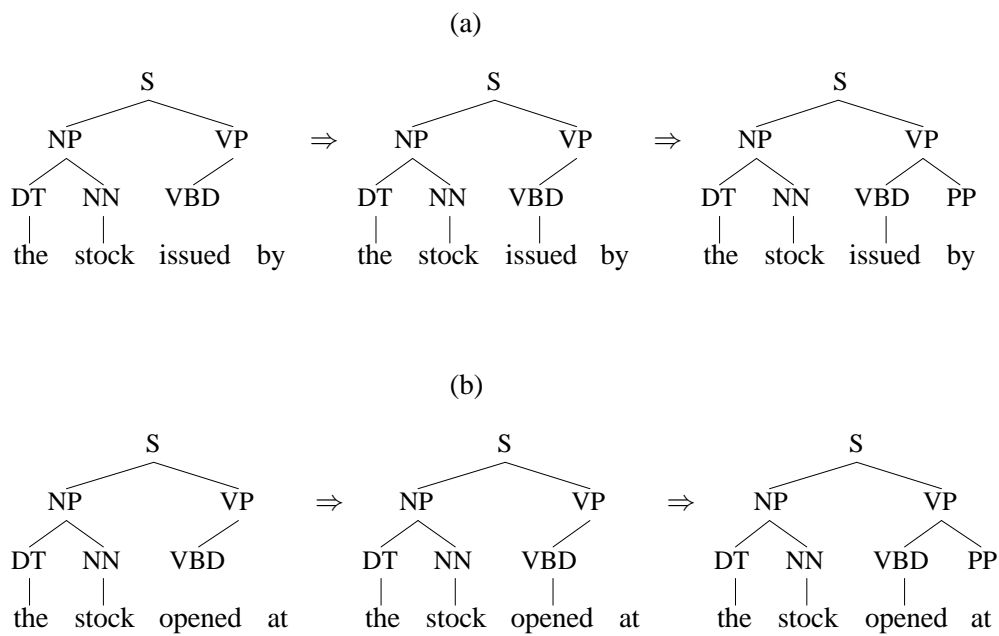
\begin{figure}[t]
\begin{picture}(364,100)(0,-100)
\put(166,-8){(a)}
\put(49,-30){\small S}
\drawline(52,-34)(21,-44)
\put(14,-52){\small NP}
\drawline(21,-56)(7,-66)
\put(-0,-74){\small DT}
\drawline(7,-78)(7,-88)
\put(1,-96){the}
\drawline(21,-56)(35,-66)
\put(27,-74){\small NN}
\drawline(35,-78)(35,-88)
\put(24,-96){stock}
\drawline(52,-34)(84,-44)
\put(77,-52){\small VP}
\drawline(84,-56)(69,-66)
\put(58,-74){\small VBD}
\put(56,-96){issued}
\put(93,-96){by}
\put(113,-52){$\Rightarrow$}
\put(180,-30){\small S}
\drawline(183,-34)(152,-44)
\put(145,-52){\small NP}
\drawline(152,-56)(138,-66)
\put(131,-74){\small DT}
\drawline(138,-78)(138,-88)
\put(131,-96){the}
\drawline(152,-56)(166,-66)
\put(158,-74){\small NN}
\drawline(166,-78)(166,-88)
\put(154,-96){stock}
\drawline(183,-34)(214,-44)
\put(207,-52){\small VP}
\drawline(214,-56)(200,-66)
\put(188,-74){\small VBD}
\drawline(200,-78)(200,-88)
\put(187,-96){issued}
\put(223,-96){by}
\put(243,-52){$\Rightarrow$}
\put(311,-30){\small S}
\drawline(313,-34)(282,-44)
\put(275,-52){\small NP}
\drawline(282,-56)(268,-66)
\put(261,-74){\small DT}
\drawline(268,-78)(268,-88)
\put(262,-96){the}
\drawline(282,-56)(296,-66)
\put(288,-74){\small NN}
\drawline(296,-78)(296,-88)
\put(285,-96){stock}
\drawline(313,-34)(345,-44)
\put(338,-52){\small VP}
\drawline(345,-56)(331,-66)
\put(319,-74){\small VBD}
\drawline(331,-78)(331,-88)
\put(317,-96){issued}
\drawline(345,-56)(359,-66)
\put(354,-74){\small PP}
\put(354,-96){by}
\end{picture}

\vspace*{.5in}

\begin{picture}(372,100)(0,-100)
\put(166,-8){(b)}
\put(51,-30){\small S}
\drawline(54,-34)(21,-44)
\put(14,-52){\small NP}
\drawline(21,-56)(7,-66)
\put(-0,-74){\small DT}
\drawline(7,-78)(7,-88)
\put(1,-96){the}
\drawline(21,-56)(35,-66)
\put(27,-74){\small NN}
\drawline(35,-78)(35,-88)
\put(24,-96){stock}
\drawline(54,-34)(86,-44)
\put(79,-52){\small VP}
\drawline(86,-56)(71,-66)
\put(60,-74){\small VBD}
\put(56,-96){opened}
\put(97,-96){at}
\put(115,-52){$\Rightarrow$}
\put(184,-30){\small S}
\drawline(187,-34)(154,-44)
\put(147,-52){\small NP}
\drawline(154,-56)(140,-66)
\put(133,-74){\small DT}
\drawline(140,-78)(140,-88)
\put(134,-96){the}
\drawline(154,-56)(168,-66)
\put(160,-74){\small NN}
\drawline(168,-78)(168,-88)
\put(157,-96){stock}
\drawline(187,-34)(219,-44)
\put(212,-52){\small VP}
\drawline(219,-56)(205,-66)
\put(193,-74){\small VBD}
\drawline(205,-78)(205,-88)
\put(189,-96){opened}
\put(230,-96){at}
\put(248,-52){$\Rightarrow$}
\put(317,-30){\small S}
\drawline(320,-34)(287,-44)
\put(281,-52){\small NP}
\drawline(287,-56)(274,-66)
\put(266,-74){\small DT}
\drawline(274,-78)(274,-88)
\put(267,-96){the}
\drawline(287,-56)(301,-66)
\put(293,-74){\small NN}
\drawline(301,-78)(301,-88)
\put(290,-96){stock}
\drawline(320,-34)(352,-44)
\put(345,-52){\small VP}
\drawline(352,-56)(338,-66)
\put(326,-74){\small VBD}
\drawline(338,-78)(338,-88)
\put(323,-96){opened}
\drawline(352,-56)(367,-66)
\put(362,-74){\small PP}
\put(363,-96){at}
\end{picture}
\caption{The same two steps in derivations for two strings}\label{fig:drv}
\end{figure}

Consider, for example, the two derivations in figure \ref{fig:drv}.
The two steps in these two derivations are identical: (i) extending
the verb POS to the verb, and (ii) building a PP child of the VP.  The
two verbs are more-or-less equiprobable in the Penn Treebank.  The
first of the two steps will involve estimating a probability for the
verb given its POS tag.  It might be useful to additionally condition
this probability on the head of the subject NP 
({\it stock\/}), since stocks are unlikely agents of {\it issue\/} and
not-so-unlikely agents of {\it open\/}.  In addition, we might want to
condition the VP expansion (to PP) on the head of the VP that
has just been found: {\it open\/} may be more likely to occur with PP
modification than {\it issue\/}.  Each derivation step in our top-down
parser carries 
with it the derivation to that point, which can be used to provide
relevant conditioning information for future steps in the derivation. 

We do not use the entire left-context to condition the rule
probabilities, but rather ``pick-and-choose'' which events in the 
left-context we would like to condition on.  One can think of the
conditioning events as functions, which take the partial tree
structure as an argument and return a value, upon which the rule
probability can be conditioned.  Each of these functions is an
algorithm for walking the provided tree and returning a value.  
For example, suppose that we want to condition the probability of the
rule $A~\rightarrow~\alpha$.  We might write a function that takes the
partial tree, finds the parent of the left-hand side
of the rule and returns its node label.  If the left-hand side has no
parent, i.e. it is at the root of the tree, the function returns the
null value (NULL).  We might write another function that returns the
non-terminal label of the closest sibling to the left of $A$, and NULL
if no such node exists.  We can then condition the probability of the
production on the values that were returned by the set of functions.

Recall that we are working with a factored grammar, so some of the
nodes in the factored tree have non-terminal labels that were created by the
factorization, and may not be precisely what we want for conditioning
purposes.  In order to avoid
any confusions in identifying the non-terminal label of a particular
rule production in either its factored or non-factored version, we
introduce the function $\CON(A)$ for every
non-terminal in the factored grammar $\LF(G)$, which is simply the label of the 
constituent whose factorization results in $A$.  For example, in figure
\ref{fig:bin}(e), $\CON(\mathrm{NP-DT,JJ})$ is simply NP.

Note that a function can return different values depending upon the
location in the tree of the non-terminal that is being expanded.  For
example, suppose that we have a function that returns the label of the
closest sibling to the left of $\CON(A)$ or
NULL if no such node exists.  Then a subsequent function could be
defined as follows:  return the parent of the parent (the grandparent)
of $\CON(A)$ {\it only if\/} 
$\CON(A)$ has no sibling to the left; 
{\it otherwise\/} return the 2nd closest sibling to the left of
$\CON(A)$, or, as always, NULL if no such
node exists.  If the function returns, for example, ``NP'', this could
either mean that the grandparent is NP or the 2nd closest sibling is
NP; yet there is no ambiguity in the meaning of the function,
since the result of the previous function disambiguates between the
two possibilities.  

\begin{figure}[t]
\begin{center}
\begin{picture}(306,236)(0,-236)
\put(-30,-10){\underline{\bf For all rules $A \rightarrow \alpha$}}
\begin{footnotesize}
%\put(0,-10){Where $A$ = $\CON(A)$-[$\ldots$]}
\put(144,-7){\circle{6}}
\put(142.5,-8.5){\tiny 0}
\put(150,-10){$A$}
\drawline(153,-12)(153,-23)
\put(69,-27){\circle{6}}
\put(67.5,-28.5){\tiny 1}
\put(75,-30){the parent, $Y_p$, of $\CON(A)$ in the derivation}
\drawline(153,-32)(153,-43)
\put(49,-47){\circle{6}}
\put(47.5,-48.5){\tiny 2}
\put(55,-50){the closest sibling, $Y_s$, to the left of $\CON(A)$ in the derivation}
\drawline(153,-52)(65,-83)
\drawline(153,-52)(235,-83)
\put(205,-70){\scriptsize\it $A$ = POS, $Y_s \neq$ NULL}
\put(-6,-87){\circle{6}}
\put(-7.5,-88.5){\tiny 3}
\put(0,-90){the parent, $Y_g$, of $Y_p$ in the
derivation}
\put(225,-90){the closest c-commanding}
\put(235,-97){lexical head to $A$}
\drawline(275,-100)(275,-111)
\put(220,-120){the next closest c-commanding}
\put(235,-127){lexical head to $A$}

\drawline(55,-93)(25,-121)
\drawline(55,-93)(135,-121)
\put(105,-110){\scriptsize\it $A$ = POS}
\put(-11,-137){\circle{6}}
\put(-12.5,-138.4){\tiny 4}
\put(0,-130){the closest sibling,}
\put(-5,-141){$Y_{ps}$, to the left of
$Y_p$}
\put(105,-130){the POS of the closest}
\put(95,-141){c-commanding lexical head to $A$}
\drawline(35,-143)(35,-160)
\put(-16,-172){\circle{6}}
\put(-17.5,-173.5){\tiny 5}
\put(-10,-165){If $Y_s$ is CC, the leftmost child}
\put(-10,-175){of the conjoining category; else NULL}
\drawline(165,-143)(225,-165)
\put(165,-176){the closest c-commanding lexical head to $A$}
\drawline(35,-180)(35,-195)
\put(-16,-207){\circle{6}}
\put(-17.5,-208.5){\tiny 6}
\put(-10,-200){the lexical head of $\CON(A)$ if already seen;}
\put(-10,-210){otherwise the lexical head of the closest}
\put(-10,-220){constituent to the left of $A$ within $\CON(A)$}
\drawline(255,-185)(255,-200)
\put(225,-210){the next closest c-commanding}
\put(235,-220){lexical head to $A$}
\end{footnotesize}
\end{picture}
\end{center}
\caption{Conditional probability model represented as a decision tree,
identifying the location in the partial parse tree of the conditioning
information}\label{fig:code}
\end{figure}
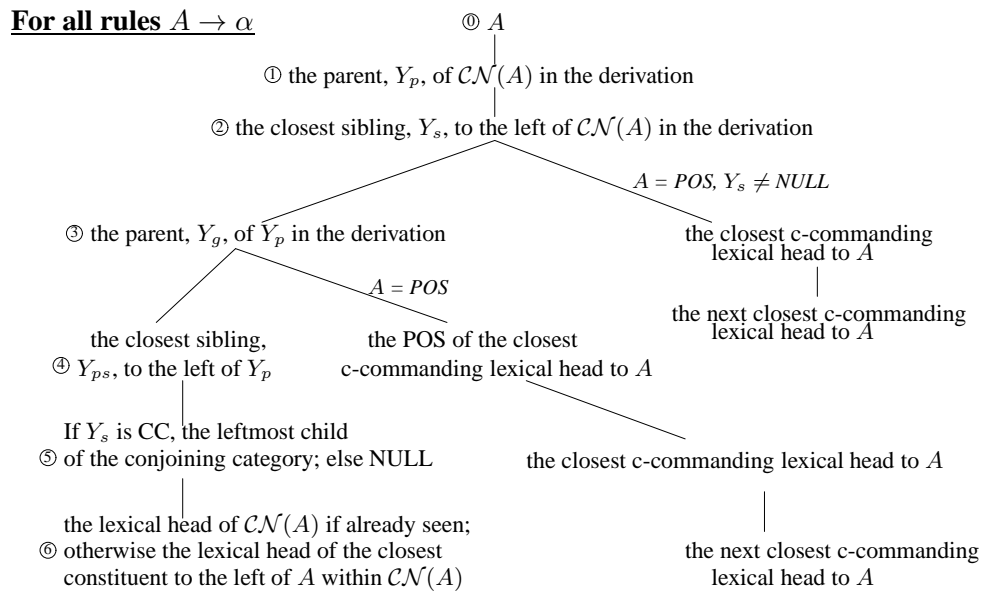

The functions that were used for the present study to condition the
probability of the rule, $A~\rightarrow~\alpha$, are
presented in figure \ref{fig:code}, in a tree structure.  
This is a sort of decision tree for a tree-walking algorithm to decide
what value 
to return, for a given partial tree and a given depth.  For example,
if the algorithm is asked for the value at level 0, it will return $A$,
the left-hand side of the rule being expanded\footnote{Recall that $A$
can be a composite non-terminal introduced by grammar factorization.
When the function is defined in terms of
$\CON(A)$, the values returned are obtained
by moving through the non-factored tree.}.  Suppose the algorithm is
asked for the value at level 4.  After level 2 there is a branch in
the decision tree.  If the left-hand side of the rule is a POS, and 
there is a sibling to the left of $\CON(A)$
in the derivation, then the algorithm takes the right branch of the
decision tree to decide what value to return; otherwise
the left branch.  Suppose it takes the left branch.  Then after level
3, there is another branch in the decision tree.  If the left-hand
side of the production is a POS, then the algorithm takes the right
branch of the decision tree, and returns (at level 4) the POS of the
closest c-commanding lexical head to $A$, which it finds by walking the
parse tree; if the left-hand side of the rule is not a POS, then the
algorithm returns (at level 4) the closest sibling to the left of
the parent of $\CON(A)$.

The functions that we have chosen for this chapter follow from the
intuition (and experience) that what helps parsing is different
depending on the constituent that is being expanded.  POS
nodes have lexical items on the right-hand side, and hence can bring
some of the head-head dependencies into the model that have been shown
to be so effective.  If the POS is leftmost within its constituent,
then very often the lexical item is sensitive to the governing
category to which it is attaching.  For example, if the POS is a
preposition, then its probability of expanding to a particular word is
very different if it is attaching to a noun phrase versus a verb
phrase, and perhaps quite different depending on the head of the
constituent to which it is attaching.  Subsequent POSs within
a constituent are likely to be open class words, and less
dependent on these sorts of attachment preferences.

Conditioning on parents and siblings of the left-hand side has proven
to be very useful.  To understand why this is the case, one need
merely to think of VP expansions.  If the parent of a 
VP is another VP (i.e. if an auxiliary or modal verb is
used), then the distribution over productions is different than if the
parent is an S.  Conditioning on head information, both
POS of the head and the lexical item itself, has proven
useful as well, although given our parser's left-to-right orientation,
in many cases the head has not been encountered within the particular
constituent.  In such a case, the head of the last child within the
constituent is 
used as a proxy for the constituent head.  All of our conditioning
functions, with one exception, return either parent or sibling node
labels at some specific distance from the left-hand side, or head
information from c-commanding constituents.  The exception is the
function at level 5 along the left branch of the tree in figure
\ref{fig:code}.  Suppose that 
the node being expanded is being conjoined with another node, which we
can tell by the presence or absence of a CC node.  In that
case, we want to condition the expansion on how the conjoining
constituent expanded.  In other words, this attempts to capture 
a certain amount of parallelism between the expansions of conjoined
categories. 

The conditioning events that we have made use of in this model have a
linear order, which, in general, corresponds to a distance from the
left-hand side of the rule being expanded, i.e. the parent of the
newly hypothesized category.  By adding this conditioning information,
we effectively encode finer and finer distinctions into the tag set,
which in its original form is quite coarse.  For example, the POS tag
``IN'' is both preposition and complementizer.  One way to make the
distinction between these two subsets is by annotating the parent of
the POS tag within the tree already built --
either IN$^\uparrow$PP versus IN$^\uparrow$SBAR, which nicely
separates these two subclasses.  Since both PPs and SBARs can occur in
similar contexts (both NP and VP modification), if we jump over the
parent of the POS, and instead condition the probability of the word
(e.g. \textttt{that}) on events farther away in the
left-context, we may end up conditioning on evidence that is pulled
from very different syntactic contexts.  The grandparent, for example,
i.e. the parent of the parent, can be a VP or NP in both cases.  Thus
any evidence for a VP grandparent bias versus an NP grandparent bias
for a complementizer like \textttt{`that'} (or any lexical attachment
preferences) will be brought to bear in support of VP attachment for
both SBAR and PP constituents, despite the fact that this particular
complementizer never participates in a PP.  Hence the general guiding
principle in building these conditioning events is that events closer
to the rule occur earlier in the linear set of conditioning events;
also tags occur before specific lexical items.  This is intended to
ensure that when information is used to support a structure, it has
actually occurred with that structure, and not some other.

In presenting the parsing results, we will systematically vary the
amount of conditioning information, so as to get an idea of the
behavior of the parser.  We will refer to the amount of conditioning
by specifying the deepest level from which a value is returned for each
branching path in the decision tree, from left to right in figure
\ref{fig:code}: the first number is for left-contexts where the left
branch of the decision tree is always followed (non-
POS non-terminals on the left-hand side); the second number for a left
branch followed by a right branch (POS nodes that are
leftmost within their constituent); and the third number for the
contexts where the right branch is always followed (POS nodes
that are not leftmost within their constituent).  For
example, (4,3,2) would represent a conditional probability model that
(i) returns NULL for all functions below level four in all contexts;
(ii) returns NULL for all functions below level three if the
left-hand side is a POS; and (iii) returns NULL for all
functions below level two for non-leftmost POS expansions.

\begin{small}
\begin{center}
\begin{table*}[t]
\begin{tabular} {|p{.8in}|p{.8in}|p{3.2in}|}
\hline
{Conditioning} & {Mnemonic label} & {Information level}\\\hline 
{0,0,0} & {none} & {Simple PCFG}\\\hline
{2,2,2} & {par+sib} & {Small amount of structural context}\\\hline
{5,2,2} & {NT struct} & {All structural (non-lexical) context for non-POS}\\\hline
{6,2,2} & {NT head} & {Everything for non-POS expansions}\\\hline
{6,3,2} & {POS struct} & {More structural info for leftmost POS
expansions}\\\hline 
{6,5,2} & {attach} & {All attachment info for leftmost POS
expansions}\\\hline 
{6,6,4} & {all} & {Everything}\\\hline
\end{tabular}
\caption{Levels of conditioning information, mnemonic labels, and a
brief description of the information level for empirical results}\label{tab:meaning}
\end{table*}
\end{center}
\end{small}

Table \ref{tab:meaning} gives a breakdown of the different levels of
conditioning information used in the empirical trials, with a mnemonic
label that will be used when presenting results.  These different
levels were chosen as somewhat natural points at which to observe how
much of an effect increasing the conditioning information has.  We
first include structural information from the context, i.e. node
labels from constituents in the left context.  Then we add lexical
information, first for non-POS expansions, then for leftmost
POS expansions, then for all expansions.

All of the conditional probabilities are linearly interpolated.  For
example, the probability of a rule conditioned on six 
events is the linear interpolation of two probabilities: (i) the
empirically observed relative frequency of the rule when the six
events co-occur; and (ii) the probability of the rule conditioned on
the first five events (which is in turn interpolated).  The
interpolation coefficients are a function of the frequency of the set
of conditioning events, and are estimated by iteratively adjusting the
coefficients so as to maximize the likelihood of the rules observed in
a held out corpus.

This was an outline of the conditional probability model that we used
for the PCFG.  The model allows us to assign probabilities to
derivations, which can be used by the parsing algorithm to decide
heuristically which candidate analyses are promising and should be
expanded, and which are less promising and should be pruned.

\section{Empirical results II}
The trials in this chapter are intended to examine the accuracy and
efficiency that can be achieved by the basic parser under a variety of
conditions. Later trials will look at: (i) further modifications to
this basic model; (ii) test corpus perplexity and recognition
performance; (iii) the effect of beam variation on these performance
measures; and (iv) performance with different corpora.  The results in
this section will give us a baseline against which we can compare
further results.

These results look at the performance of the parser on the
standard corpora for statistical parsing trials: sections 2-21
(989,860 words, 39,832 sentences) of the Penn Treebank serving as the
training 
data, section 24 (34,199 words, 1,346 sentences) as the held-out
data for parameter estimation, and section 23 (59,100 words, 2,416 
sentences) as the test data. 
Section 22 (41,817 words, 1,700 sentences) served as the development
corpus, on which the parser was tested until stable versions were
ready to run on the test data, to avoid developing the parser to fit
the specific test data.

\begin{table*}[t]
\begin{tabular}{|l|c|c|c|c|c|c|r|c|}
\hline
{\small Conditioning} & {\small LR} & {\small LP} & {\small CB} & 
{\small 0 CB} & {\small $\leq$ 2} & {\small Pct.} &
{\small Avg. rule\ \ } & {\small Average}\\
{} & {} & {} & {} & {} & {\small CB} & {\small failed} & {\small
expansions\ } & {\small analyses}\\
{} & {} & {} & {} & {} & {} & {} & {\small considered${}^{\dag}$} & {\small advanced${}^{\dag}$}\\\hline
\multicolumn{9}{|c|}{section 23: 2245 sentences of length $\leq$ 40}\\\hline
{none} & {71.1} & {75.3} & {2.48} & {37.3} & {62.9} & {0.9} & 
{14,369} & {516.5}\\\hline
{par+sib} & {82.8} & {83.6} & {1.55} & {54.3} & {76.2} & {1.1} & {9,615}
& {324.4}\\\hline
{NT struct} & {84.3} & {84.9} & {1.38} & {56.7} & {79.5} & {1.0} & {8,617}
& {284.9}\\\hline
{NT head} & {85.6} & {85.7} & {1.27} & {59.2} & {81.3} & {0.9} & {7,600}
& {251.6}\\\hline
{POS struct} & {86.1} & {86.2} & {1.23} & {60.9} & {82.0} & {1.0} & {7,327}
& {237.9}\\\hline
{attach} & {86.7} & {86.6} & {1.17} & {61.7} & {83.2} & {1.2} & {6,834}
& {216.8}\\\hline
{all} & {86.6} & {86.5} & {1.19} & {62.0} & {82.7} & {1.3} & {6,379}
& {198.4}\\\hline
\multicolumn{9}{|c|}{section 23: 2416 sentences of length $\leq$ 100}\\\hline
{attach} & {85.8} & {85.8} & {1.40} & {58.9} & {80.3} & {1.5} & {7,210}
& {227.9}\\\hline
{all} & {85.7} & {85.7} & {1.41} & {59.0} & {79.9} & {1.7} & {6,709}
& {207.6}\\\hline
\end{tabular}\\
\begin{footnotesize}
${}^{\dag}$per word
\end{footnotesize}
\caption{Results conditioning on various contextual events, standard training and testing corpora}\label{tab:res1}
\end{table*}

Table \ref{tab:res1} shows trials with increasing amounts of
conditioning information from the left-context.  
There are a couple of things to notice from these results.  First, and 
least surprising, is that the accuracy of the parses improved as we
conditioned on more and more information.  Like the non-lexicalized
parser in the previous results, we found that the search
efficiency, in terms of number of rule expansions considered 
or number of analyses advanced, also improved as we increased the amount
of conditioning.  Unlike those results, however,
our coverage did not substantially drop as the amount of
conditioning information increased, and in some cases improved
slightly.  For the earlier results, we did not smooth the conditional probability
estimates, and then blamed sparse data for our decrease in coverage as
we increased the conditioning information.  These results appear to
support this, since the smoothed model showed no such tendency.

\begin{figure*}[t]
\begin{center}
\epsfig{file={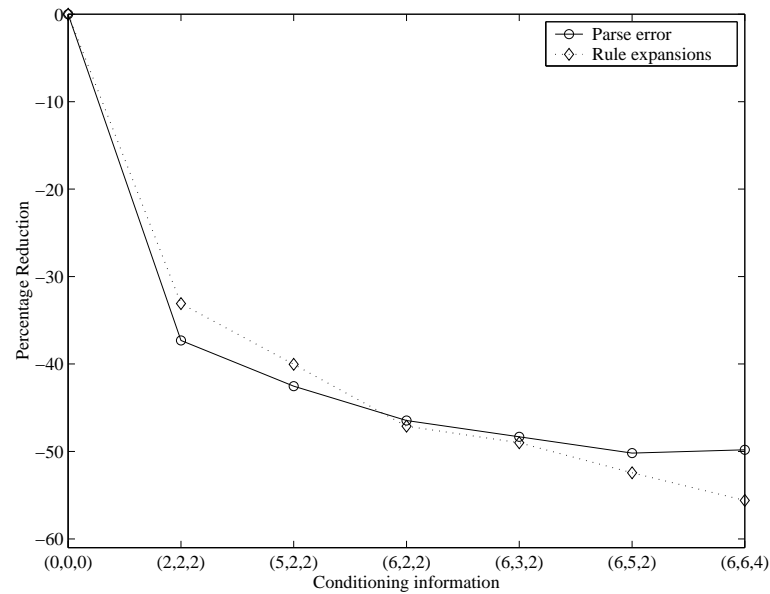}, width = 4in}
\end{center}
\caption{Reduction in average precision/recall error and in number of
rule expansions per word as conditioning increases, for sentences of length
$\leq$ 40} \label{fig:err_red}
\end{figure*}

Figure \ref{fig:err_red} shows the
reduction in parser error, $1-\frac{\mathrm{LR+LP}}{2}$, and 
the reduction in rule expansions considered as the conditioning
information increased.  The bulk of the
improvement comes from simply conditioning on the labels of the parent
and the closest sibling to the node being expanded.  Interestingly,
conditioning all POS expansions on two c-commanding heads
made no accuracy difference compared to conditioning only leftmost
POS expansions on a single c-commanding head; but it did
improve the efficiency.

These results, achieved using simple conditioning events
and considering only the left-context, are within 1-4 percentage
points of the best published accuracies cited above.  Table
\ref{tab:bestparse} compares our parser with these published
results.  All of the parsers that we are comparing with are off-line,
multi-pass parsers, which have the benefit of seeing the entire string
while building structure, and hence the ability to exploit certain
dependencies that are unobserved as we build our structure
incrementally.

Of the 2416 sentences in the section, our parser found the totally
correct parse for 728, a  
30.1 percent tree accuracy.  Also, the parser returns a set of  
candidate parses, from which we have been choosing the top ranked;  if 
we use an oracle to choose the parse with the highest accuracy from
among the candidate parses (which averaged 70.0 in number per
sentence), we find an average labeled precision/recall of 94.1, for 
sentences of length $\leq$ 100.  The parser, thus,
could be used as a front end to some other model, with the hopes of
selecting a more accurate parse from among the final candidate parses.

\begin{table*}[t]
\begin{center}
\begin{tabular}{|l|c|c|c|c|c|c|}
\hline
{Model} & {LR} & {LP} & {CB} & 
{0 CB} & {$\leq$ 2} & {Pct.} \\
{} & {} & {} & {} & {} & {CB} & {failed} \\\hline
\multicolumn{7}{|c|}{section 23: 2416 sentences of length $\leq$ 100}\\\hline
{Our parser} & {85.7} & {85.7} & {1.41} & {59.0} & {79.9} & {1.7} \\\hline\hline
{Charniak (1997)} & 86.7 & 86.6 & 1.20 & 59.9 & 83.2 & 0 \\\hline
{Ratnaparkhi (1999)} & 86.3 & 87.5 &&&& 0 \\\hline
{Collins (1999)} & 88.1 & 88.3 & 1.06 & 64.0 & 85.1 & 0 \\\hline
{Charniak (2000)} & 89.6 & 89.5 & 0.88 & 67.6 & 87.8 & 0 \\\hline
{Collins (2000)} & 89.6 & 89.5 & 0.88 & 67.6 & 87.8 & 0 \\\hline
\end{tabular}\\
\end{center}
\caption{Comparison of parsing results with the best in the literature.}\label{tab:bestparse}
\end{table*}

\begin{figure*}[t]
\begin{center}
\epsfig{file=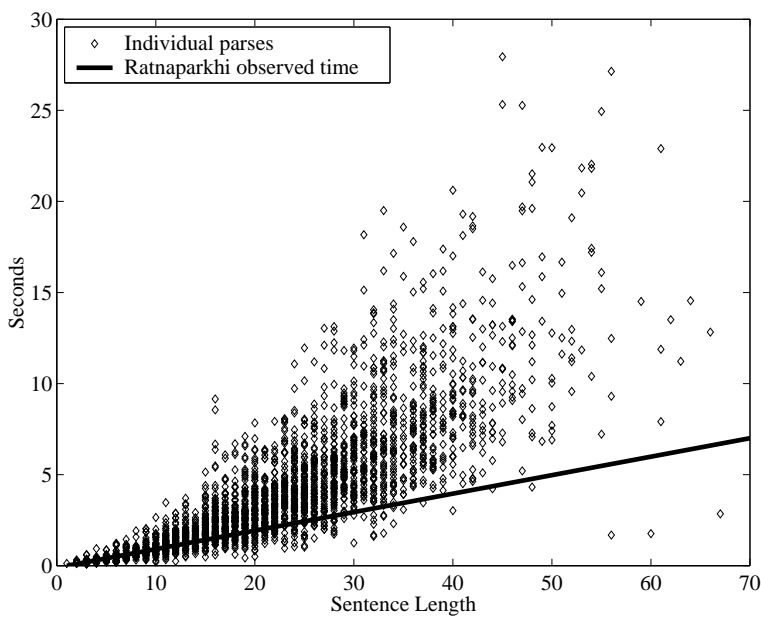, width=4in}
\end{center}
\caption{Observed running time on section 23 of the Penn treebank,
with the full conditional probability model and beam of $10^{-11}$,
using one 300 Mhz UltraSPARC processor and 256MB of RAM of a Sun
Enterprise 450}\label{fig:time}
\end{figure*}

While we have shown that the conditioning information improves the
efficiency in terms of rule expansions considered and analyses
advanced, what does the efficiency of such a parser look like in
practice?  Figure \ref{fig:time} shows the observed time at our
standard base beam of $10^{-11}$ with the full conditioning regimen,
alongside an approximation of the reported observed (linear) time in 
\namecite{Ratna97}.  Our observed times look polynomial, which is to
be expected given our pruning strategy.  We define our beam search in
terms of a probability range that narrows with the number of
successful analyses found; nonetheless, for any given number of
analyses, there is a probability range.  Hence, the denser the competitors
within a narrow probability range of the best analysis, the more time
will be spent working on these competitors.  The farther along in
the sentence, the more chance for ambiguities that can lead to such a
situation.  In other words, the density of competitors within a narrow
probability range of the best analysis will tend to grow with the
length of the sentence, leading to the increased time that we
observe.  That increase is very slight, and is bounded, since the
number of analyses that succeed will eventually raise the threshold to
the probability of the best analysis itself.  While our observed times
are not linear, and are clearly slower than his times (even with a
faster machine), they are quite respectably fast.  The differences
between a k-best and a beam-search parser (not to mention the use of
dynamic programming) make a running time difference unsurprising,
since best first can push good analyses through, without waiting to
fill the beam at each word. What is
perhaps surprising is that the difference is not greater.
Furthermore, this is quite a large beam (see discussion below),
so that very large improvements in efficiency can be had at the
expense of the number of analyses that are retained.

\section{Chapter summary}
In this chapter we have presented and tested a broad-coverage incremental
probabilistic parser that maintains fully connected structures in the
left-context.  In other words, this parser is consistent with models
of incremental interpretation, yet can cover freely occurring language
with very high accuracy, efficiency, and coverage.  We have found
that, as the quality of the statistical model improves, i.e. as more
relevant conditioning information is included in the model, the search
becomes more efficient as the parses get more accurate.  The smoothing
techniques that were used in the full model gave some relief to the
sparse data problems reported in the preliminary results, without
eliminating the efficiency gain associated with the richer models.

We will be putting this parser and model through more trials and tests
when we begin to look at language modeling for speech recognition.
There are, however, a number of remaining issues that need to be
investigated to improve the performance of this model.  In particular,
we will look at better smoothing techniques for the grammar itself; we
will take a detailed look at the parser's performance in the face of
left-recursion, and examine several methods for improving performance;
we will examine new conditioning features for parsing spoken language
transcripts; and we will investigate our model's ability to produce
empty nodes.

We have established a baseline performance level for a predictive
parser working in this very large search space.  The next chapter will
improve upon this baseline performance in all respects: coverage,
accuracy, and efficiency.  This will be followed by a detailed
examination of applying this parsing model to language modeling for
speech recognition.

\chapter{Model enhancements and modifications}
This chapter will focus upon potential improvements in three areas that are
either problematic or under-exploited in the parsing model that we
have outlined in the previous chapter: (i) the PCFG backbone; (ii)
left-recursion; and (iii) empty nodes.  Some of the modifications that
we will outline were motivated by applying the parser to spoken
language, rather than the text processing that we have presented to
this point.  To examine their effect in detail, however, we will be
comparing first with results from the previous chapter, before moving
on to parsing speech.  Later in the chapter we will present
parsing results of an improved model on the Switchboard corpus of
transcribed telephone conversations.  The subsequent chapter will explore
the use of the parser as a language model for speech recognition.

This chapter will be organized as follows.  The first section will
look at smoothing the PCFG backbone of the parser, leading to complete
coverage and improved accuracy over the results presented in the
previous chapter.  The second section will examine in detail the issue
of left-recursion, in an attempt to evaluate how much of a problem it
is for the top-down parser, and investigate some ways in which any
problems be ameliorated.  This section will 
include results from the application of the selective left-corner
transform, as well as other techniques.  We will then examine parsing
with the new Switchboard treebank, which has some new non-terminals
that will warrant changing the probability model slightly.
The last section will
examine the specification of a certain number of pre-terminal nodes as
empty, which is a simple and straightforward extension of the parser.
This could come in handy for parsing when the training data contains
terminals that are omitted in the testing data, such as punctuation
when parsing speech.  We will show parsing results when punctuation is
left in the training corpus, and treated as empty for parsing
purposes, compared to removing punctuation from both training and
testing. 

\section{Smoothed PCFG}
One of the problems that we continued to experience, even with the
smoothing regime that accompanied lexicalization at the end of the
last chapter, was a failure to find a parse for about one out of every
hundred sentences, even for this highly edited text.  While garden
pathing in this way is something that 
may recommend our model psycholinguistically, since people
seem to garden path under certain circumstances, the causes of garden
pathing that we experienced in the previous chapter were not generally
of the class that cause problems in people.  Rather, the failure to
parse was quite frequently due to rare uses of punctuation that happen
not to have been observed in particular syntactic contexts in the
training corpus.  More generally, many of the rules in the 
treebank are very flat, so that long sequences of children categories
are observed, with none of the intermediate constituents that would be
observed in more detailed hierarchical structures.  For example, noun
phrases can consist of a determiner followed by a string of nouns,
adjectives and other ``nouny'' things, in nearly every permutation.
Given limited training data, the chance of not observing some
particular permutation is quite high.  We will discuss a method of
smoothing to allow for unseen rules to be assigned a probability. 

To give an idea of how much of a problem this might be, of the 15
thousand or so PCFG rules that can be induced from the Penn Wall
St. Journal Treebank
in the standard way, over 3,600 of these are NP rules with only
pre-terminals on the right-hand side.  The pre-terminal categories on
the right-hand side of these base NPs include determiners, adjectives,
common and proper nouns, gerunds, and punctuation, among other things.  
While there are some ordering constraints -- e.g. determiners occur
typically first and only once -- productive noun compounding and
variations in punctuation result in many possible combinations.  Some
of these are observed, but many are not.  For example, the following
rule occurs in the training data with a probability of approximately
0.00000285: 
\begin{examples}
\item NP $\rightarrow$ DT JJ JJ NN NN NNS
\end{examples}
The POS tag DT is for determiners, JJ for adjectives, NN for singular
common nouns, and NNS for plural common nouns, so this rule would
cover something like \textttt{`the delicious black duck beak soups'}.
Unfortunately, the following two rules are not observed in the training 
corpus, and thus have a probability of zero with the current grammar
estimation technique:
\begin{examples}
\item NP $\rightarrow$ DT JJ JJ JJ NN NN NNS
\item NP $\rightarrow$ DT `` JJ '' JJ NN NN NNS
\end{examples}
Hence, there is no way to cover, in a flat NP rule, something like
\textttt{`the so-called delicious black duck beak soups'}, nor
something like \textttt{`the {\rm ``}delicious{\rm ''} black duck beak
soups'}.  This may be 
argued to be a short-coming of the grammar formalism, and that may be
true; more hierarchical structure would help with some of this, to the
extent that there would be more exemplars of shorter
rules\footnote{\namecite{Johnson98b} showed, however, that the perhaps
more linguistically well-motivated structures may not be as effective
as flat structures for modeling the dependencies, since the additional
level in the hierarchy carries an independence assumption that appears
to be false.}.  However, this is the grammar that has been provided in
the treebank, and the 
grammar estimation techniques that have been used up to now do not
provide sufficient probability to unseen rules of the sort we have
given in the example.  

A method that has been adopted for treebank parsing in the past
\cite{Collins97,Charniak00} is what Charniak has termed a Markov
grammar.  The basic idea is to make a Markov assumption about the
dependency between the children, i.e. that the probabilities of
children are independent of their siblings when the distance between
them is beyond some fixed $n$.  

Perhaps the easiest way to see how this would work in practice is
through the chain rule.  To simplify the notation, let us assume that
all rules in the grammar are of the form
A~$\rightarrow$~B$_0$~\ldots~B$_k$, and that B$_0$ is always some
start symbol, and B$_k$ is always 
some stop symbol.  Since every rule begins with B$_0$, its probability
is always 1.  A PCFG assigns probability to a rule as follows:
\begin{eqnarray}
\Pr(\rm{A} \rightarrow \rm{B}_0 \ldots \rm{B}_k) &=&
\prod_{i=1}^k \Pr(\rm{B}_i|\rm{A},\rm{B}_0, \ldots, \rm{B}_{i-1}) 
\end{eqnarray}
A Markov assumption of order $n$ would change each component of the
previous equation as follows: 
\begin{eqnarray}
\Pr(\rm{B}_i|\rm{A},\rm{B}_0, \ldots, \rm{B}_{i-1}) &=&
\Pr(\rm{B}_i|\rm{A},\rm{B}_{i-n}, \ldots, \rm{B}_{i-1})
\end{eqnarray}
This results in a new decomposition of the probability of a PCFG
rule.  For example, suppose we chose a Markov grammar of order 1:
\begin{eqnarray}
\Pr(\rm{A} \rightarrow \rm{B}_0 \ldots \rm{B}_k) &=&
\prod_{i=1}^k \Pr(\rm{B}_i|\rm{A},\rm{B}_{i-1})
\end{eqnarray}
To get back to our unobserved rules above, the probabilistic grammar
estimation for a Markov grammar of order 1 no longer asks how
frequently the entire sequence of children has been observed with that
parent, but rather how frequently each child has been observed with
that previous sibling and that parent.

The move to estimating PCFG rule probabilities in this manner
simplifies some things and complicates others.  We will be able to
do away with keeping track of specific rules, and simply evaluate the
probability of subsequent children as we grow the trees, conditioned
on events in the left-context as before, now including some number of
previous children in the production.  Complicating matters slightly is
the fact that constituent head identification is now no longer tied to
specific rules, and this will force us to include head identification
in our probabilistic model.

We have adopted, following \namecite{Charniak00}, a smoothed
third-order Markov grammar.  This means that a certain amount of
probability mass (depending on the smoothing parameters) is reserved
for arbitrary permutations of children that have been observed under
a particular parent.  In estimating the grammar in this way, we have
moved from a PCFG with some 15,000 possible productions to one with an
infinite number of possible productions.  At each point in the rule
expansion, some probability mass is reserved for producing any child.
This process can continue indefinitely.

\begin{figure}
\begin{algorithm}{PAR-SIB}{node,m,n}
\begin{FOR}{i \= 1 \TO m \hspace*{1in}\COMMENT{Move up $m$ nodes}}
  \begin{IF}{node \neq \text{NULL}}
    node \= node.parent
  \end{IF}
\end{FOR}\\
\begin{FOR}{i \= 1 \TO n \hspace*{1in}\COMMENT{Move left $n$ nodes}}
  \begin{IF}{node \neq \text{NULL}}
    node \= node.leftsib
  \end{IF}
\end{FOR}\\
\begin{IF}{node \neq \text{NULL}}
  \RETURN node.label
\ELSE
  \RETURN \text{NULL}
\end{IF}
\end{algorithm}

\begin{algorithm}{LEFTMOST-PS}{node,m,n}
\begin{IF}{node.leftsib \neq \text{NULL}\hspace*{.5in}\COMMENT{Only for
leftmost children}}
  \RETURN \text{NULL}
\ELSE
  \RETURN \CALL{PAR-SIB}(node,m,n)
\end{IF}
\end{algorithm}

\begin{algorithm}{LEX-HEAD}{node,m}
\begin{IF}{node \neq \text{NULL}}
  node \= node.head\hspace*{1.3in}\COMMENT{Go to a node's head child}
\end{IF}\\
\begin{WHILE}{node \neq \text{NULL} \text{and} node.child \neq
\text{NULL}\hspace*{.5in}\COMMENT{Until node is a leaf}}
  node \= node.head
\end{WHILE}\\
\begin{FOR}{i \= 1 \TO m\hspace*{1in}\COMMENT{Move up $m$ nodes}}
  \begin{IF}{node \neq \text{NULL}}
    node \= node.parent
  \end{IF}
\end{FOR}\\
\begin{IF}{node \neq \text{NULL}}
  \RETURN node
\ELSE
  \RETURN \text{NULL}
\end{IF}
\end{algorithm}

\begin{algorithm}{CURR-HEAD}{node,m}
\begin{IF}{node = \text{NULL}}
  \RETURN \text{NULL}
\end{IF}\\
headnode \= \CALL{LEX-HEAD}(node.parent,m)\\
\begin{IF}{headnode \neq \text{NULL}\hspace*{1in}\COMMENT{If parent's
head has been found, return it}}
   \RETURN headnode.label
\ELSE
   headnode \= \CALL{LEX-HEAD}(node.leftsib,m)\hspace*{.2in}
\COMMENT{Else, left-sibling head}\\ 
   \begin{IF}{headnode \neq \text{NULL}}
     \RETURN headnode.label
   \ELSE
     \RETURN \text{NULL}
   \end{IF}
\end{IF}
\end{algorithm}
\caption{Tree-walking functions to return conditioning values for the
probability model.  {\it node\/} is a pointer to a node in the tree,
which is a data structure with five fields: {\it label\/} which is a
pointer to a character string; and {\it parent\/}, {\it child\/}, {\it
leftsib\/}, and {\it head\/}, which are pointers to other nodes in the
tree. The symbol $\triangleright$ precedes comments.}\label{fig:twalg1}
\end{figure}

\begin{figure}[t]
\begin{algorithm}{LEFT-CCOMMAND}{node}
\begin{WHILE}{node \neq \text{NULL} \text{and} 
node.leftsib = \text{NULL}\hspace*{.3in}\COMMENT{node is leftmost child}}
  node \= node.parent
\end{WHILE}\\
\begin{IF}{node = \text{NULL}}
  \RETURN \text{NULL}
\end{IF}\\
parenthead \= node.parent.head\\
\begin{IF}{parenthead \neq \text{NULL}\hspace*{.3in}\COMMENT{Go to
head of constituent, if found}}
  node \= parenthead
\ELSE
  node \= node.leftsib\hspace*{.3in}\COMMENT{Else, left-sibling}
\end{IF}\\
\RETURN node
\end{algorithm}

\begin{algorithm}{CC-HEAD}{node,m,n}
\begin{FOR}{i \= 1 \TO m}
  \begin{IF}{node \neq \text{NULL}}
    node \= \CALL{LEFT-CCOMMAND}(node)
  \end{IF}
\end{FOR}\\
\RETURN \CALL{CURR-HEAD}(node,n)
\end{algorithm}

\begin{algorithm}{LEFTMOST-CCH}{node,m,n}
\begin{IF}{node.leftsib \neq \text{NULL}\hspace*{1in}\COMMENT{Only for
leftmost children}}
  \RETURN \text{NULL}
\ELSE
  \RETURN \CALL{CC-HEAD}(node,m,n)
\end{IF}
\end{algorithm}

\begin{algorithm}{CONJ-PARALLEL}{node}
\begin{IF}{node \neq \text{NULL} \text{and}
node.leftsib = \text{NULL}}
  node \= node.parent
\end{IF}\\
\begin{IF}{node = \text{NULL}}
  \RETURN \text{NULL}
\end{IF}\\
thislabel \= node.label\\
siblabel \= \CALL{PAR-SIB}(node,0,1)\\
\begin{IF}{siblabel = \text{`CC'}\hspace*{.2in}\COMMENT{If parent is
being conjoined}}
  node \= node.leftsib\\
  \begin{WHILE}{node \neq \text{NULL} node.label \neq
thislabel}
    node \= node.leftsib\hspace*{.2in}\COMMENT{Find first category with same label}
  \end{WHILE}\\
  \begin{IF}{node \neq \text{NULL}}
     node \= node.child\\
     \RETURN node.label\hspace*{.2in}\COMMENT{Return label of first
child of conjoined node}
  \end{IF}
\end{IF}\\
\RETURN \text{NULL}
\end{algorithm}
\caption{More tree-walking functions to return conditioning values for
the probability model.  {\it node\/} is a pointer to a node in the tree,
which is a data structure with five fields: {\it label\/} which is a
pointer to a character string; and {\it parent\/}, {\it child\/}, {\it
leftsib\/}, and {\it head\/}, which are pointers to other nodes in the
tree. The symbol $\triangleright$ precedes comments.}\label{fig:twalg2}
\end{figure}

\begin{figure}[t]
\begin{picture}(385,144)(0,-144)
\put(54,-8){(a)}
\put(52,-30){\small S}
\drawline(54,-34)(20,-44)
\put(13,-52){\small NP}
\drawline(20,-56)(7,-66)
\put(0,-74){\small DT}
\drawline(7,-78)(7,-88)
\put(1,-96){the}
\drawline(20,-56)(32,-66)
\put(25,-74){\small NN}
\drawline(32,-78)(32,-88)
\put(25,-96){cop}
\drawline(54,-34)(89,-44)
\put(82,-52){\small VP}
\drawline(89,-56)(62,-66)
\put(50,-74){\small VBD}
\drawline(62,-78)(62,-88)
\put(54,-96){saw}
\drawline(89,-56)(116,-66)
\put(109,-74){\small NP}
\drawline(116,-78)(87,-88)
\put(80,-96){\small NP}
\drawline(87,-100)(74,-110)
\put(67,-118){\small DT}
\drawline(74,-122)(74,-132)
\put(72,-140){a}
\drawline(87,-100)(100,-110)
\put(92,-118){\small NN}
\drawline(100,-122)(100,-132)
\put(89,-140){thief}
\drawline(116,-78)(146,-88)
\put(140,-96){\small PP}
\put(263,-8){(b)}
\put(261,-30){\small S}
\drawline(264,-34)(219,-44)
\put(212,-52){\small NP}
\drawline(219,-56)(207,-66)
\put(199,-74){\small DT}
\drawline(207,-78)(207,-88)
\put(200,-96){the}
\drawline(219,-56)(232,-66)
\put(224,-74){\small NN}
\drawline(232,-78)(232,-88)
\put(224,-96){cop}
\drawline(264,-34)(310,-44)
\put(303,-52){\small VP}
\drawline(310,-56)(261,-66)
\put(250,-74){\small VBD}
\drawline(261,-78)(261,-88)
\put(253,-96){saw}
\drawline(310,-56)(299,-66)
\put(292,-74){\small NP}
\drawline(299,-78)(287,-88)
\put(280,-96){\small DT}
\drawline(287,-100)(287,-110)
\put(285,-118){a}
\drawline(299,-78)(312,-88)
\put(304,-96){\small NN}
\drawline(312,-100)(312,-110)
\put(302,-118){thief}
\drawline(310,-56)(358,-66)
\put(352,-74){\small PP}
\end{picture}

\vspace*{.5in}

\begin{tabular}{|l|l|l|c|c|}
\hline
\multicolumn{2}{|l|}{Conditioning Function} & Description & \multicolumn{2}{|c|}{Value Returned}\\
\multicolumn{2}{|l|}{}&& Tree (a) & Tree (b) \\\hline
0&PAR-SIB({\it node,1,0}) & Left-hand side (LHS), i.e. parent & NP & VP\\\hline
1&PAR-SIB({\it node,0,1}) & Last child of LHS & NP & NP\\\hline
2&PAR-SIB({\it node,0,2}) & 2nd last child of LHS & NULL & VBD\\\hline
3&PAR-SIB({\it node,0,3}) & 3rd last child of LHS & NULL & NULL\\\hline
4&PAR-SIB({\it node,2,0}) & Parent of LHS (PAR) & VP & S\\\hline
5&PAR-SIB({\it node,1,1}) & Last child of PAR & VBD & NP\\\hline
6&PAR-SIB({\it node,3,0}) & Parent of PAR (GPAR) & S & NULL\\\hline
7&PAR-SIB({\it node,2,1}) & Last child of GPAR & NP & NULL\\\hline
8&CONJ-PARALLEL({\it node}) & First child of conjoined category & NULL &
NULL \\\hline
9&CURR-HEAD({\it node,0}) & Lexical head of current constituent & {\it
thief} & {\it saw}\\\hline
\end{tabular}
\caption{Two trees, to illustrate the tree-walking functions for
non-POS expansions.  The newly hypothesized node in trees (a) and (b)
is the `PP'.  The labels of these new nodes are the conditioned
variables.}\label{fig:twf1} 
\end{figure}

Our conditional probability model is now greatly simplified, given the
uniformity of the conditioning events.  Instead of beginning with
a PCFG rule identifier, which encodes a composite of the parent and the
previous children (recall the non-terminals introduced by
left-factorization), then continuing with values returned from 
tree-walking functions, now all of the conditioning events can be
encoded as values returned from tree-walking functions.  In addition,
our look-ahead probability will also be defined in terms of these
tree-walking functions,
and our new head probability will also be defined in this way.  Hence,
our grammar estimation routine now involves simply taking a given set
of functions and performing maximum likelihood estimation of the
conditioned variable given the values returned from these conditioning 
functions.  This is the same for all three components of our model.

To make this explicit, let us briefly define a small set of treewalking
functions, and give the conditional probability models used in the
trials that will follow.  These functions take a pointer to a node
in the tree as an argument.  Each node in the tree contains, as a part
of its structure, pointers to: (i) its label; (ii) its parent node
({\it parent\/}); (iii) its first child node ({\it child\/}); (iv)
its sibling to the left ({\it leftsib}); and (v) its designated head
child node ({\it head\/}).  The function then moves the node
pointer to other locations in the tree, and
returns a value from the final position of the pointer.  We use these
functions as follows:  hypothesize a new arc and node in the tree, and
pass the function a pointer to the new node.  
Hence all values returned from the functions are relative to
the newly hypothesized node, the label of which is the conditioned
variable.  Most of the functions are given additional parameters, so
each individual function will be identified by a function name and
up to two parameter values.  Figures \ref{fig:twalg1} and
\ref{fig:twalg2} give the algorithms for the tree-walking functions.

Figures \ref{fig:twf1} and \ref{fig:twf2} give the conditional
probability models for non-POS expansions and POS expansions,
respectively.  Each model is a linear order of functions, and the
probability of the conditioned event is conditioned on the values
returned by these functions.  The figures also provide two example
trees each, with a newly hypothesized node (the conditioned variable),
and the values that would be returned from each of the tree-walking
functions.  As before, the conditional probability estimate with $n$
features is the linear interpolation of the MLP relative frequency
estimate for $n$ features and the conditional probability estimate
with $n-1$ features.  The order in which these models is presented is
the order of interpolation.

\begin{figure}[t]
\begin{picture}(385,144)(0,-144)
\put(54,-8){(a)}
\put(52,-30){\small S}
\drawline(54,-34)(20,-44)
\put(13,-52){\small NP}
\drawline(20,-56)(7,-66)
\put(0,-74){\small DT}
\drawline(7,-78)(7,-88)
\put(1,-96){the}
\drawline(20,-56)(32,-66)
\put(25,-74){\small NN}
\drawline(32,-78)(32,-88)
\put(25,-96){cop}
\drawline(54,-34)(89,-44)
\put(82,-52){\small VP}
\drawline(89,-56)(62,-66)
\put(50,-74){\small VBD}
\drawline(62,-78)(62,-88)
\put(54,-96){saw}
\drawline(89,-56)(116,-66)
\put(109,-74){\small NP}
\drawline(116,-78)(87,-88)
\put(80,-96){\small NP}
\drawline(87,-100)(74,-110)
\put(67,-118){\small DT}
\drawline(74,-122)(74,-132)
\put(72,-140){a}
\drawline(87,-100)(100,-110)
\put(92,-118){\small NN}
\drawline(100,-122)(100,-132)
\put(89,-140){thief}
\drawline(116,-78)(146,-88)
\put(140,-96){\small PP}
\drawline(146,-100)(130,-110)
\put(124,-118){\small IN}
\drawline(130,-122)(130,-132)
\put(120,-140){with}
\put(263,-8){(b)}
\put(261,-30){\small S}
\drawline(264,-34)(219,-44)
\put(212,-52){\small NP}
\drawline(219,-56)(207,-66)
\put(199,-74){\small DT}
\drawline(207,-78)(207,-88)
\put(200,-96){the}
\drawline(219,-56)(232,-66)
\put(224,-74){\small NN}
\drawline(232,-78)(232,-88)
\put(224,-96){cop}
\drawline(264,-34)(310,-44)
\put(303,-52){\small VP}
\drawline(310,-56)(261,-66)
\put(250,-74){\small VBD}
\drawline(261,-78)(261,-88)
\put(253,-96){saw}
\drawline(310,-56)(299,-66)
\put(292,-74){\small NP}
\drawline(299,-78)(287,-88)
\put(280,-96){\small DT}
\drawline(287,-100)(287,-110)
\put(285,-118){a}
\drawline(299,-78)(312,-88)
\put(304,-96){\small NN}
\drawline(312,-100)(312,-110)
\put(302,-118){thief}
\drawline(310,-56)(358,-66)
\put(352,-74){\small PP}
\drawline(358,-78)(342,-88)
\put(337,-96){\small IN}
\drawline(342,-100)(342,-110)
\put(332,-118){with}
\end{picture}

\vspace*{.5in}

\begin{tabular}{|l|l|l|c|c|}
\hline
\multicolumn{2}{|l|}{Conditioning Function} & Description & \multicolumn{2}{|c|}{Value Returned}\\
\multicolumn{2}{|l|}{}&& Tree (a) & Tree (b) \\\hline
0&PAR-SIB({\it node,1,0}) & Left-hand side (LHS), i.e. parent & IN & IN\\\hline
1&PAR-SIB({\it node,2,0}) & Parent of LHS (PAR) & PP & PP\\\hline
2&PAR-SIB({\it node,1,1}) & Last child of PAR & NULL & NULL\\\hline
3&LEFTMOST-PS({\it node,3,0}) & Parent of PAR (GPAR) & NP & VP\\\hline
4&LEFTMOST-CCH({\it node,1,1}) & POS of C-Commanding head & NN & VBD \\\hline
5&CC-HEAD({\it node,1,0}) & C-Commanding head & {\it thief}
& {\it saw} \\\hline
6&CC-HEAD({\it node,2,0}) & Next C-Commanding head & {\it saw} &
{\it cop} \\\hline
\end{tabular}
\caption{Two trees, to illustrate the tree-walking functions for
POS expansions.  The newly hypothesized node in both trees (a) and (b)
is the word `with'.  The labels of these new nodes are the conditioned
variables.}\label{fig:twf2} 
\end{figure}

The second part of the probability model is the probability that the
previous child of the constituent is the head of the constituent.
There are three possible cases: (i) the head has already been found to
the left of the previous child; (ii) the previous child is the head;
or (iii) none of the previous children is the head of the constituent.
For example, when the new node is built in the figure \ref{fig:twf1}a,
the probability for (i) above is zero, since the previous child of the
NP is the first child; the probabilities for (ii) and (iii) must be
estimated.   The conditioned variable is one of the three above
alternatives.  The head probability model that we used in these trials
consisted entirely of values returned by the PAR-SIB function with the
following parameters: (0,1), (1,0), (0,0), (0,2), and (0,3).  In
words, we are conditioning the head location on: (0) the label of the
previous child; (1) the left-hand side (i.e. parent label of the newly
hypothesized node); (2) the label of the newly hypothesized
node; (3) the label of the 2nd child to the left; and (4) the label of
the 3rd child to the left.  Once the head is identified as the
previous child, that selection is fixed for that candidate analysis
from that point forward.  For every
rule expansion, more than one analysis must be considered, depending
on the range of possibilities for head assignment.

One possible concern would be the use of the new node label to
condition the head probability, and also the lexical head of the
constituent to condition the 
probability of the new node label.  The way that the CURR-HEAD
function is defined however, is that, if the head of the constituent
has not been assigned yet, it selects the head of the previous child.
Hence the head probability can be evaluated after the rule expansion
probability.

The look-ahead probability (LAP) is defined in exactly the way it was
in the previous section, except that instead of being conditioned on
the composite category created through factorization, it is
conditioned on the label of the current category and the three
previously emitted children.  

Table \ref{tab:smoo1} gives results using this new probability model
with the same training and testing sections presented in the previous
section, along with the results with these same conditioning features
from the previous chapter\footnote{The conditional probability models
that are presented here are, with the exception of the Markov grammar
smoothing, identical to the models used at the end of the previous
chapter.}.  Even though the conditioning events are 
the same, our accuracy improves by nearly one percentage point, while
our coverage goes to 100 percent.  The rule expansions considered for
the same beam definition increases by a third, but this is hardly
surprising.  The number of productions in the original form
(i.e. after being de-transformed) that now have probability mass is
infinite, as opposed to the previous, unsmoothed grammar of about
15,000 rules.

\begin{table*}[t]
\begin{tabular}{|l|c|c|c|c|c|c|r|c|}
\hline
{\small Grammar} & {\small LR} & {\small LP} & {\small CB} & 
{\small 0 CB} & {\small $\leq$ 2} & {\small Pct.} &
{\small Avg. rule\ \ } & {\small Average}\\
{} & {} & {} & {} & {} & {\small CB} & {\small failed} & {\small
expansions\ } & {\small analyses}\\
{} & {} & {} & {} & {} & {} & {} & {\small considered${}^{\dag}$} & {\small advanced${}^{\dag}$}\\\hline
\multicolumn{9}{|c|}{section 23: 2416 sentences of length $\leq$ 100}\\\hline
{PCFG} & {85.7} & {85.7} & {1.41} & {59.0} & {79.9} & {1.7} & {6,709}
& {207.6}\\\hline
{Smoothed} & {86.4} & {86.8} & {1.31} & {59.5} & {81.6} & {0} & {9,008}
& {198.9}\\\hline
\end{tabular}\\
\begin{footnotesize}
${}^{\dag}$per word
\end{footnotesize}
\caption{Parsing results using the conditional probability model from chapter 3, with a smoothed (Markov) grammar of order 3, versus a PCFG backbone.  Results are trained on sections 2-21 and tested on section 23.}\label{tab:smoo1}
\end{table*}

\begin{table*}[t]
\begin{tabular}{|l|c|c|c|c|c|c|r|c|}
\hline
{\small Base Beam} & {\small LR} & {\small LP} & {\small CB} & 
{\small 0 CB} & {\small $\leq$ 2} & {\small Pct.} &
{\small Avg. rule\ \ } & {\small Average}\\
{Factor} & {} & {} & {} & {} & {\small CB} & {\small failed} & {\small
expansions\ } & {\small analyses}\\
{} & {} & {} & {} & {} & {} & {} & {\small considered${}^{\dag}$} & {\small advanced${}^{\dag}$}\\\hline
\multicolumn{9}{|c|}{section 23: 2416 sentences of length $\leq$ 100}\\\hline
{$10^{-11}$} & {86.4} & {86.8} & {1.31} & {59.5} & {81.6} & {0} &
{9,008} & {198.9}\\\hline
{$10^{-10}$} & {86.2} & {86.5} & {1.34} & {59.2} & {81.4} & {0} &
{5,528} & {120.0}\\\hline
{$10^{-9}$} & {86.1} & {86.4} & {1.36} & {59.0} & {81.1} & {0} &
{3,439} & {72.6}\\\hline
{$10^{-8}$} & {85.6} & {85.9} & {1.41} & {58.3} & {80.3} & {0} &
{2,159} & {43.9}\\\hline
{$10^{-7}$} & {85.3} & {85.0} & {1.49} & {56.8} & {79.3} & {0} &
{1,374} & {26.6}\\\hline
{$10^{-6}$} & {84.2} & {84.5} & {1.59} & {55.3} & {77.9} & {0} &
{898} & {16.2}\\\hline
\end{tabular}\\
\begin{footnotesize}
${}^{\dag}$per word
\end{footnotesize}
\caption{Parsing results using the new conditional probability model,
with a variety of base beam factors.  Results are trained on sections
2-21 and tested on section 23.}\label{tab:smoo2} 
\end{table*}

Table \ref{tab:smoo2} gives results with a variety of base beam
factors.  Recall that the beam threshold is defined as a variable
probability range.  For a given base beam factor $\gamma$, we define
the beam as $\gamma|\scH_{i+1}|^{3}$, i.e. the range narrows with the
cube of the number of analyses advanced.  The results in table
\ref{tab:smoo2} indicate that, with the new model, the beam can be
greatly narrowed without losing much accuracy, and maintaining
complete coverage.  At a $\gamma$ = $10^{-9}$, the parser loses less
than half a point of either precision or recall, while considering
fewer than forty percent of the rule expansions that were considered at
the widest beam.  Recall that this measure correlates nearly perfectly
with time \cite{Roark00b}, so there is an equivalent speedup.

Because of the improvement that this approach provides over the
unsmoothed PCFG approach in the previous chapter, we will consider the
smoothed grammar parser our {\it standard\/} parser from this point
forward. We will refer to the parser from the previous chapter as the
{\it base\/} parser for comparison purposes.

\section{Left-recursion}
Left-recursion, as has been mentioned several times throughout the
course of this thesis, is a problem for top-down parsers.  Our results
up to this point have demonstrated that it is possible for a top-down
parser to efficiently build enough structure to find good parses, even
in the face of left-recursion.  This is because of the nature of the
beam-search.  It is permissive enough to retain analyses with a
certain number of left-recursive expansions, and this number seems
generally sufficient.  What we have not done is investigate this issue
in detail.  How much of a problem is left-recursion, and if it is a
problem, how can we improve performance in the face of it?

Let us consider this question in terms of the length of left-child
chains, i.e. chains of leftmost children.  At each terminal item in
the tree, we can count the number of consecutive non-terminals above
it until we reach one that is not the leftmost child within its
constituent (or the root).  We will call this the left-child chain for that
particular word. For example, in the tree in figure \ref{fig:twf2}a,
both determiners have left-child chains with three categories: {\it
the\/} at the beginning of the sentence has DT, NP, and S above it;
{\it a\/} in the object NP has DT, NP, and NP above it.  Because
lowest non-terminal in a chain is always a POS non-terminal, we will
generally omit them from the left-child chains.  Hence, we would count
each of these left-child chains to be of length 2. 

These left-children chains are where the left-recursion will occur.
Because of the beam-search, long chains of left-recursive categories
-- e.g. seven consecutive NP left children -- are not a priori more of
a problem than other long sequences of left children -- e.g. seven
consecutive left children of mixed categories -- since a threshold
exists by which all sequences over some bound will eventually be
pruned.  It could be, however, that the bulk of the long left-child
chains include left-recursive productions.  

The first step to investigate this issue is to evaluate the parser
with respect to its performance upon these left-child chains.
We want the parser to build left-child chains as long as needed, but
no longer.   Ideally, the parser would build exactly enough to find the
actual parse, although this ideal is clearly not achievable with an
incremental parser, which cannot know in advance how many will be
needed.  There are two ways in which the parser can fall short of this
ideal:  by not building left-child chains of sufficient length to
find the correct parse; and by building chains that are too long,
i.e. that can never be used, and thus wasting effort.

\begin{table}[t]
\begin{tabular}{|l|l|c|c|c|c|c|c|c|}
\hline
position & type of & total & \multicolumn{6}{|c|}{Percent with depth}\\
in sentence & chain & count & 2 & 3 & 4 & 5 & 6 & $>$6 \\\hline
all words & with recursion & 84,954 & 70.4 & 14.9 & 12.7 & 1.9 & 0.15 & 0.005 \\\cline{2-9}
(950,028) & no recursion & 95,138 & 68.1 & 31.0 & 0.9 & 0.02 & 0 & 0 \\\hline
first word & with recursion & 11,357 & 0 & 7.3 & 78.3 & 13.2 & 1.1 & 0.04 \\\cline{2-9}
(39,832) & no recursion & 28,475 & 20.4 & 76.6 & 2.9 & 0.07 & 0 & 0 \\\hline
not first & with recursion & 73,597 & 81.2 & 16.0 & 2.6 & 0.2 & 0.01 & 0 \\\cline{2-9}
(910,196) & no recursion & 66,663 & 88.5 & 11.4 & 0.05 & 0.001 & 0 & 0 \\\hline
\end{tabular}
\caption{Left-child chain counts from sections 2-21 of the Penn Wall
St. Journal Treebank.  Counts for the position in the sentence are
total number of words in that position. Counts for the chains are for
chains of length $>$ 1, where only non-POS categories are counted in a
chain.} \label{tab:221lr}
\end{table}

What is the extent of left-child chains in the training data that we
are attempting to model?  We attempted to answer this question by
collecting all of the left-child chains consisting of two or more
non-POS non-terminals, from sections 2-21 of the Penn Wall St. Journal
Treebank.  Table \ref{tab:221lr} summarizes these counts, with counts
at all word positions, at just the first word of each sentence, and
at all words except the first word.  The first word will always have a
left-child chain in the way that we have defined it, because the chain
will of necessity include the root category and the non-terminal
spanning the entire string, e.g. S.  We have split the data into cases
where some non-terminal category occurs more than once in the chain --
i.e. there is left-recursion -- and those where no such recursion
occurs.  Looking first at words other than the first word of the
sentence, we can see that there are slightly more left-child chains
with left-recursion than without, and those with recursion tend to be
slightly longer.  Even so, very few such chains occur with a depth
greater than four.  Very few chains occur with a depth greater than
three, unless left-recursion is involved.  At the first word, there
are longer chains, with some number occurring at depth six with
recursion.

\begin{table}[t]
\begin{tabular}{|l|l|r|r|r|r|r|r|r|}
\hline
type of & consecutive & total & Percent of & \multicolumn{5}{|c|}{Count
with depth}\\
recursion & non-terminal & count & rec. chains & 2 & 3 & 4 & 5 & $>$5
\\\hline
non-consecutive & & 58 & 0.07 & 0 & 38 & 15 & 4 & 1\\\hline
consecutive & NP & 74,826 & 88.08 & 69,566 & 5,100 & 157 & 3 & 0\\\cline{2-9}
& S & 4,740 & 5.58 & 4,608 & 132 & 0 & 0 & 0\\\cline{2-9}
& VP & 3,664 & 4.31 & 3,655 & 9 & 0 & 0 & 0\\\cline{2-9}
& ADJP & 728 & 0.86 & 704 & 24 & 0 & 0 & 0\\\cline{2-9}
& PP & 651 & 0.77 & 647 & 5 & 0 & 0 & 0\\\cline{2-9}
& NX & 406 & 0.48 & 398 & 8 & 0 & 0 & 0\\\cline{2-9}
& ADVP & 338 & 0.40 & 330 & 7 & 1 & 0 & 0\\\cline{2-9}
& SBAR & 337 & 0.40 & 337 & 0 & 0 & 0 & 0\\\cline{2-9}
& Other & 101 & 0.12 & 99 & 2 & 0 & 0 & 0\\\hline
\end{tabular}
\caption{Recursive left-child chain counts from sections 2-21 of the
Penn Wall St. Journal Treebank.  Percentage may sum to more than 100,
due to the fact that a single recursive chain may hold consecutive
recursion for more than category.} \label{tab:221r}
\end{table}

Let us look in more detail at the recursive chains.  Table
\ref{tab:221r} gives counts for recursive chains.  We split the
recursion into those containing only consecutive recursion --
i.e. where the same category occurs as its own leftmost child -- and
that which is non-consecutive, i.e. where a recursive category occurs
not as its own child, but as the child of a descendent.
Interestingly, this latter type of recursive chain occurs very rarely,
and then most frequently with certain short chains, such as 
SBAR~$\rightarrow$~S~$\rightarrow$~SBAR and
NP~$\rightarrow$~ADJP~$\rightarrow$~NP.  Consecutive NP recursion
accounts for over 88 
percent of all recursive chains, and NP, S, and VP together account
for 98 percent of them!  Only consecutive NP recursive chains include
significant counts beyond a depth of two.

The question that we might attempt to answer now is whether or not our
probability model as it currently stands sufficiently models the
probability of these left-child chains.  Let us focus upon the
consecutive NP case, since this is where the bulk of the deepest
recursion occurs.  If we can successfully model this, the probability
of continuing an NP left-child chain should drop off the longer the
chain gets.  By virtue of conditioning our rule probabilities on the
parent and closest sibling up to the grandparent of the left-hand
side, our model does drop the rule probability for a left-recursive
child the deeper the chain -- up to a point.  

Let us calculate the probabilities for some chains given our model.
Recall that we will calculate the child given the left-hand side of
the rule ({\it lhs\/}), the previous three children of the {\it
lhs\/} ({\it c\/}), the parent of {\it lhs\/} ({\it p\/}), the
previous child of {\it p\/} ({\it sib\/}), the parent of {\it p\/}
({\it gp\/}), and the previous child of {\it gp\/} ({\it gsib\/}).
The probability of an NP occurring as the left-child of another NP is,
overall, about .25, so we can use this as our starting point.  Suppose
the parent of the first NP in the chain is an S, and that it is the
first child of the S.  Let $\emptyset$ be the NULL value.
Then, using simple relative frequency from f2-21:
\begin{eqnarray}
\Pr(\mathrm{\small NP}|\mathrm{\it lhs}=\mathrm{\small NP},\mathrm{\it
c}=\emptyset,\mathrm{\it p}=\mathrm{\small S},\mathrm{\it sib}=\emptyset,\ldots)
&=& .25 \\
\Pr(\mathrm{\small NP}|\mathrm{\it lhs}=\mathrm{\small NP},\mathrm{\it
c}=\emptyset,\mathrm{\it p}=\mathrm{\small NP},\mathrm{\it sib}=\emptyset,\mathrm{\it gp}=\mathrm{\small S},\mathrm{\it gsib}=\emptyset)
&=& .07 \\
\Pr(\mathrm{\small NP}|\mathrm{\it lhs}=\mathrm{\small NP},\mathrm{\it
c}=\emptyset,\mathrm{\it p}=\mathrm{\small NP},\mathrm{\it sib}=\emptyset,\mathrm{\it gp}=\mathrm{\small NP},\mathrm{\it gsib}=\emptyset)
&=& .03 
\end{eqnarray}
Note that this decrease in probability as the chain grows is already
in our model.  However, each subsequent link in the chain after this
last one will have the same probability, since our conditional
probability model forgets about the chain after the grandparent of the
left-hand side.  This may or may not be a problem, since the drop off
in probability is quite large already.  If the lookahead word is, for
example, \textttt{`the'}, let the NP analysis with no recursion have a
probability $p_0$.  Then with one level of recursion, the probability
would be $.25p_0$; with two, $.0175p_0$; three, $.0005p_0$; four,
approximately $10^{-5}p_0$.  Each subsequent expansion would reduce
this by a factor of approximately $10^{-2}$.  Hence, these structures
should fall off of even a fairly wide beam relatively quickly.  

Which brings up the issue of how well the parser actually does in
dealing with these left-child chains.  To measure this, we collected
the left-child chains of length greater than or equal to two non-POS
non-terminals from the test corpus, and looked for whether or not they
were built and stayed within the beam.  Of the 56,684 lexical items in
the test set, 10,837 had left-child chains of length two or more.  We
looked for the presence of these chains in any of the candidate
analyses of our standard parser (i) at the word where they should have
been constructed; and (ii) at the word where they should have been
closed.  If the chains are present in our set of candidate analyses,
then they are evaluated with our probability model, which is all we
can ask.  Of the 10,837 chains, only 184 were missing from our set of
candidate analyses at the word 
where they should have been closed; of these, 87 were missing at the
word where they should have been constructed, i.e. they were never in
the beam at all.  Note that we do not know the reason why they were
missing from the beam.  It could be that the parser garden pathed at
some other 
location, creating a syntactic context in which these chains are no
longer viable.  Of the chains that were not present, 36 were with
non-S nodes at the root of the tree, e.g. an NP or VP rather than a
full S;  only 6 of these were not in the beam at the first word.
The parser did return as the most likely parse 306 such trees, so it
is not completely S biased.  

\begin{figure*}[t]
\begin{center}
\epsfig{file=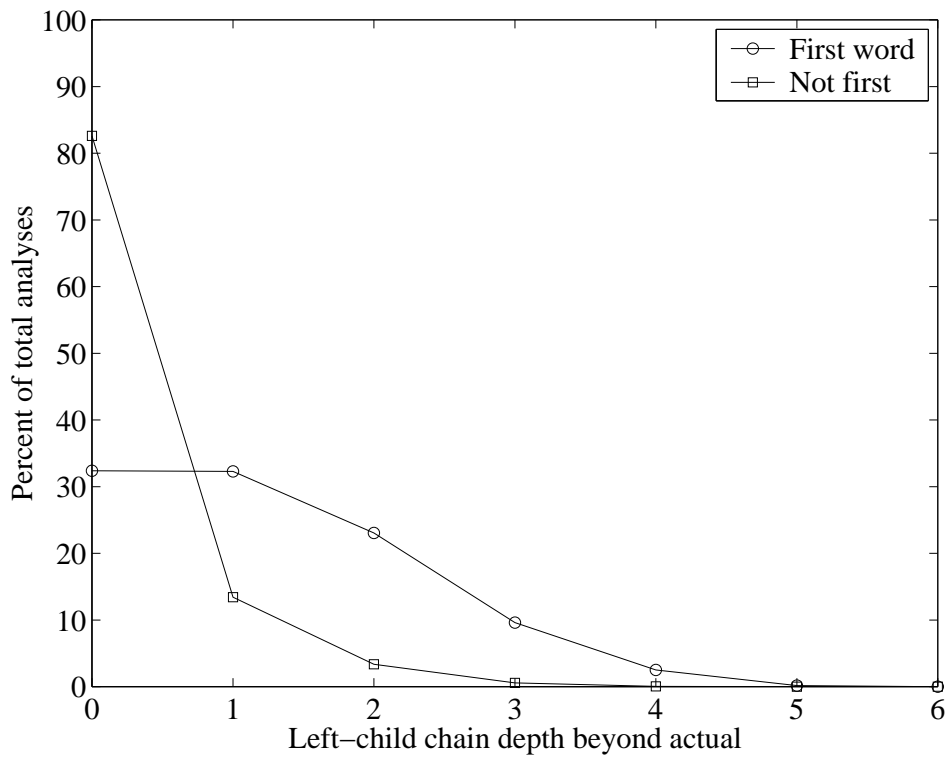, width = 5in}
\end{center}
\caption{Percentage of candidate analyses at a left-child chain
depth beyond the correct parse, at the first word in the sentences and
at other words of the sentences}\label{fig:lcd1}
\end{figure*}

Thus, overall 98.3 percent of correct left-child chains of length
greater than one were in the set of candidate parses when they were
closed.  Thus we do not seem to be building too few left-child chains
to find the correct analysis.  We may be poorly modeling these chains,
however, and build too many.  To evaluate this, we counted the depth
of left-child chains being built for each analysis staying within the
beam threshold at each word in the test corpus.  As shown above, the
length of left-child chains at the first word of the sentence is
typically longer than at other points in the sentence, so all of the
data that we collected is divided between the first position in the
sentence and all other positions.  Figure \ref{fig:lcd1} plots the
percentage of candidate analyses that built a left-child chain at a
word at a depth $n$ beyond the depth of the correct parse.  
Over 80 percent of candidate analyses at word positions that are not
sentence initial built the right depth or less.  However, nearly 70
percent of all sentence initial candidate analyses built left-child
chains beyond the length necessary for the correct analysis.

\begin{figure*}[t]
\begin{center}
\epsfig{file=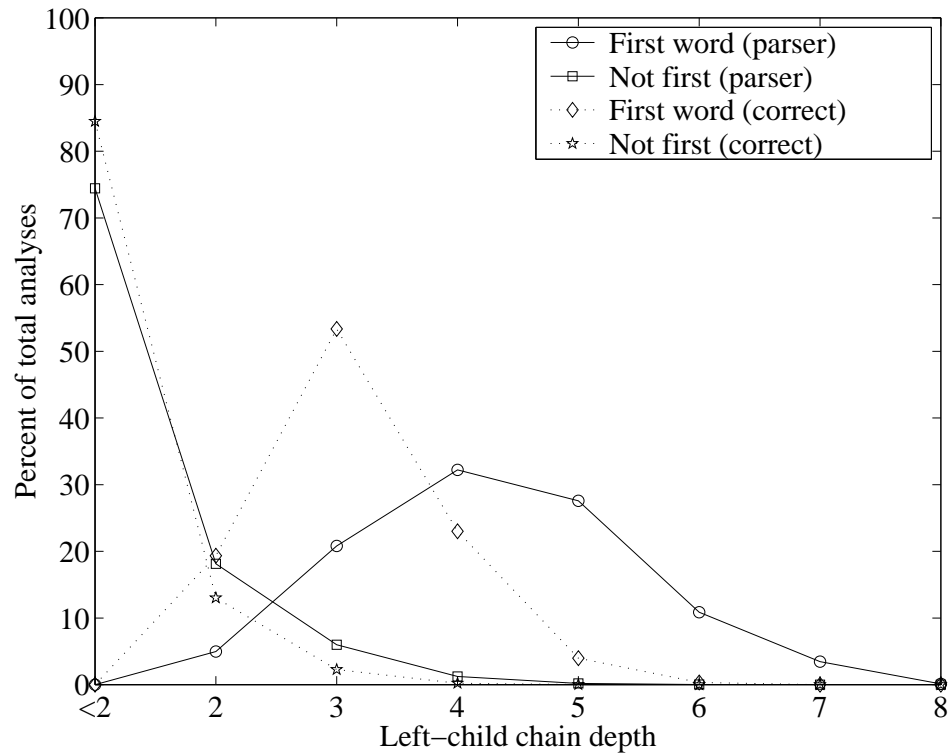, width = 5in}
\end{center}
\caption{Percentage of candidate analyses and percentage of correct
parses with a left-child chain depth, at the first word in the
sentences and at other words of the sentences}\label{fig:lcd2}
\end{figure*}

Figure \ref{fig:lcd2} shows the percentages at particular depths of
left-child chains for both the correct parses and our candidate
analyses, again split by whether or not the word is sentence initial.
From this we can see that the parser seems to be spending its time
building left-child chains in approximately the right proportion away
from the sentence initial position, by which we mean that it is
building the most analyses with the most common depths.  At the
sentence initial position, however, the parser is building more
candidate analyses at less frequent depths than at more frequent
depths.  This is perhaps not surprising, given that the combinatorics
dictate that the possibilities at depth 3 are exponentially greater in
number than at depth 2.  Nevertheless, it seems that more effort than
is necessary is being spent on long left-child chains in the sentence
initial position.

Given that the bulk of left-recursion comes from consecutive NP
left-children, and that the longer left-child chains result from
left-recursive chains, one approach to spending less time on building
lengthy chains is to perform a selective left-corner transform on
productions with an NP parent and left-child.  The transform schemata
was presented in chapter 3, on page 66.  The basic idea is that NP~$\rightarrow$~NP~$\alpha$ productions will be recognized left-corner,
and all other productions top-down.  As mentioned earlier, the
left-corner transform turns left-recursion into right-recursion, so
that long NP left-child chains will no longer be built.  The negative
to this transform is that it underspecifies the immediate dominance
links, and hence some of the conditioning information that is used in
our model will be unavailable with these productions.  The hope is
that we can disrupt the immediate dominance links for only very few
productions, while taking care of the bulk of the left-recursion
problem. 

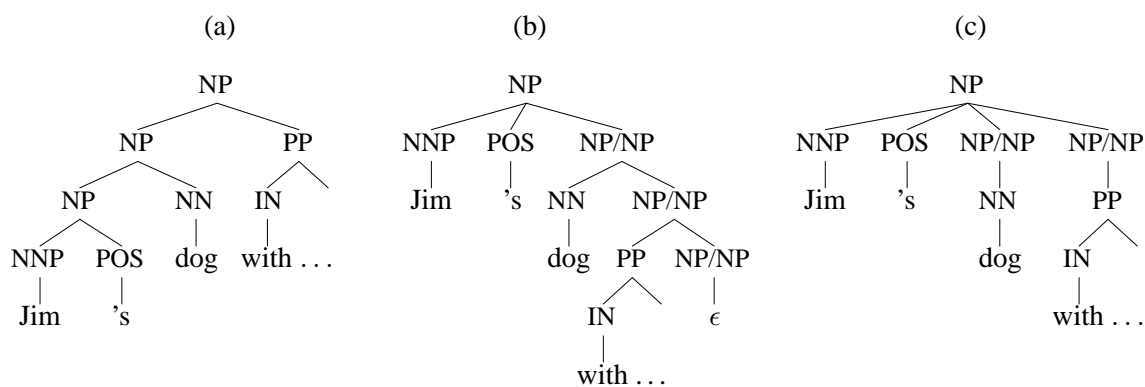
\begin{figure}[t]
\begin{picture}(441,144)(0,-144)
\put(73,-8){(a)}
\put(71,-30){\small NP}
\drawline(78,-34)(48,-44)
\put(41,-52){\small NP}
\drawline(48,-56)(26,-66)
\put(19,-74){\small NP}
\drawline(26,-78)(11,-88)
\put(0,-96){\small NNP}
\drawline(11,-100)(11,-110)
\put(3,-118){Jim}
\drawline(26,-78)(42,-88)
\put(32,-96){\small POS}
\drawline(42,-100)(42,-110)
\put(38,-118){'s}
\drawline(48,-56)(70,-66)
\put(62,-74){\small NN}
\drawline(70,-78)(70,-88)
\put(62,-96){dog}
\drawline(78,-34)(109,-44)
\put(103,-52){\small PP}
\drawline(109,-56)(97,-66)
\put(92,-74){\small IN}
\drawline(97,-78)(97,-88)
\put(87,-96){with $\ldots$}
\drawline(109,-56)(120,-66)

\put(190,-8){(b)}
\put(188,-30){\small NP}
\drawline(195,-34)(159,-44)
\put(148,-52){\small NNP}
\drawline(159,-56)(159,-66)
\put(151,-74){Jim}
\drawline(195,-34)(189,-44)
\put(180,-52){\small POS}
\drawline(189,-56)(189,-66)
\put(186,-74){'s}
\drawline(195,-34)(231,-44)
\put(215,-52){\small NP/NP}
\drawline(231,-56)(211,-66)
\put(203,-74){\small NN}
\drawline(211,-78)(211,-88)
\put(203,-96){dog}
\drawline(231,-56)(251,-66)
\put(235,-74){\small NP/NP}
\drawline(251,-78)(235,-88)
\put(229,-96){\small PP}
\drawline(235,-100)(224,-110)
\put(218,-118){\small IN}
\drawline(224,-122)(224,-132)
\put(214,-140){with \ldots}
\drawline(235,-100)(246,-110)
\drawline(251,-78)(266,-88)
\put(251,-96){\small NP/NP}
\drawline(266,-100)(266,-110)
\put(264,-118){$\epsilon$}

\put(357,-8){(c)}
\put(355,-30){\small NP}
\drawline(362,-34)(308,-44)
\put(297,-52){\small NNP}
\drawline(308,-56)(308,-66)
\put(300,-74){Jim}
\drawline(362,-34)(339,-44)
\put(329,-52){\small POS}
\drawline(339,-56)(339,-66)
\put(335,-74){'s}
\drawline(362,-34)(374,-44)
\put(359,-52){\small NP/NP}
\drawline(374,-56)(374,-66)
\put(366,-74){\small NN}
\drawline(374,-78)(374,-88)
\put(366,-96){dog}
\drawline(362,-34)(415,-44)
\put(400,-52){\small NP/NP}
\drawline(415,-56)(415,-66)
\put(409,-74){\small PP}
\drawline(415,-78)(404,-88)
\put(398,-96){\small IN}
\drawline(404,-100)(404,-110)
\put(394,-118){with $\ldots$}
\drawline(415,-78)(426,-88)
\end{picture}
\caption{Three representations of the NP modifications: (a) the
original grammar representation; (b) Selective left-corner
representation; and (c) a flat structure that is unambiguously equivalent to (b)}\label{fig:npslc} 
\end{figure}

Consider figure \ref{fig:npslc}, which gives three representations of
an NP constituent.  The first (figure \ref{fig:npslc}a) is the tree
with the original grammar.  The second tree (figure \ref{fig:npslc}b)
is the result of the selective left-corner transform on
NP~$\rightarrow$~NP~$\alpha$ productions.  The selective left-corner
transform turns the left-branching NP structure in figure
\ref{fig:npslc}a to the right-branching structure in \ref{fig:npslc}b,
by first building the categories from inside of the lowest NP in the
structure, then nesting the further modifications into a 
right-branching structure, through the use of the slash category,
(NP/NP).  Since the only production that we are transforming are these
NP productions, the slash categories will always be NP/NP.  A second
transform can deterministically flatten this structure to that in
figure \ref{fig:npslc}c.  We will call this flattened transform the
{\em flattened selective left-corner transform} with respect to
a non-terminal $A$, or $\FLC_A$.  We transform the trees in the
training corpus, estimate the parameters (using the smoothed Markov
grammar approach), then de-transform the parses returned from the
parser.

This transform with respect to NP does remove the consecutive NP
left-child chains that make up such a large proportion of the
left-child chains that we observed.  Note that it does not remove all
left-recursion from the grammar, not even all NP left-recursion.  The
left-child categories of the NP can rewrite to other non-terminals as
left-child, e.g. S, which can then have an NP as their left-child.  We
have seen, however, that this kind of non-consecutive recursion is
relatively rare. 

One issue that is important to keep in mind when using a grammar of
this sort is that the slash categories must always be preceded by some
non-slash categories, and not followed by anything other than
subsequent slash categories.  If, however, we condition the probability of
these rules using the smoothed Markov PCFG approach that we outlined
in the previous section, then some probability mass will be reserved
for rules in which the NP/NP slash categories precede another category,
such as NN.  This does not correspond to anything in the original
grammar.  In order to remove the probability mass from these
un-interpretable structures, we can model these as a linear
interpolation in which the mixing parameter when going from a
1st-order Markov grammar to a 0th-order is set to zero, i.e. no
probability mass is contributed by the 0-order model.

\begin{figure}[t]
\begin{algorithm}{LC-CHAIN}{node,m}
\begin{FOR}{i \= 1 \TO m \hspace*{1in}\COMMENT{Move up $m$ nodes, if
no left-sibling}}
  \begin{IF}{node \neq \text{NULL} \text{and} node.leftsib = \text{NULL}}
    node \= node.parent
  \ELSE
    \RETURN \text{NULL}
  \end{IF}
\end{FOR}\\
\begin{IF}{node \neq \text{NULL}}
  \RETURN node.label
\ELSE
  \RETURN \text{NULL}
\end{IF}
\end{algorithm}
\caption{A new tree-walking function to condition on values in the
left-child chain beyond the grandparent of the left-hand
side.  {\it node\/} is a pointer to a node in the tree,
which is a data structure with five fields: {\it label\/} which is a
pointer to a character string; and {\it parent\/}, {\it child\/}, {\it
leftsib\/}, and {\it head\/}, which are pointers to other nodes in the
tree. The symbol $\triangleright$ precedes comments.}\label{fig:lcchain} 
\end{figure}

This transform is one way to try to improve the efficiency with which
the parser deals with left-child chains.  There are two other simple
modifications to the original model that we also tried, which do not
change the structure of the grammar.  The first has to do with the
fact that our model forgets about ancestors beyond the grandparent of
the left-hand side.  For leftmost children, we could easily add in
features which look for links in a left-child chain beyond the
grandparent of the left-hand side.  These features would not interfere
with the conditioning provided by other features beyond the
grandparent of the left-hand side, since those functions
(CONJ-PARALLEL and CURR-HEAD) only provide non-NULL values when there
is a left-sibling somewhere below the grandparent.  Figure
\ref{fig:lcchain} provides the function LC-CHAIN to provide values
along the left-child chain.  We augmented our conditional probability
model for non-POS expansions with two functions after all of the
PAR-SIB functions: LC-CHAIN({\it node},4), and LC-CHAIN({\it
node},5).  

The final method that we investigated for improving our efficiency in
the face of left-recursion was very simple.  Since the parser seems to
be spending too much time on long left-child chains at the sentence
initial position, we narrowed our base beam factor only at the
sentence initial position.  In these trials, the base beam factor is
the standard $10^{-11}$ everywhere except in sentence initial
position, where it is $10^{-7}$ in one trial and $10^{-6}$ in another.
Everything else in these trials is left as it was in the original
smoothed grammar trials.

\begin{table*}[t]
\begin{tabular}{|l|c|c|c|c|c|c|r|c|}
\hline
{\small Model} & {\small LR} & {\small LP} & {\small CB} & 
{\small 0 CB} & {\small $\leq$ 2} & {\small Pct.} &
{\small Avg. rule\ \ } & {\small Average}\\
{} & {} & {} & {} & {} & {\small CB} & {\small failed} & {\small
expansions\ } & {\small analyses}\\
{} & {} & {} & {} & {} & {} & {} & {\small considered${}^{\dag}$} & {\small advanced${}^{\dag}$}\\\hline
\multicolumn{9}{|c|}{section 23: 2416 sentences of length $\leq$ 100}\\\hline
{Original Smoothed} & {86.4} & {86.8} & {1.31} & {59.5} & {81.6} & {0} & {9,008} & {198.9}\\\hline
{$\FLC_{\mathrm{NP}}\ \ (10^{-11})$} & {86.6} & {87.1} & {1.27} & {60.4} & {82.3} & {0.04} & {9,388} & {179.0}\\\hline
{$\FLC_{\mathrm{NP}}\ \ (10^{-10})$} & {86.4} & {86.8} & {1.28} & {60.0} & {82.0} & {0.04} & {5,625} & {104.0}\\\hline
{LC-CHAIN features} & {86.4} & {86.8} & {1.31} & {59.4} & {81.5} & {0.04} & {9,006} & {198.8}\\\hline
{First beam $10^{-7}$} & {86.4} & {86.8} & {1.30} & {59.6} & {81.6} & {0} & {8,829} & {193.6}\\\hline
{First beam $10^{-6}$} & {86.4} & {86.8} & {1.30} & {59.6} & {81.6} & {0} & {8,769} & {192.2}\\\hline
\end{tabular}\\
\begin{footnotesize}
${}^{\dag}$per word
\end{footnotesize}
\caption{Parsing results using the $\FLC_\mathrm{NP}$ transform at two
different base beam factors, the
LC-CHAIN features, and a narrow beam at the sentence initial position
only, compared with the original smoothed grammar results.  Results are trained on sections 2-21 and tested on section 23.}\label{tab:smoo3}
\end{table*}

Table \ref{tab:smoo3} reports the results of our four trials to
improve the efficiency with which our parser deals with left-child
chains, alongside the original smoothed grammar results.  The
flattened selective left-corner transform at a base beam factor of
$10^{-11}$ (the parameter for our standard parser) resulted in a slight
improvement in the accuracy over the original grammar, but more
expansions per word.  One reason why the accuracy might have improved
has to do with the way that our beam search is defined.  As explained
in chapter 3, the beam is a function of the base beam factor and the
number of successful analyses:  as the number of good parses grows,
the probability range narrows.  With the $\FLC_\mathrm{NP}$ transform,
the number of these left-child chains is dramatically reduced, and
hence the number of competing parses at each word.  If the number of
competing analyses drops, then the probability range does not narrow
as quickly, so that more analyses actually are retained on the beam.
To test this, we ran the parser with the $\FLC_\mathrm{NP}$ and a
narrower base beam factor of $10^{-10}$.  On this trial, the parser
performed nearly identically with the standard parser, but considered
40 percent fewer expansions per word.  Hence this transform is doing
what it is supposed to do, and it makes a fairly large difference.

\begin{figure*}[t]
\begin{center}
\epsfig{file=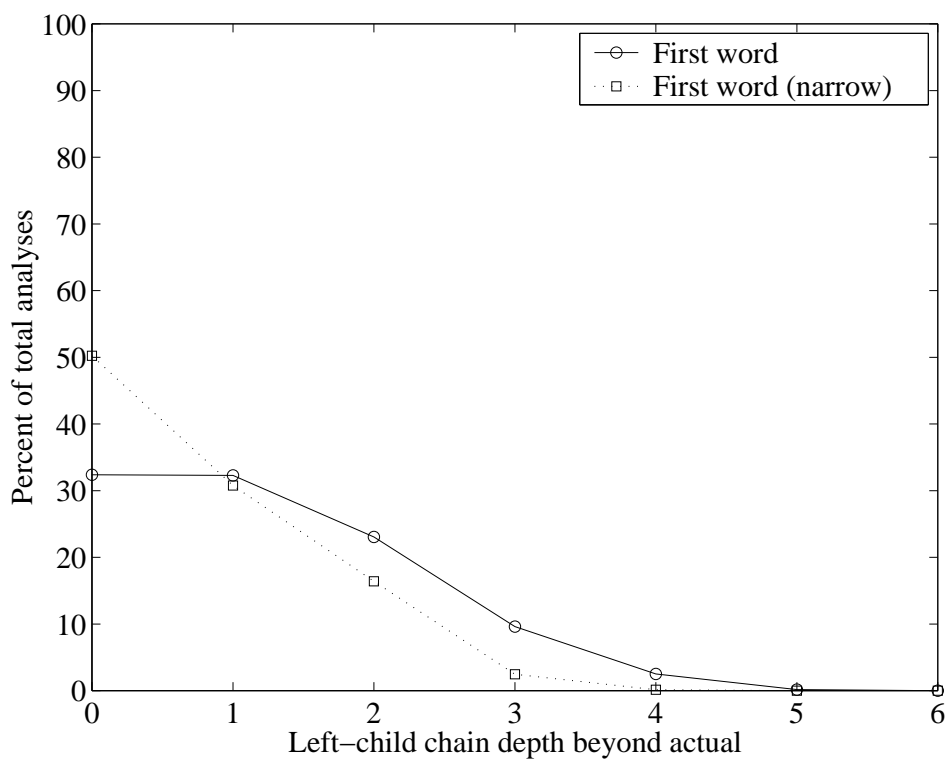, width = 5in}
\end{center}
\caption{Percentage of candidate analyses at a left-child chain
depth beyond the correct parse at the first word in the sentences, for
the original base beam factor, and with a narrow beam at the first
word.}\label{fig:lcd3} 
\end{figure*}

Adding the LC-CHAIN features did not result in a noticeable change in
performance, except for the failure to find a parse for one sentence.  The
number of left-child chains in the correct parses of the test set that
were missing from the candidates with this model was exactly the same
as with the previous model.  This indicates that additional
conditioning information beyond the grandparent of the left-hand side
of the rule being expanded is not going to help model the left-child
chains much better than the original model did.

\begin{figure*}[t]
\begin{center}
\epsfig{file=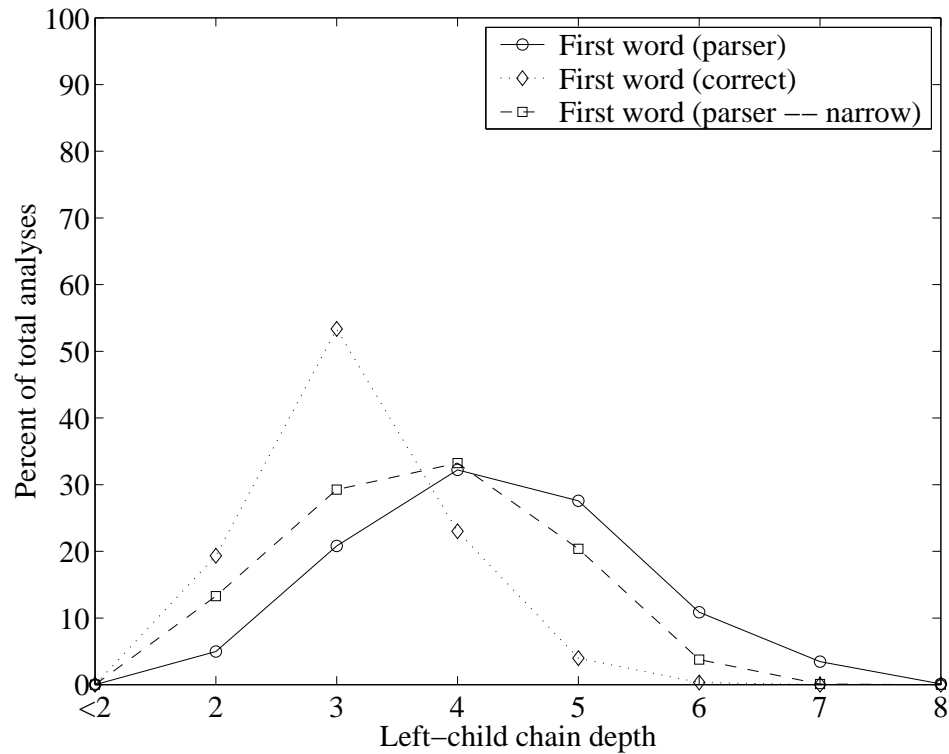, width = 5in}
\end{center}
\caption{Percentage of candidate analyses for wide and narrow beams
and percentage of correct parses with a left-child chain depth at the
first word in the sentences}\label{fig:lcd4}
\end{figure*}

The final simple technique that we tried, of narrowing the beam at the
initial word only, did improve things somewhat.  The accuracy of
the parses remained the same, but the number of expansions considered
was reduced by two percent when the beam factor at the first word was
narrowed to $10^{-7}$, and a bit more when it was narrowed to
$10^{-6}$.  The number of left-child chains in the 
correct parses of the test set that were missing from the candidates
with this model increased from 184 to 195 at $10^{-7}$, and to 208 at
$10^{-6}$, but this didn't seem to 
impact the overall accuracy of the parser.  Figures \ref{fig:lcd3} and
\ref{fig:lcd4} show the left-child chain modeling improvements with
the sentence initial beam at $10^{-6}$.  The percentage of the
parser's time that is being spent on likely depths of left-child
chains at the first word has gone up, moving 
closer to the distribution that we see in the correct parses.
Nevertheless, the overall efficiency gain of this is relatively
small. 

To conclude this section, we have investigated the extent of the
classic problem of top-down parsing, namely dealing with
left-recursion.  We characterized the extent of the problem, and
evaluated the performance of the existing model.  We then implemented
three potential solutions, one of which made things quite a lot better
($\FLC_\mathrm{NP}$), one left things the same, and one improved the
performance only slightly.  All in all, despite the potential gravity
of the problem of left recursion, even the standard top-down parser
performs reasonably well in the face of it.

\section{Parsing transcribed speech}
Up to this point, we have been parsing edited newspaper text.  In the 
next chapter, we will be discussing the application of our
probabilistic parser as a language model for statistical speech
recognition.  The ultimate applicability of the methods that we will
describe depends on whether or not a parser such as this can
effectively parse spontaneous speech.  This section will examine
parsing spontaneous telephone conversation transcripts.

As we have seen, treebank parsers can be amazingly effective on edited
newspaper text.  Parsing spontaneous speech, however, is a different
matter.  False starts, sentence and word fragments, and
ungrammaticality are quite common, all of which, needless to say, pose
a problem for any parser, but particularly for a statistical parser
trained on 
written, edited text.  The release of a new Penn Treebank version,
including a large treebank of Switchboard telephone speech, is thus a
great opportunity for examining how well treebank techniques can be
made to handle these kinds of phenomena.  It was viewing this treebank
that spurred us to investigate the smoothed Markov grammar approach
presented in section one of this chapter.

\begin{figure}
\begin{picture}(396,188)(0,-188)
\put(115,-8){\small S}
\drawline(118,-12)(26,-22)
\put(20,-30){\small PP}
\drawline(26,-34)(7,-44)
\put(1,-52){\small IN}
\drawline(7,-56)(7,-66)
\put(0,-74){for}
\drawline(26,-34)(46,-44)
\put(39,-52){\small NP}
\drawline(46,-56)(31,-66)
\put(23,-74){\small CD}
\drawline(31,-78)(31,-88)
\put(23,-96){two}
\drawline(46,-56)(60,-66)
\put(49,-74){\small NNS}
\drawline(60,-78)(60,-88)
\put(49,-96){years}
\drawline(118,-12)(87,-22)
\put(80,-30){\small NP}
\drawline(87,-34)(87,-44)
\put(78,-52){\small PRP}
\drawline(87,-56)(87,-66)
\put(81,-74){we}
\drawline(118,-12)(209,-22)
\put(202,-30){\small VP}
\drawline(209,-34)(149,-44)
\put(142,-52){\small VP}
\drawline(149,-56)(116,-66)
\put(104,-74){\small AUX}
\drawline(116,-78)(116,-88)
\put(109,-96){did}
\drawline(149,-56)(145,-66)
\put(137,-74){\small RB}
\drawline(145,-78)(145,-88)
\put(139,-96){n't}
\drawline(149,-56)(181,-66)
\put(162,-74){\small EDITED}
\drawline(181,-78)(181,-88)
\put(178,-96){\small S}
\drawline(181,-100)(168,-110)
\put(161,-118){\small CC}
\drawline(168,-122)(168,-132)
\put(160,-140){and}
\drawline(181,-100)(195,-110)
\put(188,-118){\small NP}
\drawline(195,-122)(195,-132)
\put(185,-140){\small PRP}
\drawline(195,-144)(195,-154)
\put(189,-162){we}
\drawline(209,-34)(269,-44)
\put(255,-52){\small SBAR}
\drawline(269,-56)(227,-66)
\put(211,-74){\small WHNP}
\drawline(227,-78)(227,-88)
\put(215,-96){\small WDT}
\drawline(227,-100)(227,-110)
\put(214,-118){which}
\drawline(269,-56)(312,-66)
\put(309,-74){\small S}
\drawline(312,-78)(312,-88)
\put(305,-96){\small VP}
\drawline(312,-100)(262,-110)
\put(250,-118){\small AUX}
\drawline(262,-122)(262,-132)
\put(254,-140){was}
\drawline(312,-100)(304,-110)
\put(284,-118){\small EDITED}
\drawline(304,-122)(304,-132)
\put(297,-140){\small NP}
\drawline(304,-144)(304,-154)
\put(296,-162){\small DT}
\drawline(304,-166)(304,-176)
\put(301,-184){a}
\drawline(312,-100)(362,-110)
\put(349,-118){\small ADJP}
\drawline(362,-122)(340,-132)
\put(325,-140){\small ADVP}
\drawline(340,-144)(328,-154)
\put(321,-162){\small RB}
\drawline(328,-166)(328,-176)
\put(318,-184){kind}
\drawline(340,-144)(353,-154)
\put(345,-162){\small RB}
\drawline(353,-166)(353,-176)
\put(348,-184){of}
\drawline(362,-122)(383,-132)
\put(375,-140){\small NN}
\drawline(383,-144)(383,-154)
\put(370,-162){stupid}
\end{picture}
\caption{A sample parse tree from the Penn Switchboard Treebank.}\label{fig:swbdtree}
\end{figure}
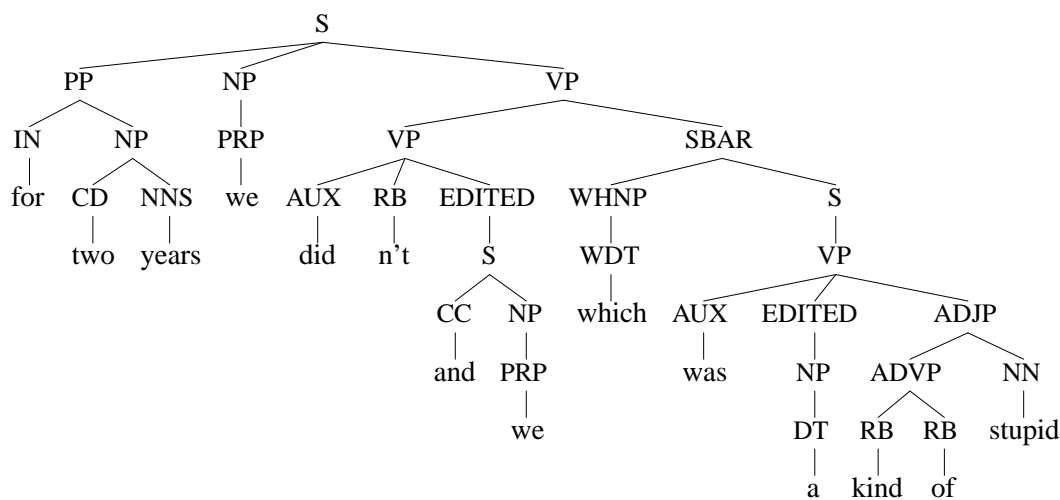

Figure \ref{fig:swbdtree} gives an example parse tree from the new
treebank.  There is a new non-terminal, `EDITED', which is used for
false starts.  For example, in the tree in figure \ref{fig:swbdtree},
a conjoined clause was begun (\textttt{`and we'}), but the VP is then
continued with a subordinate clause.  The words in the falsely started
clause are placed under an EDITED constituent, with as much internal
structure as is evident from the input.  A second false start occurs
further along in the string.  These EDITED nodes provide a way to fit
disfluencies into a parse structure, and hence we can apply a parser
trained on a treebank of such trees directly to strings of spontaneous
speech, without pre-processing.

Following \namecite{Charniak01a}, we designated all of sections 2 and
3 (92,536 sentences, 945,294 words) as the training corpus; files
sw4004 through sw4153 (6,051 sentences, 67,050 words) as the test
corpus; files sw4154 through sw4483 (6,021 sentences, 68,543 words) as
the held out corpus; and files sw4519 through sw4936 (5,895 sentences,
69,597 words) as the development corpus.  These transcriptions have
had punctuation inserted by annotators, to delimit interjections and
false starts.  Hence, a sentence such as 
\begin{examples}
\item \textttt{`Uh well we 're we have one on the way'}
\end{examples}
was transcribed
\begin{examples}
\item \textttt{`Uh , well we 're , we have one on the way .}
\end{examples}
The word counts above include punctuation.

\begin{figure}[t]
\begin{picture}(397,166)(0,-166)
\put(63,-8){(a)}
\put(61,-30){\small S}
\drawline(64,-34)(11,-44)
\put(0,-52){\small INTJ}
\drawline(11,-56)(11,-66)
\put(3,-74){\small UH}
\drawline(11,-78)(11,-88)
\put(3,-96){Oh}
\drawline(64,-34)(51,-44)
\put(32,-52){\small EDITED}
\drawline(51,-56)(51,-66)
\put(44,-74){\small NP}
\drawline(51,-78)(51,-88)
\put(42,-96){\small PRP}
\drawline(51,-100)(51,-110)
\put(49,-118){I}
\drawline(64,-34)(88,-44)
\put(81,-52){\small NP}
\drawline(88,-56)(88,-66)
\put(78,-74){\small PRP}
\drawline(88,-78)(88,-88)
\put(85,-96){I}
\drawline(64,-34)(118,-44)
\put(111,-52){\small VP}
\drawline(118,-56)(118,-66)
\put(107,-74){\small VBP}
\drawline(118,-78)(118,-88)
\put(107,-96){start}
\put(216,-8){(b)}
\put(214,-30){\small S}
\drawline(217,-34)(166,-44)
\put(160,-52){\small NP}
\drawline(166,-56)(166,-66)
\put(157,-74){\small PRP}
\drawline(166,-78)(166,-88)
\put(156,-96){You}
\drawline(217,-34)(267,-44)
\put(260,-52){\small VP}
\drawline(267,-56)(197,-66)
\put(186,-74){\small VBP}
\drawline(197,-78)(197,-88)
\put(187,-96){stay}
\drawline(267,-56)(237,-66)
\put(217,-74){\small EDITED}
\drawline(237,-78)(237,-88)
\put(231,-96){\small PP}
\drawline(237,-100)(218,-110)
\put(212,-118){\small IN}
\drawline(218,-122)(218,-132)
\put(203,-140){within}
\drawline(237,-100)(255,-110)
\put(248,-118){\small NP}
\drawline(255,-122)(255,-132)
\put(243,-140){\small PRP\$}
\drawline(255,-144)(255,-154)
\put(244,-162){your}
\drawline(267,-56)(279,-66)
\put(268,-74){\small INTJ}
\drawline(279,-78)(279,-88)
\put(271,-96){\small UH}
\drawline(279,-100)(279,-110)
\put(272,-118){uh}
\drawline(267,-56)(338,-66)
\put(332,-74){\small PP}
\drawline(338,-78)(310,-88)
\put(304,-96){\small IN}
\drawline(310,-100)(310,-110)
\put(295,-118){within}
\drawline(338,-78)(365,-88)
\put(358,-96){\small NP}
\drawline(365,-100)(347,-110)
\put(335,-118){\small PRP\$}
\drawline(347,-122)(347,-132)
\put(336,-140){your}
\drawline(365,-100)(383,-110)
\put(375,-118){\small NN}
\drawline(383,-122)(383,-132)
\put(368,-140){means}
\end{picture}
\begin{picture}(522,126)(0,-166)
\put(95,-8){(c)}
\put(93,-30){\small S}
\drawline(96,-34)(11,-44)
\put(0,-52){\small INTJ}
\drawline(11,-56)(11,-66)
\put(3,-74){\small UH}
\drawline(11,-78)(11,-88)
\put(3,-96){Uh}
\drawline(96,-34)(43,-44)
\put(32,-52){\small INTJ}
\drawline(43,-56)(43,-66)
\put(35,-74){\small UH}
\drawline(43,-78)(43,-88)
\put(32,-96){well}
\drawline(96,-34)(86,-44)
\put(66,-52){\small EDITED}
\drawline(86,-56)(86,-66)
\put(83,-74){\small S}
\drawline(86,-78)(70,-88)
\put(63,-96){\small NP}
\drawline(70,-100)(70,-110)
\put(61,-118){\small PRP}
\drawline(70,-122)(70,-132)
\put(63,-140){we}
\drawline(86,-78)(102,-88)
\put(95,-96){\small VP}
\drawline(102,-100)(102,-110)
\put(90,-118){\small AUX}
\drawline(102,-122)(102,-132)
\put(95,-140){'re}
\drawline(96,-34)(126,-44)
\put(119,-52){\small NP}
\drawline(126,-56)(126,-66)
\put(116,-74){\small PRP}
\drawline(126,-78)(126,-88)
\put(119,-96){we}
\drawline(96,-34)(182,-44)
\put(175,-52){\small VP}
\drawline(182,-56)(157,-66)
\put(145,-74){\small AUX}
\drawline(157,-78)(157,-88)
\put(146,-96){have}
\drawline(182,-56)(207,-66)
\put(200,-74){\small NP}
\drawline(207,-78)(185,-88)
\put(178,-96){\small NP}
\drawline(185,-100)(185,-110)
\put(177,-118){\small CD}
\drawline(185,-122)(185,-132)
\put(176,-140){one}
\drawline(207,-78)(229,-88)
\put(223,-96){\small PP}
\drawline(229,-100)(210,-110)
\put(204,-118){\small IN}
\drawline(210,-122)(210,-132)
\put(204,-140){on}
\drawline(229,-100)(247,-110)
\put(240,-118){\small NP}
\drawline(247,-122)(233,-132)
\put(226,-140){\small DT}
\drawline(233,-144)(233,-154)
\put(226,-162){the}
\drawline(247,-122)(261,-132)
\put(253,-140){\small NN}
\drawline(261,-144)(261,-154)
\put(251,-162){way}
\end{picture}
\caption{Typical disfluencies from the Switchboard treebank}\label{fig:disfl}
\end{figure}
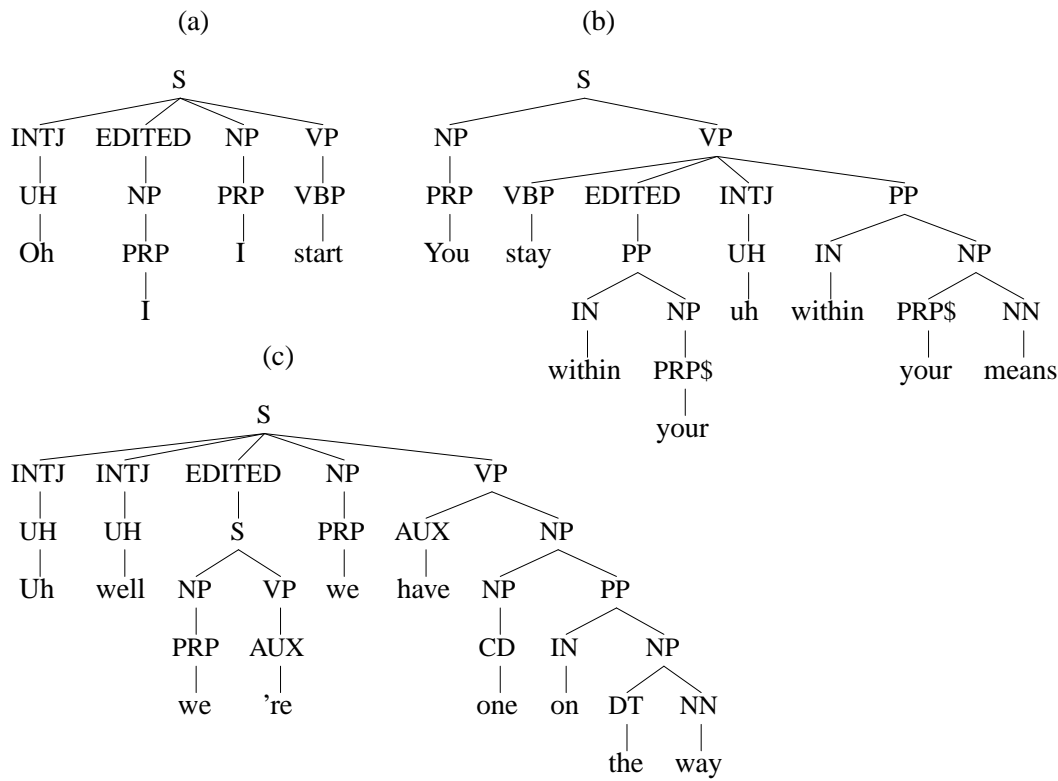

The first thing that we tried was to simply leave the model as it was
for the Wall St. Journal parsing trials, train on the new treebank in
exactly the same way, parse and evaluate\footnote{These trials will be
reported a bit later in the section.}.  It is clear, however, by
inspecting some of the disfluencies found in the corpus, that there is
some relationship between the false start and what replaces it.
Figure \ref{fig:disfl} shows two common kinds of false starts, in
which the EDITED node is either followed by the identical constituent,
or by a very similar constituent.  In the same way that we were able
to effectively model parallelism in conjoined constituents, we could
condition the probability of structures on values returned from a
function looking at these EDITED constituents.

\begin{figure}[t]
\begin{algorithm}{EDIT-SKIP}{node}
\begin{IF}{\CALL{PUNCTUATION}(node)\hspace*{.2in}\COMMENT{If punctuation}}
  \RETURN \CALL{TRUE}
\end{IF}\\
\begin{IF}{node.label = \text{`PRN'} \text{or} node.label =
\text{`INTJ'}\hspace*{.1in}\COMMENT{If a parenthetical or interjection}}
  \RETURN \CALL{TRUE}
\end{IF}\\
\RETURN \CALL{FALSE}
\end{algorithm}

\begin{algorithm}{EDIT-CHILD}{node}
\begin{IF}{node \neq \text{NULL} \text{and}
node.leftsib = \text{NULL} \text{and}
node.parent \neq \text{NULL}}
  node \= node.parent
\ELSE
  \RETURN \text{NULL}
\end{IF}\\
thislabel \= node.label\hspace*{1in}\COMMENT{Save left-hand side label}\\
node \= node.leftsib\\
\begin{WHILE}{ node \neq \text{NULL} \text{and} \CALL{EDIT-SKIP}(node) }
  node \= node.leftsib\hspace*{1in}\COMMENT{Move left past skip categories}
\end{WHILE}\\
\begin{IF}{node = \text{NULL} \text{or} node.label \neq \text{`EDITED'}}
  \RETURN \text{NULL}  
\end{IF}\\
parentlabel \= node.parent.label\hspace*{1in}\COMMENT{Save parent label}\\
node \= node.child\hspace*{2in}\COMMENT{Move to first child}\\
\begin{IF}{node.label = parentlabel\hspace*{1in}\COMMENT{If same label
as parent, keep going}}
  node \= node.child
\end{IF}\\
\begin{IF}{node.label \neq thislabel\hspace*{1in}\COMMENT{If category
is not same as left-hand side}}
  \RETURN \text{NULL}
\end{IF}\\
node \= node.child\hspace*{2in}\COMMENT{Go to first child}\\
\begin{IF}{node = \text{NULL}}
  \RETURN \text{NULL}
\ELSE
  \RETURN node.label
\end{IF}
\end{algorithm}
\caption{Functions for conditioning probabilities so as to capture
EDITED node parallelism. {\it node\/} is a pointer to a node in the tree,
which is a data structure with five fields: {\it label\/} which is a
pointer to a character string; and {\it parent\/}, {\it child\/}, {\it
leftsib\/}, and {\it head\/}, which are pointers to other nodes in the
tree. The symbol $\triangleright$ precedes comments.}\label{alg:edit}
\end{figure}

There are two observations that we could easily exploit to model the
likelihood of an EDITED constituent: (i) the first child of the first
constituent {\it after\/} the EDITED constituent tends to be the same
as the first child of the category {\it inside\/} of the EDITED
constituent; and (ii) the first word after the EDITED constituent
tends to be the same as the first word of the EDITED constituent.
There are a couple of provisos that need to made to these
observations.  First, interjections (e.g. \textttt{`uh'}) and
parentheticals (e.g. \textttt{`you know'} or \textttt{`I mean'}) as
well as punctuation frequently stand between the disfluency and the
continuation.  Hence, any algorithm that wants to link a disfluency
and its continuation should skip these categories.  Second, as
evidenced by the disfluency in figure \ref{fig:disfl}c, the category
directly under the EDITED node may not be the next produced category,
but rather the category under which the disfluency occurs.  In this
case, in order to get the parallelism, one must look at the category
under the EDITED node, and, if it is the same as the parent of the
EDITED node, go to its first child to find the parallel constituent.

Two new functions were written, one to match the first constituent
following the EDITED node with a constituent under the EDITED node,
and condition its expansion on the first child of this edited
category (EDIT-CHILD).  This function is presented in figure
\ref{alg:edit}.  The next is to condition the probability of the first
lexical item after an EDITED node with the left-corner lexical item in
the EDITED node (EDIT-LEX).  This function is presented in figure
\ref{alg:edit1}.  The revised conditional probability models are
presented in figure \ref{fig:editcond}.

\begin{figure}[t]
\begin{algorithm}{EDIT-LEX}{node,m}
\begin{WHILE}{node \neq \text{NULL} \text{and} node.leftsib =
\text{NULL}\hspace*{.3in}\COMMENT{Move up left-child chain}} 
  node \= node.parent
\end{WHILE}\\
\begin{IF}{node = \text{NULL}}
  \RETURN \text{NULL}
\end{IF}\\
node \= node.leftsib\\
\begin{WHILE}{ node \neq \text{NULL} \text{and} \CALL{EDIT-SKIP}(node) }
  node \= node.leftsib\hspace*{1in}\COMMENT{Move left past skip categories}
\end{WHILE}\\
\begin{IF}{node = \text{NULL} \text{or} node.label \neq \text{`EDITED'}}
  \RETURN \text{NULL}  
\end{IF}\\
\begin{WHILE}{node.child \neq \text{NULL}}
  node \= node.child\hspace*{1in}\COMMENT{Move down to the left-corner}
\end{WHILE}\\
\begin{FOR}{i \= 1 \TO m\hspace*{2in}\COMMENT{Move up $m$ nodes}}
  \begin{IF}{node \neq \text{NULL}}
    node \= node.parent
  \end{IF}
\end{FOR}\\
\begin{IF}{node \neq \text{NULL}}
  \RETURN node
\ELSE
  \RETURN \text{NULL}
\end{IF}
\end{algorithm}
\caption{Another function for conditioning probabilities so as to capture
EDITED node parallelism. {\it node\/} is a pointer to a node in the tree,
which is a data structure with five fields: {\it label\/} which is a
pointer to a character string; and {\it parent\/}, {\it child\/}, {\it
leftsib\/}, and {\it head\/}, which are pointers to other nodes in the
tree. The symbol $\triangleright$ precedes comments.}\label{alg:edit1}
\end{figure}

\begin{figure}[t]
\begin{tabular}{|l|l|l|}
\hline
\multicolumn{3}{|c|}{Conditional Probability Model for non-POS expansions}\\\hline
\multicolumn{2}{|l|}{Conditioning Function} & Description \\\hline
0&PAR-SIB({\it node,1,0}) & Left-hand side (LHS)\\\hline
1&PAR-SIB({\it node,0,1}) & Last child of LHS\\\hline
2&PAR-SIB({\it node,0,2}) & 2nd last child of LHS\\\hline
3&PAR-SIB({\it node,0,3}) & 3rd last child of LHS\\\hline
4&PAR-SIB({\it node,2,0}) & Parent of LHS (PAR)\\\hline
5&EDIT-CHILD({\it node}) & First Child of category under EDITED node\\\hline
6&PAR-SIB({\it node,1,1}) & Last child of PAR\\\hline
7&PAR-SIB({\it node,3,0}) & Parent of PAR (GPAR)\\\hline
8&PAR-SIB({\it node,2,1}) & Last child of GPAR \\\hline
9&CONJ-PARALLEL({\it node}) & First child of conjoined category\\\hline
10&CURR-HEAD({\it node,0}) & Lexical head of current constituent\\\hline
\end{tabular}

\vspace*{.5in}

\begin{tabular}{|l|l|l|}
\hline
\multicolumn{3}{|c|}{Conditional Probability Model for POS expansions}\\\hline
\multicolumn{2}{|l|}{Conditioning Function} & Description\\\hline
0&PAR-SIB({\it node,1,0}) & Left-hand side (LHS)\\\hline
1&PAR-SIB({\it node,2,0}) & Parent of LHS (PAR)\\\hline
2&PAR-SIB({\it node,1,1}) & Last child of PAR\\\hline
3&EDIT-LEX({\it node,1}) & POS of EDITED left-corner lexical item\\\hline
4&EDIT-LEX({\it node,0}) & EDITED left-corner lexical item\\\hline
5&LEFTMOST-PS({\it node,3,0}) & Parent of PAR (GPAR)\\\hline
6&LEFTMOST-CCH({\it node,1,1}) & POS of C-Commanding head\\\hline
7&CC-HEAD({\it node,1,0}) & C-Commanding head\\\hline
8&CC-HEAD({\it node,2,0}) & Next C-Commanding head \\\hline
\end{tabular}
\caption{The modified conditional probability models used for
Switchboard parsing.  These are identical to those in figures
\ref{fig:twf1} and \ref{fig:twf2}, except for the EDIT-CHILD and
EDIT-LEX functions.}\label{fig:editcond}
\end{figure}

Table \ref{tab:swbd1} gives parsing results both with the conditional
probability models from the previous sections, and with the new EDITED
node functions.  The look-ahead and head probability models remained
the same.  Overall, both models do pretty well.  The new
functions provide a half a percentage point improvement in accuracy,
and about a four percent decrease in expansions considered.  As far as
correctly finding EDITED nodes, the old model, inherited from the Wall
St. Journal parser, gets 56.5 percent recall and 67.0 percent
precision for these nodes; the new model gets 63.5 percent recall and
71.0 percent precision.  Thus our new functions do seem to be buying us
some improvement in detecting disfluencies, which translates to
overall accuracy improvements.

Table \ref{tab:swbd2} gives the performance with our Switchboard
conditional probability model at a variety of base beam factors.  It
would be surprising if the same parameterization worked equally well
both for Wall St. Journal text and spoken language.  Indeed,  between
$10^{-9}$ and $10^{-11}$ there is virtually no difference in accuracy,
yet there is a sixty percent reduction in the number of rule 
expansions considered\footnote{The parser was correspondingly faster,
from 3.7 words per second to 10.1 words per second, which is a 63.3
percent speed up.  The expansions considered metric is directly
proportional to time, and provides a machine-independent metric, as
argued in \namecite{Roark00b}.}.

\begin{table*}[t]
\begin{tabular}{|l|c|c|c|c|c|c|r|c|}
\hline
{\small Model} & {\small LR} & {\small LP} & {\small CB} & 
{\small 0 CB} & {\small $\leq$ 2} & {\small Pct.} &
{\small Avg. rule\ \ } & {\small Average}\\
{} & {} & {} & {} & {} & {\small CB} & {\small failed} & {\small
expansions\ } & {\small analyses}\\
{} & {} & {} & {} & {} & {} & {} & {\small considered${}^{\dag}$} & {\small advanced${}^{\dag}$}\\\hline
\multicolumn{9}{|c|}{sw4004-sw4153: 6051 sentences of length $\leq$ 100}\\\hline
{WSJ Model} & {84.6} & {85.2} & {0.64} & {83.6} & {92.1} & {0} & {9,987} & {176.6}\\\hline
{w/ {\small EDITED} functions} & {85.2} & {85.6} & {0.61} & {83.8} & {92.4} & {0} & {9,562} & {170.1}\\\hline
\end{tabular}\\
\begin{footnotesize}
${}^{\dag}$per word
\end{footnotesize}
\caption{Parsing results using the model from previous sections, and
the new model with functions to condition probabilities on EDITED node
parallelism.  The parser was trained on sections 2 and 3 of the
switchboard treebank, and tested on files sw4004 through
sw4153.}\label{tab:swbd1} 
\end{table*}

\begin{table*}
\begin{tabular}{|l|c|c|c|c|c|c|r|c|}
\hline
{\small Base Beam} & {\small LR} & {\small LP} & {\small CB} & 
{\small 0 CB} & {\small $\leq$ 2} & {\small Pct.} &
{\small Avg. rule\ \ } & {\small Average}\\
{\small Factor} & {} & {} & {} & {} & {\small CB} & {\small failed} & {\small
expansions\ } & {\small analyses}\\
{} & {} & {} & {} & {} & {} & {} & {\small considered${}^{\dag}$} & {\small advanced${}^{\dag}$}\\\hline
\multicolumn{9}{|c|}{sw4004-sw4153: 6051 sentences of length $\leq$ 100}\\\hline
{$10^{-11}$} & {85.2} & {85.6} & {0.61} & {83.8} & {92.4} & {0} & {9,562} & {170.1}\\\hline
{$10^{-10}$} & {85.1} & {85.6} & {0.61} & {83.9} & {92.5} & {0} & {5,799} & {101.7}\\\hline
{$10^{-9}$} & {85.0} & {85.5} & {0.61} & {83.8} & {92.4} & {0} & {3,574} & {61.1}\\\hline
{$10^{-8}$} & {84.8} & {85.4} & {0.62} & {83.7} & {92.3} & {0} & {2,219} & {36.8}\\\hline
{$10^{-7}$} & {84.4} & {85.0} & {0.64} & {83.5} & {92.1} & {0} & {1,404} & {22.4}\\\hline
{$10^{-6}$} & {83.5} & {84.1} & {0.68} & {82.7} & {91.9} & {0.02} & {915} & {13.6}\\\hline
\end{tabular}\\
\begin{footnotesize}
${}^{\dag}$per word
\end{footnotesize}
\caption{Parsing results using the new model with functions to
condition probabilities on EDITED node parallelism, at various base
beam factors.  The parser was trained on sections 2 and 3 of the 
switchboard treebank, and tested on files sw4004 through
sw4153.}\label{tab:swbd2} 
\end{table*}

Since the Switchboard treebank is relatively new, there is only one
other parsing result that we are aware of, to which we could compare
these results.  This is the two-stage architecture presented in
\namecite{Charniak01a}, which first runs a high-precision classifier
to decide whether lexical items are EDITED.  If they are, then they
are removed for input into a statistical parser.  The
labeled precision and recall percentages presented in that paper were
measured according to a special definition, which has three
modifications to the definition that we have been using until now.
First, all internal structure of EDITED nodes is removed, creating
flat constituents.  Second, two EDITED nodes with no intermediate
non-EDITED material are merged.  Third, the beginning and ending
positions of the EDITED constituents are treated as equivalent for
scoring purposes (like punctuation).

Table \ref{tab:cj} presents their results and ours with this modified
precision and recall metric.  We present precision and recall, as well
as the F-measure, since the precision and recall can be rather far apart.
We also measured the performance of our parser just with respect to
these modified EDITED nodes.  These results indicate that we do pretty
well on the internal structure of edited nodes, so that our
performance drops somewhat when that structure is omitted.  With the
pre-processing, the Charniak parser outperforms ours by a point and a
half.

In summary, we have taken our existing parser and applied it
unmodified to transcribed speech with quite good results.  With the
additional conditioning information, we eke out an additional half a
point of accuracy.

\begin{table}
\begin{center}
\begin{tabular}{|l|c|c|c|}
\hline
Parser & LR & LP & F-measure\\\hline
Charniak and Johnson & 86.5 & 85.3 & 85.9\\\hline
Our parser & 84.7 & 84.9 & 84.8\\\hline
Our EDITED nodes & 63.9 & 67.5 & 65.6\\\hline
\end{tabular}
\end{center}
\caption{Results using the Charniak and Johnson modified labeled
precision and recall metric, of their parser, our parser, and EDITED
nodes from our parser.}\label{tab:cj}
\end{table}

\section{Empty punctuation}
Transcribed speech may have punctuation, but hypotheses from a speech
recognition system typically do not\footnote{They might be modified to
produce punctuation tokens at places where transcribers are likely to
place them, but the systems that we will be using the output from do
not do this.}.  Two approaches will be investigated for parsing spoken
language without the transcribed punctuation: (i) removal of
punctuation from both training and testing corpora; and (ii) removal
of punctuation from testing data, and treating the punctuation
categories as empty.  In both cases we can compare the performance
with the results presented in the previous section.

Why might we want to keep the punctuation in the training data?  One
situation that will occur when punctuation is removed is that distinct
productions will be collapsed into the same productions.  Collapsing
distinct rules versus keeping them distinct can make a difference to a
parser. It could happen that neither of two separate readings is in 
the maximum likelihood parse, but when their probability mass is
combined in a single production, they are.  For example,
the productions 
\begin{examples}
\item NP $\rightarrow$ NP SBAR
\item NP $\rightarrow$ NP , SBAR ,
\end{examples}
are very different kinds of constructions, restrictive versus
non-restrictive modification, as in the following two strings, which
are examples of the previous two rules, respectively:
\begin{examples}
\item \textttt{`legislation that would restrict how the \ldots'}
\item \textttt{`the bill , whose backers include \ldots'}
\end{examples}
An example from our spoken language corpus is parentheticals, which
can be either the sort of vacuous interjection represented by
\textttt{`you know'} or \textttt{`I mean'}, or an actual parenthetical
construction, such as \textttt{`at least'}.  The former are delimited
by commas in the transcription, but the latter are not.

Our top-down parsing algorithm is already able to handle
$\epsilon$-productions.  In practice, if our empty categories are {\it
always\/} 
empty (as in the case with punctuation categories when there is no
punctuation in the input), when an empty category is at the top of the
stack, it is simply popped off, and the derivation is pushed back onto
the current heap.  The look-ahead probability treats the punctuation
as an empty node, so that the definition as given can be applied.  

\begin{table*}[t]
\begin{tabular}{|l|c|c|c|c|c|c|r|c|}
\hline
{\small Model} & {\small LR} & {\small LP} & {\small CB} & 
{\small 0 CB} & {\small $\leq$ 2} & {\small Pct.} &
{\small Avg. rule\ \ } & {\small Average}\\
{\small Factor} & {} & {} & {} & {} & {\small CB} & {\small failed} & {\small
expansions\ } & {\small analyses}\\
{} & {} & {} & {} & {} & {} & {} & {\small considered${}^{\dag}$} & {\small advanced${}^{\dag}$}\\\hline
\multicolumn{9}{|c|}{section 23: 2416 sentences of length $\leq$ 100}\\\hline
{With punctuation} & {86.4} & {86.8} & {1.31} & {59.5} & {81.6} & {0} & {9,008} & {198.9}\\\hline
{No punctuation} & {83.4} & {84.1} & {1.69} & {54.4} & {76.6} & {0.04} & {11,146} & {229.0}\\\hline
{Empty punctuation} & {81.8} & {82.8} & {1.88} & {51.3} & {74.5} & {0.29} & {19,728} & {252.5}\\\hline
\end{tabular}\\
\begin{footnotesize}
${}^{\dag}$per word
\end{footnotesize}
\caption{Parsing results on section 23 of the Penn Wall St. Journal
Treebank: (i) with punctuation; (ii) with no punctuation; and (iii)
with punctuation treated as an empty node.}\label{tab:emp1} 
\end{table*}

Table \ref{tab:emp1} presents parsing results under three conditions:
(i) with punctuation as it is given in the treebank; (ii) with all
punctuation removed from the training and testing corpora; and (iii)
with punctuation removed from the testing corpora, but not the
training corpus, and punctuation treated as an empty node.  From these
results, we can see that punctuation provides much disambiguating
information, since when it is removed, the labeled precision and
recall drops by about three percentage points, and the number of
expansions considered increases by over twenty percent.  However,
treating punctuation as empty nodes does not help.  In fact, it
worsens performance dramatically, particularly in terms of efficiency.
What seems to be going on is that the ability to predict punctuation
as needed leads to a proliferation of competitor analyses.  Since the
beam is defined as a function of both the probability of the best
analysis and the number of successful candidate analyses, this
proliferation leads to substantially narrower beams.  In other words,
the large number of analyses crowds out good analyses.

\begin{table*}[t]
\begin{tabular}{|l|c|c|c|c|c|c|r|c|}
\hline
{\small Model} & {\small LR} & {\small LP} & {\small CB} & 
{\small 0 CB} & {\small $\leq$ 2} & {\small Pct.} &
{\small Avg. rule\ \ } & {\small Average}\\
{\small Factor} & {} & {} & {} & {} & {\small CB} & {\small failed} & {\small
expansions\ } & {\small analyses}\\
{} & {} & {} & {} & {} & {} & {} & {\small considered${}^{\dag}$} & {\small advanced${}^{\dag}$}\\\hline
\multicolumn{9}{|c|}{sw4004-sw4153: 6051 sentences of length $\leq$ 100}\\\hline
{With punctuation} & {85.2} & {85.6} & {0.61} & {83.8} & {92.4} & {0} & {9,562} & {170.1}\\\hline
{No punctuation} & {84.0} & {84.6} & {0.68} & {82.1} & {91.5} & {0.08} & {12,051} & {182.2}\\\hline
{Empty punctuation} & {79.9} & {81.8} & {0.79} & {79.4} & {90.8} & {0.36} & {19,180} & {201.7}\\\hline
\end{tabular}\\
\begin{footnotesize}
${}^{\dag}$per word
\end{footnotesize}
\caption{Parsing results on section 23 of the Penn Switchboard
Treebank: (i) with punctuation; (ii) with no punctuation; and (iii)
with punctuation treated as an empty node.}\label{tab:emp2} 
\end{table*}

Table \ref{tab:emp2} performs the same experiment on our Switchboard
testing corpus.  Here, interestingly, the removal of punctuation has
much less impact on the accuracy of the parser -- labeled precision
and recall drops by only a percentage point.  Critically, precision
and recall for EDITED nodes goes from a 67.0 F-measure with
punctuation to 65.3 without, which is perhaps less of a drop than
might be expected, given that the punctuation is largely used to
delimit false starts.  Treating punctuation as empty nodes leads to
a far worse drop in performance than in the WSJ experiment, although
the increase in expansions considered follows the same pattern.
It appears that in the case of Switchboard, the punctuation is
providing far less guidance than in the case of the newspaper text, so
that not only do good analyses get crowded out by the proliferation of
candidate analyses, but the punctuation does not perform much useful
extra disambiguation.

In sum, while it is possible to use empty nodes within this framework,
it does not appear, at least when it comes to punctuation, to provide
any benefit.

\section{Chapter summary}
In this chapter, we have presented several modifications to the parser
architecture and model, some of which have improved performance
substantially (smoothed Markov grammar and conditioning functions for
EDITED parallelism), others of which have had no noteworthy positive
effect (attempts to improve efficiency of processing left-child
chains), and still others which hurt performance (empty punctuation).
As we enter the final chapter, we have a model which is capable of
handling spontaneous spoken language effectively, in terms of the
accuracy with which it identifies constituents, the efficiency with
which it finds them, and the coverage it achieves.

\chapter{Language modeling with a top-down parser}
With certain exceptions, computational linguists have in the
past generally formed a separate
research community from speech recognition researchers, despite some
obvious overlap of interest.  Perhaps one reason for this is that,
until relatively recently, few methods have come out of the natural 
language processing community that were shown to 
improve upon the very simple language models still standardly in use in 
speech recognition systems.  In the past few years, however, some
improvements have been made over these language models through
the use of statistical methods of natural language processing; and
the development of innovative, linguistically well-motivated
techniques for improving language models for speech recognition is
generating more interest among computational linguists.  While
language models built around shallow local dependencies are still the
standard in state-of-the-art speech recognition systems, there is
reason to hope that better language models can and will be developed
by computational linguists for this task.

This chapter will examine language modeling for speech recognition from
a natural language processing point of view.  Some of the recent
literature investigating approaches that use syntactic structure in an  
attempt to capture long-distance dependencies for language modeling
will be reviewed.  A new language model, based on probabilistic
top-down parsing, will be outlined and compared with the previous
literature, and extensive empirical results will be presented which
demonstrate its utility.

Two features of our top-down parsing approach will emerge as key
to its success.  First, the top-down parsing algorithm builds a set of
{\it rooted} candidate parse trees from left-to-right over the string,
which allows it to calculate a generative probability for each prefix
string from the probabilistic grammar, and hence a conditional
probability for each word given the previous words and the
probabilistic grammar.  A left-to-right parser whose derivations are
not rooted, i.e. with derivations that can consist of disconnected
tree fragments, such as an LR or shift-reduce parser, cannot simply
use PCFG probabilities to incrementally calculate a generative
probability of each prefix string, because their derivations include
probability mass from unrooted structures.  In order to get generative
probabilities from a bottom-up parser, additional calculations
beyond the parsing itself must be done \cite{Jelinek91}.  However,
if the probabilities are on parser {\it operations\/}, and not on the
rules in the grammar, then a shift-reduce parser can be generative, as
in the Structured Language Model of \namecite{Chelba98a}.  In our
approach, the derivations are rooted, so the generative probability is
simply the sum of the probability of all parses, using the PCFG.

A parser that is not left-to-right, but which has rooted derivations,
e.g. a head-first parser, will be able to calculate generative joint
probabilities for entire strings; however it will not be able to
calculate probabilities for each word conditioned on previously
generated words, unless each derivation generates the words in the
string in exactly the same order.  For example, suppose that there are
two possible verbs that could be the head of a sentence.  For a
head-first parser, some derivations will have the first verb as the
head of the sentence, and the second verb will be generated after the
first; hence the second verb's probability will be conditioned on the
first verb. Other derivations will have the second verb as the head of
the sentence, and the first verb's probability will be conditioned on
the second verb.  In such a scenario, there is no way to decompose the
joint probability calculated from the set of derivations into the 
product of conditional probabilities using the chain rule.  Of course,
the joint probability can be used as a language model, but it cannot
be interpolated on a word-by-word basis with, say, a trigram model,
which we will demonstrate is a useful thing to do. 

Thus, our top-down parser allows for the incremental calculation of
generative conditional word probabilities, a property it shares with
other left-to-right parsers with rooted derivations such as Earley parsers
\cite{Earley70} or left-corner parsers \cite{Rosenkrantz70}.  

A second key feature of our approach is that top-down guidance
improves the efficiency of the search as more and more conditioning
events are extracted from the derivation for use in the probabilistic
model.  Because the rooted partial derivation is fully connected, all
of the conditioning information that might be extracted from the
top-down left context has already been specified, and a conditional
probability model built on this information will not impose any
additional burden on the search.  In contrast, an Earley or
left-corner parser will underspecify certain connections between
constituents in the left-context, and if some of the underspecified
information is used in the conditional probability model, the state
will have to be split.  Of course, this can be done, but at the expense
of search efficiency; the more that this is done, the less of a
benefit there is to be had from the underspecification.  A top-down
parser will, in contrast, derive an efficiency benefit from precisely
the information that is left underspecified in these other
approaches. 

Thus, our top-down parser makes it very easy to condition the
probabilistic grammar on an arbitrary number of values extracted from
the rooted, fully specified derivation.  This has lead us to a
formulation of the conditional probability model in terms of values
returned from tree-walking functions that themselves are contextually
sensitive.  The top-down guidance that is provided makes this approach
quite efficient in practice.

The next section of the chapter will provide a brief introduction to
language modeling for speech recognition, as well as a brief
discussion of some recent approaches to using syntactic structure to
this end.  It will be followed by empirical results from the use of
our parser for language modeling.

\section{Background}
\subsection{Language modeling for speech recognition}
This section will briefly introduce language modeling for statistical
speech recognition, with such topics as the chain rule, n-gram language
modeling, and interpolation for smoothing\footnote{For a detailed
introduction to statistical speech recognition, see
\namecite{Jelinek98}.}.

A speech recognition system can be thought of as a function that takes 
an acoustic signal as input and outputs a string of words; a ``good''
function outputs a ``good'' string of words, i.e. a string of words
that largely matches the string of words intended by the speaker.  A
statistical approach to speech
recognition tries to find the string of words which is most likely
given the observed speech signal, i.e. the string with the maximum
a posteriori probability.  Given an acoustic signal $A$, the
system attempts to find a string of words $S_{max}$ in the language
$L$ such that
\begin{equation}
S_{max} = \mathrm{arg}\!\max_{\scriptscriptstyle S} \mathrm{P}(S | A)\ 
=\ \mathrm{arg}\!\max_{\scriptscriptstyle S} \frac{\mathrm{P}(A |
S)\mathrm{P}(S)}{\mathrm{P}(A)}\ 
=\ \mathrm{arg}\!\max_{\scriptscriptstyle S} \mathrm{P}(A | S)\mathrm{P}(S)\label{eq:argmax}
\end{equation}
The first component of this model, $\mathrm{P}(A | S)$, is the probability of
the acoustic signal given the string, which is known as the acoustic
model.  The second component of the model, $\mathrm{P}(S)$, is the prior
probability of the string itself, which is known as the language
model.  The quality of a statistical speech recognizer will jointly
depend on the quality of these two models. We will put acoustic
modeling to the side and consider language modeling in
isolation. 

In language modeling, we assign probabilities to strings of words.
A string, taken as a whole, may never have been seen 
before, and, in fact, very likely has not been seen.  To assign a
probability, the chain rule is generally invoked.  The chain rule
states, for a string of {\it k+1\/} words, $w_{0}^{k}$
\begin{eqnarray}
\mathrm{P}(w_{0}^{k}) &=&
\mathrm{P}(w_{0})\prod_{i=1}^{k}\mathrm{P}(w_{i}|w_{0}^{i-1})\label{eq:chain}
\end{eqnarray}
This corresponds to the left-to-right ordering of most speech
recognition systems, in that the probability of any particular word
is conditioned on the words to its left in the string.  However, even
if the permitted sentence length were bounded to some fixed
finite length, this formulation would require an infeasibly
large conditional probability estimate for the words at the end of
the string.  In order to estimate 
these conditional probabilities, the conditioning substrings
(i.e. $w_{0}^{i-1}$) are clustered into equivalence classes,
and the probability of a word is conditioned on the equivalence class
of the string to its left.  

One interesting way in which the models that we will be
discussing can and do differ is in how they define these equivalence
classes.  The most prevalent class of models makes the assumption that
language is (more-or-less) a Markov process.  That is, they assume
that the probability of a word is dependent only on the $n$ closest
words on either side, for some stipulated $n$, and is independent of
anything else.

In terms of equivalence classes, a Markov language model of order $n$
stipulates that two prefix strings of words belong to the same
equivalence class if their final $n$ words are the same.  
The effect of such a model is that the
conditioning information in the chain rule 
is truncated to include only the previous $n$ words.
\begin{eqnarray}
\mathrm{P}(w_{0}^{k}) &=&
\mathrm{P}(w_{0})\mathrm{P}(w_{1}|w_{0})\ldots \mathrm{P}(w_{n-1}|w_{0}^{n-2})\prod_{i=n}^{k}\mathrm{P}(w_{i}|w_{i-n}^{i-1})
\end{eqnarray}
These models are commonly called {\it n-gram\/} models\footnote{The
$n$ in {\it n-gram\/} is one more than the order of the Markov model,
since the $n$-gram includes the word being conditioned.}, because their 
probabilities are defined in terms of strings of words of length $n$.
The standard language model used in many speech recognition systems is
the trigram model, i.e. a Markov model of order 2, which can be
characterized by the following equation:
\begin{eqnarray}
\mathrm{P}(w_{0}^{n-1}) &=&
\mathrm{P}(w_{0})\mathrm{P}(w_{1}|w_{0})\prod_{i=2}^{n-1}\mathrm{P}(w_{i}|w_{i-2}^{i-1})
\end{eqnarray}

As presented above, however, $n$-gram models have a serious problem with
parameter estimation.  The number of 
$n$-gram probabilities that must be estimated for a vocabulary of size $|V|$ is
$|V|^{n}$.  Thus, even for a moderately sized vocabulary and a large
corpus, the  
number of parameters that must be estimated quickly outgrows the
number of tokens in the training corpus, i.e. the data is too sparse
for the number of parameters.  For example, a
trigram model for a modest vocabulary of 10,000 words must assign a
probability to $10^{12}$ trigrams.  A simple maximum likelihood
estimate, which counts the number of occurrences of the trigram in the 
training corpus, will assign a probability of zero to all unobserved
trigrams.  Many of these unobserved trigrams, however, do not really
have zero probability of occurring;  the training data is just not
large enough to begin converging to the true distribution, i.e. it is
sparse. 

Of course, unigram probabilities, i.e. word probabilities conditioned
on none of the preceding words (Markov model of order 0), do not face
sparse data problems of 
nearly the same severity, with merely $|V|$ parameters. The word
probabilities could be estimated with this model.  Unfortunately, it
turns out that the conditional probability of a word given some number 
of previous words, when there is enough training data, is a {\it
far\/} better estimate of the probability of a word.  Consider the
word {\it sprouts\/}.  If one were going to estimate a probability for
this word, without knowing anything else, the estimate would
probably be relatively low.  If, however, one knew that the word {\it
Brussels\/} had just occurred, a good probability estimate would be
relatively high. So it seems that there is a dilemma here:  on the one
hand, there is not enough data to estimate all parameters of an $n$-gram 
model with $n>1$;  on the other hand, there is not enough information
in a unigram probability model to accurately estimate word
probabilities within a string.

The most common solution to this dilemma is to smooth the probability
estimates of higher order Markov models with lower order Markov
models.  The idea is that, if there is enough observed data to
accurately estimate the probability for a higher order Markov model,
that estimate will be relied upon heavily.  If there is not enough,
the estimates of lower order Markov models will be relied upon.
There are several methods for doing this;  one very common method is 
interpolation \cite{Jelinek80}.  The idea behind interpolation is
simple and has been shown to be very effective.  For an interpolated
($n+1$)-gram 

\begin{eqnarray}
\mathrm{P}(w_{i}|w_{i-n}^{i-1}) &=&
\lambda_{n}(w_{i-n}^{i-1})\hat{\mathrm{P}}(w_{i}|w_{i-n}^{i-1}) +
(1-\lambda_{n}(w_{i-n}^{i-1}))\mathrm{P}(w_{i}|w_{i-n+1}^{i-1}) \label{eq:int}
\end{eqnarray}

Here $\hat{\mathrm{P}}$ is the empirically observed relative frequency, and
$\lambda_{n}$ is a function from $V^{n}$ to [0,1].  What the equation
says is that the probability estimate of a word conditioned on the $n$
previous words is the sum of two parts:  the first part, which
makes up $\lambda_{n}$ of the estimate, is the empirically observed
probability of the word given its $n$ predecessors;  the second part,
which contributes the remaining $1-\lambda_{n}$, is the probability
estimate of the word conditioned on the ($n-1$) previous words.  This
interpolation is recursively applied to the smaller order $n$-grams
until the bigram is finally interpolated with the unigram,
i.e. $\lambda_{0}$ = 1.

An interpolated trigram model performs very well, better than an
interpolated bigram model, and only marginally worse than an
interpolated 4-gram model for typical training data sizes.  It is the
standard in speech recognition, and most work on 
language modeling is focused on providing models that improve on the
trigram.  This may seem a surprising fact, given the obvious
limitations of looking only two words back.  

\subsection{Previous work}
There have been attempts to jump over adjacent words to words farther
back in the left-context, without the use of dependency links or
syntactic structure, for example \namecite{Saul97} and Rosenfeld
\shortcite{Rosenfeld96,Rosenfeld97}.  We will focus our very brief review,
however, on those which use grammars or parsing for their language
models.  These can be divided into two rough groups:  those that use
the grammar as a language model; and those that use a parser to
uncover phrasal heads standing in an important relation (c-command) to
the current word.  The approach that we will subsequently present uses
the probabilistic grammar as 
its language model, but only includes probability mass from those
parses that are found, i.e. it uses the parser to find a subset of the
total set of parses (hopefully most of the high probability parses)
and uses the sum of their probabilities as an estimate of the true
probability given the grammar.

\subsubsection{Grammar models}
As mentioned in section \ref{sec:gramm}, a PCFG defines a
probability distribution over strings of words.  One approach to
syntactic language modeling is to use this distribution directly as a
language model.  There are efficient algorithms in the literature
\cite{Jelinek91,Stolcke95} for calculating exact string prefix marginal
probabilities given a PCFG.  The algorithms both utilize a left-corner
matrix, which can be calculated in closed form through matrix
inversion.  Exact calculation is limited, therefore, to grammars where
the non-terminal set is small enough to permit inversion.  String
prefix probabilities can be straightforwardly used to compute
conditional word probabilities by definition: 
\begin{eqnarray}
\Pr(w_{j+1}|w_0^j) &=& \frac{\Pr(w_0^{j+1})}{\Pr(w_0^j)}\label{eq:wd_prst}
\end{eqnarray}

\namecite{Stolcke94} and \namecite{Jurafsky95} used these basic ideas
to estimate bigram probabilities from hand-written PCFGs,
which were then used in language models.  Interpolating observed
bigram probabilities with calculated bigrams led, in both cases, 
to improvements in word error rate over using the observed bigrams
alone, demonstrating that there is some benefit to using these syntactic
language models to generalize beyond observed $n$-grams.  

\subsubsection{Finding phrasal heads}\label{sec:SLM}
Another approach that uses syntactic structure for language modeling
has been to use a shift-reduce parser to identify preceding
c-commanding phrasal head words or part-of-speech (POS) tags from 
arbitrarily far back in the prefix string, for use in a trigram-like
model.  

A shift-reduce parser\footnote{For details, see
e.g. \namecite{Hopcroft79}.} operates from left-to-right using a stack and
a pointer to the next word in the input string.  Each stack entry
consists minimally of a non-terminal label.  The parser performs two
basic operations: (i) {\it shifting\/}, which involves pushing the
POS label of the next word onto the stack
and moving the 
pointer to the following word in the input string; and (ii) {\it
reducing\/}, which takes the top $k$ stack entries and replaces them
with a single new entry, the non-terminal label of which is the left-hand
side of a rule in the grammar which has the $k$ top stack entry labels on
the right-hand side.  For example, if there is a rule
NP~$\rightarrow$~DT~NN, and the top two stack entries are NN and DT,
then those two entries can be popped off of the stack and an entry
with the label NP pushed onto the stack.  

\namecite{Goddeau92} used a robust deterministic shift-reduce parser
to condition word probabilities by extracting a specified number of
stack entries from the top of the current state, and conditioning on
those entries in a way similar to an $n$-gram.  In empirical trials,
Goddeau used the top 2 stack entries to condition the word
probability.  He was able to reduce both sentence and word error rates
on the ATIS corpus using this method.

The ``Structured Language Model'' (SLM) used in Chelba and
Jelinek \shortcite{Chelba98a,Chelba98b,Chelba99},
\namecite{Jelinek99}, and \namecite{Chelba00} is similar to that of
Goddeau, except that (i) their shift-reduce parser follows a
non-deterministic beam 
search, and (ii) each stack entry contains, in addition to the
non-terminal node label, the head-word of the constituent.  The
SLM is like a trigram, except that the conditioning words
are taken from the tops of the stacks of candidate parses in the beam,
rather than from the linear order of the string.

Their parser functions in three stages.  The first stage assigns a
probability to the word given the left-context (represented by the
stack state).  The second stage predicts the POS given the word and
the left-context.  The last stage performs parser operations (shifting
the new category and reducing the top-two stack entries) until this
can no longer be done.  When there are no more parser operations
to be done (or, in their case, when the beam is full), the following
word is predicted.  And so on until the end of the string.

Each different POS assignment or parser operation is
a step in a derivation.  Each distinct derivation path within the beam 
has a probability and a stack state associated with it.  Every stack
entry has a non-terminal node label and a designated head
word of the constituent.  When all of the
parser operations have finished at a particular point in the string,
the next word is predicted as follows.  For each derivation in the
beam, the head words of the two
topmost stack entries form a trigram with the conditioned word.  This
interpolated trigram probability is then multiplied by the normalized
probability of the derivation, to provide that derivation's
contribution to the probability of the word.  More precisely, for a
beam of derivations $D_{i}$
\begin{eqnarray}
\Pr(w_{i+1}|w_0^i) &=& \frac{\sum_{d \in D_{i}}
\Pr(w_{i+1}|h_{0d},h_{1d}) \Pr(d)}{\sum_{d \in D_{i}}\Pr(d)}
\end{eqnarray}
where $h_{0d}$ and $h_{1d}$ are the lexical heads of the top two
entries on the stack of $d$.

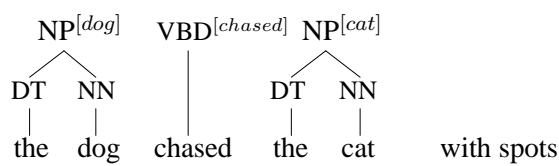
\begin{figure}[t]
\begin{picture}(106,70)(-80,-110)
\put(5,-52){NP$^{[dog]}$}
\drawline(15,-56)(2,-66)
\put(-5,-74){\small DT}
\drawline(2,-78)(2,-88)
\put(-4,-96){the}
\drawline(15,-56)(27,-66)
\put(20,-74){\small NN}
\drawline(27,-78)(27,-88)
\put(20,-96){dog}
%\put(82,-25){VP$^{[bit]}$}
%\drawline(88,-29)(62,-66)
\put(50,-52){\small VBD$^{[chased]}$}
\drawline(62,-56)(62,-88)
\put(49,-96){chased}
%\drawline(88,-29)(104,-44)
\put(105,-52){NP$^{[cat]}$}
\drawline(114,-56)(101,-66)
\put(93,-74){\small DT}
\drawline(101,-78)(101,-88)
\put(94,-96){the}
\drawline(114,-56)(127,-66)
\put(119,-74){\small NN}
\drawline(127,-78)(127,-88)
\put(120,-96){cat \hspace*{.3in}with spots}
\end{picture}
\caption{Tree representation of a derivation state}\label{fig:derst}
\end{figure}

Figure \ref{fig:derst} gives a partial tree representation of a
potential derivation state for the string \textttt{`the dog chased the
cat with spots'}, at the point when the word \textttt{`with'} is to be
predicted.  The shift-reduce parser will have, perhaps, built the
structure shown, and the stack state will have an NP entry with the
head \textttt{`cat'} at the top of the stack, and a VBD entry with
the head \textttt{`chased'} second on the stack.  In the Chelba and
Jelinek model, the probability of \textttt{`with'} is conditioned on
these two head words, for this derivation.

Since the specific results of the SLM will be compared in detail with
our model when the empirical results are presented, at this point we
will simply state that they have achieved a reduction in both perplexity
and WER over a standard trigram using this model.

\subsection{Evaluation}\label{sec:eval}
This section will present standard methods for the  evaluation of
language models. 
Perplexity is a standard measure within the speech recognition
community for comparing language models.  In principle, if two models
are tested on the same
test corpus, the model that assigns the lower perplexity to the 
test corpus is the model closest to the true distribution of the
language, and thus better as a prior model for
speech recognition.  Perplexity is the exponential of the cross
entropy, which we will define next.

Given a random variable $X$ with distribution $p$ and a probability
model $q$, the cross entropy, $H(p,q)$ is defined as follows: 
\begin{eqnarray}
H(p,q) &=& -\sum_{x \in X} p(x) \log q(x)
\end{eqnarray}
Let $p$ be the true distribution of
the language.  Then, under certain assumptions\footnote{See
\namecite{Cover91} for a discussion of the Shannon-McMillan-Breiman
theorem.}, given a large enough sample, the sample mean of the
negative log probability of a model will converge to its cross entropy
with the true model. That is
\begin{eqnarray}
H(p,q) &=& - \lim_{n \rightarrow \infty} \frac{1}{n} \log
q(w_0^n)
\end{eqnarray}
where $w_0^n$ is a string of the language $L$. In practice, one takes
a large sample of the language, and calculates the negative log 
probability of the sample, normalized by its size\footnote{It is
important to remember to include the end marker in the strings of the
sample.}.  The lower the cross entropy (i.e. the higher the
probability the model assigns to the sample), the better the model. 
Usually this is reported in terms of perplexity, which we will do as
well.  Perplexity ($\mathcal{P}$), as stated above, is the exponential
of the cross entropy:
\begin{eqnarray}
\mathcal{P} &=& \exp\left(-\frac{\log q(w_0^n)}{n}\right)\ \ = \ \ 
\exp\left(\log \frac{1}{q(w_0^n)}\right)^{\frac{1}{n}}\ \ = \ \ 
\left(\frac{1}{q(w_0^n)}\right)^{\frac{1}{n}}
\end{eqnarray}
Hence it is the inverse of the geometric mean word probability -- the
mean contribution, per word, to the probability of the sample.

Some of the trials discussed below will report results in terms of
word and/or sentence error rate, which are obtained when the language
model is embedded in a speech recognition system.  Word error rate is
the number of 
deletion, insertion, or substitution errors per 100 words.  Sentence
error rate is the number of sentences with one or more errors per 100
sentences.

A statistical speech recognizer attempts to find the string which
maximizes the posterior probability.  Standard practice is to add more
terms into the formulation of the posterior.  One is a language model
weight factor $\beta$, which is a power to which the prior is raised.
The higher $\beta$, the more the prior language model is relied upon.
A second term is a constant term $\exp(-\gamma)$, which is raised to
the power of the size of the hypothesis string.  The idea is to favor
short strings 
of words over longer strings of words.  The constant $\gamma$ is known as
the word insertion penalty.  Let $|S|$ be the number of words in $S$.
Then the best hypothesis is:
\begin{equation}
\mathrm{arg}\!\max_{\scriptscriptstyle S} \mathrm{P}(A |
S)\mathrm{P}(S)^\beta\exp(-\gamma |S|)\ \ = \ \
\mathrm{arg}\!\max_{\scriptscriptstyle S} \log\mathrm{P}(A | S) + 
\beta\log\mathrm{P}(S) - \gamma |S|
\end{equation}
These parameters are reported with each recognition trial.  In
general, unless otherwise noted, we will use the same parameters as
the lattice trigram for the trial.

The rest of this chapter will present the application of our parsing
model to language modeling for speech recognition.
The first three sections of empirical results were generated with
the base parsing model presented in chapter three.  Then we will
present results using the improvements to the model presented in
chapter four, as well as some additional improvements to the way the
language model is trained, and how it is mixed with other models.

\section{Empirical results}\label{lmemp:sec}
The modifications to the base parser presented in chapter four were
evaluated with respect to parsing.  Now that our parser is to be
evaluated as a language model, it is of interest to see the effect of
the move from the base to the standard parsing model.  For that
reason, results will be presented first for the base model (sections
\ref{lmemp:sec}.1-3), then for the standard model (sections
\ref{lmemp:sec}.4-6).

\subsection{Perplexity results}
The next set of results will highlight what recommends our parsing
approach most: the ease with which one 
can estimate string probabilities in a single pass from left-to-right
across the string.  By definition, a PCFG's estimate of a
string's probability is the sum of the probabilities of all trees that
produce the string as terminal leaves (see equation \ref{eq:prst}). In 
the beam-search approach used by our parser, we can estimate the string's
probability in the same manner, by summing the probabilities of the
parses that the algorithm finds.  Since this is not an exhaustive
search, the parses that are returned will be a subset of the total set 
of trees that would be used in the exact PCFG estimate;
hence the estimate thus arrived at will be bounded above by the
probability that would be generated from an exhaustive search.  The
hope is that a large amount of the probability mass will be accounted
for by the parses in the beam.  The method cannot overestimate the
probability of the string. 

A PCFG also defines a marginal probability distribution over string
prefixes, and we will present this in terms of partial derivations.
A partial derivation (or parse) $d$ is defined with respect to a
string $w_0^j$  as follows: it is the leftmost
derivation\footnote{Recall our presentation of derivations in chapter
3.  A leftmost derivation is a derivation in which the leftmost
non-terminal is always expanded.} of 
the string, with $w_j$ on the right-hand side of the last expansion in
the derivation.  Let $D_{w_0^j}$ be the set of all partial derivations
for a prefix string $w_0^j$.  Then the marginal probability $\Pr_M$ is
the probability of all strings which begin with the prefix string
$w_0^j$, and is defined as
\begin{eqnarray}
\Pr_M(w_0^j) &=& \sum_{d \in D_{w_0^j}}\Pr(d) \label{eq:pr_prst}
\end{eqnarray}
This is the same as summing the probability of all complete trees
which have $w_0^j$ as the first $j+1$ words of their terminal yield.
Let $T_w$ be the set of all complete trees rooted at the start symbol,
with the string of terminals $w\ =\ w_0^jw^\prime$ as the terminal
yield, for some $w^\prime \in T^*$.   Then 
\begin{eqnarray}
\Pr_M(w_0^j) &=& \sum_{t \in T_w}\Pr(t) 
\end{eqnarray}
By definition, 
\begin{eqnarray}
\Pr(w_{j+1}|w_0^j) &=& \frac{\Pr_M(w_0^{j+1})}{\Pr_M(w_0^j)}
\hspace*{.1in}=\hspace*{.1in} \frac{\sum_{d \in
D_{w_0^{j+1}}}\Pr(d)}{\sum_{d \in 
D_{w_0^j}}\Pr(d)}\label{eq:my_prst} 
\end{eqnarray}
Note that the numerator at word $w_j$ is the denominator at word
$w_{j+1}$, so that the product of all of the word probabilities is the 
numerator at the final word, i.e. the marginal string prefix probability.  

We can make a consistent estimate of the string probability by
similarly summing over all of the trees within our beam.  Let
$\scH_{i}^{init}$ be the priority queue $\scH_{i}$ before any
processing has begun with word $w_{i}$ in the look-ahead.  This is a
subset of the possible leftmost partial derivations with respect to
the prefix string $w_0^{i-1}$.  Since $\scH_{i+1}^{init}$ is produced
by expanding only analyses on priority queue 
$\scH_{i}^{init}$, the set of complete trees consistent with the partial
derivations on priority queue $\scH_{i+1}^{init}$ is a subset of the set of
complete trees consistent with the partial derivations on priority queue
$\scH_{i}^{init}$, i.e. the total probability mass represented by the
priority queues are monotonically decreasing.  Thus conditional word
probabilities defined in a way consistent with equation
\ref{eq:my_prst} will always be between zero and one.  Our conditional 
word probabilities are calculated as follows:
\begin{eqnarray}
\Pr(w_{i}|w_{0}^{i-1}) &=& \frac{\sum_{d\in
\scH_{i+1}^{init}}\Pr(d)}{\sum_{d\in \scH_{i}^{init}}\Pr(d)} 
\end{eqnarray}
The probability of the end-of-string symbol
$\langle/\mathrm{s}\rangle$ is the sum of the probabilities of the
completed trees divided by the sum of the partial derivations for the
string though the last word.

As mentioned above, the model cannot overestimate the probability 
of a string, because the string probability is simply the sum over the
beam, which is a subset of the possible derivations.  By utilizing a
figure-of-merit to identify promising analyses, we are simply placing
our attention on those parses which are likely to have a high
probability, and thus we are increasing the amount of probability mass
that we do capture, of the total possible.  It is not part of the
probability model itself.  

Since each word
is (almost certainly, because of our pruning strategy) losing some
probability mass, the probability model is not ``proper'', i.e. the
sum of the probabilities over the vocabulary is less than one.  In
order to have a proper probability distribution, we would need to
renormalize by dividing by some factor.  Note, however, that this
renormalization 
factor is necessarily less than one, and thus would uniformly 
increase each word's probability under the model, i.e. any perplexity
results reported below will be higher than the ``true'' perplexity
that would be assigned with a properly normalized distribution.  In
other words, renormalizing would make our perplexity measure lower 
still.  The hope, however, is that the improved parsing model provided
by our conditional probability model will cause the distribution over
structures to be more peaked, thus enabling us to capture more of the
total probability mass, and making this a fairly tight upper bound on
the perplexity.

One final note on assigning probabilities to strings:
because this parser does garden path on a small percentage of
sentences, this must be interpolated with another estimate, to ensure
that every word receives a probability estimate.  In our trials, we
used the unigram, with a very small mixing coefficient:
\begin{eqnarray}
\Pr(w_{i}|w_{0}^{i-1}) &=& \lambda(w_{0}^{i-1})\frac{\sum_{d\in
\scH_{i+1}^{init}}\Pr(d)}{\sum_{d\in \scH_{i}^{init}}\Pr(d)} +
(1-\lambda(w_{0}^{i-1}))\Prhat(w_{i}) 
\end{eqnarray}
If $\sum_{d\in H_{i}^{init}}\Pr(d)$ = 0 in our model, then our model
provides no distribution over following words, since the denominator
is zero.  Thus,
\begin{eqnarray}
\lambda(w_{0}^{i-1}) &=& \left\{ \begin{array}{ll} 0 & 
\mathrm{\it if}\ \sum_{d\in H_{i}^{init}}\Pr(d) = 0\\ .999 & otherwise 
\end{array}
\right.
\end{eqnarray}

\begin{table*}[t]
\begin{tabular}{|l|c|c|c|c|c|r|c|}
\hline
{\small Corpora} & {\small Condi-} & {\small LR} & {\small LP} &
{\small Pct.} & {Perplexity} & 
{\small Avg. rule\ \ } & {\small Average}\\
{} & {\small tioning} & {} & {} & {\small failed} & {} & {\small
expansions\ } & {\small analyses}\\
{} & {} & {} & {} & {} & {} & {\small considered${}^{\dag}$} & {\small advanced${}^{\dag}$}\\\hline
\multicolumn{8}{|c|}{sections 23-24: 3761 sentences $\leq$ 120}\\\hline
{unmodified} & {all} & {85.2} & {85.1} & {1.7} & {} & {7,206} &
{213.5}\\\hline
{no punct} & {all} & {82.4} & {82.9} & {0.2} & {} & {9,717} &
{251.8}\\\hline\hline
{C\&J corpus} & {par+sib} & {75.2} & {77.4} & {0.1} & {310.04} & {17,418} &
{457.2}\\\hline
{C\&J corpus} & {{\small NT} struct} & {77.3} & {79.2} & {0.1} & {290.29} & {15,948} &
{408.8}\\\hline
{C\&J corpus} & {{\small NT} head} & {79.2} & {80.4} & {0.1} & {255.85} & {14,239} &
{363.2}\\\hline
{C\&J corpus} & {POS struct} & {80.5} & {81.6} & {0.1} & {240.37} & {13,591} &
{341.3}\\\hline
{C\&J corpus} & {all} & {81.7} & {82.1} & {0.2} & {152.26} & {11,667} &
{279.7}\\\hline
\end{tabular}\\
\begin{footnotesize}
${}^{\dag}$per word
\end{footnotesize}
\caption{Results conditioning on various contextual events, sections
23-24, modifications following Chelba and Jelinek}\label{tab:res2}
\end{table*}

Chelba and Jelinek \shortcite{Chelba98a,Chelba98b} also used a parser
to help assign word probabilities, via the Structured Language Model
outlined in section \ref{sec:SLM}.  They trained and tested the SLM on
a modified, more ``speech-like'' version of the Penn Treebank.  
Their modifications included: (i) removing orthographic cues to
structure (e.g. punctuation); (ii) replacing all numbers with the
single token {\it N\/}; and (iii) closing the vocabulary at 10,000,
replacing all other words with the UNK token.  They
used sections 00-20 (929,564 words) as the development set, sections
21-22 (73,760 words) as the check set (for interpolation coefficient
estimation), and tested on sections 23-24 (82,430 words).  We obtained
the training and testing corpora from 
them (which we will denote \textttt{C\&J corpus}), and also created 
intermediate corpora, upon which only the first two modifications were
carried out (which we will denote \textttt{no punct})\footnote{Since
the \textttf{C\&J} and \textttf{no punct} corpora do not have a fixed
vocabulary, we still must deal with unknown words in the test
corpora.  This is done by estimating a distribution over POS tags for
unknown words from a held-out corpus.}. Differences in
performance will give an indication of the impact on parser 
performance of the different modifications to the corpora.  All trials 
in this section used sections 00-20 for counts, held out 21-22, and
tested on 23-24, with our base parser from chapter 3.

Table \ref{tab:res2} shows several things.  First, it shows relative
performance for unmodified, no punct, and C\&J 
corpora with the full set of conditioning information.  We can see
that removing the punctuation causes (unsurprisingly) a dramatic drop
in the accuracy and efficiency of the parser.  Interestingly, it also
causes coverage to become nearly total, with failure on just two
sentences per thousand on average.  

\begin{figure*}[t]
\begin{center}
\epsfig{file=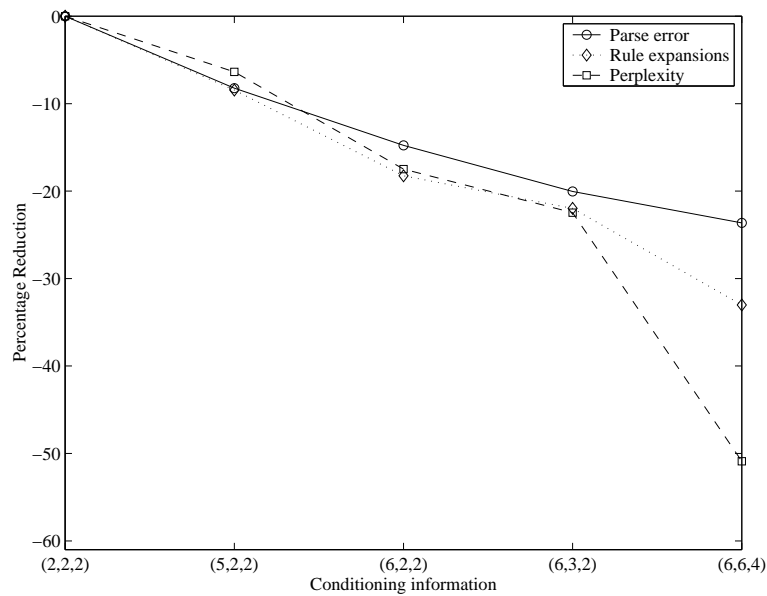, width = 4in}
\end{center}
\caption{Reduction in average precision/recall error, number of
rule expansions, and perplexity as conditioning increases} \label{fig:perp_red}
\end{figure*}

We see the familiar pattern, in the C\&J corpus results, of
improving 
performance as the amount of conditioning information grows.  In this
case we have perplexity results as well, and figure \ref{fig:perp_red} 
shows the reduction in parser error, rule expansions, and perplexity as
the amount of conditioning information grows.  While all three seem to 
be similarly improved by the addition of structural context
(e.g. parents and siblings), the addition of c-commanding heads has
only a moderate effect on the parser accuracy, but a very large effect 
on the perplexity.  The fact that the efficiency was improved more
than the accuracy in this case (as was also seen in figure
\ref{fig:err_red}), seems to indicate that this additional information is
causing the distribution to become more peaked, so that fewer analyses 
are making it into the beam.

\begin{table*}[t]
\begin{tabular} {|p{1.9in}|p{.9in}|p{.9in}|p{.9in}|}
\hline
{\small Paper} & \multicolumn{3}{|c|}{Perplexity}\\\cline{2-4}
{} & {\small Trigram Baseline} &
{\small Model} & {\small Interpolation, $\lambda$=.36}\\\hline
{\namecite{Chelba98a}} & {167.14} & {158.28} & {148.90}\\\hline
{\namecite{Chelba98b}} & {167.14} & {153.76} & {147.70}\\\hline
{Current results} & {167.02} & {152.26} & {137.26}\\\hline
\end{tabular}
\caption{Comparison with previous perplexity results}\label{tab:res3}
\end{table*}

Table \ref{tab:res3} compares the perplexity of our base model with
Chelba and Jelinek \shortcite{Chelba98a,Chelba98b} on the same
training and testing corpora.  We built an interpolated trigram model
to serve as a baseline (as they did), and also interpolated our
model's perplexity with the trigram (replacing the unigram
interpolation), using the same mixing coefficient as
they did in their trials (taking 36 percent of the estimate from the
trigram)\footnote{Our optimal mixture level was closer to 40 percent,
but the difference was negligible.}.  The trigram model was also
trained on sections 00-20 of the C\&J corpus.  Trigrams and bigrams
were binned by the total count of the conditioning words in the
training corpus, and maximum likelihood mixing coefficients were
calculated for each bin, to mix the trigram with bigram and unigram
estimates.  Our trigram model performs at almost exactly the 
same level that theirs does, which is what we would expect.  Our
parsing model's perplexity improves upon \namecite{Chelba98a}
fairly substantially, but is only slightly better than
\namecite{Chelba98b}.  However, when we interpolate with the trigram,
we see that 
the additional improvement is greater than the one they experienced.
This is not surprising, since our conditioning information is in many
ways orthogonal to that of the trigram, insofar as it includes the
probability mass of the derivations;  in contrast, their model in
some instances is very close to the trigram, by conditioning on two
words in the prefix string, which may happen to be the two adjacent
words.

These results are particularly remarkable, given that we did not
build our model as a language model {\it per se\/}, but rather as a
parsing model.  The perplexity improvement was achieved by simply taking the
existing parsing model and applying it, with no extra training beyond
that done for parsing.

The hope was expressed above that our reported perplexity would be
fairly close to the ``true'' perplexity that we would achieve if the
model were properly normalized, i.e. that the amount of probability
mass that we lose by pruning is small.  One way to test this is the
following\footnote{Thanks to Ciprian Chelba for this suggestion.}: at
each point in the sentence, calculate the conditional probability of
each word in the vocabulary given the previous words, and sum them.
If there is little loss of probability mass, the sum should be close
to one.  We did this for the first 10 sentences in the test corpus, a
total of 213 words (including the end-of-sentence markers).  The
parser failed to parse one of
the sentences, so that 12 of the word probabilities (all
of the words after the point of the failure) were not estimated by our
model.  Of the remaining 201 words, the
average sum of the probabilities over the 10,000 word vocabulary was
0.9821, with a minimum of 0.7960, and a maximum of 0.9997.
Interestingly, at the word where the failure occurred, the sum of the
probabilities was 0.9301.

\subsection{Word error rate}
In order to get a sense of whether these perplexity reduction results
can translate to improvement in a speech recognition task, we
performed a very small preliminary experiment on N-best lists.  The
DARPA `93 HUB1 test 
setup consists of 213 utterances read from the Wall St. Journal, a
total of 3446 words.  The corpus comes with a baseline trigram model,
using a 20,000 word open vocabulary, and trained on approximately 40
million words.  We used Ciprian Chelba's A$^\star$
decoder\footnote{See \namecite{Chelba00} for details.} to find the
50 best hypotheses from each lattice, along with the acoustic and
trigram scores.  Given the idealized circumstances of the
production (text read in a lab), the lattices are relatively sparse,
and in many cases 50 distinct string hypotheses were not found in a
lattice.  We reranked an average of 22.9 hypotheses with our
language model per utterance.

\begin{table*}[t]
\begin{tabular} {|p{1.3in}|p{.6in}|p{.7in}|p{.4in}|p{.7in}|p{.6in}|}
\hline
{Model} & {Training Size} & {Vocabulary Size} & {LM Weight} &
{Word Error Rate \%} & {Sentence Error}\\
{} & {} & {} & {} & {} & {Rate \%} \\\hline 
{Lattice trigram} & {40M} & {20K} & {16} & {13.7} & {69.0}\\\hline
{\namecite{Chelba00} ($\lambda$=.4)} & {20M} & {20K} & {16} & {13.0} &
{}\\\hline 
{Current model} & {1M} & {10K} & {15} & {15.1} & {73.2}\\\hline
{Treebank trigram} & {1M} & {10K} & {5} & {16.5} & {79.8}\\\hline
{No language model} & {} & {} & {0} & {16.8} & {84.0}\\\hline
\end{tabular}
\caption{Word and sentence error rate results for various models, with
differing training and vocabulary sizes, for the best language model
factor for that particular model} \label{tab:wer}
\end{table*}

One complicating issue has to do with the tokenization in the Penn
Treebank versus that in the HUB1 lattices.  In particular,
contractions (e.g. \textttt{he's}) are split in the Penn Treebank
(\textttt{he 's}) but not in the HUB1 lattices.  Splitting of the
contractions is critical for parsing, since the two parts oftentimes
(as in the previous example) fall in different constituents.  We
follow \namecite{Chelba00} in dealing with this problem: for parsing
purposes, we use the Penn Treebank tokenization;  for interpolation
with the provided trigram model, and for evaluation, the lattice
tokenization is used.  If we are to interpolate our model with the
lattice trigram, we must wait until we have our model's estimate for
the probability of both parts of the contraction; their product can
then be interpolated with the trigram estimate.  In fact, 
interpolation in these trials made no improvement over the better of
the uninterpolated models, but simply resulted in performance
somewhere between the better and the worse of the two models, so we
will not present interpolated trials here.

Table \ref{tab:wer} reports the word and sentence error rates for
five different models:  (i) the trigram model that comes with the
lattices, trained on approximately 40M words, with a vocabulary of
20,000; (ii) the best performing model from \namecite{Chelba00}, which
was interpolated with the lattice trigram at $\lambda$=0.4; (iii) 
our parsing model, with the same training and vocabulary as the
perplexity trials above; (iv) a trigram model with the same training
and vocabulary as the parsing model; and (v) no language model at all.
This last model shows the performance from the acoustic model
alone, without the influence of the language model.  Recall that the
log of the language model score is multiplied by the language model (LM)
weight when summing the logs of the language and acoustic scores, as a
way of increasing the relative contribution of the language model to
the composite score.  We followed \namecite{Chelba00} in using an
LM weight of 16 for the lattice trigram.  For our model and the
treebank trigram model, the LM weight that resulted in the
lowest error rates is given.

The small size of our training 
data, as well as the fact that we are rescoring N-best lists, rather
than working directly on lattices, makes comparison with the other
models not particularly informative.  What is more informative is the
difference between our model and the trigram trained on the same
amount of data.  We achieved a 1.4 percent improvement in
word error rate, and a 6.6 percent improvement in sentence
error rate over the treebank trigram.  Interestingly, as mentioned
above, interpolating two models 
together gave no improvement over the better of the two, whether our
model was interpolated with the lattice or the treebank trigram.  This
contrasts with our perplexity results reported above, as well as with
the recognition experiments in \namecite{Chelba00}, where the best
results resulted from interpolated models.  We will see an improvement
in WER with interpolation in the next section.

The point of this small experiment was to see if our parsing model could
provide useful information even in the case that recognition errors
occur, as opposed to the (generally) fully grammatical strings upon which
the perplexity results were obtained.  It has been pointed out that, given
that our model relies so heavily on context, it may have difficulty
recovering from even one recognition error, perhaps more difficulty
than a more locally-oriented trigram.  While the improvements over the
trigram model in 
these trials are modest, they do indicate that our model is robust
enough to provide good information even in the face of noisy input.
Subsequent chapters will include more substantial word recognition
experiments.  

\subsection{Beam variation}
\begin{table*}[t]
\begin{tabular}{|l|c|c|c|c|c|r|r|}
\hline
{\small Base} & {\small LR} & {\small LP} &
{\small Pct.} & {Perplexity} & {Perplexity} & 
{\small Avg. rule\ \ } & {\small Words per}\\
{\small Beam} & {} & {} & {\small failed} & {\small $\lambda$=0} &
{\small $\lambda$=.36} & {\small expansions\ } & {\small second\ \ \ \ }\\
{\small Factor} & {} & {} & {} & {} & {} & {\small
considered${}^{\dag}$} & {}\\\hline 
\multicolumn{8}{|c|}{sections 23-24: 3761 sentences $\leq$ 120}\\\hline
{\small $10^{-11}$} & {81.7} & {82.1} & {0.2} & {152.26} & {137.26} & {11,667} &
{3.1}\\\hline
{\small $10^{-10}$} & {81.5} & {81.9} & {0.3} & {154.25} & {137.88} & {6,982} &
{5.2}\\\hline
{\small $10^{-9}$} & {80.9} & {81.3} & {0.4} & {156.83} & {138.69} & {4,154} &
{8.9}\\\hline
{\small $10^{-8}$} & {80.2} & {80.6} & {0.6} & {160.63} & {139.80} & {2,372} &
{15.3}\\\hline
{\small $10^{-7}$} & {78.8} & {79.2} & {1.2} & {166.91} & {141.30} & {1,468} &
{25.5}\\\hline
{\small $10^{-6}$} & {77.4} & {77.9} & {1.5} & {174.44} & {143.05} & {871} &
{43.8}\\\hline
{\small $10^{-5}$} & {75.8} & {76.3} & {2.6} & {187.11} & {145.76} & {517} &
{71.6}\\\hline
{\small $10^{-4}$} & {72.9} & {73.9} & {4.5} & {210.28} & {148.41} & {306} &
{115.5}\\\hline
{\small $10^{-3}$} & {68.4} & {70.6} & {8.0} & {253.77} & {152.33} & {182} &
{179.6}\\\hline
\end{tabular}\\
\begin{footnotesize}
${}^{\dag}$per word
\end{footnotesize}
\caption{Results with full conditioning on the C\&J corpus
at various base beam factors}\label{tab:res4}
\end{table*}

The next set of results that we will present addresses the question of
how wide the beam must be for adequate results.  The base beam factor
that we have used to this point is $10^{-11}$, which is quite wide.
It was selected with the goal of high parser accuracy; but in this new 
domain, parser accuracy is a secondary measure of performance.  
To determine the effect on perplexity, we
varied the base beam factor in trials on the Chelba and Jelinek
corpora, keeping the level of conditioning
information constant, and table \ref{tab:res4} shows the
results across a variety of factors.

\begin{figure*}[t]
\begin{center}
\epsfig{file=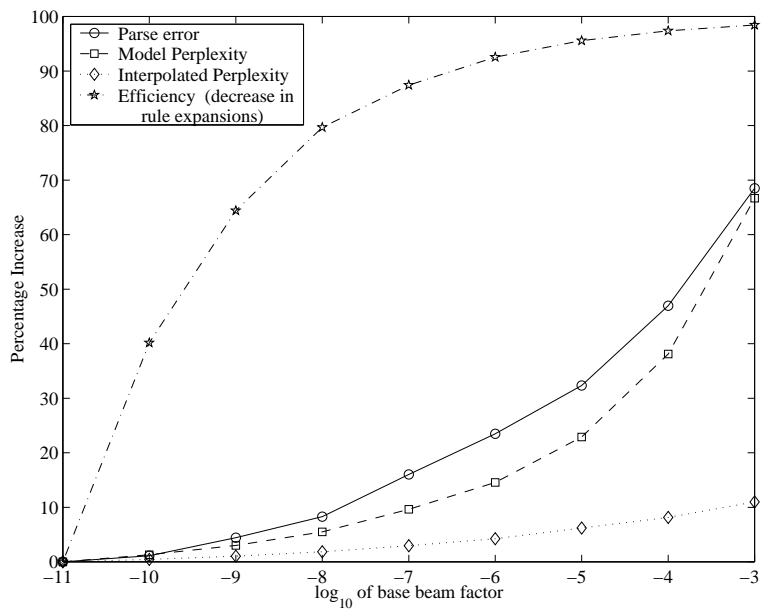, width = 4in}
\end{center}
\caption{Increase in average precision/recall error, model perplexity, 
interpolated perplexity, and efficiency (i.e. decrease in rule expansions
per word) as base beam factor decreases} \label{fig:beam}
\end{figure*}

The parser error, parser coverage, and the uninterpolated model
perplexity ($\lambda$ = 1) all suffered substantially from a narrower
search,  
but the interpolated perplexity remained quite good even at the
extremes.  Figure \ref{fig:beam} plots
the percentage increase in parser error, model perplexity,
interpolated perplexity, and efficiency (i.e. decrease in rule expansions
per word) as the base beam factor decreased.  Note that the model
perplexity and parser accuracy are quite similarly effected, but that
the interpolated perplexity remained far below the trigram baseline,
even with extremely narrow beams.

\subsection{Further perplexity reduction}
The first question that we can ask is, what kind of perplexity
reduction from the results presented in previous sections do we get
by virtue of the model improvements from chapter 
four -- in particular smoothing the grammar.  We took the identical
training and test set and ran the standard parser over it.  The results are
presented in table \ref{tab:resL1}.  This change in the parsing model
resulted in a four point reduction in perplexity over the results
generated by the previous version of the parser.

A second way to improve perplexity that we investigated was a
modification to the way in which the language model scores are mixed.
As it stands, following \namecite{Chelba98a}, we mix our scores on a
word by word basis with a fixed mixing parameter of $\lambda = .36$,
which was the same factor that they used.  To understand how we want
to improve the mixing, let us first discuss the trigram model.

\begin{table*}[t]
\begin{tabular} {|p{1.9in}|p{.9in}|p{.9in}|p{.9in}|}
\hline
{\small Model} & \multicolumn{3}{|c|}{Perplexity}\\\cline{2-4}
{} & {\small Trigram Baseline} &
{\small Model} & {\small Interpolation, $\lambda$=.36}\\\hline
{\namecite{Chelba98a}} & {167.14} & {158.28} & {148.90}\\\hline
{\namecite{Chelba98b}} & {167.14} & {153.76} & {147.70}\\\hline
{PCFG model} & {167.02} & {152.26} & {137.26}\\\hline
{Smoothed grammar} & {167.02} & {148.12} & {134.22}\\\hline
\end{tabular}
\caption{Comparison with previous perplexity results}\label{tab:resL1}
\end{table*}

As mentioned earlier, the trigram model is smoothed through deleted
interpolation.  The n-grams are grouped into sets (or buckets) and
treated as equivalent for the purpose of mixing parameter estimation.
A separate 
mixing parameter is estimated for each bucket, and it is used for all
events that fall in that bucket.  For the trigram used to this point,
we bucketed based on the raw frequency of the event.  In what follows,
we used the score advocated in \namecite{Chen96}, which is the total count 
of the conditioning event, divided by the number of distinct
conditioned events with which the conditioning event was found (which
we will call `Average Count').  For example, in a bigram model, the
`Average Count' of the conditioning word is 
the total count of that word, divided by the number of distinct
bigrams within which it is the first word.  This bucketing score has
been shown, and is shown here, to outperform raw frequency quite
dramatically.  We will use this score to create buckets for the
estimation of the parameters for mixing the raw bigram with the
unigram, the raw trigram with the smoothed bigram, and the smoothed
trigram with the grammatical language models.

We propose an improved method for mixing the parsing language model
and the trigram model together that is a simple extension of the
interpolation that we are already using for the trigram.  We will
decide upon the mixing of the models based upon the ``reliability'' of
the n-gram estimate, in much the same way as we decide upon how much
to mix the raw trigram, bigram, and unigram estimates.  The
idea is to use the buckets defined for the trigram interpolation, and
assign each smoothed trigram estimate a new bucket;  then, using a
held out corpus in the standard way, estimate for each bucket 
the optimal mixing coefficient for mixing the smoothed trigram and the
parser's language model score.  The more reliable the smoothed trigram
estimate, presumably the more it will be relied upon.

The new buckets for the smoothed trigrams are not simply the original
trigram buckets, because an unobserved 
trigram will be totally smoothed to the bigram score,
which has a bucket of its own.  Instead of having a single index, each
bucket will be identified by a pair of indices (i,j).  We reserve
(0,0) for sentence initial prediction.  If the trigram bucket is
greater than zero (i.e. the two previous words have been seen
together), i = 0 and j is the trigram bucket; otherwise, i = 1 and j
is the bigram bucket. This gives us a measure of the reliability of
the smoothed trigram. For example, if the bucket for smoothing the
trigram to bigram is bucket 10, the bucket for smoothing from trigram
to the grammatical language model is (0,10).  If the bucket for
smoothing from trigram to bigram is 0, and the bucket for smoothing
from bigram to unigram is 5, the bucket for smoothing from trigram
to the grammatical language model is (1,5).  Using these buckets, we
then estimated the optimal mixing coefficients for the two language
model probability estimates.

\begin{table*}[t]
\begin{tabular} {|l|l|r|r|r|}
\hline
{\small Trigram model} & {Mixing} & \multicolumn{3}{|c|}{Perplexity}\\\cline{3-5}
{} & {} & {\small Trigram Baseline} &
{\small Parsing Model} & {\small Interpolation}\\\hline
{Frequency buckets} & {$\lambda = .36$} & {167.02} & {148.12} & {134.22}\\\hline
{Average Count buckets} & {$\lambda = .36$} & {158.49} & {148.12} & {131.64}\\\hline
{Average Count buckets} & {Variable} & {158.49} & {148.12} & {130.24}\\\hline
\end{tabular}
\caption{Perplexity results when mixing the new parsing model with (i)
the old trigram; (ii) a better trigram with Average Count bucketing;
and (iii) the better trigram with variable mixing.}\label{tab:resL2}
\end{table*}

Table \ref{tab:resL2} gives the results of our trials with different
mixing methods.  Moving from the raw frequency based bucketing to the
Average Count bucketing improved our trigram performance, and
correspondingly the performance of the mixture.  Mixing the parsing
model and the improved trigram with a variable mixing coefficient
based on the trigram bucketing did improve perplexity slightly, but
not as much as we might have hoped.  One possible reason for this is
the fact that in this circumstance the trigram is the weaker of the
two models, so that we are deciding when and how much to use the
stronger of the two models, rather than the other way around.  It
would probably be advisable to try to find some way to decide, given
the parser state, how much to smooth to the trigram.  This, however,
is not quite as simple as using the n-gram buckets.

Next we perform more substantial n-best re-ranking trials, to see if
our language model can improve word error rate over a trigram model.

\subsection{Further word error rate trials}
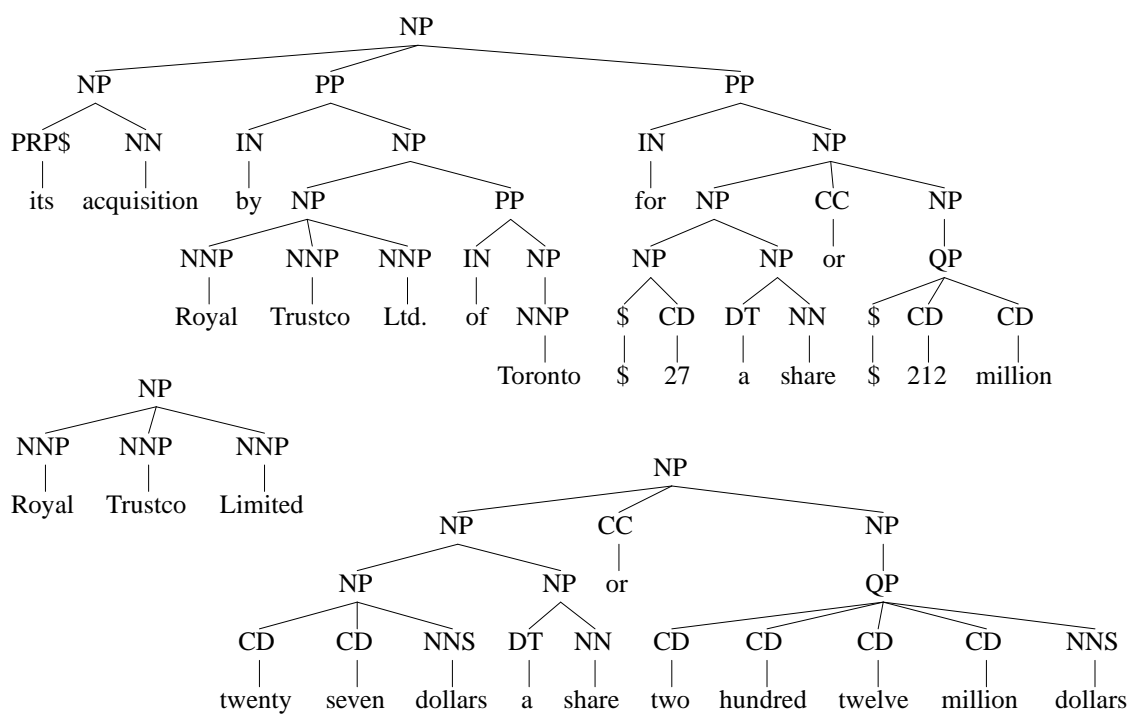
\begin{figure}[t]
\begin{small}
\begin{picture}(397,144)(0,-144)
\put(147,-8){\small NP}
\drawline(154,-12)(32,-22)
\put(25,-30){\small NP}
\drawline(32,-34)(12,-44)
\put(-0,-52){\small PRP\$}
\drawline(12,-56)(12,-66)
\put(7,-74){its}
\drawline(32,-34)(51,-44)
\put(43,-52){\small NN}
\drawline(51,-56)(51,-66)
\put(27,-74){acquisition}
\drawline(154,-12)(121,-22)
\put(115,-30){\small PP}
\drawline(121,-34)(90,-44)
\put(85,-52){\small IN}
\drawline(90,-56)(90,-66)
\put(85,-74){by}
\drawline(121,-34)(151,-44)
\put(144,-52){\small NP}
\drawline(151,-56)(112,-66)
\put(106,-74){\small NP}
\drawline(112,-78)(75,-88)
\put(64,-96){\small NNP}
\drawline(75,-100)(75,-110)
\put(62,-118){Royal}
\drawline(112,-78)(114,-88)
\put(104,-96){\small NNP}
\drawline(114,-100)(114,-110)
\put(98,-118){Trustco}
\drawline(112,-78)(150,-88)
\put(139,-96){\small NNP}
\drawline(150,-100)(150,-110)
\put(141,-118){\small Ltd.}
\drawline(151,-56)(189,-66)
\put(183,-74){\small PP}
\drawline(189,-78)(176,-88)
\put(171,-96){\small IN}
\drawline(176,-100)(176,-110)
\put(172,-118){of}
\drawline(189,-78)(202,-88)
\put(195,-96){\small NP}
\drawline(202,-100)(202,-110)
\put(191,-118){\small NNP}
\drawline(202,-122)(202,-132)
\put(184,-140){Toronto}
\drawline(154,-12)(276,-22)
\put(270,-30){\small PP}
\drawline(276,-34)(242,-44)
\put(237,-52){\small IN}
\drawline(242,-56)(242,-66)
\put(236,-74){for}
\drawline(276,-34)(310,-44)
\put(303,-52){\small NP}
\drawline(310,-56)(266,-66)
\put(259,-74){\small NP}
\drawline(266,-78)(242,-88)
\put(235,-96){\small NP}
\drawline(242,-100)(232,-110)
\put(229,-118){\$}
\drawline(232,-122)(232,-132)
\put(229,-140){\$}
\drawline(242,-100)(252,-110)
\put(245,-118){\small CD}
\drawline(252,-122)(252,-132)
\put(247,-140){27}
\drawline(266,-78)(290,-88)
\put(283,-96){\small NP}
\drawline(290,-100)(277,-110)
\put(270,-118){\small DT}
\drawline(277,-122)(277,-132)
\put(275,-140){a}
\drawline(290,-100)(302,-110)
\put(294,-118){\small NN}
\drawline(302,-122)(302,-132)
\put(291,-140){share}
\drawline(310,-56)(311,-66)
\put(304,-74){\small CC}
\drawline(311,-78)(311,-88)
\put(307,-96){or}
\drawline(310,-56)(353,-66)
\put(347,-74){\small NP}
\drawline(353,-78)(353,-88)
\put(347,-96){\small QP}
\drawline(353,-100)(326,-110)
\put(324,-118){\$}
\drawline(326,-122)(326,-132)
\put(324,-140){\$}
\drawline(353,-100)(347,-110)
\put(339,-118){\small CD}
\drawline(347,-122)(347,-132)
\put(339,-140){212}
\drawline(353,-100)(381,-110)
\put(373,-118){\small CD}
\drawline(381,-122)(381,-132)
\put(365,-140){million}
\end{picture}
\begin{picture}(424,122)(0,-122)
\put(48,0){NP}
\drawline(55,-4)(13,-14)
\put(2,-22){NNP}
\drawline(13,-26)(13,-36)
\put(0,-44){Royal}
\drawline(55,-4)(52,-14)
\put(41,-22){NNP}
\drawline(52,-26)(52,-36)
\put(36,-44){Trustco}
\drawline(55,-4)(96,-14)
\put(85,-22){NNP}
\drawline(96,-26)(96,-36)
\put(79,-44){Limited}

\put(243,-30){NP}
\drawline(250,-34)(169,-44)
\put(162,-52){NP}
\drawline(169,-56)(131,-66)
\put(124,-74){NP}
\drawline(131,-78)(94,-88)
\put(86,-96){CD}
\drawline(94,-100)(94,-110)
\put(79,-118){twenty}
\drawline(131,-78)(131,-88)
\put(123,-96){CD}
\drawline(131,-100)(131,-110)
\put(119,-118){seven}
\drawline(131,-78)(167,-88)
\put(156,-96){NNS}
\drawline(167,-100)(167,-110)
\put(153,-118){dollars}
\drawline(169,-56)(208,-66)
\put(201,-74){NP}
\drawline(208,-78)(196,-88)
\put(188,-96){DT}
\drawline(196,-100)(196,-110)
\put(193,-118){a}
\drawline(208,-78)(221,-88)
\put(213,-96){NN}
\drawline(221,-100)(221,-110)
\put(209,-118){share}
\drawline(250,-34)(230,-44)
\put(222,-52){CC}
\drawline(230,-56)(230,-66)
\put(225,-74){or}
\drawline(250,-34)(330,-44)
\put(323,-52){NP}
\drawline(330,-56)(330,-66)
\put(323,-74){QP}
\drawline(330,-78)(250,-88)
\put(243,-96){CD}
\drawline(250,-100)(250,-110)
\put(242,-118){two}
\drawline(330,-78)(286,-88)
\put(278,-96){CD}
\drawline(286,-100)(286,-110)
\put(268,-118){hundred}
\drawline(330,-78)(328,-88)
\put(320,-96){CD}
\drawline(328,-100)(328,-110)
\put(313,-118){twelve}
\drawline(330,-78)(369,-88)
\put(361,-96){CD}
\drawline(369,-100)(369,-110)
\put(352,-118){million}
\drawline(330,-78)(410,-88)
\put(399,-96){NNS}
\drawline(410,-100)(410,-110)
\put(395,-118){dollars}
\end{picture}
\end{small}
\caption{Original treebank tree, and the restructured constituents.
Everything except the changes shown remained the same in our transform
of the training corpus.}\label{fig:align}
\end{figure}

There were several factors that made the word error rate results in
the previous section not a full-blown test of the model.  First, the
parsing model used was the unsmoothed PCFG model, not the more
effective model with the smoothed Markov grammar.  Next, the
vocabulary size was kept at 10,000 words, rather than the 20,000 for
the other models.  Third, in ways that we will explain, the format of
the training corpus was not perfectly aligned with the format of the
test corpus.  Lastly, we only trained on the approximately one million
word treebank, whereas the other models were trained on 20-40 times
that much data.

To remedy this, we first created a training corpus that was aligned
with the spoken language input from the n-best lists.  We wanted to
change the lexical items of the treebank, while preserving the phrase
structure intact.  The major differences between the newspaper text
and the read text are: (i) numbers, currencies, and dates; (ii) common
abbreviations; (iii) acronyms; and (iv) hyphen and period delimited
items.  Figure \ref{fig:align} shows the changes to two constituents
due to the tree transformations that we performed.  The following list
illustrates some common changes that were made.

\begin{itemize}
\item currency: \$ 3.75 $\rightarrow$ three dollars and seventy five cents
\item numbers: 2.1 \% $\rightarrow$ two point one percent
\item dates: Nov. 30 , 1999 $\rightarrow$ November thirtieth ,
nineteen ninety nine
\item abbreviations: Inc. $\rightarrow$ Incorporated
\item acronyms: IBM $\rightarrow$ I. \ B. \ M.
\item hyphens: top-yielding $\rightarrow$ top yielding
\item periods: U.S. $\rightarrow$ U. \ S.
\end{itemize}

With the exception of the hyphens, these transformations can be
carried out deterministically.  This is because the read format
for numbers are easily modeled with a small number of rules, and the
parts-of-speech for all output are predictable from the original
part-of-speech.  For 
example, acronyms are typically tagged NNP, and the resulting initials
also are tagged NNP.  The hyphenated words, however, are most commonly
adjectives which decompose into non-adjective words.  We chose the
most common tag for each of the words, or the original tag of the
hyphenated item for previously unobserved words.

As in the previous trials, there is still a tokenization mis-alignment
between what is required for parsing and what is standardly used in
language models.  Namely, contractions and possessives
(e.g. \textttt{John's}) are split for parsing but are a single token
for the language model.  As stated previously, we will split these for
parsing, but for interpolation with the lattice trigram or for
evaluation, they were treated as a single token.

We expanded our vocabulary to a 20,000 word open vocabulary, which is
the standard from NIST for this test set, replacing all items outside
of the lexicon with UNK.  The
DARPA `93 HUB1 test 
setup consists of 213 utterances read from the Wall St. Journal, a
total of 3446 words.  We then parsed the n-best lists extracted from
the HUB1 test set, training on the
new transformed training corpus for sections 0-20, and transformed
held-out sections 21-22.  We used the smoothed Markov grammar model
given in the previous chapter.  Table \ref{tab:werii1} gives word
error rate results for the model at various interpolation values with
the lattice trigram, where the $\lambda$ contribution is from the
trigram, and 1-$\lambda$ from our parsing model.  Recall that the
lattice trigram is trained on 40 million words, while our parser is
trained on a 961,786 word training corpus with 76,563 words held out.
For all WSJ trials, we used a language model weight of 16
and no word insertion penalty.

\begin{table}
\begin{center}
\begin{tabular}{|l|c|c|c|c|c|c|}
\hline
$\lambda$ & 0.0 & 0.2 & 0.4 & 0.6 & 0.8 & 1.0\\\hline
WER \% & 14.0 & 14.0 & 13.7 & 13.6 & 13.4 & 13.7\\\hline
\end{tabular}
\end{center}
\caption{Word Error Rate results on the WSJ HUB1 test set, with a 20k
vocabulary, trained on the WSJ treebank, at various mixtures with the
lattice trigram.  The model is $\lambda$ times the trigram plus
(1-$\lambda$) times the parsing model. }\label{tab:werii1} 
\end{table}

Our model on its own gets a WER percent of 14.0, a 1.1 percent error
reduction from the performance presented in the previous section.  As we
interpolate with the lattice trigram, the performance improves, so
that at $\lambda$ = 0.8, we are improving on the lattice trigram with
a modest 0.3 percent reduction in error.

The last problem with the previous word error rate results has to do
with the amount of training data.  We are limited by the amount of
annotated data, which is about a million words.  To increase the
amount of training data, we performed a single pass of Expectation
Maximization (EM) on an additional 1.2 million words (for more on EM
and the Inside/Outside algorithm, see e.g. \egcite{Manning99}.  While the
potential amount of training data is very large (the lattice trigram
is trained on 40 million words, and \namecite{Chelba00} used EM to
train on 20 million words), the number of parameters in our model made
the memory requirements of training on that much additional data
prohibitive.  A trigram model with a 20,000 word vocabulary has on the
order of $10^{13}$ parameters.  Our conditional probability model,
with approximately 80 non-terminals and 20,000 words, has on the order
of $10^{22}$ parameters.  One future area of research is to reduce the
number of parameters without sacrificing the level of performance, and
so be able to train on much larger data sets.  For the purposes of
this thesis, however, we will investigate EM on just a million more
words.

EM works by taking an estimate of the parameters, and finding the
expected counts of hidden data.  It then uses the expected counts of
the hidden data to find a maximum likelihood estimate for the
parameters.  In our case, the hidden data are parse trees.  The initial
parameter values are those estimated from the treebank; the expected
count for a hidden event is the probability of the event divided by
the probability of the string.  Since the probability of the string is
estimated in our model by the sum of the probabilities of all parse
trees found for the string, the expected count for a particular parse
tree is its probability normalized by the total probability of the
trees in the beam.  

In practice, we train the model as follows.  First, we train the
parser as usual with the given training corpus, and parse the
additional data, in our case 1.2 million words.  For each string in
this additional data, parse trees are saved with their normalized
probabilities\footnote{To save space, we saved all parse trees until
99 percent of the probability mass is accounted for.  All remaining
candidate trees were discarded.}.  These normalized probabilities are
the expected frequency of the parse tree.
We then train the model with both
the given treebank and the parse trees from this additional data,
which we will call the estimated treebank.  The count
for each event in the given treebank is 1; each event in the estimated
treebank is given as its count the expected frequency of the parse
tree within which it occurs.  Thus, if we observe the PCFG rule
S~$\rightarrow$~NP~VP in the given treebank, we increment its count by
1; if we observe it in the estimated treebank, we increment its count
by the normalized probability of the tree within which it occurs.
With these counts we then perform maximum likelihood relative
frequency estimation.  This is the standard EM method.

\begin{table}[t]
\begin{center}
\begin{tabular}{|l|c|c|c|c|c|c|}
\hline
$\lambda$ & 0.0 & 0.2 & 0.4 & 0.6 & 0.8 & 1.0\\\hline
WER \% & 13.2 & 13.1 & 12.8 & 12.7 & 13.0 & 13.7\\\hline
\end{tabular}
\end{center}
\caption{Word Error Rate results on the WSJ HUB1 test set, with a 20k
vocabulary, trained on the WSJ treebank plus the hidden data from an
additional 1.2 million words, at various mixtures with the
lattice trigram.  The model is $\lambda$ times the trigram plus
(1-$\lambda$) times the parsing model. }\label{tab:werii2} 
\end{table}

Table \ref{tab:werii2} gives the word error rate results on the HUB1
test set with the additional 1.2 million words of training.  EM
reduced the word error rate of our model by 0.8 percent.
Uninterpolated, our model now improves upon the lattice trigram, reducing
the error by 0.5 percent.  Interpolated with the lattice trigram, the
best model (at $\lambda$ = 0.6) reduces the word error rate by 1.0
percent from the lattice trigram rate.  Further, our 12.7 percent is
better than the best performing model reported in \namecite{Chelba00},
which had a 13.0 percent WER.  This despite only training on a
fraction of the training data of the other models.

For the Switchboard trials, we tested on 2,427 utterances (20,639 words)
from the CLSP WS97 test set \cite{CLSP97}.  Lattices from each of the
test utterances were provided by CLSP, from which the 50 best
hypotheses were extracted, along with their lattice trigram and
acoustic scores.  Some of the test utterances were part of the new
Switchboard treebank, so their trees were removed from the training
data for these trials.  The parsing model that we used for these
trials is the one presented in chapter four, with the additional
conditioning functions for EDITED parallelism.

Table \ref{tab:werii3} gives the word error rate results for the
parsing model trained on the given treebank, which, after punctuation
and utterances from the test section were removed, contained 766,268
words.  The held out corpus was the same as in the parsing trials,
56,293 words with punctuation removed.  The language models were used
with a language model weighting factor of 12, and a word insertion
penalty of 10.  The results do not improve
upon the lattice trigram.  Table \ref{tab:werii4} gives the results
with a parsing model trained with an additional 900,682 words using
EM.  Here we do get a small improvement over the trigram model, when
the two models are mixed at $\lambda$ = 0.8, a 0.2 percent
reduction in the error rate.

\begin{table}[t]
\begin{center}
\begin{tabular}{|l|c|c|c|c|c|c|}
\hline
$\lambda$ & 0.0 & 0.2 & 0.4 & 0.6 & 0.8 & 1.0\\\hline
WER \% & 42.4 & 40.0 & 40.1 & 40.0 & 40.0 & 39.1\\\hline
\end{tabular}
\end{center}
\caption{Word Error Rate results on the Switchboard test set, with a 22k
vocabulary, trained on the Switchboard treebank, at various mixtures with the
lattice trigram.  The model is $\lambda$ times the trigram plus
(1-$\lambda$) times the parsing model. }\label{tab:werii3} 
\end{table}

\begin{table}[t]
\begin{center}
\begin{tabular}{|l|c|c|c|c|c|c|}
\hline
$\lambda$ & 0.0 & 0.2 & 0.4 & 0.6 & 0.8 & 1.0\\\hline
WER \% & 39.4 & 39.1 & 39.0 & 39.0 & 38.9 & 39.1\\\hline
\end{tabular}
\end{center}
\caption{Word Error Rate results on the Switchboard test set, with a 22k
vocabulary, trained on the Switchboard treebank plus additional 900k words, at
various mixtures with the lattice trigram.  The model is $\lambda$
times the trigram plus (1-$\lambda$) times the parsing
model. }\label{tab:werii4}  
\end{table}

\subsection{Grammaticality bias}
As mentioned earlier, it is critical to evaluate a parser-based
language model in terms of how well it reduces word error rate,
because the parser is likely to be sensitive to recognizer errors, and
hence be of limited use when scoring a typically very noisy set of
hypotheses.  The question is: does the parser-based model have a gross
preference for well-formed input, limiting its applicability?  We have
shown that the parser can provide useful information, and reduce word
error rate in the face of noisy hypotheses, but this section will take
a closer look at this issue via a simple additional experiment. 

What we want to do is to look at the difference in the performance of
the models when the correct string is present and when the correct
string is not present.  If the drop in performance of the parsing
language model is greater than the drop in performance of the lattice
trigram, then this is evidence for a grammaticality bias.  To do this,
we narrowed our test set to just those sentences with the correct
string among the n-best hypotheses.  We then tested the models both
with the correct string included and with the correct string
excluded. 

The number of strings in the test sets for which the correct string
was among the 50 best was about half of the total for both the WSJ and
the Switchboard trials.  In the case of Switchboard, however, the
largest number of such strings were single word utterances, which is
too short to be germane to the question at hand.  For that reason, we
set lower bounds on the sentence length, and tested at various
lower bounds.  Given the number of very short strings, it is
perhaps surprising that a grammatical model can improve on an n-gram
model at all in this domain.

\begin{table}[t]
\begin{center}
\begin{small}
\begin{tabular}{|l|c|c|c|c|c|c|c|c|}
\hline
Test & Sentences & Minimum & \multicolumn{3}{|c|}{Trigram Model} &
\multicolumn{3}{|c|}{Parsing Model (EM)}\\
&& Length & \multicolumn{2}{|c|}{with correct} & without &
\multicolumn{2}{|c|}{with correct} & without\\
&&& WER & SER & WER & WER & SER & WER \\\hline
HUB1 (WSJ) & 100 &  & 5.1 & 42.0 & 9.5 & 4.7 & 43.0 &
9.1\\\hline\hline
Switchboard & 187 & 7 & 13.0 & 67.4 & 17.2 & 13.8 & 69.0 &
17.7\\\hline 
Switchboard & 146 & 8 & 12.1 & 68.5 & 15.9 & 12.9 & 69.9 &
16.4\\\hline
Switchboard & 118 & 9 & 11.5 & 69.5 & 15.2 & 11.4 & 69.5 &
14.9\\\hline 
Switchboard & 86 & 10 & 10.7 & 70.9 & 13.7 & 11.3 & 70.9 &
14.0\\\hline 
\end{tabular}
\end{small}
\end{center}
\caption{Word error rate (WER) and Sentence error rate (SER) for both
HUB1 and Switchboard, when the correct hypothesis was included in the
50 best, and also excluded from those same lists.  Results were
grouped by the minimum length of the correct strings within the set.}
\label{tab:grammtest}
\end{table}

Table \ref{tab:grammtest} summarizes the results of our trials.  The
parsing model that we used for the trials was after one iteration of
EM training.  One
notices several things from this test.  First, the WER for these
trials is very much below the WER for the entire test set, in both
domains.  The reason for this presumably has to do with the fact that
the correct hypothesis falls within the 50 best.  If that is the case,
then presumably many close neighbors to the correct string are also
within the 50 best, so 
that the overall performance, even without selecting the correct
string, will be better than when the correct hypothesis falls outside
of the 50-best list.  Also note that the sentence error rate (SER) for
the trigram is always equal to or less than that for the parsing
model, which is the opposite of what you would expect if there were a
strong grammaticality bias in the parsing model.  

In the Wall St. Journal trial, the change in WER when the correct
string was removed was identical for both the trigram and the parsing
model.  For the various Switchboard trials, the change was in fact
0.2 or 0.3 percent greater for the trigram model than for the parsing
model, which is in the wrong direction for a strong grammaticality
bias in the parsing model.  Given the small number of strings in these
trials, these 
results cannot be called conclusive, but they add to the evidence that
there is not a large grammaticality bias in the model.

One last comment about the results in table \ref{tab:grammtest}.  In
the Switchboard trials, the WER actually improves as the set is
narrowed to include only longer strings.  It appears that the longer
sentences in our smaller test set are ``easier'' in some sense than
the shorter sentences.  In order to see if this generalizes to the
test set as a whole, we looked at the 816 strings in the Switchboard
test set that 
had a length greater than or equal to 10.  On this subset, the lattice
trigram scored a WER of 39.3, 0.2 percent worse than on the test set
as a whole.  The parsing model, in contrast, scored 39.0 on this
subset, 0.4 percent better than its overall score, and better than the
lattice trigram for this subset.  The improvement on the very small
sets that we reported on in table \ref{tab:grammtest} was much larger
than this as the length of the sentences grew, so this does not 
appear to be a general phenomenon, but rather something about this
subset.  Perhaps the reason is that, the longer the string, the more
densely packed the competitors, so that if the correct string makes
the 50-best list, the acoustic scores must really favor the good
words, and hence improve the scores in general.

To sum up this section, it appears unlikely that the parsing model
that we have been exploring has a gross bias towards grammaticality
that would hinder its use within speech recognition systems.

\section{Chapter summary}
In summary, we have defined a language model for speech recognition
that is a direct application of our existing parser model.  We have
demonstrated both perplexity and word error rate reduction over
trigram models.  In the case of the WSJ trials, we improved on the
trigram even with just 5 percent of its training data.  In the
Switchboard trials, we successfully used EM to improve our model to
the point where we could improve on the lattice trigram.  Improvements
in our ability to model the Switchboard utterances syntactically may
lead to further improvements.  Overall, this model has proven useful
even in the face of ill-formed input.

\chapter{Conclusion}
We have presented a parser in this thesis that provides fully
connected syntactic structure incrementally.  It is a model that is
consistent with models of human sentence processing \cite{Jurafsky96},
yet which handles syntactic complexity without resorting to dynamic
programming, while scaling up to handle freely occurring language.  In
addition to its psycholinguistic motivations, we found 
a large computational motivation in the model's applicability to
language modeling for speech recognition.  Our robust parser can cover
spontaneous spoken language input without the need for pre-processing,
and the language model that it provides yields substantial
reductions in perplexity and word error rate on standard test corpora.

There are several directions to go with this research in the future.
One is to attempt to improve the model, by reducing the number of
parameters and better modeling the dependencies.  The linear order of
interpolation, and even the use of interpolation as opposed to other
smoothing methods, is one clear area of potential weakness.  Maximum
entropy modeling is an obvious candidate to replace our current
approach.

Another direction is to modify the model to better fit the
needs of statistical speech recognition.  In this thesis, we have
applied the parser straight ``out of the box'', yet the desideratum is
quite different from that which the model was built to deliver.  If
our goal is language modeling and not parsing, perhaps we can modify
the structure of the grammar in ways that do not preserve the original
structure, but which capture the dependencies as well or better.
Perhaps some of these modifications may also reduce the search problem
as well.
We demonstrated the efficacy of the selective left-corner transform,
in a flattened form that preserved constituent structure.  Perhaps
other transforms that do not preserve constituent structure would be
even more beneficial.  If recovery of Penn Treebank style structure is
not the goal, then such modifications can be carried out without worry.

Given the results of this thesis, as well as what has been seen
recently in the literature from Chelba and Jelinek, it is clear that
syntactic parsing can provide access to dependencies that elude
standard language models.  We speculate that parsing will play a role
in the language models of commonly available speech recognition
systems within the next five to ten years.  Clearly there is much work
to be done to make such models viable under the processing demands of
speech recognition.  This thesis should illustrate that this
additional work will be well worth the effort.

\clearpage
\bibliographystyle{fullname}
\addcontentsline{toc}{chapter}{References}
\bibliography{ber}
\end{document}